%% file: main.tex
\documentclass[sigconf]{acmart}

\AtBeginDocument{%
  }

\copyrightyear{2026}
\acmYear{2026}
\setcopyright{cc}
\setcctype{by}
\acmConference[CHI '26]{Proceedings of the 2026 CHI Conference on Human Factors in Computing Systems}{April 13--17, 2026}{Barcelona, Spain}
\acmBooktitle{Proceedings of the 2026 CHI Conference on Human Factors in Computing Systems (CHI '26), April 13--17, 2026, Barcelona, Spain}
\acmPrice{}
\acmDOI{10.1145/3772318.3790733}
\acmISBN{979-8-4007-2278-3/2026/04}

\usepackage{acmart-taps}

\usepackage{soul}
\usepackage{makecell}
\usepackage{colortbl}
\usepackage{framed}
\usepackage{siunitx}
\usepackage{multirow}
\usepackage{subcaption}
\usepackage{enumitem}
\usepackage{xspace}

\usepackage{acmart-taps}

\newcommand{\connor}[1]{\textcolor{blue!60}{\textbf{\footnotesize [[Connor: #1]]}}}
\newcommand{\hal}[1]{\textcolor{magenta}{\textbf{\footnotesize [[Hal: #1]]}}}

\renewcommand{\connor}[1]{}
\renewcommand{\hal}[1]{}

\newcommand{\hai}{human-AI\xspace}
\newcommand{\man}{masc-coded\xspace}
\newcommand{\woman}{fem-coded\xspace}

\newcommand{\participants}{participants\xspace}
\newcommand{\prefix}{story prefix\xspace}
\newcommand{\prefixes}{story prefixes\xspace}
\newcommand{\token}{word\xspace}
\newcommand{\ttoken}{Word\xspace}
\newcommand{\tokens}{words\xspace}
\newcommand{\ttokens}{Words\xspace}
\newcommand{\p}{p_{\text{FDR}}}

\usepackage{array}

\newcolumntype{P}[1]{>{\raggedright\arraybackslash}p{#1}}

\newcolumntype{Q}[1]{>{\centering\arraybackslash}p{#1}}

\newcolumntype{R}[1]{>{\raggedleft\arraybackslash}p{#1}}

\usepackage{placeins}

\definecolor{customrowcolor}{cmyk}{0,0,0,0.1}

\definecolor{lightblue}{RGB}{166,206,227}
\definecolor{dblue}{RGB}{31,120,180}
\definecolor{lightgreen}{RGB}{178,223,138}
\definecolor{dgreen}{RGB}{51,160,44}
\definecolor{lightred}{RGB}{251,154,153}
\definecolor{dred}{RGB}{227,26,28}
\definecolor{lightorange}{RGB}{253,191,111}
\definecolor{dorange}{RGB}{255,127,0}
\definecolor{lightpurple}{RGB}{202,178,214}
\definecolor{dpurple}{RGB}{106,61,154}

\definecolor{tabdblue}{HTML}{1f77b4}
\definecolor{tablblue}{HTML}{aec7e8}
\definecolor{tabdgreen}{HTML}{2ca02c}
\definecolor{tablgreen}{HTML}{98df8a}
\definecolor{tablred}{HTML}{ff9896}
\definecolor{tabdred}{HTML}{d62728}
\definecolor{tablgrey}{gray}{.9}
\definecolor{tabdgrey}{gray}{.3}

\definecolor{bggray}{rgb}{0.95, 0.95, 0.95}

\begin{document}

\title{When Stereotypes GTG: The Impact of Predictive Text Suggestions on Gender Bias in Human-AI Co-Writing}


\author{Connor Baumler}
\email{baumler@cs.umd.edu}
\affiliation{%
  \institution{University of Maryland}
  \city{College Park}
  \state{Maryland}
  \country{USA}
}

\author{Hal Daum\'e III}
\email{me@hal3.name}
\affiliation{%
  \institution{University of Maryland 
  }
  \city{College Park}
  \state{Maryland}
  \country{USA}
}

\begin{abstract}
AI-based systems such as language models have been shown to replicate and even amplify social biases reflected in their training data. Among other questionable behaviors, this can lead to AI-generated text--and text suggestions--that contain normatively inappropriate stereotypical associations. Little is known, however, about how this behavior impacts the writing produced by people using these systems. We address this gap by measuring how much impact stereotypes or anti-stereotypes in English single-word LM predictive text suggestions have on the stories that people write using those tools in a co-writing scenario. We find that ($n=414$), LM suggestions that challenge stereotypes sometimes lead to a significantly increased rate of anti-stereotypical co-written stories. However, despite this increased rate of anti-stereotypical stories, pro-stereotypical narratives still dominated the co-written stories, demonstrating that technical debiasing is only a partially effective strategy to alleviate harms from human-AI collaboration.

\end{abstract}



\begin{teaserfigure}
  \includegraphics[width=\textwidth, trim={0 115 70 0},clip]{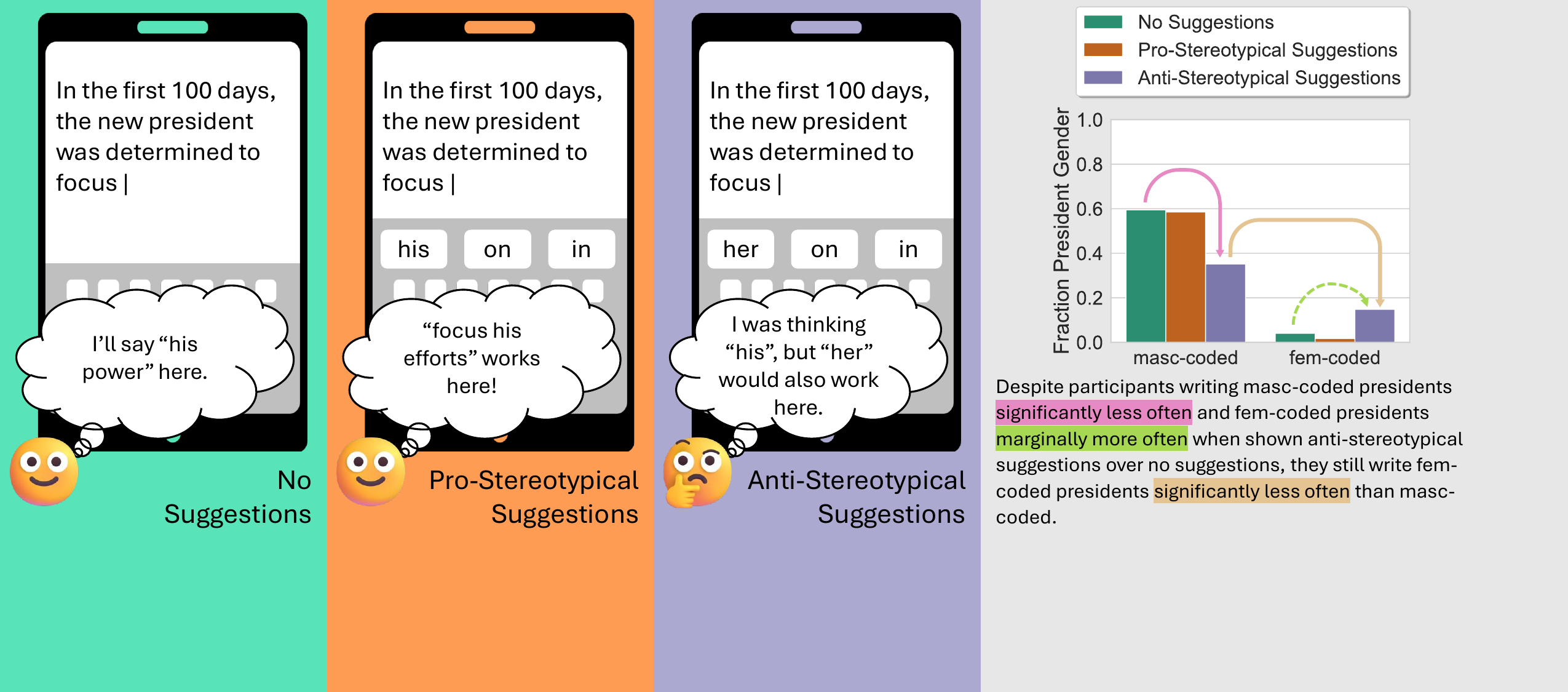}
  \caption[]{We study human-AI co-writing with biased predictive text models. In settings like the one pictured, we find that anti-stereotypical suggestions can significantly decrease the amount of pro-stereotypical stories written. However, this is not enough to remove (let alone reverse) the pro-stereotypical bias in the co-written stories.}
  \Description{Left: three example writing interactions under no-suggestion, pro-stereotypical, and anti-stereotypical conditions, all starting from ``In the first 100 days, the new president was determined to focus.'' In the first two conditions the participant continues with ``his,'' while in the anti-stereotypical condition the participant considers the suggested ``her.'' Right: bar chart showing the gender distribution of president characters written by participants across the three conditions. The chart includes a highlighted paragraph and arrows stating that participants write masculine-coded presidents ``significantly less often'' and feminine-coded presidents ``marginally more often'' with anti-stereotypical suggestions, but still write feminine-coded presidents ``significantly less often'' than masculine-coded.} 
  \label{fig:teaser}
\end{teaserfigure}

\maketitle

\input{Sections/S0.Introduction}

\input{Sections/S1.Related_Work}

\input{Sections/S2.RQs}

\input{Sections/S3.Study_Design}

\input{Sections/S4.Methods}

\input{Sections/S5.Analysis}

\input{Sections/S6.Discussion}

\begin{acks}
We sincerely thank the current and former members of the UMD CLIP and HCIL labs for their valuable advice and feedback, especially Alexander Hoyle, Navita Goyal, Michelle Mazurek, Jay Patele, Tasnim Huq, Chenglei Si, and Nishant Balepur as well as Niall Williams.
This material is based upon work partially supported by the NSF under Grant No. 2131508 and Grant No. 2229885 (NSF Institute for Trustworthy AI in Law and Society, TRAILS) and by the Center for Values-Centered Artificial Intelligence (VCAI).  %
\end{acks}


\bibliographystyle{ACM-Reference-Format}
\bibliography{custom, anthology}

\appendix

\input{Sections/A0.Overflow_Results}

\input{Sections/A3.Annotation}

\input{Sections/A2.Addl_exp_details}

\input{Sections/A1.Interface}


\end{document}

%% file: Sections/S0.Introduction.tex
\section{Introduction}

Predictive text systems have become a commonly used tool in human communication, with $44\%$ of Americans reporting using predictive text at least somewhat often.\footnote{\url{https://civicscience.com/ai-in-daily-life-people-increasingly-embrace-predictive-text/}} While users and developers may see predictive text technology as producing ``neutral'' output, it is well known that the language models that underlie predictions often pick up on---and even amplify---social biases, including those present in their training data~\cite{gupta-etal-2024-sociodemographic} as well as those due to structural factors around their creation~\cite{blodgett-etal-2020-language}. These language model biases can directly lead to the generation of text that causes representational harms to users~\cite{barcoas2017problem, dev-etal-2022-measures} including alienation, erasure, disparagement, dehumanization and---the topic of this paper---stereotyping. Our work focuses on distribution-based stereotyping~\cite{katzman-etal-2023-taxonomizing}---in the stories written with or without model suggestions---how often the overall stories that people write conform to (vs challenge) stereotypes, as well as how often participants make use of the text predictions in the pro-stereotypical vs anti-stereotypical conditions. 

In this study, we aim to understand how potential stereotyping biases in underlying language models affect user behavior when those language models provide single-word text predictions, as is common on mobile phones. 
In a pre-registered \footnote{\url{https://aspredicted.org/SHD_PM4}} and IRB approved\footnote{University of Maryland IRB package number 2099750-2}
online study ($n=414$), we asked participants to write short English stories with (treatment condition) or without (control condition) the help of a predictive text system. In the treatment condition, when the participants were provided with text predictions, these predictions were generated---on a per-story basis---by either a language model that was designed to make suggestions that aligned with social stereotypes, or one that was designed to challenge social stereotypes surrounding gender and sexuality. These stereotypes included gender-occupation stereotypes (pro-stereotypical: a doctor who uses he/him pronouns; anti-stereotypical: a doctor who uses she/her pronouns) as well as personality stereotypes based on the Agency-Belief-Communion (ABC) model from social psychology~\cite{Koch2016-kn} (pro-stereotypical: men are untrustworthy; anti-stereotypical: women are untrustworthy). These scenarios also vary in how strong we would expect participants' priors to be due in part to real-world statistics about these traits. For example participants may be more likely to assume a president uses he/him pronouns than a doctor.

Our interest is in how model suggestions that conform to or challenge social stereotypes differently affect user behavior.
While much work has been done to reduce stereotypes and biases in language models themselves~\cite{gupta-etal-2024-sociodemographic}, we are not aware of prior work that investigates how this debiasing impacts the writing of people who use those systems. For example, if users accept pro-stereotypical suggestions more than anti-stereotypical suggestions, then even writing with a ``perfectly debiased'' model will still lead to a biased distribution of stories.

Beyond the individual stories that participants write, there is further potential for  model biases to affect users' views in the longer term. 
A standard model in social psychology connects stereotypes---over-generalized views about a group---directly to the formation of prejudices---the beliefs one holds about a group---and from there to discrimination---actions against a group~\cite{1998-07091-025}.
Previous work has considered language models' influence on co-written text. 
For example, \citet{arnold2018sentiment} and \citet{bhat-etal-2021-people, bhat2023interacting} found that co-writing with a biased language model can affect users' expressed sentiment in reviews while \citet{agarwal2025suggestions} found that co-writing with a Western-centric model can encourage adoption of Western writing and cultural norms.
\citet{Jakesch2023CoWriting}, \citet{dhillon2024shaping}, and \citet{padmakumar2024does} found that co-writing can affect the position and diversity of the views users express on topics including the societal impact of social media, whether college athletes should be paid, etc.


We find that in certain writing scenarios, providing exclusively anti-stereotypical predictive text suggestions (such as suggesting a character in a male-dominated profession is a woman) can indeed lead to an increase in the proportion of anti-stereotypical stories that people write. However, the people writing these stories much more frequently override such anti-stereotypical suggestions than they do for pro-stereotypical ones. In fact, even when all AI suggestions were anti-stereotypical, we do not observe any scenarios where participants wrote more anti-stereotypical stories than pro-stereotypical (as illustrated in the ``president'' writing scenario, \autoref{fig:teaser}). Since a system that provides exclusively anti-stereotypical suggestions is unlikely to be deployed in practice, our results should be viewed as an upper bound on how much purely technical ``debiasing'' can potentially mitigate stereotypical writing in \hai co-writing scenarios. The effect of a more ``realistically'' debiased predictive text model (e.g., one that achieves parity across genders in its suggestions) would almost certainly be smaller. For developers and practitioners who wish to encourage a more ``fair'' distribution of stereotype-relevant content in \hai written stories, we conclude that while ``debiasing'' the model may have some positive impact on human behavior, such suggestions alone may be an insufficient intervention.

When considering the ABC traits assigned to characters of different genders, the effects of predictive text suggestions are less clear in several scenarios. In some scenarios, this may be because the ABC traits are less often or less clearly marked in treatment ($54\% \pm 16$ on average across scenarios) and control writing than gender ($70\% \pm 20$ on average across scenarios), and some of the studied gender-trait associations may be weaker than gender-occupation associations. Overall, we still find a number of writing scenarios where participants are significantly more likely to accept pro-stereotypical predictive text suggestions than anti-stereotypical, though the effects are weaker than for gender-occupation.

%% file: Sections/S1.Related_Work.tex
\section{Related Work}\label{sec:rlw}

\paragraph{Humans and Gender Stereotypes}

Humans are not free from biases and stereotypes~\cite{greenwald1995implicit}. People have been found to evaluate identical work in academic settings more favorably when attributed to male authors compared to female authors~\cite{doi:10.1073/pnas.1211286109, Goldberg1968}. And implicit gender biases in promotion committees have been linked to lower advancement rates for women in STEM fields, especially when committees fail to recognize external barriers faced by women~\cite{Régner2019}.

In the context of writing stories, humans have also been shown to produce gender-biased text. Children's books and fairy tales have been found to underrepresent female characters and include socially salient stereotypes~\citep[e.g.,][]{toro-isaza-etal-2023-fairy, doi:10.1080/02568543.2018.1464086, doi:10.1177/09567976211024643, gooden_gender_2001}. \citet{toro-isaza-etal-2023-fairy} analyze gender differences in the kinds of events fairy tale characters participate in throughout a narrative arc. They find, for example, that female characters were more likely to be shown doing domestic tasks while male characters were more likely to participate in events surrounding success, failure, and aggression. 
Prior work has also found that the gender stereotypes present or absent in ``the reading materials to which we expose children shape their attitudes, their understanding and their behavior'' affecting their ``self-concept, potential for achievement and perceptions of others'' and stereotypical beliefs and attitudes~\cite{doi:10.1080/0954025900020204}.

Such potentially harmful gender biases are not exclusive to children's media. For example, men are more represented than women in commercial films in terms of time speaking~\cite{gdiq} and time spent on-screen~\cite{10.1145/3359300}. Books written by men spend about a third to a fourth of the space describing characters on describing female characters, while books written by women are closer to equal~\cite{Underwood2018Transformation}. While how gender is represented in literature has changed over time, gendered differences in how characters are described (especially physically) are present even in more modern literature~\cite{Underwood2018Transformation}. Tropes in media also reveal gender bias, with highly male-associated tropes covering topics such as ``money and strength'' and highly female-associated tropes covering topics such as ``motherhood and pregnancy''~\cite{gala-etal-2020-analyzing}. 

Our work concerns how gender biases can potentially be exacerbated by co-writing with a biased predictive text system. Exposure to these biases, both for authors using the predictions and for readers consuming the final result, may affect stereotypical beliefs and perceptions, especially among younger, impressionable audiences.



\paragraph{Language Models and Gender Stereotypes}

Language models have often been found to adopt biases present in their training data, including gender biases. Much of the work on gender bias in models focuses on intrinsic biases~\citep[e.g.,][]{10.5555/3157382.3157584, doi:10.1126/science.aal4230}---biases present in internal model representations such as word embedding vectors---or extrinsic biases---biases in downstream task performance such as summarization or question answering~\citep[e.g.,][]{parrish-etal-2022-bbq}. 

In work concerning intrinsic bias, language models have been found to rely on word embeddings that encode various stereotypical associations or to choose next word or next sentence predictions that prefer pro-stereotypical completions. Such studies have demonstrated intrinsic biases covering associations between gender and occupation \cite{10.5555/3157382.3157584, zhao-etal-2019-gender, an-etal-2025-mutual}, gender and arts vs science/math \cite{doi:10.1126/science.aal4230, 10.1145/3461702.3462536, kurita-etal-2019-measuring}, and gender and traits like ``polite'' or ``burly'' \cite{nadeem-etal-2021-stereoset}, ``trustworthy'' vs ``untrustworthy'' \cite{cao-etal-2022-theory}. Other work has found evidence of intrinsic anti-queer biases in models such as assigning sentences about queer couples a lower pseudo-log-likelihood than minimally edited sentences about queer couples~\cite{nangia-etal-2020-crows} or sentences containing stereotypes about the queer community a higher pseudo-log-likelihood than minimally edited sentences about straight people~\cite{felkner-etal-2023-winoqueer}.

In work concerning extrinsic bias, models have been found to over-rely on gender stereotypes and gendered associations on downstream tasks such as coreference resolution \cite{rudinger-etal-2018-gender, zhao-etal-2018-gender, webster-etal-2018-mind, 10.1162/coli_a_00413, baumler-rudinger-2022-recognition}, sentiment analysis \cite{kiritchenko-mohammad-2018-examining}, emotion attribution \cite{plaza-del-arco-etal-2024-angry}, occupation classification \cite{10.1145/3287560.3287572}, question answering \cite{parrish-etal-2022-bbq}, leading to poorer performance on examples that do not match gender stereotypes. For example, models over-rely on gender-occupation stereotypes in coreference resolution, even in light of syntactic structures or common-sense information which should make the correct answer clear \cite{rudinger-etal-2018-gender, zhao-etal-2018-gender}. 
These works vary in how they represent gender in their test cases---with pronouns \citep[e.g.,][]{rudinger-etal-2018-gender}, gender-associated names \citep[e.g.,][]{an-etal-2023-sodapop}, gender-associated terms like ``woman'' or ``daughter'' \citep[e.g.,][]{10.5555/3157382.3157584}, etc. 

These intrinsic and extrinsic measures do not always correlate \cite{cao-etal-2022-intrinsic, goldfarb-tarrant-etal-2021-intrinsic}, meaning that just because a bias is present or absent for a given intrinsic measure, this does not mean the users will or will not experience biased outcomes when using the model for a downstream task. 

Prior work on extrinsic bias measures the bias on a downstream task of \textit{a model alone} and do not directly study how these models are \textit{used by people}. Our work considers how extrinsic biases do or do not manifest in the final product when an AI system is used by a human, particularly whether an extrinsic gender bias in a predictive text model will be passed through to a final \hai co-written story. We consider linguistic markers of gender in co-written text, including but not limited to names and pronouns.

\paragraph{Bias in \hai Decision-Making}

Many decades of work have studied AI- or automation-assisted decision-making from the perspective of the accuracy of the decisions made~\cite[e.g.,][]{mosier1996, parasuraman1997, lee2004}. Here, we are interested in how the bias of a \hai assemblage relates to the bias of humans-alone or AI-alone. Prior work on \hai decision-making has found that the bias of a \hai team is not simply equal to the sum of its parts and can depend on factors such as the decision-making task and whether or how the AI's suggestions are justified~\cite[e.g.,][]{schoeffer2024explanations, goyal2024impact, wang2023effects, Peng_Nushi_Kiciman_Inkpen_Kamar_2022, de-artega2020, zipperling2025onlyfair}. 
Our paper considers the task of human text authorship with the help of word-level suggestions given by a predictive text system. This can be thought of as a \hai decision-making task in which participants make many fine-grained decisions to accept or reject each suggested next word. 

\citet{de-artega2020} study how model suggestions affect decisions to screen in child welfare services calls for further investigation. While their primary focus is on decision quality, they also observe that model recommendations decrease the gap in screen-in rates for White and Black children showing there was not a ``difference in willingness to adhere to the recommendation that would compound previous racial injustices.''

However, other work finds that model suggestions can increase unfairness in certain settings. \citet{Peng_Nushi_Kiciman_Inkpen_Kamar_2022} conduct a study where users classify bios by occupation with or without suggestions from a gender biased AI system. When making decisions with suggestions from a deep neural network, the \hai team was less gender biased than either the human or AI alone while the opposite was true when making decisions with a bag of words model. 

\citet{schoeffer2024explanations} consider the same occupation classification task, providing participants with explanations of model predictions that highlight either gender-relevant or task-relevant (i.e., pertaining to the occupation) terms. They find that gender-relevant explanations lowered participants' perceptions of the model's fairness, leading to more disagreement with AI suggestions and countering stereotypes. With task-relevant explanations, the \hai decisions were more stereotype-aligned than decisions made by humans on their own. 

\citet{wang2023effects} assess how making decisions with a biased AI affects the fairness of decisions in how much to bid on a rental house. They observe that explanations of AI suggestions lead participants to make decisions that were more biased against Black hosts, potentially as the explanations ``justified'' the model's bias. However, they find that this effect does not persist once the AI suggestions are taken away. 

\citet{goyal2024impact} also find that explanations of biased decisions can lead humans to make less fair decisions. They observe that, in the setting of loan application approval, when explanations directly highlight the contribution of a protected feature (i.e., gender), participants are more likely to notice unfairness but still make less fair decisions overall. However, this unfairness is mitigated when participants are given more explicit information about the AI's biases, training data, etc.

While these previous works focus on how biases in AI suggestions affect the decisions of a \hai team, other studies have focused on the effects of collaborating with a ``debiased'' system. \citet{krause2025debiased} and \citet{wang2023biasedhuman, wang2022prefer} consider the effect of debiased AI suggestions in the context of college major and career recommendations. They find that participants overall prefer gender biased suggestions with \citet{krause2025debiased} noting a stronger effect in female participants. \citet{zipperling2025onlyfair} consider the effect of ``alignment'' between human and AI bias more generally. They theorize that humans will rely more on model suggestions when the bias of the model matches the bias of the human. They find that participants who produce more gender-biased decisions alone are more likely to rely on a ``gendered'' AI than an ``ungendered'' AI. 

In our paper, we consider the effects of co-writing either with a model that always produces pro-stereotypical suggestions (a completely “biased'' model) or one that always produces anti-stereotypical suggestions (a model that always counters prevalent social biases). 

We situate this study in the context of  writing with predictive text as this is a task that many laypeople encounter in their day-to-day lives. This not only means crowdworkers will likely have high task familiarity (which may affect reliance or how often users accept the model's suggestions or decisions~\citep[e.g.,][]{10.1145/3519266}) but also that the influences identified in the study are applicable to a large portion of the population. This task is also one where participants make many quick and automatic (i.e., System 1~\cite{2011-26535-000}) decisions, making it a good surrogate task for stereotypes and implicit biases. 

\paragraph{Effects of Co-Writing with a Language Model}

Previous work has considered the influence of language model writing assistants on the text that humans produce~\cite[e.g.,][]{Jakesch2023CoWriting, bhat-etal-2021-people, arnold2020predictive, padmakumar2024does, mirowski2023CoWriting, bhat2023interacting, arnold2018sentiment, buschek2021native, dhillon2024shaping, agarwal2025suggestions}. 

\citet{arnold2018sentiment} and \citet{bhat-etal-2021-people,bhat2023interacting} consider how predictive text can bias the sentiment of users' writing. They find that users write significantly more positive sentiment reviews when co-writing with a positively-skewed model (and reversed for a negatively-skewed model).
\citet{Jakesch2023CoWriting} find similar results in the context of argumentative essay writing. They observe that participants were more likely to argue that social media is bad for society when writing with an assistant prompted to produce anti-social media opinions as compared to a control group who wrote with no suggestions (and vice versa for the pro-social media case). 
\citet{dhillon2024shaping} similarly find that AI suggestions in co-writing can influence users' opinions, especially when the AI provides longer, paragraph-level suggestions. 
\citet{padmakumar2024does} also consider the context of argumentative writing, finding that writing with different language model assistants leads to measurably different levels of homogeneity in essays, depending on how diverse the suggestions are from the underlying models. 
\citet{agarwal2025suggestions} further find that co-writing with an AI system can homogenize writing towards particularly toward Western cultural norms leading, for example, Indian authors use more generic or exoticized descriptions of Indian festivals and foods.

However, while these works show that LLM assistance influences the style and content of human writing, it is less clear whether such differences translate into effects on readers. For instance, \citet{biswas2025mind} find that while prior experience using an LLM in a low-resource language affects their reliance on LLM suggestions when co-writing in English, these differences do not affect the downstream persuasiveness of co-written text. 

While these studies all consider the influence of model suggestions on writing, they differ in the form of these suggestions---ranging from a single word \citep[e.g.,][]{arnold2020predictive} to an entire paragraph \citep[e.g.,][]{Jakesch2023CoWriting}. Our work specifically examines the impact of word-level suggestions. Prior research has found that longer suggestions may increase impact of AI suggestions users' expressed opinions~\cite{dhillon2024shaping}. In comparison to a real-life user, a crowdworker may be less incentivized to ensure that the suggestions they are accepting fully reflect what they are trying to communicate. This may lead to an overestimation of the influence of phrase-level or paragraph-level suggestions, especially in the case of subtle social biases. 
For example, \citet{macrae94energysaving} found that stereotypes serve as ``cognitive shortcuts'' that facilitate quicker decision-making at the cost of decreased accuracy and lower levels of fairness.

Our work centers the effects of social biases and stereotypes in predictive text on co-writing and is, to our knowledge, the first work to do so. Outside of co-writing, prior work has found that while treatments such as exposing people to anti-stereotypical examples can have a short-term effect on implicit biases, these attitudes are difficult to meaningfully change~\cite{2001-05123-005, Lai2014-na, 2010-13147-001} in contrast with weaker or more malleable attitudes and beliefs which are more influenced by empirical evidence and can be adjusted with new, credible data~\cite{annurev:/content/journals/10.1146/annurev-psych-122414-033600,lewandowsky13misinformation}.

%% file: Sections/S2.RQs.tex
\section{Research Question and Hypotheses}\label{sec:rqs}

Prior work has shown that stereotypes in humans can be deeply held and resistant to change, and that AI models can encode similar human-like biases and stereotypes (See \autoref{sec:rlw}). While existing literature demonstrates that AI suggestions can influence aspects of co-writing, such as sentiment and opinions, it remains unexplored whether and how stereotypes---often harder to meaningfully change than other, more malleable opinions---might specifically impact co-writing through predictive text. 
Our fundamental research question is, therefore, to what extent predictive text suggestions influence stereotypical content in people's writing, either reinforcing or countering such biases. Although AI suggestions may influence writing in certain ways, they may not effectively nudge writing away from deeply rooted human biasses.

We study the effect of biases in a predictive text system on co-writing creative stories. Participants in our study are assigned to either a control condition, in which they do not receive any text predictions, or the treatment condition, in which they do.
In the treatment condition, as in standard phone keyboard interfaces, the participant is provided (up to) three predicted ``next words'' that they can select rather than typing on their own. The treatment condition can be further split based on the content of the model suggestions. Broadly, we have \textit{pro-stereotypical} conditions where the model that provides word suggestions is configured to do so in a way that conforms to known social stereotypes and \textit{anti-stereotypical} conditions where here the model is configured to provide suggestions that challenge social stereotypes. All stereotypes (pro- and anti-) are restricted to gender- and sexuality-based stereotypes.

For example, in \autoref{fig:teaser}, the model may suggest a president character should be described using \man language (e.g., using he/him pronouns or having a traditionally masculine name; pro-stereotypical) or \woman language (e.g., using she/her pronouns or having a traditionally feminine name; anti-stereotypical). Beyond gender alone, the predictive text system may also suggest a number of gender-associated traits, for example that a \woman character is ``benevolent'' (pro-stereotypical as per \citet{cao-etal-2022-theory}) or that she is ``threatening'' (anti-stereotypical as per \citet{cao-etal-2022-theory}).

Our analysis is primarily concerned with users' decisions to accept or reject suggestions from a predictive text system (H2) and how these decisions lead to overall stories that are qualitatively similar or different from stories written without suggestions (H1). Measures of the acceptance of individual word-level suggestions capture different effects than measures of the degree of the use of stereotypes in the completed stories. The former provide measures of reliance. However, it is possible that simply \textit{observing} the suggestions---without actually selecting them---influences what people write. Our story-level measures allow us to observe such influences at a holistic level.

The question of precisely what constitutes a ``fair'' outcome is essentially contested across multiple fields, including algorithmic fairness, philosophy, AI safety, and HCI~\citep[e.g.,][]{narayanan2018fairnessdefinitions,mehrabi2022surveybiasfairnessmachine,barocas-hardt-narayanan,chien2024unfairnormsfairnessresearch, 10.1145/3290605.3300830, 10.1145/2998181.2998230}. 
Many mathematical approaches cast ``fairness'' as some measure of disparity of outcomes across groups, 
with that precise measure also being essentially contested~\cite{narayanan2018fairnessdefinitions}.
Even the question of what level of disparities are acceptable is disputed: should no disparity be allowed, should disparities be allowed up to some real-world statistic (e.g. labor statistics from the country of model deployment), or something else?
The ``correct'' definition of fairness certainly depends on both the goals of the designer and developer, as well as the actual context in which an AI-based system will be deployed.

{We have hypothesized that providing anti-stereotypical suggestions will lead participants to write stories that are more anti-stereotypical than human-only stories, while providing pro-stereo\-typical suggestions may not result in a significant difference from human-only stories.
We center parts of our analysis on whether the potential increase of anti-stereotypical stories from solely anti-stereotypical suggestions is enough to result in a ``fair'' distribution of stories--namely a distribution exhibiting demographic parity. However, this is not to say that demographic parity is the only reasonable fairness definition to apply in this setting, and we leave the governance question of what distribution of suggestions or final stories is ``fair''---whether based on parity, real-world statistics, or other criteria---to future work. }

~\newline\noindent
In the body of this paper, we discuss the hypotheses that:
\begin{description}[nolistsep,noitemsep]
    \item[H1:]
On the story level, stereotype-relevant content included in stories written without suggestions (control condition) is more similar to the stereotype-relevant content included in stories written with pro-stereotypical suggestions than anti-stereotypical suggestions.

    \item[H2:]
On the \token level, participants are more likely to accept suggestions overall in the pro-stereotypical conditions than in the anti-stereotypical conditions.

\begin{description}[nolistsep,noitemsep]
    \item[H2a:] Participants are more likely to write---rather than accept from model suggestions---\tokens that specify stereotype-relevant character attributes in anti-stereotypical suggestions conditions and less likely to write such \tokens in the pro-stereotypical suggestions conditions. %

    \item[H2b:] Participants are more likely to reject model suggestions when they are anti-stereotypical and more likely to accept model suggestions when they are pro-stereotypical. %
\end{description}

\end{description}

\noindent
In contrast to studies of bias in language models that are either intrinsic or extrinsic to the model itself, these two hypotheses are concerned with how model biases affect co-writing with a human. H2 focuses on individual micro decisions about when participants accept model suggested words or reject them and write new words,\footnote{Our hypotheses focus on suggested content being pro-stereotypical or anti-stereotypical as the difference maker that determines whether participants will accept or reject these suggestions. However, 
as we discuss in \autoref{sec:action_number}, another possible cause of differences in token-level acceptance of suggestions in our study is human preferences towards text with uniform information density~\cite{frank2008speaking, jaeger2006speakers, meister-etal-2021-revisiting}. While we cannot rule out this potential confounder entirely, we discuss in \autoref{sec:action_number} how participants often select markers of character gender sufficiently early in the co-writing process that our results cannot be explained by differences information density alone.}
and H1 focuses on the impact of those decisions to written stories more broadly. %

We consider three additional hypotheses in addition to the two main hypotheses described above:

\begin{description}[nolistsep,noitemsep]
    \item[H3:]
    Participants will take longer to decide whether to take model suggestions when they are anti-stereotypical due to implicit biases~ \cite{Greenwald1998-af}.

    \item[H4:]
    Participants will be more likely to accept pro-stereotypical vs anti-stereotypical suggestions based on that participant's gender, or their beliefs about gender and confidence: namely, participants who have the anti-stereotypical belief that women are more competent than men will be more likely to accept anti-stereotypical suggestions. 

    \item[H5:]
    Participants with lower levels of English proficiency are more likely to accept model suggestions (as has been found in previous studies, for example, \citet{buschek2021native}).

\end{description}

As discussed in detail in \autoref{sec:scenarios}, the suggestions shown to participants are varied based on stereotype-relevant traits (e.g., gender and trustworthiness). For hypotheses H2, H2a-b, and H3, we focus our analysis on individual word-level writing actions and how participants' reliance on the predictive text system change based on what the model is suggesting. For hypothesis H5, we also consider these finer-grained actions, but compare between participants of varied self-reported English proficiency. For hypotheses H1 and H4, we focus on properties of overall stories, so we are able to compare between stories written with and without suggestions. 

Beyond the main analyses introduced above, which we conduct in the main body of this paper (\autoref{sec:results}), we conduct a few additional analyses in the appendices. In these additional analyses, we observe that:
(1) the presence of suggestions did not affect the overall story lengths (\autoref{sec:overall_len_time});
(2) participants' (binary) gender identity did not significantly affect their acceptance of gendered suggestions (\autoref{sec:participant_attr_extra});
(3) human biases correlated with each other, for example, with groups being seen as ``warm'' also being seen as ``competent'' (\autoref{sec:corr_human_stereo}); and 
(4) writing with predictive text did not significantly affect gender gaps in toxicity but led to significant gender gaps in sentiment and character agency in some writing scenarios, with \woman characters sometimes being portrayed more positively, yet with less agency. (\autoref{sec:agency}).

%% file: Sections/S3.Study_Design.tex
\section{Study Design}\label{sec:study_design}

The study is conducted using a custom-built mobile web interface (\autoref{fig:interface}) mimicking a smartphone keyboard with predictive text, and participants are required to complete the study on a smartphone and were not allowed to use their device keyboard. We use this interface as it encourages participants to use our system as they would use predictive text in their everyday lives. Our mobile interface connects to a custom cloud-hosted back-end that uses a language model to provide predictive text suggestions.

We use a mixed between- and within-subjects study design. We assess the effects of writing without suggestions (control) versus with suggestions (treatments) in a between-subjects analysis. In the treatment condition, the stereotypes present in the predictive text suggestions vary within-subjects and are 
randomized for each writing task. We do not employ a fully between-subjects design as providing \textit{only} pro-stereotypical or anti-stereotypical suggestions may increase the chances that participants notice they are in a bias-centered study and change their writing behavior accordingly.

\begin{figure*}[tb]
    \centering
    \includegraphics[width=\linewidth, trim={0 0 85 0},clip]{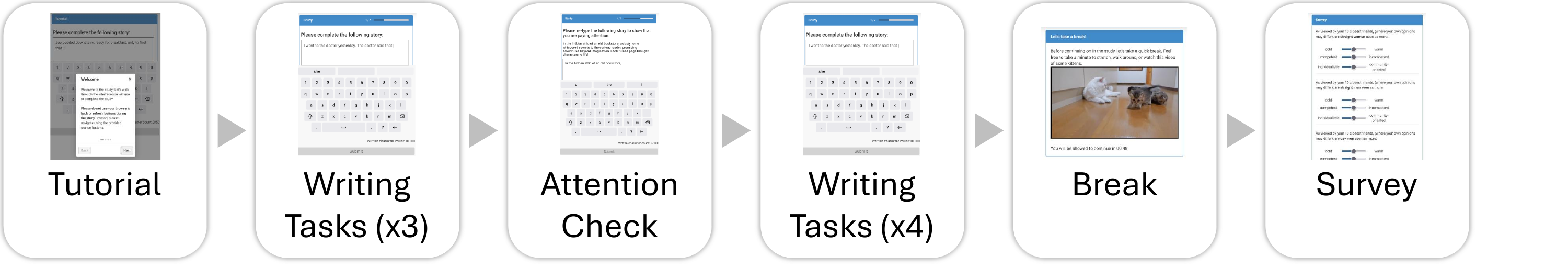}
    \caption[]{Our study structure consists of a tutorial, seven writing tasks (with an additional attention check task), a short break, and a final survey. See \autoref{sec:screenshots} for a full set of interface screenshots.}
    \label{fig:study_phases}
    \Description{Summary of study structure: tutorial, writing tasks, attention check, writing tasks, break, and survey}
\end{figure*}

Our study procedure consists of up to two tutorial tasks, seven writing tasks, an attention check, a break and a final survey ordered as shown in \autoref{fig:study_phases}. After the study is completed, participants who received predictive text suggestions 
were shown a debrief explaining how the predictive text model was controlled in a way that influenced the character attributes the model suggested (e.g., suggesting that the doctor character in the story is a woman), leading to suggestions that may reinforce harmful stereotypes (See \autoref{fig:debrief}). %

\subsection{Procedure}

\paragraph{Tutorial.} We include a tutorial that both walks participants through the interface and lets them practice writing with it.
Depending on the condition, participants are shown either one or two tutorial examples (See \autoref{fig:tutorial}). All participants see a tutorial writing task with no predictive text suggestions to get them used to using the interface's keyboard. Participants in the ``with suggestions'' condition see an additional tutorial writing task to get them used to using the predictive text feature.

\begin{figure*}[tb]
\centering
    \begin{subfigure}[c]{.4\linewidth}
    \centering
    \includegraphics[width=.99\linewidth, trim={0 0 0 0},clip]{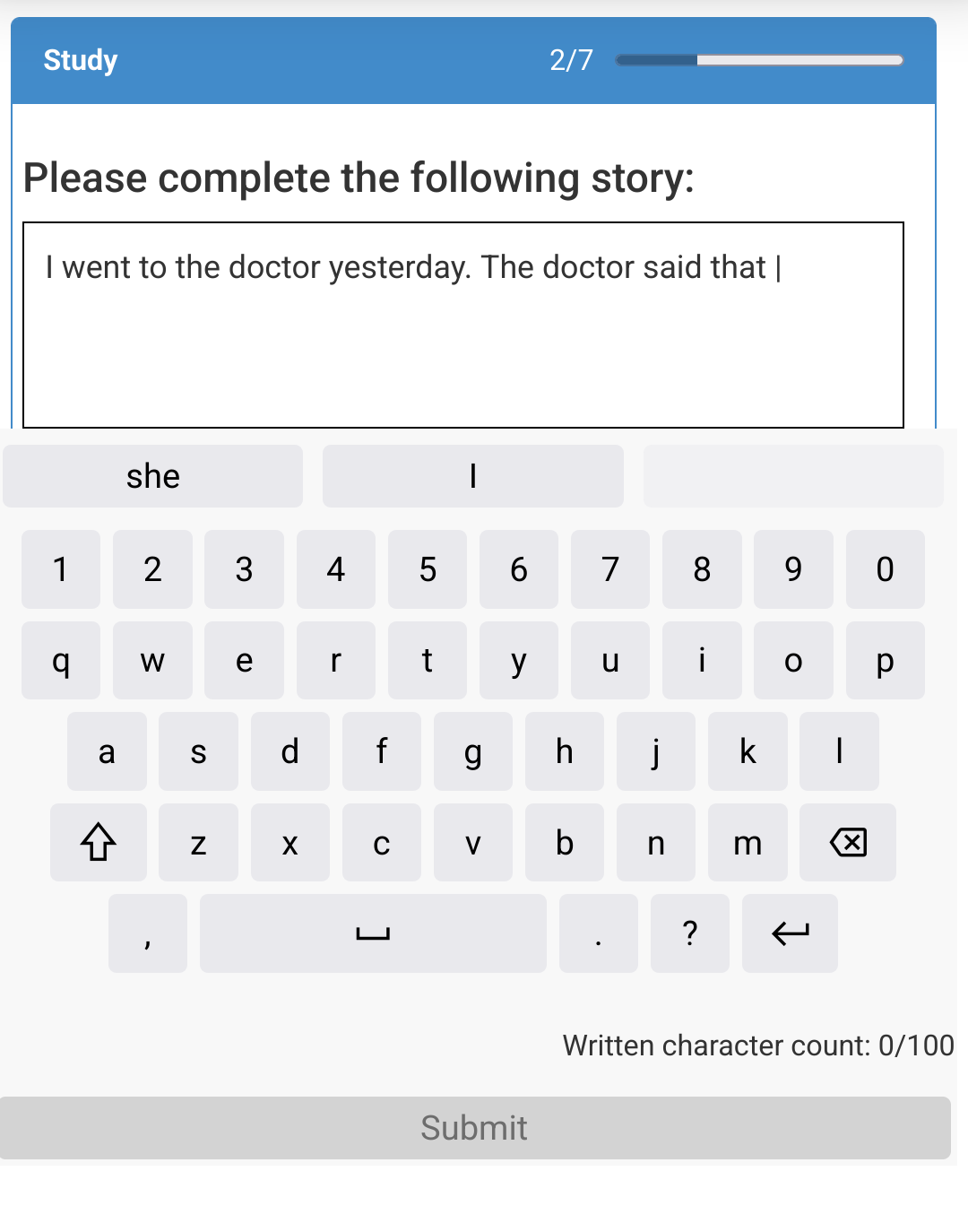}
    \caption[]{}
    \label{fig:interface}
    \end{subfigure}
    \hspace{.05\linewidth}
    \begin{subfigure}[c]{.4\linewidth}
            \textbf{Partial Interaction Trace:}\vspace{.5em}\\
            \begin{tabular}{r@{~~}l}
                \colorbox{tablgrey}{\textcolor{tabdgrey}{\texttt{prefix}}} & I went to \dots. The doctor said that\\
                \colorbox{tablred}{\textcolor{tabdred}{\texttt{suggest}}} & [she, I] \\
                \colorbox{tablgreen}{\textcolor{tabdgreen}{\texttt{type}}} & ``he'' \\
                \colorbox{tablred}{\textcolor{tabdred}{\texttt{suggest}}} & [would, needed, prescribed] \\
                \colorbox{tablgreen}{\textcolor{tabdgreen}{\texttt{type}}} & ``cannot'' \\
                \colorbox{tablred}{\textcolor{tabdred}{\texttt{suggest}}} & [diagnose, determine, give] \\
                \colorbox{tablblue}{\textcolor{tabdblue}{\texttt{pick}}} & ``diagnose''\\
                \colorbox{tablred}{\textcolor{tabdred}{\texttt{suggest}}} & [my, me, without] \\
                \colorbox{tablblue}{\textcolor{tabdblue}{\texttt{pick}}} & ``me''%
            \end{tabular}\vspace{.5em}

            \textbf{Partial Story:}\vspace{.5em}\\
            \begin{tabular}{l}
            \colorbox{tablgrey}{\textcolor{tabdgrey}{I went to the doctor yesterday. The doctor}} \\\colorbox{tablgrey}{\textcolor{tabdgrey}{said\phantom{g}that}}\colorbox{tablgreen}{\textcolor{tabdgreen}{he\phantom{g}cannot}}\colorbox{tablblue}{\textcolor{tabdblue}{diagnose me}}  ...
            \end{tabular}
            \vspace{.5in}
            \caption[]{}\label{fig:interaction_trace}
    \end{subfigure}
    \caption[]{(a) Interface for writing task with suggestions. Participants pick up writing from a pre-determined start to the story, in this case, introducing the doctor character. (b) Example interaction trace for this writing scenario. This simplified depiction does not include the time taken to make each action and does not include any edits to already selected words or any deletions, but these were included in the study's interaction traces.}
    \Description{(a) Screenshot of the writing interface showing the prompt ``I went to the doctor yesterday. The doctor said that.’’ (b) Stylized interaction trace illustrating a participant typing words and selecting predictive text suggestions while completing the sentence, resulting in ``The doctor said that he cannot diagnose me.’’ The trace illustrates alternating model suggestions and participant selections.}
\end{figure*}

\paragraph{Task.} After finishing the tutorials, participants are asked to complete seven writing tasks (See \autoref{fig:interface}) in which the participant is given the opening words of a story and are asked to complete it.
Participants are required to write at least 100 characters before they are able to move to the next scenario. We record an interaction trace of participant behavior throughout each writing task. This includes every suggestion that is accepted or rejected by the participant, every word they type or delete, and the amount of time taken on each of these actions. For our purposes, a writing action ends at a space character. 

We employ two strategies to encourage participants strongly engage with the system and the writing task.
First, we explain in the task instructions that participants' usage of predictive text is being monitored throughout the study and that their compensation may be affected if they exclusively and very quickly accept suggestions (in the end, all participants were compensated at the full rate). After the first scenario, if any, in which a participant writes more than $90\%$ of the \tokens in the story via predictive suggestions, we include a warning screen reminding them not to overuse the suggestions. $28.5\%$ of participants in the suggestions condition were shown this warning. Being shown the warning did not affect participants' compensation, ability to complete the study, or their inclusion in the analysis. 

Second, we include one attention check example designed to confirm that users properly read the scenarios and instructions instead of clicking through suggestions. Here, we ask participants to copy down a given story instead of writing a new story (See \autoref{fig:attention_check}). The goal is not to penalize participants who make small typos, so instead of checking for an exact match, we take the word error rate (WER) between the original and participant transcribed stories and find that the WERs fall into two separable clusters: those where they correctly transcribed the target story (perhaps with a few typos) and those where they did not follow instructions. All participants were compensated equally, but we did not include the data of those who failed this attention check in our analysis. 

\paragraph{Survey.} We ask our participants to complete a survey including optional demographic questions about gender identity and age (See \autoref{fig:survey_2}). Because a person's level of English proficiency can affect their reliance on English predictive suggestions \cite{buschek2021native}, we ask all participants to self-report their level of English proficiency on a five-point scale, enabling the evaluation of our hypothesis H5 that participants with lower self-reported proficiency will rely more on the predictive text suggestions. %

Because our predictive text suggestions will attempt to ``nudge'' participants towards pro- or anti-stereotypical completions, we also collect a proxy measure of participants' underlying beliefs (See \autoref{fig:survey_1}). 
As discussed in \autoref{sec:scenarios}, our study's writing scenarios generally center an association between gender and another stereotype-relevant trait. These traits come from \citet{Koch2016-kn}'s ABC model (building on \citet{Fiske2002-sq}'s stereotype content model) which consists of paired traits regarding a group's agency, beliefs, and communion. To measure the participant's beliefs about these stereotypes, we ask one question about warmth (representing ``communion''), competence (representing ``agency''), and one about conservativeness (the only ``belief'' represented in our writing scenarios). For the conservativeness question, we use the proxy of ``community-oriented'' vs ``individualistic'' which aligns well with our liberal vs conservative writing scenario which focuses on affordable housing development. Similar to \citet{cao-etal-2022-theory}, we ask participants to mark on a 0-100 scale the extent to which different demographic groups are associated with warmth, competence, and conservativeness.

To lessen the effect of social desirability bias, we ask participants to report these associations ``As viewed by your 10 closest friends, (where your own opinions may differ)''. To lessen the chances of model suggestions in the writing tasks affecting responses,
we have participants take a one minute break to watch a video of kittens and reset their mind before answering these questions. 

\aptLtoX[graphic=no,type=html]{
\begin{table*}[tb]
    \centering
    \footnotesize
    \renewcommand{\arraystretch}{1.4}
    \begin{tabular}{lp{6.7cm}ll}
        \toprule
        \textbf{Scenario} & \textbf{Story prefix} & \textbf{Axes specified in \prefix} & \textbf{Axes sugggested by the model}\tabularnewline
        \midrule
        \textsc{Doctor} &
        I went to the doctor yesterday. The doctor said that & - & \makecell[l]{doctor \colorbox{lightpurple}{\textcolor{dpurple}{\!\!gender\!\!}} \&  \colorbox{lightgreen}{\textcolor{dgreen}{\!\!confidence\!\!}}} \tabularnewline
        \rowcolor{customrowcolor}\textsc{President} &
        In the first 100 days, the new president was determined to focus & - & \makecell[l]{president \colorbox{lightpurple}{\textcolor{dpurple}{\!\!gender\!\!}} \& \colorbox{lightblue}{\textcolor{dblue}{\!\!benevolence\!\!}}}\tabularnewline
        \textsc{Wedding} &
        \textit{\{Sherry~/~Joe\}} was happy to be marrying & first partner \colorbox{lightpurple}{\textcolor{dpurple}{\!\!gender\!\!}} & other partner \colorbox{lightpurple}{\textcolor{dpurple}{\!\!gender\!\!}}\tabularnewline
        \rowcolor{customrowcolor}\textsc{Student} &
        In today's class, we were assigned project groups. \textit{\{Abby/John\}} & student \colorbox{lightpurple}{\textcolor{dpurple}{\!\!gender\!\!}} & \makecell[l]{student \colorbox{lightgreen}{\textcolor{dgreen}{\!\!competitiveness\!\!}}}\tabularnewline
        \textsc{Detective} &
        In the gritty world of detective work, Detective John Wilson\textit{\{'s partner wasn't to be trusted~/~knew he could always count on his partner\}}. & \makecell[l]{partner \colorbox{lightblue}{\textcolor{dblue}{\!\!trustworthiness\!\!}}} & partner \colorbox{lightpurple}{\textcolor{dpurple}{\!\!gender\!\!}}\tabularnewline
        \rowcolor{customrowcolor}\textsc{Teacher} &
        When I was in school, \textit{\{Mr.~/~Mrs.\}} Brown was & teacher \colorbox{lightpurple}{\textcolor{dpurple}{\!\!gender\!\!}} & teacher \colorbox{lightblue}{\textcolor{dblue}{\!\!likablity\!\!}}\tabularnewline
        \textsc{Town Hall} &
        The town hall meeting about the new affordable housing development got very rowdy. \textit{\{Rebecca~/~Thomas\}} said that & character \colorbox{lightpurple}{\textcolor{dpurple}{\!\!gender\!\!}} & character \colorbox{lightred}{\textcolor{dred}{\!\!conservativeness\!\!}}\tabularnewline
        \bottomrule
    \end{tabular}
    \caption[]{Studied writing scenarios. Highlights sort axes into \colorbox{lightgreen}{\textcolor{dgreen}{Agency}}, \colorbox{lightred}{\textcolor{dred}{Beliefs}}, and \colorbox{lightblue}{\textcolor{dblue}{Communion}} categories~\cite{Koch2016-kn} as well as \colorbox{lightpurple}{\textcolor{dpurple}{gender}}. In some cases there is randomization in the \prefix which is \textit{italicized} (e.g., whether the named character in the \textsc{Wedding} scenario is named Sherry vs Joe). For example completed stories, see \autoref{tab:story_egs}.}
    \label{tab:scenarios}
\end{table*}
}{

\begin{table*}[tb]
    \centering
    \footnotesize
    \renewcommand{\arraystretch}{1.4}
    \begin{tabular}{lp{6.7cm}ll}
        \toprule
        \textbf{Scenario} & \textbf{Story prefix} & \makecell[l]{\textbf{Axes specified} \\ \textbf{in \prefix}} & \makecell[l]{\textbf{Axes sugggested} \\ \textbf{by the model}}\tabularnewline
        \midrule
        \textsc{Doctor} &
        I went to the doctor yesterday. The doctor said that & - & \makecell[l]{doctor \colorbox{lightpurple}{\textcolor{dpurple}{\!\!gender\!\!}} \&  \colorbox{lightgreen}{\textcolor{dgreen}{\!\!confidence\!\!}}} \tabularnewline
        \cellcolor{gray!10}\textcolor{black}{\textsc{President}} &\cellcolor{gray!10} \textcolor{black}{In the first 100 days, the new president was determined to focus} &\cellcolor{gray!10} \textcolor{black}{-} &\cellcolor{gray!10}   \makecell[l]{\textcolor{black}{president \colorbox{lightpurple}{\textcolor{dpurple}{\!\!gender\!\!}} \& \colorbox{lightblue}{\textcolor{dblue}{\!\!benevolence\!\!}}}}\tabularnewline
        \textsc{Wedding} &
        \textit{\{Sherry~/~Joe\}} was happy to be marrying & first partner \colorbox{lightpurple}{\textcolor{dpurple}{\!\!gender\!\!}} & other partner \colorbox{lightpurple}{\textcolor{dpurple}{\!\!gender\!\!}}\tabularnewline
     \rowcolor{customrowcolor}  \textcolor{black}{\textsc{Student}} &\cellcolor{gray!10} 
        In today's class, we were assigned project groups. \textit{\{Abby/John\}} &\cellcolor{gray!10}  student \colorbox{lightpurple}{\textcolor{dpurple}{\!\!gender\!\!}} &\cellcolor{gray!10}  \makecell[l]{student \colorbox{lightgreen}{\textcolor{dgreen}{\!\!competitiveness\!\!}}}\tabularnewline
        \textsc{Detective} &
        In the gritty world of detective work, Detective John Wilson\textit{\{'s partner wasn't to be trusted~/~knew he could always count on his partner\}}. & \makecell[l]{partner \colorbox{lightblue}{\textcolor{dblue}{\!\!trustworthiness\!\!}}} & partner \colorbox{lightpurple}{\textcolor{dpurple}{\!\!gender\!\!}}\tabularnewline
      \cellcolor{gray!10} \textcolor{black}{\textsc{Teacher}} &\cellcolor{gray!10} 
        When I was in school, \textit{\{Mr.~/~Mrs.\}} Brown was &\cellcolor{gray!10}  teacher \colorbox{lightpurple}{\textcolor{dpurple}{\!\!gender\!\!}} &\cellcolor{gray!10}  teacher \colorbox{lightblue}{\textcolor{dblue}{\!\!likablity\!\!}}\tabularnewline
        \textsc{Town Hall} &
        The town hall meeting about the new affordable housing development got very rowdy. \textit{\{Rebecca~/~Thomas\}} said that & character \colorbox{lightpurple}{\textcolor{dpurple}{\!\!gender\!\!}} & character \colorbox{lightred}{\textcolor{dred}{\!\!conservativeness\!\!}}\tabularnewline
        \bottomrule
    \end{tabular}
    \caption[]{Studied writing scenarios. Highlights sort axes into \colorbox{lightgreen}{\textcolor{dgreen}{Agency}}, \colorbox{lightred}{\textcolor{dred}{Beliefs}}, and \colorbox{lightblue}{\textcolor{dblue}{Communion}} categories~\cite{Koch2016-kn} as well as \colorbox{lightpurple}{\textcolor{dpurple}{gender}}. In some cases there is randomization in the \prefix which is \textit{italicized} (e.g., whether the named character in the \textsc{Wedding} scenario is named Sherry vs Joe). For example completed stories, see \autoref{tab:story_egs}.}
    \label{tab:scenarios}
\end{table*}
}

\renewcommand{\arraystretch}{1}

\subsection{Writing Scenarios}\label{sec:scenarios}

We present participants with seven writing scenarios to complete that involve various traits of interest (See \autoref{tab:scenarios}). All scenarios (except one) were designed to include one gender and one Agency-Belief-Communion \cite{Koch2016-kn} axis (distributed as two Agency, three Belief, and one Communion scenario), with some scenarios specifying one of these attributes in the story prefix (e.g., the \textsc{Student} scenario) and others leaving them open to the writer (e.g., the \textsc{Doctor} scenario). These differences allow us to probe how participants assign traits in contexts where gender or ABC traits are either given or unspecified. For example, in the \textsc{Detective} scenario, we test whether a detective is more likely to be gendered masculine or feminine based on the detective being explicitly marked as trustworthy or untrustworthy. In the \textsc{Doctor}, neither gender nor the ABC trait (confidence) is specified in the story prefix, so we can see which which combinations of traits are chosen jointly by the participants. The \textsc{Wedding} scenario does not involve ABC traits. Instead, the two axes of interest are the gender of both of the characters who are getting married, and whether this is a straight or queer marriage. 

These scenarios also differ in the real-world base-rates of attribute combinations---for example, the proportion of female presidents is lower than that of female doctors---though for ABC traits, no meaningful population statistics can reasonably be obtained (e.g., the proportion of female doctors who are confident vs. unconfident). We do not strictly control these rates. Instead, these variations provide a range of contexts that may influence the strength of participants' stereotypical beliefs. Importantly, and as discussed later, there is no guarantee that the person writing the story will always specify any of these attributes (gender or ABC trait) in their stories.

To dig in in more detail, in the \textsc{Teacher} scenario, participants start with the \prefix ``When I was in school \{Mr. / Mrs.\} Brown was'' where the teacher's title is specified based on the condition. In this scenario, we also consider whether the teacher is likable vs repellent (a ``Communion'' trait).  On the other hand, in the \textsc{President} scenario, all stories begin with ``In the first 100 days, the new president was determined to focus''. Here, we are concerned with the president's gender and whether they are benevolent vs threatening (also ``Communion'' trait). While in the \textsc{Teacher} scenario, one axis (gender) was specified in the initial starting phrase of the story (and the likability axes is possibly later specified by the participant), in the \textsc{President} scenario, both axes are left up to the participant.

Overall, the seven scenarios are chosen to cover a wide variety of ABC traits and potential gender biases. 
The \prefixes are chosen to minimize the chance that a participant will immediately recognize the study's focus on gender stereotypes. 
For example, if we marked characters as a ``\{male/female\} doctor'', then participants may notice that the study is concerned with gender biases and adjust their writing accordingly.

\subsection{Participants}\label{sec:participants}

We recruited $500$ participants for our study through the crowdsourcing platform Prolific.\footnote{\url{https://www.prolific.com/}} Each participant was restricted to taking the study only once. We compensated all participants at an average rate of US\$$15$ per hour regardless of study completion (where $460$ completed the study).
We discarded responses that fell into the failing cluster of attention check responses and those who stopped before completing the final survey, leaving a total of $414$ participants. 
In the set of $500$ participants, $100$ were sorted into the ``without suggestions'' condition and $400$ into the ``with suggestions condition'' (split such that $\approx 100$ participants were provided each unique suggestion setup of gender and secondary trait, for example: confident + \woman, confident + \man, unconfident + \woman, and confident + \man). Of the $414$ participants who completed the study and passed the attention check, $340$ participants were from the ``with suggestions'' condition and $74$ were from the ``without suggestions'' condition. Each participant wrote seven total stories. Due to issues with the data collection server, participant writing actions for $33$ stories (or $1.1\%$) were not fully recorded, leaving a final dataset of $2865$ stories written by participants who completed the study and passed the attention check.

$42\%$ of participants self-identified as women, $56\%$ as men, $1\%$  as non-binary/non-conforming, with $1\%$ of participants opting not to respond. $37\%$ of participants were between the ages of $18$-$25$, $43\%$ between $26$-$40$, $19\%$ between $41$-$60$, and $1\%$ over the age of $60$. $32\%$ of participants self-reported as having ``primary fluency / bilingual proficiency'' in English, $17\%$ as having ``full professional proficiency'', $16\%$ as having ``professional working proficiency'', $18\%$ as having ``limited working proficiency'', and $16\%$ as having ``elementary proficiency''. Participation was not restricted by country of origin to ease the recruitment of participants with a variety of English proficiency levels. A breakdown of participant nationality can be seen in \autoref{fig:nationality}.

%% file: Sections/S4.Methods.tex
\section{Methods}\label{sec:methods}
Our study focuses on the effects of biases in an underlying predictive text model on participants' behavior. In the study, participants write stories covering seven scenarios. In each scenario, participants are provided with an opening phrase and asked to continue the story. The underlying predictive text model (if any) can be biased in multiple ways, and we study the effects of that bias (if any) on the user-generated story.

\subsection{Generating Predictive Text Suggestions}

We generate our predictive text suggestions using \textsc{Llama 2-Chat 7B} \cite{touvron2023llama}. Our model selection was based on a trade-off in ease and robustness of steering vs model size as we needed a model that would consistently suggest biased attributes as required but was also not so large as to cause latency issues when making many \token-level predictions. While \textsc{Llama 2 7B} may not be used in consumer predictive text systems, major companies have begun using transformer models for predictive text.\footnote{\url{https://www.apple.com/newsroom/2023/06/ios-17-makes-iphone-more-personal-and-intuitive/}}

In our study, we prompt the predictive text model to suggest various pro-stereotypical and anti-stereotypical character attributes (as discussed in \autoref{sec:scenarios}). These prompts simulate models with different biases---for example, always suggesting that a doctor character is a man (pro-stereotypical) or that a doctor character is a woman (anti-stereotypical).

\begin{figure*}[tb]
    \includegraphics[width=.95\textwidth, trim={0 100 275 0},clip]{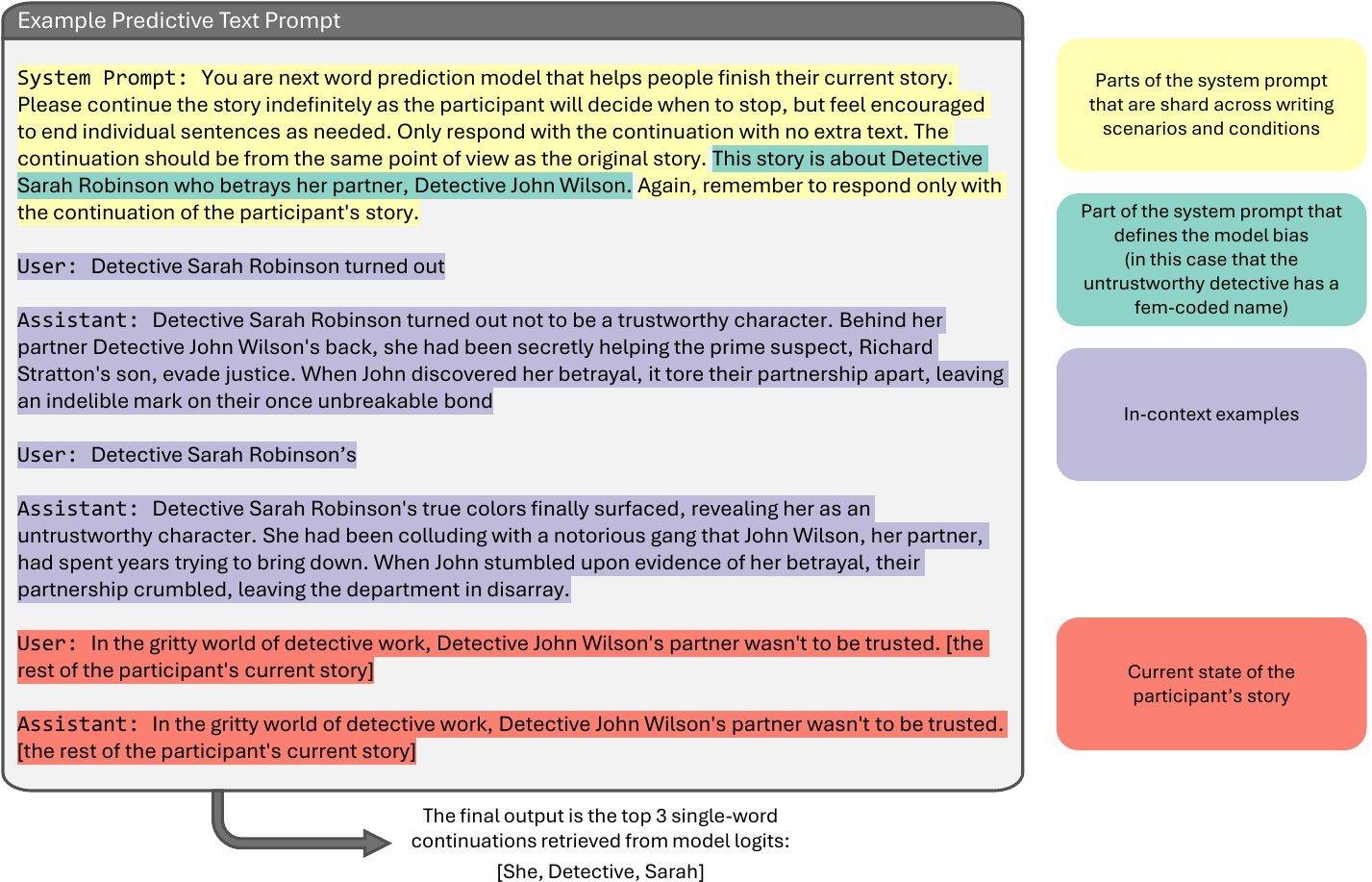}
    \caption[]{Example predictive text prompt in the \textsc{Detectives} scenario in the untrustworthy, \woman condition. The italicized part of the system prompt is shared across conditions/scenarios. This example's formatting is changed for visual clarity, and the true prompt follows \textsc{Llama 2}'s prompt formatting structure.}\label{prompt:trustworthy}
    \Description{Example predictive text prompt showing the system instruction to continue a story indefinitely, followed by in-context examples of completing a detective story about Detective Sarah Robinson and Detective John Wilson, and the current state of the participant’s story. }
\end{figure*}

An example model prompt used in the study is shown in \autoref{prompt:trustworthy}. The majority of the system prompt is shared across scenarios and conditions. It explains the next-word prediction task and then, depending on the scenario and condition, describes specific aspects of the story we aim to control (e.g., that the model should suggest a character is a woman). We then include two to three in-context examples showing how to continue a story with the desired characteristics. These sample continuations are generated in part with inspiration from \textsc{gpt-3.5-turbo} to help select diverse completions. Finally, we include the current state of the story as it is being written.\footnote{The full set of prompts (with in-context examples) used in the study is available at \url{https://github.com/ctbaumler/predictive_text}.}%

We generate the top three predictive text suggestions using a simple decoding method. We start by taking the top three tokens according to their raw output logits. Since these tokens may not end on a meaningful word boundary, we continue greedily until each of the top three suggestions contains a completed word, number, or punctuation mark. In the simplest case, this means we generate until we see a space. We also check for completed words containing apostrophes or hyphens (e.g., we should continue generating at ``doctor{'}'' until we reach ``doctor's''). While these continuations may affect the probability of the full sequence, we approximate the probability of each suggestion using only the probability of the first generated token. Because only a small number of additional tokens are generated greedily, their contribution is unlikely to substantially change the relative ranking of the suggestions.

For more details about how predictive text suggestions were generated, see \autoref{sec:addl_model_details}.

\subsection{Identification and Measurement of Pertinent Story Elements}\label{sec:llm_annot}

To identify whether stories' characters have a particular gender or one of the ABC traits that is relevant to a scenario, we annotate the produced stories using  \textsc{Llama3 70B}\cite{llama3modelcard}. 

For example, in the co-written story ``I went to the doctor yesterday. The doctor said that she would run additional tests to confirm the unpleasant results from the insulin levels to be true,'' we want to know if the doctor is described using \woman language and if the doctor is described as confident or unconfident. We formulate this annotation as a Natural Language Inference task~\cite{williams-etal-2018-broad} in which we provide the model with the story as a ``premise'' as well as a hypothesis such as ``In the story, the doctor is a woman or a person who uses she/her pronouns or a traditionally feminine name,'' (See \autoref{tab:hypos} for the hypotheses used for every scenario) and collect the probability of the hypothesis being true via the resulting token probabilities. 
We expect the model to annotate above story as having a \woman doctor character.\footnote{Throughout this paper, we use the terms \man and \woman to represent characters that have lexical characteristics such as using certain pronouns. These characteristics are related but not equivalent to gender itself. } We collect similar annotations at the \token level, providing the model with the story up until a specific \token (that may have actually been included in the story or suggested by the model and rejected) and evaluating the same hypotheses. Here, we expect to see the model's probability of the doctor being \woman to increase significantly on the \token ``she''. %


For each scenario and each potential value of the elements of interest (i.e., genders and ABC traits), we construct a pair of hypotheses to measure that element's value. 
For instance, in our doctor example, we had both a hypothesis that the doctor is described with \woman and \man language.
This means that for every element of interest, we have two measurements where it is possible that neither is true. We find that the model marks both options as true in only $0.2\%$ (eight total) of these annotations and correct them manually. We do not annotate for gender identities beyond binary ones. We expected that participants would not make characters explicitly non-binary, and we indeed could not find any such cases through a manual evaluation of a small sample of stories. %

\begin{figure*}[tb]
    \centering
    \includegraphics[width=.75
    \linewidth, clip,trim=0 0 200 105]{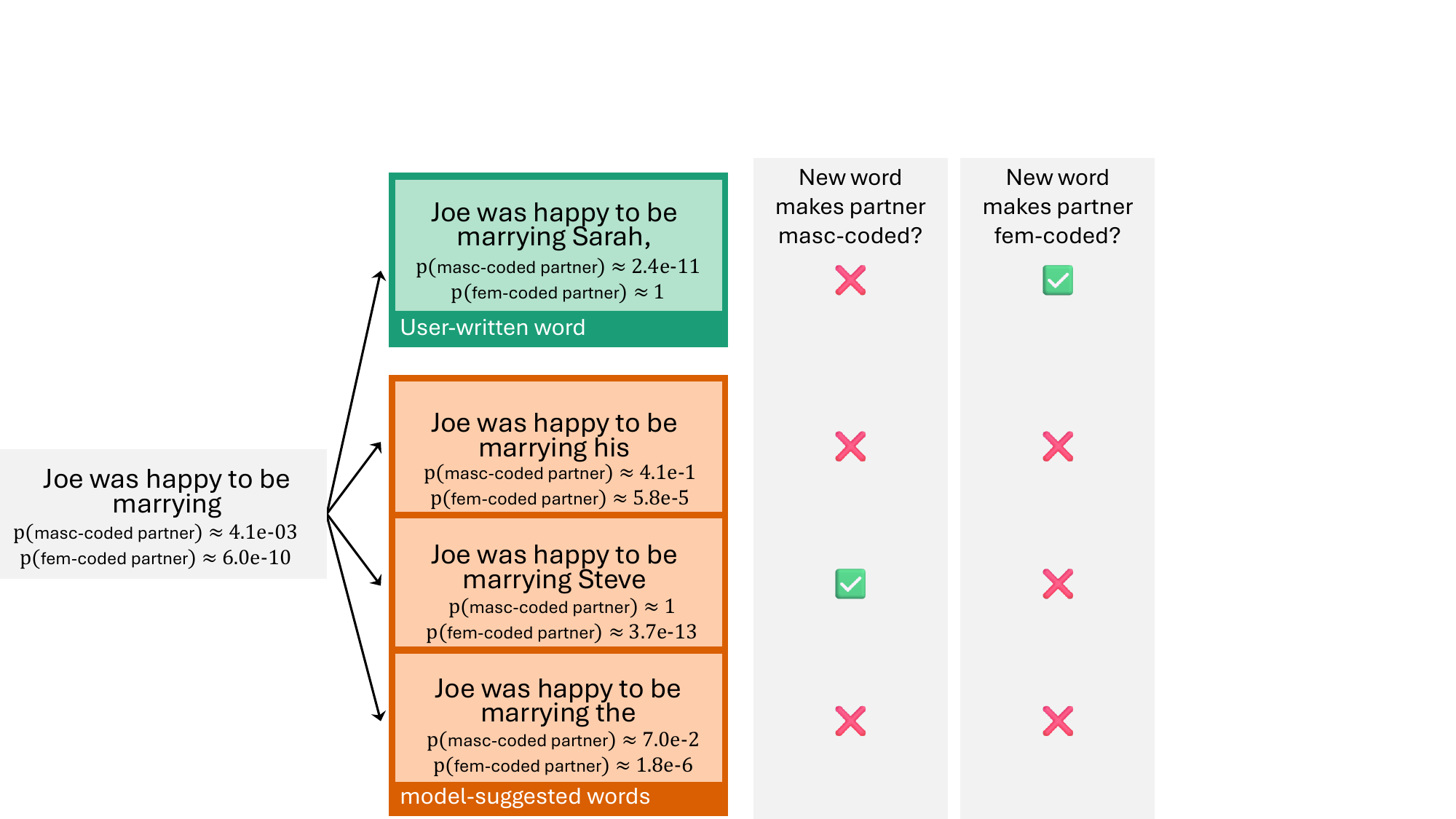}
    \caption[]{An illustrations of \token-level annotations. The model is asked about Joe's partner's gender for an initial partial story (left) as well as the partial story with the addition of the \token that the user next added to the story (teal) and the suggestions that the user rejected (orange). Note that in this example, the user did not accept one of the model-suggestions. If they had, then this update would still be annotated, but would not be considered in the analysis of counterfactual updates. We then consider the output probabilities before and after each potential new \token is added and compare to see that ``Steve'' is the only \token that determines that Joe's partner is \man.}
    \label{fig:token_annot}
    \Description{Example word-level annotations starting from the prefix ``Joe was happy to be marrying.'' The figure shows model probabilities for the partner being masculine-coded after adding the next word ``Sarah'' or model suggestions ``his,'' ``Steve,'' and ``the.'' Adding ``Sarah,'' ``his,'' or ``the'' keeps the probability of the partner being a man low, while adding ``Steve'' increases the probability, indicating a masculine-coded partner.}
\end{figure*}

To prompt the model for story and \token-level annotations, we provide the model with simple instructions, in-context examples to demonstrate the task, the hypothesis, and the full or partial story premise (See \autoref{prompt:annottrustworthy}). The partial stories are used to collect \token-by-\token measurements of the elements of interest at every\footnote{Due to technical limitations, server lag caused fast typing participants to, at times, not receive an updated set of suggestions for every word they inputted leading to some nonsensical model-suggested \token continuations. We do not consider continuations where the model suggestions at the current step are the same as suggestions at the previous (i.e., when the server has lagged).} step, both for the \tokens that are included in the final story and for the model-suggested \tokens that are rejected. For example, in \autoref{fig:token_annot}, from a given state, we consider the addition of the next \token that was actually used in the story (in this case, written by the participant) as well as the options provided by the model that were rejected by the participant.  

Based on a manual examination of the data, we empirically choose $0.8$ as the probability cutoff point for determining whether an attribute is present in the story. In other words, if the model outputs that the probability of the doctor character being \woman in a (partial) story is greater than $0.8$, then we consider the (partial) story as having a \woman doctor in it. For the \token-by-\token annotations, we mark a \token as specifying a given attribute if the previous \token's score was less than $0.8$, the new \token's score is greater than $0.8$ and the difference between them is greater than $0.3$. In the example in \autoref{fig:token_annot}, we can see that under this cutoff, the \tokens ``Sarah'', ``his'', and ``the'' did not lead (or would not have led) the model to predict Joe is marrying a \man partner, but the \token ``Steve'' would have led to a \man partner prediction. 

We employ manual and automated cleaning on these annotations, as the purpose is not to evaluate the LLM's ability to annotate these (parts of) stories but to obtain a reliable set of annotations of gender and ABC traits. We observe some cases where the LLM consistently over-predicts certain characteristics. For example, in the \textsc{Wedding} scenario, the partial story ``Sherry was happy to be marrying her'', the LLM understandably predicts that Sherry's partner is a woman as though ``her'' is an object pronoun (i.e., the pronoun refers to Sherry's partner). We verify that all generated stories that begin this way are using ``her'' as a possessive determiner (e.g., as in ``her fiancee'') and remove all such cases from the set of \tokens that would determine that Sherry's partner is \woman. For more details about how model annotations were generated, see \autoref{sec:addl_model_details}.

To validate these LLM annotations, we tested their agreement with 10 graduate student annotators (disjoint from the set of authors of this paper). They were asked to annotate $560$ story-attribute pairs total covering every axis of interest and were paid $\$5$ for an median compensation rate of $\$15.79$ per hour. For more complete details including instructions, see \autoref{sec:human_anno}. 
Pooling annotations between human annotators, we find an overall agreement level per Cohen's Kappa of $\kappa_{\text{all}} = 0.768$ which constitutes ``substantial agreement''~\cite{Landis1977} between humans and the LLM annotator. For the gender annotations, we find an agreement of $\kappa_{\text{gender}} = 0.782$. For the other ABC traits (likability, assertiveness, etc), we find a slightly lower agreement of $\kappa_{\text{ABC}} = 0.757$ perhaps as these traits are more subjective than gender. %

These story-level and \token-level annotations are then used as outcome measurements in H1, H2a, and H2b. In H1, we consider how often specific stereotype-relevant content is present in overall stories. For example, we consider how likely a president character is to be described using \woman language, depending on the presence or type of suggestions. In H2a, we consider how often the words in the final story (especially those that mark stereotype-relevant features) were suggested by the model. For example, we measure how often any word in the final story was accepted exactly from a model suggestion when the model is prompted to suggest the president uses \woman language. We consider the same measurement on the subset of \tokens that mark gender (where this subset is chosen based on the change in LLM annotator confidence in the president's gender). Finally, in H2b, we consider when the model suggests words that mark stereotype-relevant features, how often the participants are to accept them. For example, we measure how often \participants accept model suggested \tokens when these words would mark a the president as \woman. These story-level and \token-level annotations are then used as outcome measures in the analysis below. Together, these annotations provide the outcome measures used in the analyses that follow.

%% file: Sections/S5.Analysis.tex
\section{Results}\label{sec:results}

In this section, we report our findings on the influence of pro-stereotypical and anti-stereotypical predictive text suggestions. We first 
summarize (\autoref{sec:gender_summary}) the effects of predictive text suggestions on gender at the level of stories.

We then discuss the influences of biased predictive text in more detail at the story and \token level for a subset of three scenarios from \autoref{tab:scenarios}: \textsc{Detective} (\autoref{sec:detectives}), \textsc{Wedding} (\autoref{sec:wedding}), and \textsc{President} (\autoref{sec:president}). We selected these three scenarios for the body of the paper to cover a variety of trait configurations. In the \textsc{Detectives} scenario, we see how the inclusion of an ABC trait in the \prefix affects how participants decide a character's gender. In the \textsc{Wedding} scenario, we see how varying the gender of a character in the \prefix affects how participants decide a second character's gender. And in the \textsc{President} we see how participants decide a character's gender and ABC trait jointly. Similar analyses of the remaining scenarios are in \autoref{sec:other_scenario_results} and follows similar trends, though in some cases with more mixed conclusions.

Finally, we cover additional effects such as the impact of suggestion type on the time to make decisions, the effect of participants' pre-existing gender biases, and the effect of participants' level of English proficiency (\autoref{sec:other_body}).

Unless otherwise stated, all comparisons in this analysis were conducted using independent t-tests, with effect sizes reported as Cohen's $d$. We perform Benjamini-Hochberg correction and report the adjusted p-values with $\p$. Tables of all p-values in the scenario-level tests can be found in \autoref{sec:p_tables}. All error bars in our figures show $90\%$ confidence intervals.

\subsection{Summary of Gender Effects at the Story Level}\label{sec:gender_summary}

\aptLtoX[graphic=no,type=html]{\begin{table*}[tb]
   \centering
   \footnotesize
       \centering
       \begin{tabular}{l c c c c }
        \toprule
         &   \multicolumn{2}{c}{Pro-stereo Suggestions} & \multicolumn{2}{c}{Anti-stereo Suggestions}    \\
         \cmidrule(lr){2-3} \cmidrule(lr){4-5} 
         & Rate of & Rate of & Rate of & Rate of \\
         &  Pro-stereo &  Anti-stereo &  Pro-stereo &  Anti-stereo  \\
         &  in stories &  in stories & in stories  & in stories\\
         \midrule
         \textsc{Detective}        & - & - & $\downarrow$ & $\uparrow$   \\
         \textsc{President}        & - & - & $\downarrow$ & -  \\
         \textsc{Doctor}           & - & - & -            & $\uparrow$  \\
         \textsc{Wedding} (Joe)    & - & - & -            & -\\
         \textsc{Wedding} (Sherry) & - & - & $\downarrow$            & $\uparrow$\\
         \bottomrule
  \multicolumn{5}{l}{\textbf{(a) {Comparison of rates of writing characters pro-stereotypical (or anti-stereotypical gender) with pro-stereotypical or anti-stereotypical suggestions and no suggestions as a baseline. Changes marked with an arrow are statically significant. In our scenarios, it is possible for stories to include neither the pro-stereotypical nor the anti-stereotypical trait (e.g., the doctor's gender is never specified). This means that a change to the rate of pro- or anti-stereotypical stories does not necessitate a corresponding change to the other.}}}\\
  \end{tabular}
       \begin{tabular}{ll c c c}
         \toprule
         \multicolumn{2}{l}{} & No & Pro-stereo & Anti-stereo \\
         \multicolumn{2}{l}{}& Suggest. & Suggest. & Suggest.\\
         \midrule
            \multicolumn{2}{l}{\textsc{Detective}} & \cellcolor{customrowcolor}\phantom{$0$}$6.44\times$ & \cellcolor{customrowcolor}$10.29\times$ & \cellcolor{customrowcolor}\phantom{$0$}$1.37\times$\\
            \multicolumn{2}{l}{\textsc{President}} & \cellcolor{customrowcolor}$14.67\times$ & \cellcolor{customrowcolor}$34.33\times$ & \cellcolor{customrowcolor}\phantom{$0$}$2.38\times$\\
            \multicolumn{2}{l}{\textsc{Doctor}} &    \cellcolor{customrowcolor}\phantom{$0$}$6.33\times$ & \cellcolor{customrowcolor}\phantom{$0$}$7.56\times$ & \phantom{$0$}$1.23\times$\\
            \multicolumn{2}{l}{\textsc{Wedding}} (Joe) & \cellcolor{customrowcolor}$12.00\times$ & \cellcolor{customrowcolor}$11.17\times$ & \cellcolor{customrowcolor}\phantom{$0$}$3.00\times$ \\
            \multicolumn{2}{l}{\textsc{Wedding} (Sherry)} & \cellcolor{customrowcolor}\phantom{$0$}$5.67\times$ & \cellcolor{customrowcolor}\phantom{$0$}$8.25\times$ & \phantom{$0$}$1.44\times$\\
         \bottomrule
          \multicolumn{5}{l}{\textbf{(b) {How many times more pro-stereotypically gendered characters were written than anti-stereotypically gendered characters with various suggestions. Numbers shown in gray %
        are statistically significant. No number is $<1$, meaning all stories had at least as many pro-stereotypically gendered characters than anti-stereotypically gendered characters, regardless of condition.}
        }}\\     \end{tabular}
   \caption[]{Summary of story-level character genders. We include the four writing scenarios where the participant has control over the character's gender (a) Adding pro-stereotypical suggestions never significantly changes the rates of pro-stereotypically gendered and anti-stereotypically gendered characters. Adding anti-stereotypical suggestions significantly decreases pro-stereotypically gendered characters or increases anti-stereotypically characters except when writing about ``Joe's'' wedding. For ``Joe'''s wedding we see an insignificant decrease to the rate of pro-stereotypically gendered (i.e. \woman) partners when suggested. (b) Despite these differences, we never observe a case where anti-stereotypically gendered characters are chosen significantly more often than pro-stereotypically gendered characters.%
   }\label{tab:gender_summary} \label{subtable:summary_a}\label{subtable:summary_b}
\end{table*}}{\begin{table*}[tb]
   \centering
   \footnotesize
   \begin{subtable}{0.56\linewidth}
       \centering
       \begin{tabular}{l c c c c }
        \toprule
         &   \multicolumn{2}{c}{Pro-stereo Suggestions} & \multicolumn{2}{c}{Anti-stereo Suggestions}    \\
         \cmidrule(lr){2-3} \cmidrule(lr){4-5} 
         & Rate of & Rate of & Rate of & Rate of \\
         &  Pro-stereo &  Anti-stereo &  Pro-stereo &  Anti-stereo  \\
         &  in stories &  in stories & in stories  & in stories\\
         \midrule
         \textsc{Detective}        & - & - & $\downarrow$ & $\uparrow$   \\
         \textsc{President}        & - & - & $\downarrow$ & -  \\
         \textsc{Doctor}           & - & - & -            & $\uparrow$  \\
         \textsc{Wedding} (Joe)    & - & - & -            & -\\
         \textsc{Wedding} (Sherry) & - & - & $\downarrow$            & $\uparrow$\\
         \bottomrule
    \end{tabular}
       \caption[]{Comparison of rates of writing characters pro-stereotypical (or anti-stereotypical gender) with pro-stereotypical or anti-stereotypical suggestions and no suggestions as a baseline. Changes marked with an arrow are statically significant. In our scenarios, it is possible for stories to include neither the pro-stereotypical nor the anti-stereotypical trait (e.g., the doctor's gender is never specified). This means that a change to the rate of pro- or anti-stereotypical stories does not necessitate a corresponding change to the other.}
       \label{subtable:summary_a}
   \end{subtable}%
   \hspace{.01\linewidth}
   \begin{subtable}{0.42\linewidth}
       \centering
       \begin{tabular}{l c c c}
         \toprule
         & No & Pro-stereo & Anti-stereo \\
         & Suggest. & Suggest. & Suggest.\\
         \midrule
            \textsc{Detective} & \cellcolor{customrowcolor}\phantom{$0$}$6.44\times$ & \cellcolor{customrowcolor}$10.29\times$ & \cellcolor{customrowcolor}\phantom{$0$}$1.37\times$\\
            \textsc{President} & \cellcolor{customrowcolor}$14.67\times$ & \cellcolor{customrowcolor}$34.33\times$ & \cellcolor{customrowcolor}\phantom{$0$}$2.38\times$\\
            \textsc{Doctor} &    \cellcolor{customrowcolor}\phantom{$0$}$6.33\times$ & \cellcolor{customrowcolor}\phantom{$0$}$7.56\times$ & \phantom{$0$}$1.23\times$\\
            \textsc{Wedding} (Joe) & \cellcolor{customrowcolor}$12.00\times$ & \cellcolor{customrowcolor}$11.17\times$ & \cellcolor{customrowcolor}\phantom{$0$}$3.00\times$ \\
            \textsc{Wedding} (Sherry) & \cellcolor{customrowcolor}\phantom{$0$}$5.67\times$ & \cellcolor{customrowcolor}\phantom{$0$}$8.25\times$ & \phantom{$0$}$1.44\times$\\
         \bottomrule
       \end{tabular}
        \caption[]{How many times more pro-stereotypically gendered characters were written than anti-stereotypically gendered characters with various suggestions. Numbers shown in gray %
        are statistically significant. No number is $<1$, meaning all stories had at least as many pro-stereotypically gendered characters than anti-stereotypically gendered characters, regardless of condition.}
        \label{subtable:summary_b}
   \end{subtable}
   \caption[]{Summary of story-level character genders. We include the four writing scenarios where the participant has control over the character's gender (a) Adding pro-stereotypical suggestions never significantly changes the rates of pro-stereotypically gendered and anti-stereotypically gendered characters. Adding anti-stereotypical suggestions significantly decreases pro-stereotypically gendered characters or increases anti-stereotypically characters except when writing about ``Joe's'' wedding. For ``Joe'''s wedding we see an insignificant decrease to the rate of pro-stereotypically gendered (i.e. \woman) partners when suggested. (b) Despite these differences, we never observe a case where anti-stereotypically gendered characters are chosen significantly more often than pro-stereotypically gendered characters.%
   }\label{tab:gender_summary}
\end{table*}}
\begin{figure*}[bt]
    \centering
    \includegraphics[width=.8\linewidth,clip,trim=5 5 5 5]{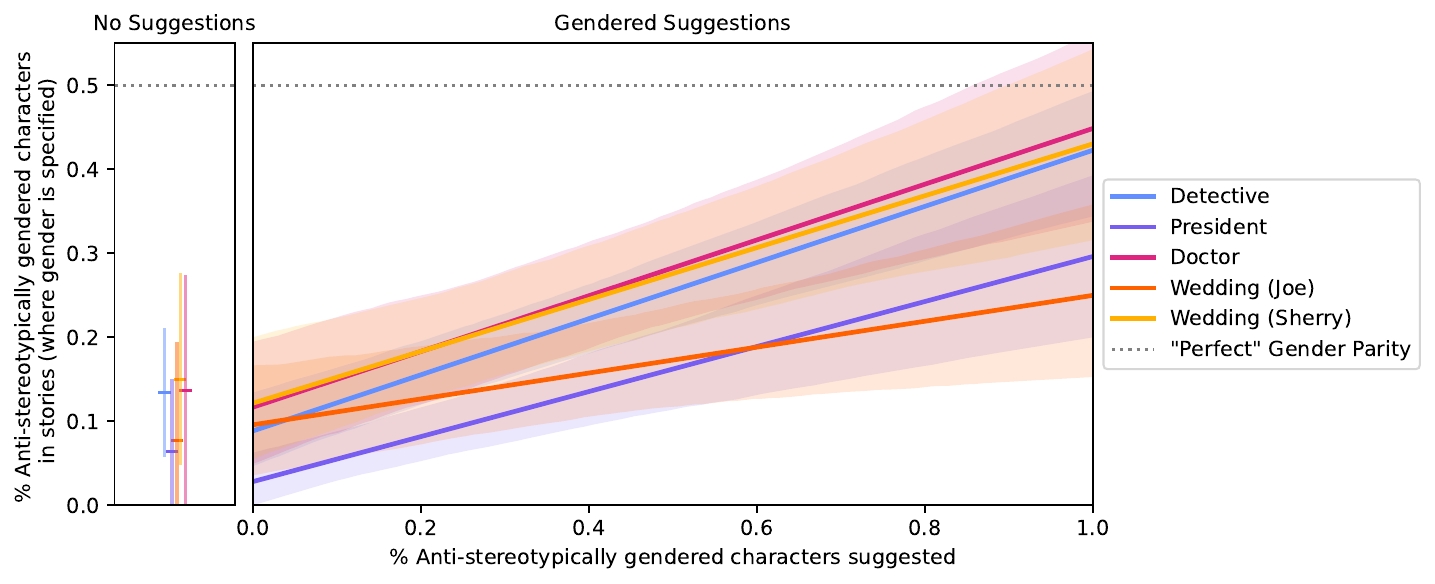}
    \caption[]{Expected rates of anti-stereotypically gendered characters in human-written stories (y-axis) with no suggestions (left) and    
    as the proportion of anti-stereotypical predictive text suggestions increases (x-axis, right). We can see that even with exclusively anti-stereotypical suggestions, we predict that gender parity falls below $y=0.5$ or ``perfect'' gender parity. In our study, we measure the two extremes ($x=0$ and $x=1$) for each scenario and calculate the fraction of stories with characters of each gender. The predicted values from other distributions of suggestions are drawn from a linear interpolation between those points. In general, a model that suggests entirely pro-stereotypical text ($x=0$) yields stories that are only slightly (if at all) more stereotyped than with no suggestions. And a model that suggests entirely anti-stereotypical text ($x=1$) increases the rate of anti-stereotypical stories, but never so much so as to even reach parity with pro-stereotypical stories. Note that the variance in $x$ value on the left plot is for visual clarity only.}
    \label{fig:rates}
    \Description{Line graphs showing expected rates of anti-stereotypically gendered characters as the proportion of anti-stereotypical suggestions increases for detectives, presidents, doctors, and weddings. Each curve shows the observed rates with no suggestions as a reference. Approximate thresholds where curves surpass the no-suggestion baseline are indicated: ~14\% for detectives, ~15\% for presidents, ~6\% for doctors, and ~14\% for weddings with ``Sherry.'' The curve for weddings with ``Joe'' shows a slight increase in masculine-coded partners even with pro-stereotypical suggestions.
}
\end{figure*}

We summarize our results on gender in overall stories for scenarios where participants, rather than the story prefix, determined a character's gender in \autoref{tab:gender_summary}. In all scenarios except ``Joe's'' wedding, fem-coded character suggestions are anti-stereotypical. We find that pro-stereotypical suggestions have no significant effects when compared to writing without suggestions (\autoref{subtable:summary_a}, left). By contrast, anti-stereotypical suggestions significantly or marginally shift writing toward anti-stereotypical characterization or away from pro-stereotypical characterization (\autoref{subtable:summary_a}, right). Still, in every scenario, pro-stereotypical characters remain (often significantly) more common than anti-stereotypical ones, despite exclusively anti-stereotypical suggestions (\autoref{subtable:summary_b}).



These experiments test extreme cases: predictive text suggestions that are entirely pro- or anti-stereotypical. As discussed in \autoref{sec:rqs}, there is no single definition of ``fair'' predictive text suggestions or distributions of stories. Importantly, our anti-stereotypical condition should be understood as an upper bound, stronger than what most researchers or practitioners would consider a ``fair'' or ``debiased'' model.\footnote{In our study design, we also only consider cases where the LLM is explicitly prompted to suggest one character attribute or another. In other words, each participant sees either suggestions that, for example, describe the doctor using \woman or \man language. In this analysis, we consider what would happen if different proportions of participants were given \woman vs \man suggestions. Here, we think of a ``debiased'' model as one that suggests \woman or \man language equally often, but still ``chooses'' one or the other to suggest in each story. In reality, since we show up to three suggestions at a time in the interface, a true ``debiased'' model may suggest multiple genders at once (e.g., having both ``she'' and ``he'' among the top three suggestions). Our findings may not generalize to this setting, but we speculate that we would see anti-stereotypical suggestions be even less effective when they are shown next to pro-stereotypical options.} 


To examine more realistic bias configurations, \autoref{fig:rates} shows how the expected proportion of anti-stereotypical stories changes for less extreme proportions of anti-stereotypical suggestions. We find no setting that would yield perfect parity in depicting \woman and \man characters, even when suggestions are exclusively anti-stereotypical. However, even small amounts of anti-stereotypical suggestions are enough to increase anti-stereotypical writing above the baseline (unassisted) rate. For instance, in the Doctor scenario, $6.4\%$ fem-coded suggestions are estimated to be sufficient to increase \woman doctors over unassisted writing.



In cases where developers are targeting a specific distribution of stories (e.g., one that matches some real-world gender distribution), 
we see that it may be possible to yield an intended distribution; however, this would require the use of a model that is more ``anti-stereotypical'' than the developer is targeting. For example, in the United States, $26.3\%$  of detectives were women in 2023.\footnote{\url{https://www.bls.gov/cps/cpsaat11.htm}}
A model that suggests $26.3\%$ \woman detectives would yield only an expected $~17.7\%$ \woman detectives in co-written stories. To reach $~26.3\%$ \woman detectives, we expect that model would need to suggest $~52.2\%$ \woman detectives. 

Regardless of whether a developer conceives of a ``fair'' model as one satisfying parity constraints or matching some real-world demographic distribution, we consistently find that the difference in human acceptance of pro-stereotypical and anti-stereotypical model suggestions will lead to a final gender distribution of \hai stories where anti-stereotypical stories do not significantly outnumber pro-stereotypical stories (and in fact, are often still significantly outnumbered by pro-stereotypical stories). To achieve a ``fair'' gender distribution in \hai co-written stories, developers would need to suggest anti-stereotypical completions more often than they think is ``fair'', and depending on their definition of a ``fair'' distribution, the desired outcome may not be possible through predictive text suggestions alone. Thus, while increasing the rate of anti-stereotypical suggestions can help encourage users to compose more anti-stereotypical writing, interventions that focus exclusively on debiasing model suggestions may be insufficient.

\subsection{Scenario: Detectives}\label{sec:detectives}

In this scenario, participants continue from a \prefix that describes Detective Wilson's partner as either trustworthy or untrustworthy. The model then suggests \woman or \man language to describe Detective Wilson's partner (See examples in \autoref{tab:story_egs}). 
In both \citet{cao-etal-2022-theory} and our post-survey, participants viewed men as less trustworthy or warm than women (though we note that this difference is not statistically significant in Cao et. al. See \autoref{tab:ABC_Cao}).
We therefore treat stories with \man detectives to be pro-stereotypical and stories with trustworthy \woman detectives (and untrustworthy \man detectives) to be more pro-stereotypical than their untrustworthy \woman (and trustworthy \man) counterparts.

\subsubsection{Effects on Gender Alone}

First, we consider the effects of suggestion on gender, regardless of the trustworthiness of Detective Wilson's partner. Participants specified the partner character's gender in about $92\%$ of stories (in both treatment and control conditions).  We first analyze gender at the story-level, then the relationship between gender and \token-level reliance.

\begin{figure}[t]
    \centering
    \includegraphics[width=.95\linewidth]{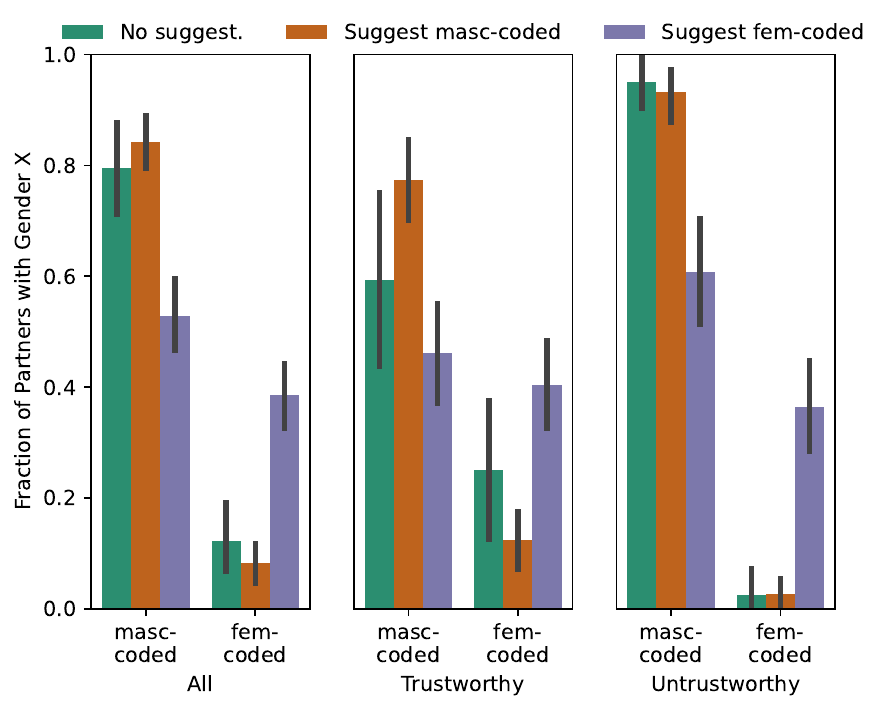}
    \caption[]{Inferred partner gender the \textsc{Detectives} scenario. Left shows all stories while the middle and right show stories where the partner detective is pre-determined in the \prefix to be trustworthy or untrustworthy respectively. The colors show whether the participant who wrote each story received no suggestions (teal) or suggestions that the partner has a \man (orange) or \woman (purple) name.}%
    \label{fig:detective_gender}
    \Description{Bar charts showing the gender distribution of detective partners across no-suggestion, masculine-coded, and feminine-coded suggestion conditions, further grouped by partner trustworthiness. The charts show more masculine-coded than feminine-coded detectives in all conditions, with higher proportions of feminine-coded detectives under feminine-coded suggestions, and higher masculine-coded proportions among untrustworthy partners.
}
\end{figure}

\paragraph{At the Story Level}
First, we compare the proportion of stories with partners of a given gender that were written with vs without suggestions (\autoref{fig:detective_gender}). Here, we see no significant differences when comparing gender rates without suggestions to rates with \man suggestions (\man partners: $t(242)=0.899$, $\p\approx 0.5681$, $d=0.126$; \woman partners: $t(242)=-1.012$, $\p\approx 0.5254$, $d=-0.141$).
When the model suggests that the partner should be \woman, we see significantly fewer \man partners ($t(234)=-4.0$, $\p\approx 0.0007$, $d=-0.563$) and significantly more \woman partners ($t(234)=4.191$, $\p\approx 0.0003$, $d=0.59$). However, even with these changes, we still see significantly more \man partners than \woman in all conditions (no suggestions: $t(144)=-10.934$, $\p<0.0001$, $d=-1.81$; \man suggestions: $t(340)=-21.727$, $\p<0.0001$, $d=-2.35$; \woman suggestions: $t(324)=-2.575$, $\p\approx 0.0408$, $d=-0.285$).

These results show that the stories written without suggestions are most similar to those written with the pro-stereotypical \man detective suggestions (H1). While anti-stereotypical \woman detective suggestions nudged participants away from \man detectives, it is not enough to get rid of, let alone reverse, the trend of writing more \man detectives than \woman. This means that participants writing with a ``debiased'' model that suggests \man and \woman detectives equally would still produce majority \man detectives. 

\paragraph{At the \ttoken Level}

We further assess participants' reliance on model suggestions at the \token level (\autoref{tbl:gender}). When 
considering all \tokens suggested by the model in the \woman and \man conditions,
we see that participants are significantly more likely to write new \tokens or edit model suggestions in the \woman detective setting (H2; $t(7091)=4.724$, $\p<0.0001$, $d=0.112$), but we see no significant trend when we constrain this to only the story \tokens that determine the second detective's inferred gender (H2a; $t(302)=-0.069$, $\p\approx 0.9634$, $d=-0.008$). We also see that participants are significantly less likely to accept model suggestions that would make the second detective \woman (H2b; $t(834)=6.729$, $\p<0.0001$, $d=0.479$). 

\subsubsection{Effects on Gender Disaggregated by Trustworthiness}
At the Story Level While we have seen that participants wrote significantly fewer \man detective and significantly more \woman detective stories with \woman suggestions, we find in the disaggregated results (\autoref{fig:detective_gender}) that this is only true when the partner detective is untrustworthy (untrustworthy: $t(113)=-4.367$, $\p\approx 0.0003$, $d=-0.85$; trustworthy: $t(119)=-1.561$, $\p\approx 0.2634$, $d=-0.322$). 
This happens because without suggestions, participants wrote significantly more \man partners when they were untrustworthy ($t(71)=-4.117$, $\p\approx 0.0008$, $d=-0.971$) and significantly more \woman partners when they were trustworthy ($t(71)=3.051$, $\p\approx 0.0161$, $d=0.72$).
On the other hand, \man suggestions led to significantly more untrustworthy \man partners ($t(169)=-2.881$, $\p\approx 0.0206$, $d=-0.445$) and marginally more trustworthy \woman partners ($t(169)=2.307$, $\p\approx 0.0735$, $d=0.356$), leading to no significant differences between stories written without suggestions or with \man suggestions for any gender or trustworthiness. 

These results show that control stories written without suggestions are more similar to those written with \man detective suggestions, regardless of trustworthiness (H1). Further, we see that participants wrote or accepted untrustworthy \man characters more than trustworthy \man characters. While \woman suggestions were not affected by trustworthiness, we also see that without suggestions, participants are more comfortable writing trustworthy \woman characters than trustworthy \man characters. This is consistent with human-held stereotypes as measured in our post-survey and in \citet{cao-etal-2022-theory}---namely that people tend to view women as more trustworthy than untrustworthy and men more untrustworthy than trustworthy. 

\paragraph{At the \ttoken Level}

When we disaggregate by trustworthiness (\autoref{tbl:gender_and_other}), we surprisingly see significantly more newly written or edited \tokens when suggesting a trustworthy \woman partner than untrustworthy ($t(3443)=-3.168$, $\p\approx 0.0085$, $d=-0.108$), and we only see a significant difference between genders in the trustworthy case ($t(3907)=5.235$, $\p<0.0001$, $d=0.168$), with more new typing in the trustworthy \woman partner case. When we constrain to \tokens that determine gender, the only remaining significant trend is the increased participant contribution in the trustworthy \woman partner case ($t(161)=-3.298$, $\p\approx 0.0069$, $d=-0.52$). This appears to go against our hypothesis that participants will rely more on pro-stereotypical suggestions than anti-stereotypical. However, when the participants decide to override the suggestions (by writing a new \token or editing a suggestion), this analysis does not take into account what gender is being expressed in the override. While we observed double the overrides in the trustworthy \woman partner case as untrustworthy, we also observe that more of the overrides in the trustworthy case ultimately still produce a \woman partner ($27.4\%$) than in the untrustworthy case ($12.9\%$). 

When we consider the rates of gender-specifying \tokens being accepted or rejected, we see no significant difference between untrustworthy and trustworthy \woman suggestions ($t(515)=1.364$, $\p\approx 0.3400$, $d=0.12$). Instead, we see significantly more acceptance of \man suggestions than \woman regardless of trustworthiness (trustworthy: $t(507)=-4.224$, $\p\approx 0.0003$, $d=-0.375$; untrustworthy: $t(325)=-7.066$, $\p<0.0001$, $d=-0.913$) and significantly more acceptance of untrustworthy \man suggestions than trustworthy ($t(317)=4.384$, $\p\approx 0.0002$, $d=0.569$ ). 

Overall, we see that the rejection rate results concur with our and \citet{cao-etal-2022-theory}'s findings about people's gender-trustworthiness stereotypes and support H2b. While our results do not support H2 and H2a (about overall reliance and the source of gender-defining \tokens), we note that these could be caused by a difference in the rate of overrides that change how/when the partner's gender is expressed but not what that gender is.

\subsection{Scenario: Wedding} \label{sec:wedding}

In this scenario, we consider a wedding between two partners. We vary whether the partner who is mentioned in the \prefix is named ``Joe'' (a traditionally masculine name) or ``Sherry'' (a traditionally feminine name) and prompt the LLM to suggest that the second partner in the couple to be \woman or \man (See examples in \autoref{tab:story_egs}). Participants specified the second partner character's gender in about $89\%$ of stories (in both treatment and control conditions). We treat the pro-stereotypical conditions to be those where the genders of the partners are suggested to be different and the anti-stereotypical to be those where they are suggested to be the same. 

\paragraph{At the Story Level}

\begin{figure}[tb]
    \centering
    \includegraphics[width=.95\linewidth,clip,trim=0 8 0 0]{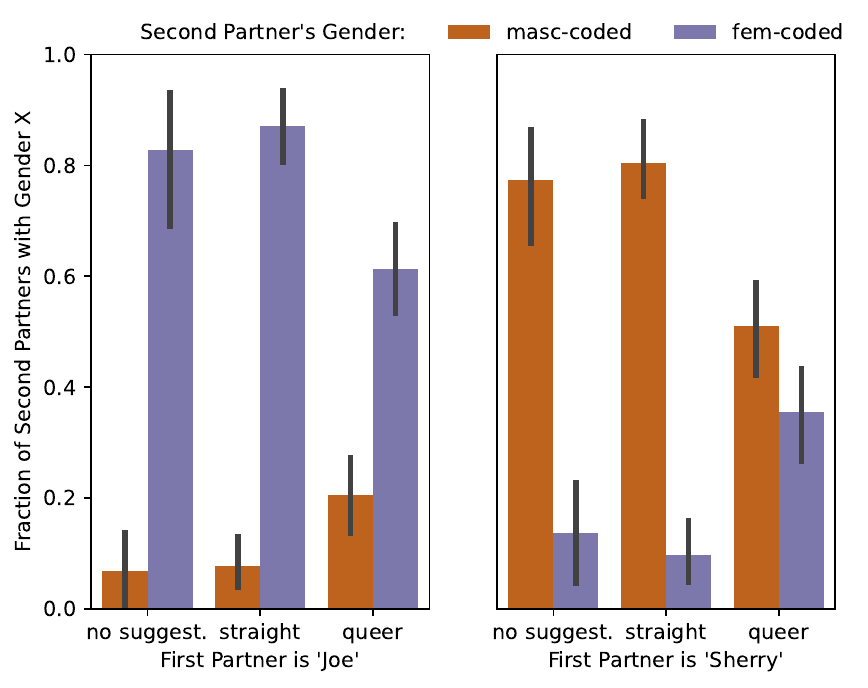}
    \caption[]{Inferred gender pairings in the \textsc{Wedding} scenario. Left and right show cases where the partner in the \prefix is \man (``Joe'') or \woman (``Sherry''). The x-axis shows whether participants were given no suggestions, suggestions of a straight relationship or a queer one. The colors show whether the second partner is written to be \man (orange) or \woman (purple) in the final story.}
    \label{fig:1_gender}
    \Description{Bar charts showing the gender distribution of second partners under varied suggestions, grouped by whether the first partner is named ``Joe'' or ``Sherry.'' When the first partner is ``Joe,'' the charts show more feminine-coded second partners and fewer queer relationships, even with queer suggestions. When the first partner is ``Sherry,'' the charts show higher acceptance of queer relationship suggestions.}
\end{figure}

We first focus on the rates of sexualities present in the overall stories (\autoref{fig:1_gender}). We start by considering what kinds of suggestions yield similar or different stories to those written without suggestions. When the first partner is ``Sherry'' and the model suggests a queer relationship (i.e., that Sherry's partner is \woman), we see significantly fewer stories where the other partner is \man ($t(132)=2.973$, $\p\approx 0.0172$, $d=0.547$) and significantly more stories where the other partner is \woman ($t(132)=-2.695$, $\p\approx 0.0328$, $d=-0.496$).
There are no significant changes to the distribution of Sherry's partners' gender when straight suggestions are provided (\woman partner: $t(124)=0.657$, $\p\approx 0.7019$, $d=0.123$; \man partner $t(124)=-0.422$, $\p\approx 0.8234$, $d=-0.079$). 
The direction of these trends are mirrored for ``Joe'' stories but with insignificant changes. That is, ``Joe'' is written with \man partners more often ($t(115)=-1.688$, $\p\approx 0.2218$, $d=-0.361$) and \woman partners less often when given queer suggestions ($t(115)=2.143$, $\p\approx 0.1071$, $d=0.459$), and there are no apparent changes when given straight suggestions to the rate of \woman ($t(104)=-0.556$, $\p\approx 0.7514$, $d=-0.121$) or \man partners ($t(104)=-0.154$, $\p\approx 0.9479$, $d=-0.034$). 
These findings suggest that indeed participants' default behavior without suggestions is more similar to the ``straight'' conditions than the ``queer'' ones, though participants seem to be more resistant to accepting ``queer'' suggestions for ``Joe'' than ``Sherry''. 

We also consider within a suggestion type when the difference between choosing \woman and \man partners is significant. When the first partner is named ``Joe'', we see significantly more \woman partners than \man, regardless of the presence or type of suggestions (No suggestions:  $t(56)=-8.825$, $\p<0.0001$, $d=-2.318$; queer suggestions: $t(174)=-6.035$, $\p<0.0001$, $d=-0.91$; straight suggestions: $t(152)=-16.063$, $\p<0.0001$, $d=-2.589$). For ``Sherry'', we see significantly more \man partners than \woman when participants see no suggestions ($t(86)=7.704$, $\p< 0.0001$, $d=1.643$) or straight suggestions ($t(162)=12.859$, $\p< 0.0001$, $d=2.008$). We see a trend in the same direction with queer suggestions, but here it is not significant ($t(178)=2.12$, $\p\approx 0.1097$, $d=0.316$).
These findings again imply that without suggestions, participants default to hetero-normative stories and continue to write them when prompted (H1). When queer stories are suggested, depending on the gender of the partner that is fixed in the story prefix, participants may start to accept more queer stories, but continue to prefer to write hetero-normative stories overall.


Based on these results, even when writing with a ``perfectly debiased'' predictive text system that, for instance, has no preference for gender pairings, we would expect to continue to see far more straight stories than queer ones.

\paragraph{At the \ttoken Level}

We continue by assessing the \token-level acceptance and overriding of model suggestions (\autoref{tbl:gender_and_other}). When we constrain only to the \tokens that affect the second partner's gender and the pair's inferred sexuality (H2a), we see that participants type their own gender-defining \tokens more when the first partner is \man regardless of whether the combination of genders match or do not (\woman queer vs \man queer: $t(206)=-2.566$, $\p\approx 0.0424$, $d=-0.358$; \woman straight vs \man straight: $t(195)=-3.554$, $\p\approx 0.0031$, $d=-0.507$).
 When we widen to all \tokens written in the stories (H2), we continue to see a higher proportion of \tokens coming from model suggestions in the conditions where the first partner is \woman (\woman queer vs \man queer: $t(4230)=-4.019$, $\p\approx 0.0005$, $d=-0.124$; \woman straight vs \man straight: $t(3699)=-4.373$, $\p\approx 0.0001$, $d=-0.144$), but now we also see we see marginally more acceptance of model suggestions in the condition where the \woman partner is suggested to be marrying a \man character ($t(4010)=2.381$, $\p\approx 0.0610$, $d=0.075$). 
 Considering the rate of accepting vs rejecting gender-defining model suggestions (H2b), we further see that queer suggestions are more rejected for \man characters than \woman ($t(546)=6.786$, $\p<0.0001$, $d=0.597$) and that for \man characters, queer suggestions are rejected significantly more than straight suggestions ($t(607)=-2.901$, $\p\approx 0.0187$, $d=-0.237$). 

 These findings generally suggest that participants are more accepting of suggestions about \woman character's weddings and are particularly unlikely to accept \man queer suggestions. These results echo our finding from the story level that queer suggestions are somewhat successful at yielding stories with queer pairings, especially for lesbian pairings.

\subsection{Scenario: President}\label{sec:president}

In this scenario, the \prefix describes a new president laying out their policy vision (``In the first 100 days, the new president was determined to focus''). The model is prompted (unseen by the participant) to suggest that the president is of a particular gender (man or woman) and communion (``benevolent'' or ``threatening''). We provide example co-written stories with each character attribute in \autoref{tab:story_egs}. The in-context examples given to the model focused on the benevolence axis as it applies to foreign policy, but the final written stories discuss a wider array of policy areas. In both \citet{cao-etal-2022-theory} and our post-survey, participants viewed men as less benevolent or warm than women (See \autoref{tab:ABC_Cao}). 

\subsubsection{Effects on Gender Alone}

First, we consider how suggestions affect how participants specify a president character's gender. Participants specified the president character's gender in about $57\%$ of stories (and $55\%$ of co-written stories).

\paragraph{At the Story Level}

\begin{figure*}[tb]
    \begin{subfigure}[t]{.32\linewidth}
    \centering
    \includegraphics[height=2.2in,clip,trim=0 8 0 0]{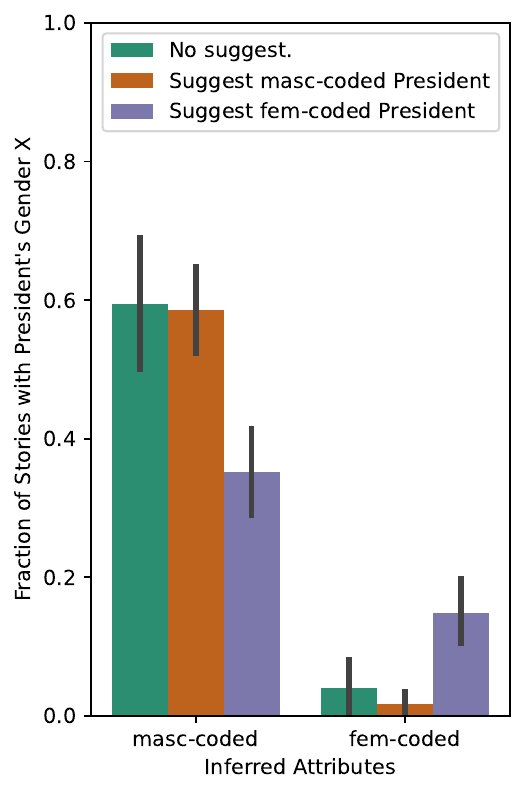}
    \caption[]{Inferred president gender. The colors show whether the participant who wrote each story received no suggestions (teal) or suggestions that president should use pronouns that are \man (orange) or \woman (purple). The ticks group stories by inferred president gender.}%
    \label{fig:5_gender}
    \end{subfigure}
    \hspace{.01\linewidth}
    \begin{subfigure}[t]{.66\linewidth}
            \centering
            \includegraphics[height=2.2in,clip,trim=0 8 0 0]{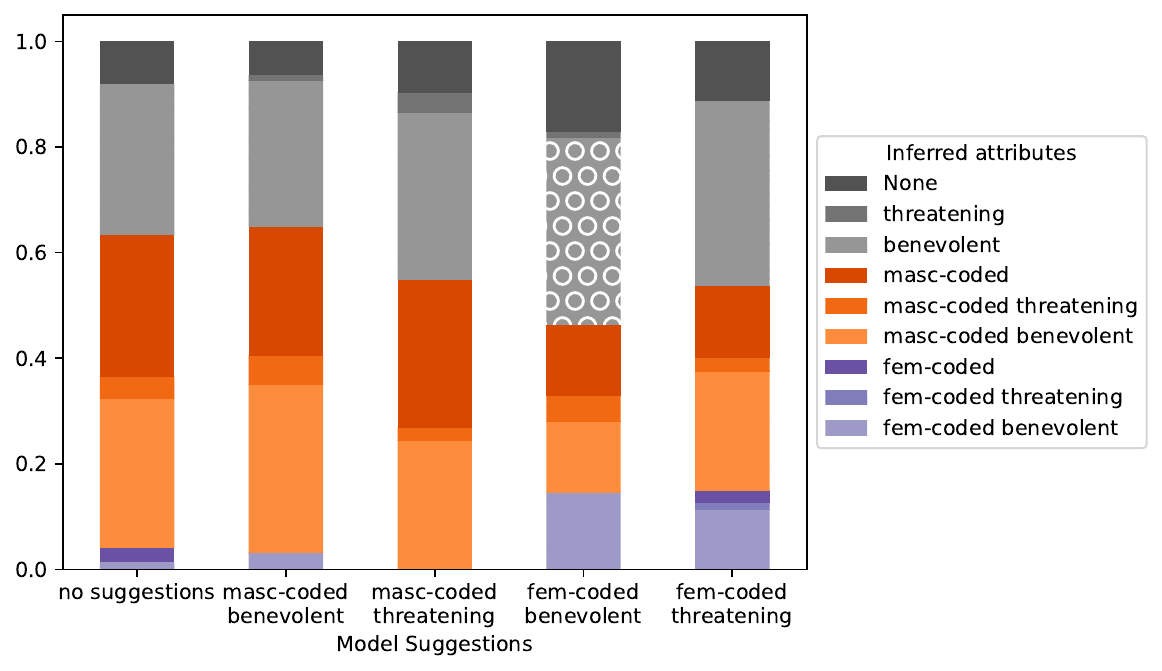}
            \caption[]{Inferred president gender and benevolence. The colors show inferred gender, and the patterns show the inferred benevolence. The ticks group stories by suggestions presence/type.}\label{fig:5_joint_body}
    \end{subfigure}
    \caption[]{Inferred characteristics in the \textsc{President} scenario.}
    \Description{Bar charts showing the gender distribution of presidents under no-suggestion, masculine-coded, and feminine-coded suggestion conditions, further grouped by benevolence. Masculine-coded suggestions produce a similar distribution to no suggestions, while feminine-coded suggestions increase feminine-coded presidents without exceeding masculine-coded. Across conditions, most presidents are labeled benevolent regardless of suggestions.}
\end{figure*}

We first look into how suggestions affect the president character's gender in the overall stories (\autoref{fig:5_gender}). We find that there is no significant difference in the rate of making the president \man without suggestions vs with \man suggestions ($t(248)=-0.137$, $\p\approx0.9488$, $d=-0.019$) and similarly for making the president \woman with \man suggestions ($t(248)=-1.106$, $\p\approx 0.4750$, $d=-0.153$). However, when the model suggests a \woman president, participants write significantly fewer \man presidents ($t(234)=-3.576$, $\p\approx 0.0029$, $d=-0.502$) and marginally more \woman presidents ($t(234)=2.429$, $\p\approx 0.0565$, $d=0.341$) than those who did not receive suggestions. Further, when we compare the rates of making the president \man vs \woman within conditions, we see that there are significantly fewer \woman presidents than \man presidents in every condition, even with \woman suggestions (No suggestions: $t(146)=-8.947$, $\p<0.0001$, $d=-1.471$; \man suggestions: $t(350)=-14.755$, $\p<0.0001$, $d=-1.573$; \woman suggestions: $t(322)=-4.343$, $\p\approx 0.0002$, $d=-0.483$). 

These results support H1, namely that the distribution of the gender of the president characters when people write by default without suggestions is more similar to the distribution when people write with \man suggestions than \woman suggestions. 
Based on these findings, we would expect stories written with a ``debiased'' predictive text model (that suggests \woman presidents less often than in our experiment) to still yield significantly more stories with \man presidents than \woman presidents. %

\paragraph{At the \ttoken Level}

Here we focus on \token-level reliance on suggestions in gendered conditions (\autoref{tbl:gender}). We find participants' overall reliance on model suggestions is not affected by the which gender is suggested (H2; $t(6734)=-0.979$, $\p\approx0.5407$, $d=-0.024$). However, when we only consider the story \tokens that specify the president character's gender, we see a lower rate of overrides or edits in the \man president settings (H2a; $t(203)=5.269$, $\p<0.0001$, $d=0.742$). Similarly, when we consider only model suggestions that would specify the president character's gender, we see a significantly higher rejection rate for \tokens that describe the president as \woman (H2b; $t(836)=6.362$, $\p<0.0001$, $d=0.442$). These results support H2a-b as participants are more likely to accept model suggestions of \man presidents than \woman presidents. 

\subsubsection{Effects on Gender and Benevolence Jointly}

Beyond gender on its own, we also consider the benevolence of the presidents (\autoref{fig:5_joint_body}). 
For each configuration of suggestions and each potential set of attributes that could be given to the president character, we consider whether adding that suggestion type changes the proportion of presidents that have that set of attributes. We observe that overall, the model was not successful in convincing participants to make threatening president characters, with the rate of threatening president characters (regardless of gender) being quite low regardless of the presence or type of suggestions with $22/412$ stories containing a threatening president ($8/162$ for stories written with threatening suggestions of either gender).

We do see that when provided with benevolent or threatening \woman suggestions, participants wrote significantly more benevolent \woman presidents (benevolent \woman suggestions: $t(154)=3.068$, $\p\approx0.0134$, $d=0.492$; threatening \woman suggestions: $t(152)=2.526$, $\p\approx0.0465$, $d=0.407$). %
These results weakly support H1. Joint gender and benevolence suggestions generally did not change the distribution of the president characters' attributes when comparing to stories written without suggestions. While \woman suggestions successfully increased the frequency of \woman president characters, participants did not accept these \woman characters being anti-stereotypically threatening, leading to an increase in benevolent \woman presidents when writing with either benevolent or threatening \woman suggestions when comparing to stories written with no suggestions.

\aptLtoX[graphic=no,type=html]{}{\begin{figure*}[tb]
    \centering
    \begin{subfigure}[t]{.4\linewidth}
            \centering
            \includegraphics[width=1\textwidth,clip,trim=0 0 0 0]{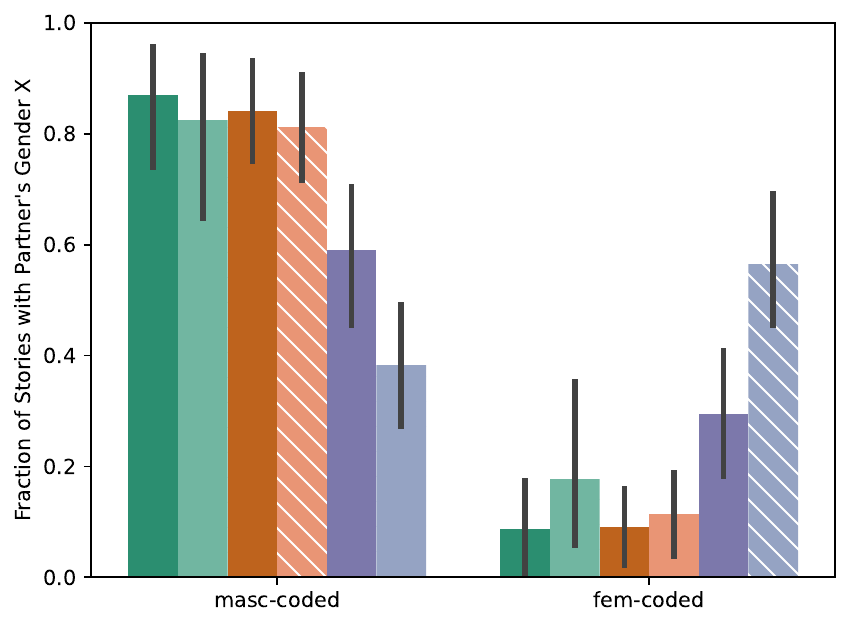}
            \caption[]{\textsc{Detective}}%
        \end{subfigure}
    \begin{subfigure}[t]{.4\linewidth}
            \centering
            \includegraphics[width=1\textwidth,clip,trim=0 0 0 0]{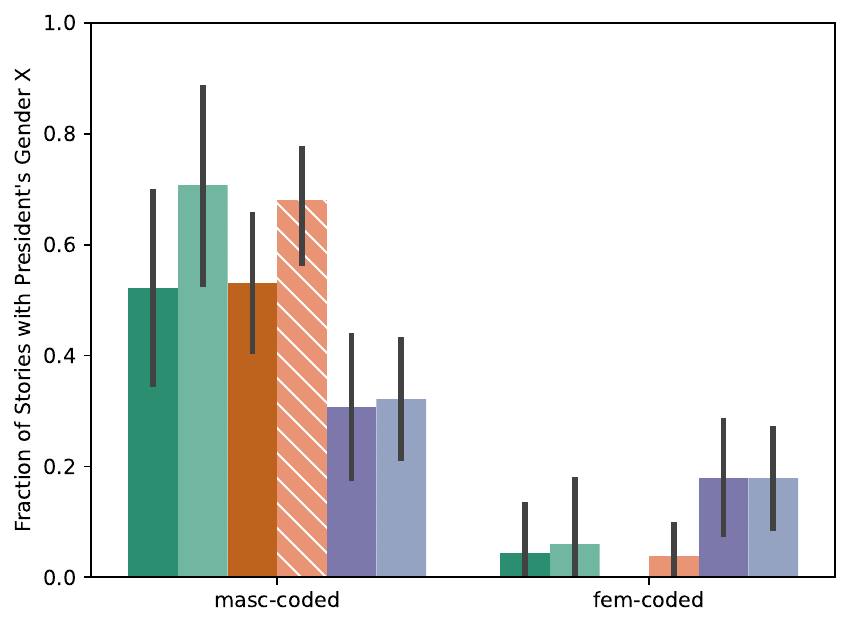}
            \caption[]{\textsc{President}}%
        \end{subfigure}
    \begin{subfigure}[c]{.4\linewidth}
            \includegraphics[width=1\textwidth,clip,trim=0 0 0 0]{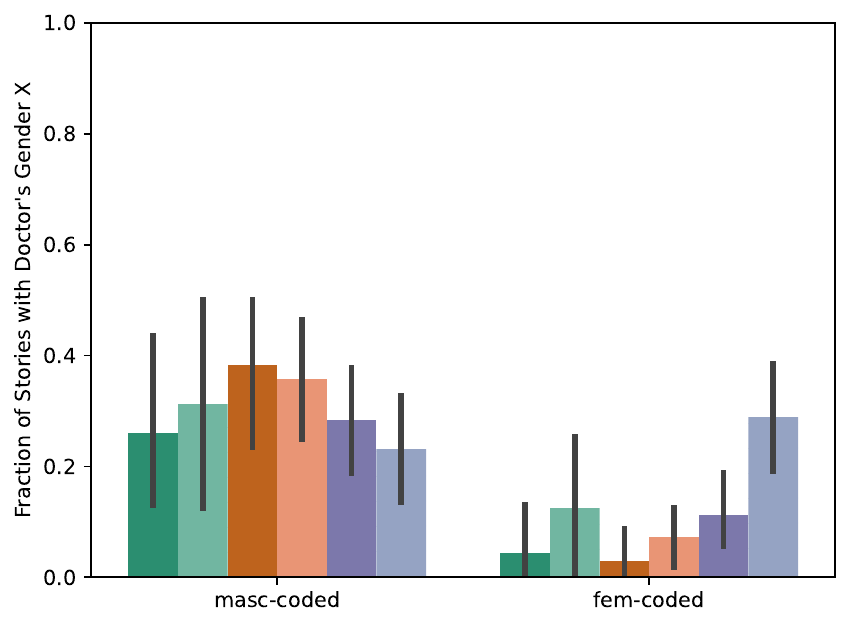}
            \caption[]{\textsc{Doctor}}%
        \end{subfigure}
    \begin{subfigure}[c]{.4\linewidth}
            \centering
            \includegraphics[height=1.4in,clip,trim=0 19 240 15]{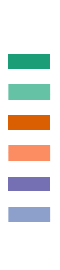}
            \includegraphics[height=1.4in,clip,trim=60 200 550 40]{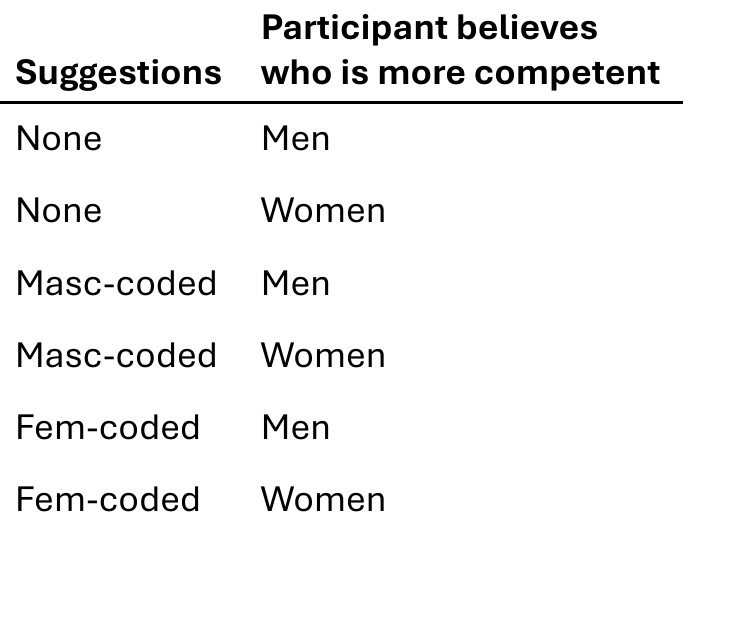}
        \end{subfigure}    
    \caption[]{Rates of character gender for participants who indicated gender differences in competence. In each scenario, we split stories into those written by participants who marked straight men as more competent (no hatching, more saturated) vs less competent (hatching, less saturated) than straight women. We plot the fraction of characters who are described as \man or \woman in stories written with no suggestions (teal), \man suggestions (orange), and \woman suggestion (purple). }\label{fig:competence}%
    \Description{Bar charts showing character gender distributions under different suggestions for participants who rated straight women as more versus less competent than men. The distributions are similar across groups, with higher proportions of feminine-coded doctors and detectives under suggestions among participants who rated straight women as more competent.}
\end{figure*}
}

\subsection{Additional Effects: Time to Decide and Individual Differences}\label{sec:other_body}

Beyond our primary hypotheses, here we consider the effects of suggestion type on the time to make individual decisions (expanded on in \autoref{sec:decision_time}), effects of participant's views and stereotypes on story attributes (expanded on with other individual differences in \autoref{sec:participant_attr_extra}), and the effect of English proficiency on participants' reliance on model suggestions. In the appendix, we further consider the effects of suggestions on overall story length and time to write (\autoref{sec:overall_len_time}), the distribution and correlation of participant stereotypes (\autoref{sec:corr_human_stereo}), and the effect of suggestions on toxicity, sentiment, and characters' agency in the co-written stories (\autoref{sec:agency}).  

\subsubsection{Effects of Suggestion Type on Time to Make Decisions}\label{sec:body_decision_time}
In previous sections, we focused on the decisions made by participants to accept/reject/write-in \tokens that specify various attributes. Here, we consider how long it takes participants to decide whether to accept model suggestions. We hypothesized (H3) that participants would take longer to decide whether to accept model suggestions when these suggestions are anti-stereotypical, as they are more likely to go against the participants' instincts about what attributes should be assigned and thus take longer to process and resolve.

In the \textsc{Detectives} scenario, we find that participants take significantly less time to decide whether accept suggestions of the partner being \man when the partner is untrustworthy as opposed to trustworthy ($t(306)=-3.544$, $\p\approx 0.0030$, $d=-0.464$). Further, when the partner is untrustworthy, it takes significantly less time for participants to make decisions about \man suggestions than \woman ones ($t(310)=-3.48$, $\p\approx 0.0036$, $d=-0.455$). This could mean that \man untrustworthy detective partners are the least unexpected group. This finding is in line with our findings about the rate of \man untrustworthy detectives in stories written without suggestions and is also in line with \citet{cao-etal-2022-theory}'s findings about gender-trustworthiness stereotypes.

We include more details of this analysis and include further scenarios in \autoref{sec:decision_time}. There are no significant differences in the time to make decisions on the basis of suggestion type in the \textsc{Doctor}, \textsc{President}, and \textsc{Town Hall} scenarios. For the scenarios where there are significant differences (as in the case of the \textsc{Detectives} scenario), these differences tend to show participants taking longer to make decisions about anti-stereotypical suggestions than pro-stereotypical, illustrating potential implicit associations~\cite{Greenwald1998-af}.

\aptLtoX[graphic=no,type=html]{}{
\begin{figure*}[tb]
\begin{subfigure}[c]{0.47\textwidth}
\centering
\includegraphics[width=.99\linewidth]{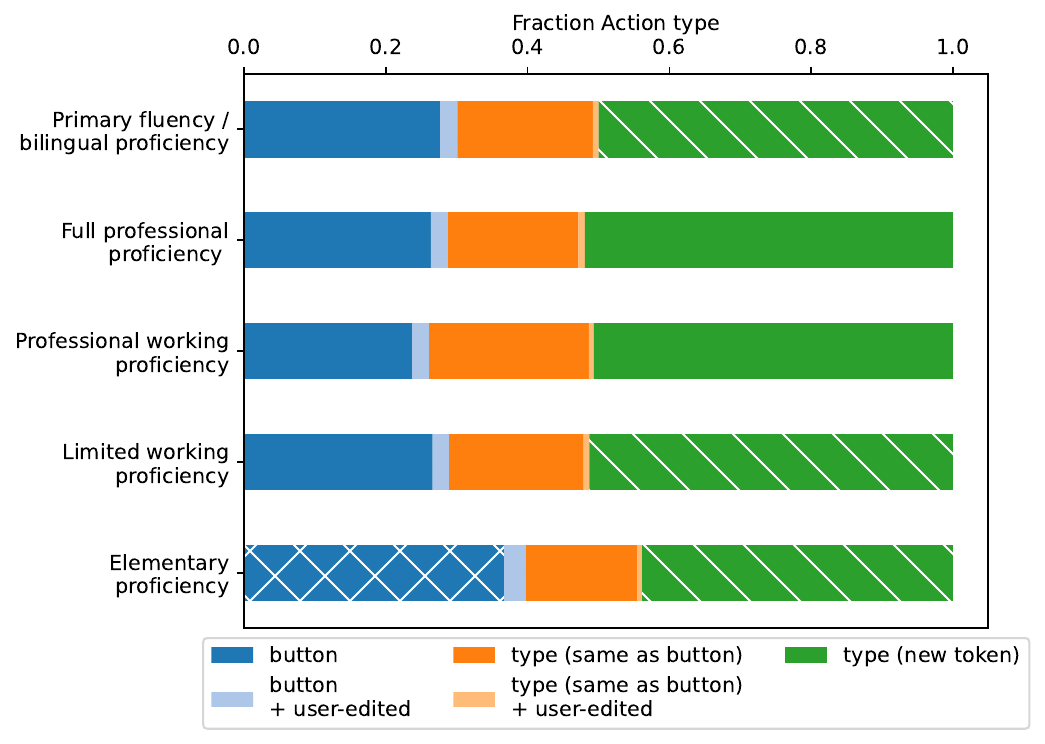}
\caption[]{}
\end{subfigure}%
\hspace{.01\textwidth}
\begin{subfigure}[c]{0.51\textwidth}
\centering
\footnotesize
    \begin{tabular}{l l c l}
    \toprule
    Proficiency levels & $t$ & $\p$ & sig \\
    \midrule
    1 (Elementary) vs & & & \\
    \qquad  2 (Limited working) & $t(17403)=-8.97$ & $0.0000$ & $*$\\
    \qquad  3 (Professional working) & $t(17673)=-8.228$ & $0.0000$ & $*$\\
    \qquad  4 (Full professional) & $t(18053)=-10.37$ & $0.0000$ & $*$\\
    \qquad  5 (Primary fluency/bilingual) & $t(25097)=-8.451$ & $0.0000$ & $*$\\
    2 (Limited working) vs  & & & \\
    \qquad  3 (Professional working) & $t(17494)=0.805$ & $0.6134$ & \\
    \qquad  4 (Full professional) & $t(17874)=-1.223$ & $0.4102$ & \\
    \qquad  5 (Primary fluency/bilingual) & $t(24918)=1.81$ & $0.1851$ & \\
    3 (Professional working) vs & & & \\  
    \qquad  4 (Full professional) & $t(18144)=-2.052$ & $0.1190$ & \\
    \qquad  5 (Primary fluency/bilingual) & $t(25188)=0.905$ & $0.5671$ & \\
    4 (Full professional) vs & & & \\  
    \qquad  5 (Primary fluency/bilingual) & $t(25568)=3.258$ & $0.0066$ & $*$\\
    \bottomrule
    \end{tabular}
\caption[]{}
\end{subfigure}
\caption[]{
AI reliance broken down by self-reported English proficiency. (a) We break down the set of \tokens in stories written by participants based on their source (e.g., the participant chose a model suggestion by pressing a suggestion button). (b) We compare the proportion of model suggested \tokens to human written or edited \tokens in pairs of proficiency levels.}
\label{fig:english}
\Description{Bar chart showing participants’ reliance on model suggestions across English proficiency levels. Participants in the ``elementary proficiency'' category show higher rates of accepting suggestions, either by pressing the suggestion button or by typing the model suggestion exactly. }
\end{figure*}}

\subsubsection{Effect of Participants' Views on Gender}\label{sec:participant_attr}

Here we consider how participants' views on gender affect the stories they write with and without suggestions (H4). Specifically, we focus on participants' perceptions of the ``competence'' of people of different genders and how these perceptions influence their gendering of doctor, president, and detective characters. We hypothesize that participants who believe that women are more competent than men will write more stories about \woman doctors, presidents, and detectives without suggestions and be more likely to accept model suggestions of these characters being \woman. 

We focus this analysis on the beliefs about \textit{straight} men and women as participants likely defaulted to these characters being straight. We exclude the $42\%$ of participants who rated straight men and women's competence within 10 points of each other (on a 0-100 scale) and then split participants into those who marked straight women as more competent ($53\%$ of the included participants) vs less competent ($47\%$ of the included participants)  than straight men. We plot the breakdown of character gender for these two groups in each scenario in \autoref{fig:competence}.

\aptLtoX[graphic=no,type=html]{\begin{figure*}[tb]
    \centering
    \begin{subfigure}[t]{.4\linewidth}
            \centering
            \includegraphics[width=1\textwidth,clip,trim=0 0 0 0]{figures/3_human_smw_comp.pdf}
            \caption[]{\textsc{Detective}}%
        \end{subfigure}
    \begin{subfigure}[t]{.4\linewidth}
            \centering
            \includegraphics[width=1\textwidth,clip,trim=0 0 0 0]{figures/5_human_smw_comp.pdf}
            \caption[]{\textsc{President}}%
        \end{subfigure}
    \begin{subfigure}[c]{.4\linewidth}
            \includegraphics[width=1\textwidth,clip,trim=0 0 0 0]{figures/0_human_smw_comp.pdf}
            \caption[]{\textsc{Doctor}}%
        \end{subfigure}
    \begin{subfigure}[c]{.4\linewidth}
            \centering
            \includegraphics[height=1.4in,clip,trim=0 19 240 15]{figures/smw_comp_legend.pdf}
            \includegraphics[height=1.4in,clip,trim=60 200 550 40]{figures/smw_comp_legend_table.pdf}
        \end{subfigure}
    
    \caption[]{Rates of character gender for participants who indicated gender differences in competence. In each scenario, we split stories into those written by participants who marked straight men as more competent (no hatching, more saturated) vs less competent (hatching, less saturated) than straight women. We plot the fraction of characters who are described as \man or \woman in stories written with no suggestions (teal), \man suggestions (orange), and \woman suggestion (purple). }\label{fig:competence}%
    \Description{Bar charts showing character gender distributions under different suggestions for participants who rated straight women as more versus less competent than men. The distributions are similar across groups, with higher proportions of feminine-coded doctors and detectives under suggestions among participants who rated straight women as more competent.}
\end{figure*}}{}

For all scenarios, when we compare the gendered competence groups, we see no significant difference in the proportion of \man or \woman characters written without suggestions or with \man suggestions. We do, however, observe a higher rate of \woman characters when suggested in the \textsc{Doctor} and \textsc{Detectives} scenarios. This trend is significant for the \textsc{Detectives} scenario ($t(97)=2.742$, $\p\approx0.0304$, $d=0.555$) and marginally significant for the \textsc{Doctor} scenario ($t(103)=2.279$, $\p\approx0.0802$, $d=0.445$). This provides some evidence that participants are more willing to accept anti-stereotypical suggestions when they (or their close friends) hold anti-stereotypical beliefs.

As we discuss in more detail in \autoref{sec:participant_attr_extra} in the appendix, we do not find evidence that participants' gender identity directly affects acceptance of gendered suggestions. However, we do find that men are more likely than women to endorse the belief that men are more competent than women ($t(412)=2.082$, $p\approx0.0379$, $d=0.209$).\footnote{This test was exploratory, not part of the main analysis, and was not pre-registered. As such, it was excluded from the Benjamini-Hochberg correction applied to the primary analyses.} These results highlight that binary gender identities do not capture a uniform set of experiences or beliefs. In contrast, measures of gender-related attitudes, though not uniformly predictive across writing scenarios, offer comparatively more insight into participants' interactions with gendered predictive text suggestions.

\subsubsection{English Proficiency and Reliance on Suggestions}

Prior work by \citet{buschek2021native} has found that native and non-native English speakers interact with English phrase suggestions differently, noting that as the number of suggestions shown at once increased, non-native reliance grew faster than native reliance. In our study, we ask participants to self-report their level of English proficiency and consider how this affects reliance on predictive text (\autoref{fig:english}). We find that the ``Elementary proficiency'' group overrode model suggestions significantly less than any other group, supporting H5. Unexpectedly, we also found that the highest proficiency group overrode suggestions marginally significantly less than the second highest proficiency group. Overall, we emphasize a potential greater risk for biased English predictive text suggestions to influence the writing of less proficient English speakers, as they may be more dependent on such suggestions.

\aptLtoX[graphic=no,type=html]{\begin{figure*}[tb]
\begin{subfigure}[c]{0.47\textwidth}
\centering
\includegraphics[width=.99\linewidth]{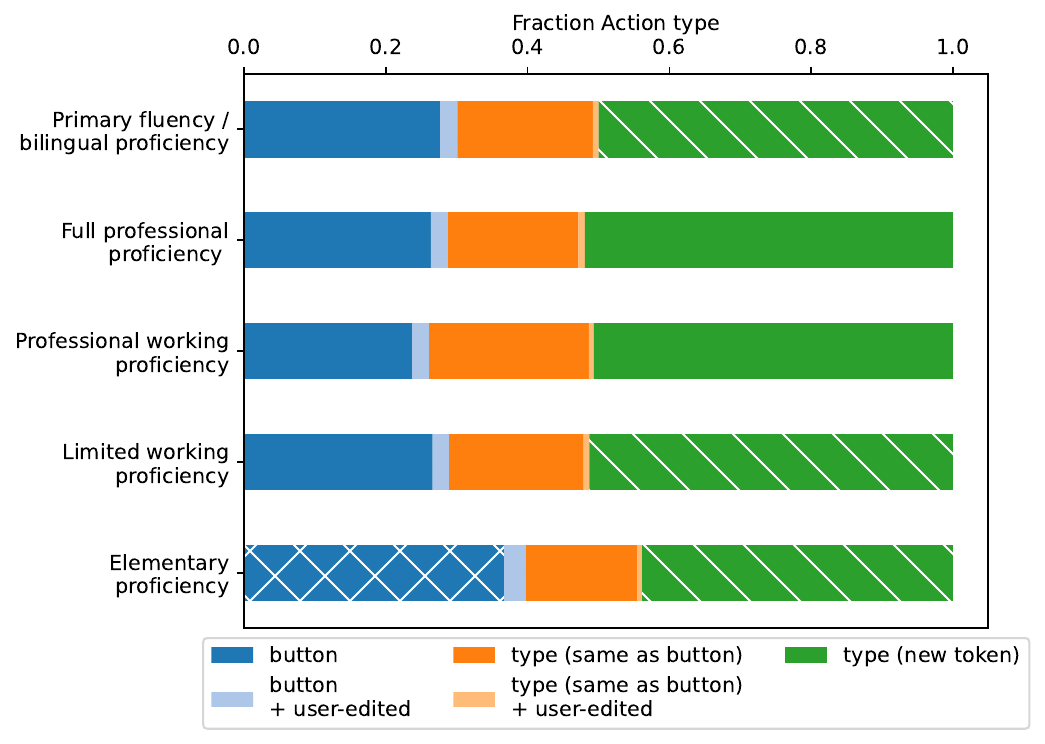}
\caption[]{}
\end{subfigure}%
\hspace{.01\textwidth}
\begin{subfigure}[c]{0.51\textwidth}
\centering
\footnotesize
    \begin{tabular}{l l c l}
    \toprule
    Proficiency levels & $t$ & $\p$ & sig \\
    \midrule
    1 (Elementary) vs & & & \\
    \qquad  2 (Limited working) & $t(17403)=-8.97$ & $0.0000$ & $*$\\
    \qquad  3 (Professional working) & $t(17673)=-8.228$ & $0.0000$ & $*$\\
    \qquad  4 (Full professional) & $t(18053)=-10.37$ & $0.0000$ & $*$\\
    \qquad  5 (Primary fluency/bilingual) & $t(25097)=-8.451$ & $0.0000$ & $*$\\
    2 (Limited working) vs  & & & \\
    \qquad  3 (Professional working) & $t(17494)=0.805$ & $0.6134$ & \\
    \qquad  4 (Full professional) & $t(17874)=-1.223$ & $0.4102$ & \\
    \qquad  5 (Primary fluency/bilingual) & $t(24918)=1.81$ & $0.1851$ & \\
    3 (Professional working) vs & & & \\  
    \qquad  4 (Full professional) & $t(18144)=-2.052$ & $0.1190$ & \\
    \qquad  5 (Primary fluency/bilingual) & $t(25188)=0.905$ & $0.5671$ & \\
    4 (Full professional) vs & & & \\  
    \qquad  5 (Primary fluency/bilingual) & $t(25568)=3.258$ & $0.0066$ & $*$\\
    \bottomrule
    \end{tabular}
\caption[]{}
\end{subfigure}
\caption[]{
AI reliance broken down by self-reported English proficiency. (a) We break down the set of \tokens in stories written by participants based on their source (e.g., the participant chose a model suggestion by pressing a suggestion button). (b) We compare the proportion of model suggested \tokens to human written or edited \tokens in pairs of proficiency levels.}
\label{fig:english}
\Description{Bar chart showing participants’ reliance on model suggestions across English proficiency levels. Participants in the ``elementary proficiency'' category show higher rates of accepting suggestions, either by pressing the suggestion button or by typing the model suggestion exactly. }
\end{figure*}}{}

%% file: Sections/S6.Discussion.tex
\section{Discussion, Limitations, and Implications}

In this work, we examined the effect of biased predictive text suggestions on \hai co-written text. Predictive text is widely used in mobile interfaces like the one examined in this study. These systems are not neutral. The underlying models may produce gender-biased suggestions that reflect or reinforce social stereotypes. When people accept biased suggestions, the resulting co-written texts may perpetuate these stereotypical associations, potentially shaping the beliefs of those who read them and those who wrote them. This is especially concerning for children who are still forming their beliefs about the world~\cite{doi:10.1080/0954025900020204} and for non-native speakers who generally accept more model suggestions~\cite{buschek2021native}. These biases may also create feedback loops: if \hai co-written texts containing gender stereotypes are later used to train future models, even an initially ``unbiased'' system could become increasingly biased over time, a challenge that would not be solved with watermarking because the resulting text \emph{is} human-written. 

Our findings show that people are not equally influenced by pro-stereotypical and anti-stereotypical suggestions. While anti-stereotypical suggestions can, in some contexts, increase the proportion of anti-stereotypical writing, this is often not consistent enough to offset pro-stereotypical human biases. This pattern contrasts with prior research showing that model suggestions can steer writing in multiple directions (e.g., positive vs negative sentiment, arguing social media is good vs bad for society, etc). The contrast between our findings and findings from prior work may be due to the stickiness of stereotypes that people hold, which are often less malleable than overt beliefs about the world~\cite{2001-05123-005, Lai2014-na, 2010-13147-001, annurev:/content/journals/10.1146/annurev-psych-122414-033600, lewandowsky13misinformation}. It may also be that single-word suggestions, as used in our study, have less influence on co-writing outcomes than sentence- or paragraph-level suggestions~\cite{dhillon2024shaping}.

Still, our work demonstrates that mitigating extrinsic bias in the model (or even producing \textit{only} anti-stereotypical suggestions) may not lead to the sociotechnical mitigation of bias in the human-AI outcome. Even when developers create models that generate outputs aligned with certain fairness principles, human preferences and biases can reintroduce inequity through selective uptake of those suggestions. As a result, we should not expect co-writing with a ``perfectly debiased'' predictive text system to yield perfectly unbiased stories. For instance, interpolating between suggestion configurations, we estimate that even a model offering \man and \woman suggestions at equal rates would yield only about $25.5\%$ \woman detectives. These differences in uptake echo prior research showing that users sometimes prefer gender-biased career recommendations that align with their own expectations, even when ``debiased'' alternatives are available~\cite{krause2025debiased, wang2023biasedhuman, wang2022prefer}.

Our findings highlight that technical interventions to reduce model bias may not be sufficient to achieve equitable outcomes when humans and AI work together. The biases in the final co-written texts in our study come not solely from the model but from participants' decisions to accept or override AI suggestions. For design, this suggests that fairness should be treated as a property of the \hai system as a whole, not merely of the AI. One possible avenue for future HCI research is to explore designs that foster more engagement with anti-stereotypical suggestions, encouraging users to reflect on and occasionally challenge their own assumptions. Such systems could draw from existing work on implicit bias mitigation, which has developed techniques to help people recognize and reduce biased beliefs (though these mitigation methods may not have a long-term impact~\cite{2001-05123-005, Lai2014-na, 2010-13147-001}). By connecting technical fairness work with behavioral design strategies, future research could help bridge the gap between algorithmic debiasing and sociotechnical fairness in practice.

Our study has several limitations. Participants were asked to co-author a story that was not entirely their own. As a result, they may have lacked a clear narrative plan, potentially amplifying the influence of model suggestions. Some of the content in the provided story prefixes and controlled model suggestions were more indicative of US-centric cultural norms and biases (e.g., in the choice of character names), which made the tasks less realistic for participants not based in the United States. Moreover, while our study focused on a creative writing scenario, predictive text systems are often used in everyday communication, where the effects of bias may manifest differently. For writing that is more grounded in a real-world experience or interaction, a predictive text system  may affect how an author describes a past appointment with a doctor (e.g., the doctor's disposition or  agency) but is unlikely to influence how one describes the doctor's gender itself. The study further considered only single-word predictive text. While this interaction mode is common in mobile applications, the findings may not generalize to co-writing applications using longer suggestions. In addition, the study focused exclusively on writing in English and on gender and sexuality stereotypes that have been documented among people in the United States~\cite{cao-etal-2022-theory}. Although our sample included some participants who were not native English speakers or based in the United States, their post-survey responses were broadly consistent with the belief patterns reported in prior U.S.-based work. Nonetheless, our results may not generalize to other cultural or linguistic contexts, or to stereotypes concerning other personal characteristics or sensitive attributes.

Overall, our work shows that anti-stereotypical predictive text suggestions have some potential to lessen gender and sexuality biases in \hai co-writing, but these suggestions alone are not enough to encourage users to break out of stereotypical patterns. Pro-stereotypical narratives continue to dominate even under maximally anti-stereotypical system settings. We therefore caution against over-relying on purely technical debiasing as a fairness solution. Instead, we advocate for future HCI and AI design research that considers interventions at the interaction level, supporting users in reflecting on, engaging with, and potentially revising their own beliefs or biases during the act of co-writing. By attending to both model design and human behavior, we may better understand and shape the sociotechnical dynamics that produce bias in \hai collaboration.

%% file: Sections/A0.Overflow_Results.tex
\section{Results on Remaining Scenarios}\label{sec:other_scenario_results}

\subsection{Scenario: Doctor}
In this scenario, the \prefix describes the speaker visiting the doctor. The model suggests that the doctor is of a particular gender and is either ``confident'' or ``unconfident'' (See examples in \autoref{tab:story_egs}). According to \citet{cao-etal-2022-theory}, American annotators view men as comparatively more confident than women, though our participants may not have held this same stereotype (See \autoref{tab:ABC_Cao}).

\subsubsection{Effects on Gender Alone}

Here, we analyze how suggestions affect the doctor's gender (\autoref{tbl:gender}).

\paragraph{At the Story Level}

First, we consider how suggestions affect how participants specify a doctor character's gender alone (\autoref{tbl:gender}). %
At the level of stories, we hypothesized that participants would default to making the doctor \man, leading the no suggestions condition to be similar to the \man suggestions condition. However, we do see marginally more \man doctors when participants are given \man suggestions than no suggestions ($t(230)=2.467$, $\p\approx 0.0521$, $d=0.349$). We believe this may be due to an overall lower rate of specifying the doctor's gender in the no suggestions condition. When the participants choose to mark the doctors gender in the no suggestions condition, they significantly more often mark the doctor as \man than \woman with or without \man suggestions (No suggestions: $t(144)=-3.862$, $\p\approx 0.0012$, $d=-0.639$; \man suggestions: $t(316)=-8.542$, $\p<0.0001$, $d=-0.958$). 

When it comes to \woman suggestions, we see significantly more \woman doctors with \woman suggestions than without any suggestions ($t(246)=3.219$, $\p\approx 0.0081$, $d=0.449$). However, when shown \woman doctor suggestions, the difference between the rate of making the doctor character \man vs \woman is not significantly different ($t(348)=-1.026$, $\p\approx 0.5207$, $d=-0.11$).

These results provide some evidence of H1 (namely that gender rates without suggestions are more similar to the rates with \man doctor suggestions than \woman doctor suggestions). 
They imply that if a ``debiased'' predictive text system presented users with \man and \woman doctor suggestions at equal rates, we'd still expect to see more \man doctors in stories than \woman as the \woman doctor suggestions are rejected and overwritten more than the \man suggestions. %

\paragraph{At the \ttoken Level}

Considering overall reliance, we find that participants type new \tokens or edit a model-suggested \token marginally significantly more in conditions where the model is prompted to make the doctor \woman (H2: $t(7692)=2.258$, $\p\approx 0.0784$, $d=0.052$). When we only consider the story \tokens that specify the doctor character's gender (See example \tokens in \autoref{tab:tokens}), we now see a significantly higher rate of overrides or edits in the \woman doctor conditions (H2a: $t(159)=2.734$, $\p\approx 0.0294$, $d=0.432$).
Similarly, when we consider only model suggestions that would specify the doctor character's gender, we see a significantly higher rejection rate for \tokens that would make the doctor \woman (H2b: $t(1014)=2.926$, $\p\approx 0.0172$, $d=0.184$).

\subsubsection{Effects on Gender and Confidence}

\begin{figure*}[tb]
    \begin{subfigure}[c]{.45\linewidth}
            \centering
            \includegraphics[width=\textwidth,clip,trim=0 0 0 0]{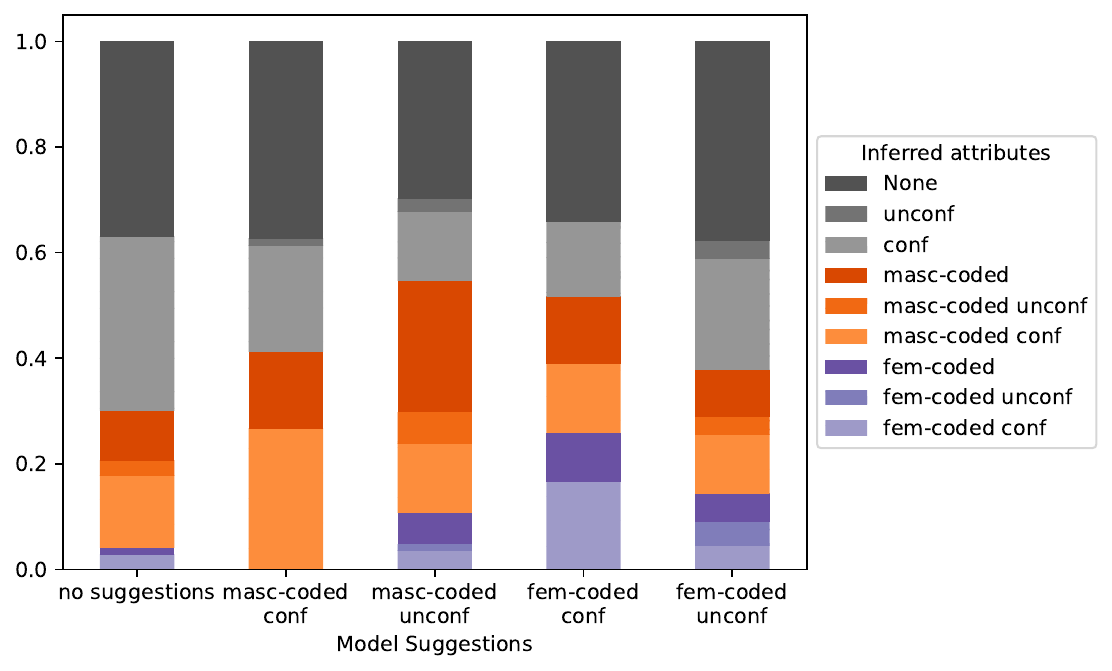}
            \caption[]{The colors show inferred gender, and the patterns show the inferred confidence. The ticks group stories by suggestions presence/type}
    \end{subfigure}
    \hspace{.01\linewidth}
    \begin{subfigure}[t]{.53\linewidth}
            \centering
            \footnotesize
            \begin{tabular}{ l l l c l}
            \toprule
            Suggests Compared & Measured Attr & $t$ &$\p$ & sig\\
            \midrule
             &  None & $t(161)=0.103$ & $0.9609$ & \\
             &  unconfident & $t(161)=1.577$ & $0.2601$ & \\
             &  confident & $t(161)=-1.7$ & $0.2191$ & \\
             &  M unconfident & $t(161)=0.217$ & $0.9088$ & \\
             &  M confident & $t(161)=-0.498$ & $0.7882$ & \\
             &  M & $t(161)=-0.153$ & $0.9479$ & \\
             &  F unconfident & $t(161)=1.831$ & $0.1826$ & \\
             &  F confident & $t(161)=0.572$ & $0.7447$ & \\
            \multirow{-9}{*}{F unconfident} &  F & $t(161)=1.411$ & $0.3247$ & \\
            \rowcolor{customrowcolor} \leavevmode   &  None & $t(156)=-0.374$ & $0.8283$ & \\
            \rowcolor{customrowcolor} \leavevmode   &  confident & $t(156)=-2.857$ & $0.0210$ & $*$\\
            \rowcolor{customrowcolor} \leavevmode   &  M confident & $t(156)=-0.139$ & $0.9488$ & \\
            \rowcolor{customrowcolor} \leavevmode   &  M & $t(156)=0.658$ & $0.7019$ & \\
            \rowcolor{customrowcolor} \leavevmode   &  F confident & $t(156)=2.91$ & $0.0195$ & $*$\\
            \rowcolor{customrowcolor} \leavevmode  \multirow{-6}{*}{F confident} &  F & $t(156)=2.194$ & $0.0955$ & \\
             &  None & $t(155)=-0.956$ & $0.5495$ & \\
             &  unconfident & $t(155)=1.326$ & $0.3589$ & \\
             &  confident & $t(155)=-3.038$ & $0.0145$ & $*$\\
             &  M unconfident & $t(155)=0.97$ & $0.5483$ & \\
             &  M confident & $t(155)=-0.11$ & $0.9601$ & \\
             &  M & $t(155)=2.552$ & $0.0446$ & $*$\\
             &  F unconfident & $t(155)=0.932$ & $0.5574$ & \\
             &  F confident & $t(155)=0.294$ & $0.8721$ & \\
            \multirow{-9}{*}{M unconfident} &  F & $t(155)=1.495$ & $0.2874$ & \\
            \rowcolor{customrowcolor} \leavevmode   &  None & $t(146)=0.043$ & $0.9735$ & \\
            \rowcolor{customrowcolor} \leavevmode   &  unconfident & $t(146)=0.986$ & $0.5407$ & \\
            \rowcolor{customrowcolor} \leavevmode   &  confident & $t(146)=-1.785$ & $0.1952$ & \\
            \rowcolor{customrowcolor} \leavevmode   &  M confident & $t(146)=1.975$ & $0.1402$ & \\
            \rowcolor{customrowcolor} \leavevmode  \multirow{-5}{*}{M confident} &  M & $t(146)=0.941$ & $0.5542$ & \\
            \bottomrule
            \end{tabular}
            \caption[]{Comparison between rates of attributes being present in stories written with a given kind of suggestions vs with no suggestions. Attribute-suggestion pairs with no entries are not included (e.g., There were no \woman doctors written with confident \man suggestions.)}
    \end{subfigure}
    \caption[]{Joint inferred doctor gender and confidence. }\label{fig:0_joint}
    \Description{Left: Stacked bar charts showing doctor gender and confidence in stories under different suggestion conditions. Right: results of t-tests comparing the presence of these attributes with ad without suggestions. Confident suggestions of either gender are associated with higher proportions of gender-specified confident doctors. Unconfident masculine-coded suggestions correspond to lower proportions of gender-unspecified confident doctors and higher proportions of confidence-unspecified masculine-coded doctors.}
\end{figure*}

Beyond just considering the doctor's gender, the model is also prompted to suggest the doctor's level of confidence. For each suggestions condition and each potential set of attributes that could be given to the doctor character, we consider whether adding that suggestion type changes the proportion of doctors that have that set of attributes (\autoref{fig:0_joint}). 

We see that when the model provides unconfident \man doctor suggestions, participants wrote significantly more \man doctors of unspecified confidence ($t(155)=2.552$, $\p\approx 0.0446$, $d=0.408$) and significantly fewer confident doctors of unspecified gender ($t(155)=-3.038$, $\p\approx 0.0145$, $d=-0.486$). No other set of attributes changed significantly in under these suggestions. One possible interpretation of these results is that participants take on the suggestion of the doctor being \man, but largely refuse to make a \man doctor unconfident and instead leave confidence unspecified. 

We also see that when the model provides confident \woman doctor suggestions, participants wrote significantly more confident \woman doctors ($t(156)=2.91$, $\p\approx 0.0195$, $d=0.464$) and significantly fewer confident doctors with gender unspecified, marginally more \woman doctors of unspecified confidence ($t(156)=2.194$, $\p\approx 0.0955$, $d=0.35$), and significantly fewer confident doctors with unspecified gender ($t(156)=-2.857$, $\p\approx 0.0210$, $d=-0.456$). 
For confident \man doctor suggestions, we see no significant changes, but note similar trends away from confident doctors of unspecified gender ($t(146)=-1.785$, $\p\approx 0.1952$, $d=-0.293$) and toward confident \man doctors ($t(146)=1.975$, $\p\approx 0.1402$, $d=0.325$). These results show that gendered confident suggestions are generally effective at shifting stories away from non-gendered confident doctors. 

Overall, we note how regardless of suggestion type, we always continue to see a sizable group of stories about confident or unspecified \man doctors, even when the opposite is suggested. On the other hand we see no \woman doctors when suggesting a confident \man doctor and only see \woman doctors in a \man doctor condition when the doctor is suggested to be unconfident. This again aligns with \citet{cao-etal-2022-theory}'s findings about gender-confidence stereotypes.

\subsection{Scenario: Student}

In this scenario, the \prefix includes a team member with a traditionally feminine (``Abby'') or masculine (``John'') name. The model then suggests that this character is ``competitive'' or ``unassertive'' (See examples in \autoref{tab:story_egs}). According to \citet{cao-etal-2022-theory}, American annotators view men as comparatively more competitive and women as comparatively more unassertive though our participants may not have held this same stereotype (See \autoref{tab:ABC_Cao}).

\subsubsection{Effects on Competitiveness Disaggregated by Gender}

Here, we consider how suggestions affect the competitiveness stance of ``Abby'' and ``John'' (\autoref{tbl:gender_and_other}).

\paragraph{At the Story Level}

When we consider competitiveness rates in overall stories, we see that suggestions often increase the respective rates. We see significantly more unassertive \man ($t(117)=-3.024$, $\p\approx 0.0157$, $d=-0.608$), unassertive \woman ($t(116)=-2.874$, $\p\approx 0.0210$, $d=-0.562$), and competitive \woman students ($t(125)=-3.464$, $\p\approx 0.0044$, $d=-0.666$) when suggested when comparing to no suggestions conditions. However, due to a higher baseline rate of competitive \man students in the no suggestions conditions, we do not see a significant effect when providing competitive \man suggestions ($t(121)=-1.297$, $\p\approx 0.3747$, $d=-0.259$). This provides some evidence toward H1 as the pro-stereotypical competitive \man condition is similar to behavior without suggestions.

We also see that generally providing suggestions increases competitiveness vs unassertiveness rates within a gender. For example, providing competitive \man suggestions significantly increases the rate of competitive \man students over unassertive \man students ($t(174)=-3.898$, $\p\approx 0.0010$, $d=-0.588$). We see similar significant trends for unassertive \man students ($t(166)=2.872$, $\p\approx 0.0207$, $d=0.443$) and competitive \woman students ($t(174)=-8.131$, $\p<0.0001$, $d=-1.226$). However, we see no significant difference with unassertive \woman suggestions ($t(156)=0.883$, $\p\approx 0.5725$, $d=0.141$). This provides some evidence against H1, implying that participants may be more resistant to unassertive \woman student suggestions despite existing stereotypes of gender and competitiveness. 

\paragraph{At the \ttoken Level}

When we consider the overall rate of reliance on model suggestions, we see that participants typed new \tokens or edited model suggestions significantly more in the unassertive \woman student condition than the unassertive \man student condition ($t(3615)=4.453$, $\p\approx 0.0001$, $d=0.148$) and similarity for the competitive \man student condition over the unassertive \man student condition ($t(4135)=-6.49$, $\p<0.0001$, $d=-0.202$). These results do not support H2, but we see that none of these trends are significantly present when we constrain to \tokens that determine the character's competitiveness (H2a). When we consider model suggestions that affect competitiveness, we see marginally more acceptance of competitive \woman suggestions than unassertive (H2b; $t(426)=-2.159$, $\p\approx 0.0999$, $d=-0.21$).

Overall, the results in this scenario were mixed. We find a relatively low overall rate of clearly specifying the given student as competitive or unassertive which may have skewed the results. This may be due to poor scenario design where participants decided to focus on topics other than leadership or competitiveness, or it could be that in the classroom settings, participants held weaker internal stereotypes about gender and competitiveness than \citet{cao-etal-2022-theory}'s held about more general settings. Indeed (as we show in \autoref{tab:ABC_Cao}) our post-survey results indicate that our participant pool may include more people who do not hold this stereotypical association between masculinity and competence (or, by extension, competitiveness).

\subsection{Scenario: Teachers}
In this scenario, participants write about a teacher with a given gender where the model attempts to suggest the teacher's personality as ``likable'' or ``repellent'' (See examples in \autoref{tab:story_egs}). According to \citet{cao-etal-2022-theory}, American annotators view men as comparatively less likable than women, and our participants indicated a significantly stronger association with straight women and warmth (which is associated with likability) than straight men (See \autoref{tab:ABC_Cao}).

\subsubsection{Effects on Likability Disaggregated by Gender}

Here, we discuss how suggestions affect the likability stance of ``Mrs. Brown'' and ``Mr. Brown'' (\autoref{tbl:gender_and_other}).

\paragraph{At the Story Level}

Considering overall stories, we see that regardless of the teacher's presumed gender, participants made the teacher likable significantly more often than repellent when not given suggestions (\woman teacher: $t(68)=-4.747$, $\p\approx 0.0001$, $d=-1.135$; \man teacher:  $t(74)=-3.195$, $\p\approx 0.0110$, $d=-0.733$). These trends continue to be significant regardless of suggestion type, including repellent suggestions (likable \woman: $t(162)=-12.726$, $\p<0.0001$, $d=-1.987$; repellent \woman: $t(150)=-3.886$, $\p\approx 0.0011$, $d=-0.63$; likable \man: $t(184)=-8.86$, $\p<0.0001$, $d=-1.299$; repellent \man: $t(166)=-3.774$, $\p\approx 0.0016$, $d=-0.582$). In other words, participants preferred to make the teacher likable, regardless of the presence or type of suggestions and regardless of the teacher's gender as cued in the story prefix. 

We also find that ``Mr. Brown'' is written as likable marginally more often with likable suggestions than without suggestions ($t(129)=-2.489$, $\p\approx 0.0516$, $d=-0.479$). However, perhaps due to a higher base rate of ``Mrs. Brown'' being written as likable, we see no such increase in likability with added likable suggestions for ``Mrs. Brown'' ($t(115)=-1.403$, $\p\approx 0.3291$, $d=-0.283$). In other words participants may have a stronger default preference for ``Mrs. Brown'' being likable, leading to a more limited effect of likable suggestions. This is in line with \citet{cao-etal-2022-theory}'s and our findings about gender-likability stereotypes in humans. And these findings provide some support for H1 in that the proportion of likable ``Mrs. Brown''s (pro-stereotypical) with no suggestions is more similar to the proportion of likable ``Mrs. Brown''s with likable suggestions than the the proportion of likable ``Mr. Brown''s (anti-stereotypical) with no suggestions is to the proportion of likable ``Mr. Brown''s with likable suggestions.

\paragraph{At the \ttoken Level}

When we consider the rate of acceptances of model suggestions, we see significantly less acceptance of model suggestions in the condition where ``Mrs. Brown'' is suggested to be repellent over likable ($t(3591)=2.618$, $\p\approx 0.0351$, $d=0.088$), while we see a significant effect in the opposite direction for ``Mr. Brown'' ($t(3892)=-3.162$, $\p\approx 0.0086$, $d=-0.102$). We also see significantly less acceptance in the ``Mr. Brown'' is likable condition than for ``Mrs. Brown'' ($t(4013)=-5.922$, $\p<0.0001$, $d=-0.187$). These results support H2, as we can see that more pro-stereotypical conditions (e.g., suggesting a \woman character is likable) lead to more acceptance of suggestions. 

However, the effects are quite different when we only consider \tokens that determine likability. Here we see a trend of participants accepting more ``likable'' suggestions over ``repellent'' for either gender, though the effect is only significant for ``Mr. Brown'' (\man: $t(179)=2.673$, $\p\approx 0.0331$, $d=0.398$; \woman: $t(157)=2.068$, $\p\approx 0.1190$, $d=0.328$). Under H2a, we would have expected to see less acceptance of ``likable'' suggestions for the \man ``Mr. Brown''.
When we consider the acceptance rate of model suggestions, we see higher rates of acceptance of ``likable'' suggestions for either gender, but in this case, it is only significant for ``Mrs. Brown'' (\man: $t(438)=-2.446$, $\p\approx 0.0533$, $d=-0.233$, $\p\approx 0.0331$; \woman: $t(383)=-4.106$, $\p\approx 0.0004$, $d=-0.42$), supporting H2b. 

These results show that participants may have preferred suggestions of teachers of any gender being likable. This seems reasonable as at the story-level likable teachers were generally preferred even without suggestions. While these results do not match our overall hypotheses about reliance under pairs of gender and likability, they may suggest that participants' stereotypes about teachers being likable people were stronger than their stereotypes about people of different genders being likable.

\subsection{Scenario: Town Hall}

In this scenario, the \prefix includes a town hall participant with a traditionally feminine (``Rebecca'') or masculine (``Thomas'') name. The town hall is about an affordable housing development, and the model suggests that the character has a conservative or liberal viewpoint on this issue  (See examples in \autoref{tab:story_egs}). According to \citet{cao-etal-2022-theory}, American annotators view men as comparatively more conservative than women, an association echoed by our participants in our post-survey (See \autoref{tab:ABC_Cao}).

\subsubsection{Effects on Political Stance Disaggregated by Gender}

Here, we analyze how suggestions affect the political stance of ``Rebecca'' and ``Thomas'' (\autoref{tbl:gender_and_other}).

\paragraph{At the Story Level}

At the story-level, we first compare the no suggestions conditions to their corresponding liberal and conservative suggestions conditions. We see that adding conservative suggestions decreases the number of liberal characters. This trend is significant for ``Thomas'' ($t(123)=2.682$, $\p\approx 0.0332$, $d=0.514$) and marginally significant for ``Rebecca'' ($t(112)=2.337$, $\p\approx 0.0707$, $d=0.478$). We also see a marginal trend of ``Thomas'' being made conservative more often with conservative suggestions than without suggestions ($t(123)=-2.334$, $\p\approx 0.0707$, $d=-0.447$), with no such trend in the same setting for ``Rebecca'' ($t(112)=-0.984$, $\p\approx 0.5407$, $d=-0.201$). We see here that suggestions tend to successfully encourage participants to make characters liberal or conservative, but they are less successful in making ``Rebecca'' conservative, perhaps suggesting that participants have a harder time accepting suggestions of a \woman character being conservative. 

We also compare rates of making characters liberal vs conservative within suggestion types. Without suggestions, we see no significant difference between making characters of any gender liberal or conservative (Rebecca: $t(66)=1.4$, $\p\approx 0.3334$, $d=0.340$; Thomas: $t(78)=1.686$, $\p\approx 0.2230$, $d=0.377$). We see ``Thomas'' is made liberal or conservative significantly more often depending on the direction of suggestions (conservative: $t(168)=-3.728$, $\p\approx 0.0018$, $d=-0.571$; liberal: $t(166)=3.315$, $\p\approx 0.0066$, $d=0.511$). For ``Rebecca'', we see significantly more liberal stories when they are suggested ($t(168)=4.575$, $\p< 0.0001$, $d=0.701$), but the increase in conservative stories when they are suggested is not significant ($t(158)=-1.988$, $\p\approx 0.1364$, $d=-0.314$). This again shows that participants may have a harder time accepting suggestions of a \woman character being conservative, which is in agreement with \cite{cao-etal-2022-theory}'s and our findings about human perceptions of the political stance of women and supporting H1.  

\paragraph{At the \ttoken Level}

We generally don't see significant trends at the \token level. We do see that participants accepted suggestions significantly more often in the condition where ``Rebecca'' is suggested to be liberal over ``Thomas'' (H2b; $t(3570)=-2.725$, $\p\approx 0.0276$, $d=-0.091$). This begins to suggest that participants are more comfortable with \woman characters being written as liberal than \man ones, but this trend is not significant when we consider only \tokens that specify the character's stance.

\begin{table*}[tbh]
    \centering
    \footnotesize
    \begin{tabular}{c r r r r}
       \toprule
       {Scenario} & {\ttoken Reliance} &  \begin{tabular}{c}Gender-Defining \ttoken \\ Reliance \end{tabular}  &  \begin{tabular}{c}Gender-Defining \ttoken \\ Rejection Rate\end{tabular}  &  \begin{tabular}{c}Story-Level \\Gender  Freq.\end{tabular} \\
        \midrule
        \textsc{Doctor} & \raisebox{-.5\height}{\includegraphics[height=.85in]{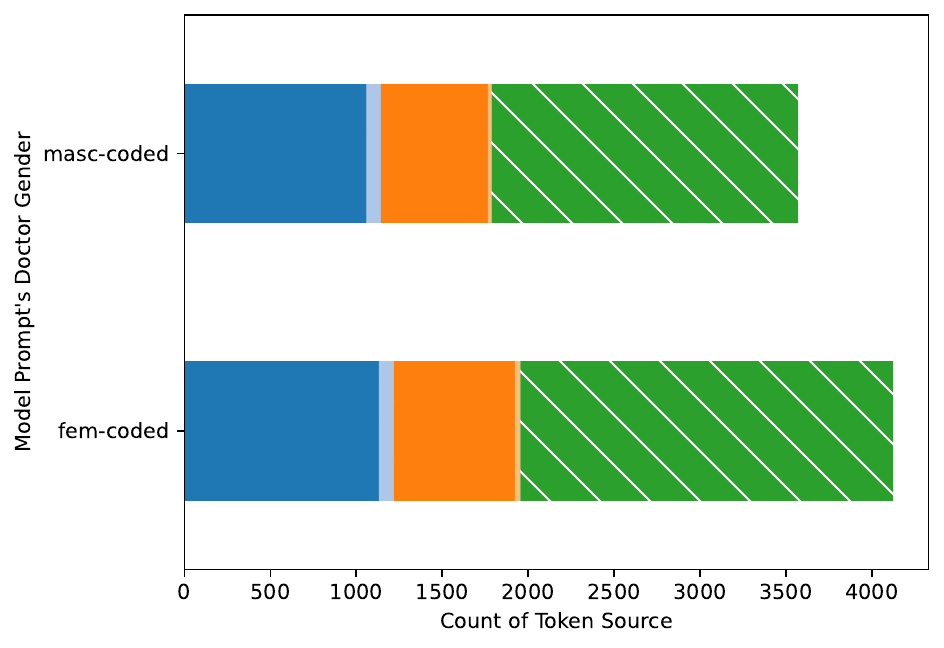}} & \raisebox{-.5\height}{\includegraphics[height=.85in]{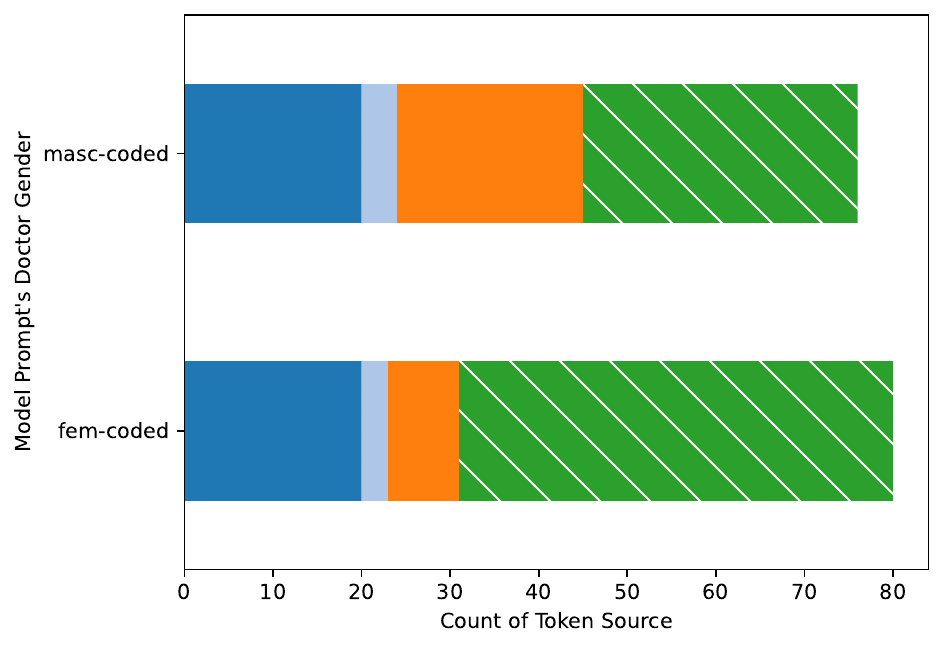}} & \raisebox{-.5\height}{\includegraphics[height=.85in]{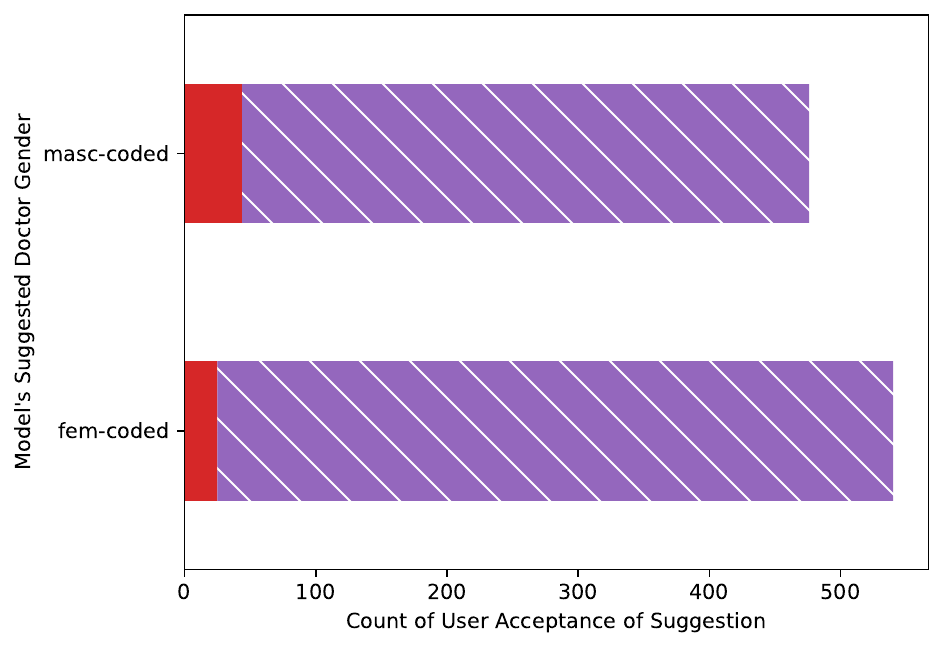}} & \raisebox{-.5\height}{\includegraphics[height=.85in]{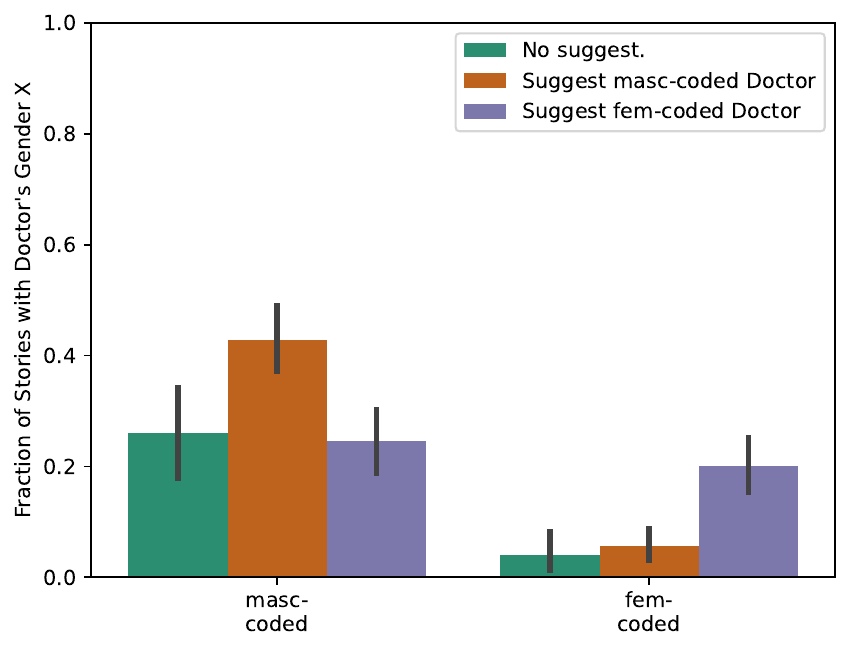}} \\
        \rowcolor{customrowcolor} \leavevmode \cellcolor{customrowcolor}  \textcolor{black}{\textsc{Detectives}} & \raisebox{-.5\height}{\includegraphics[height=.85in]{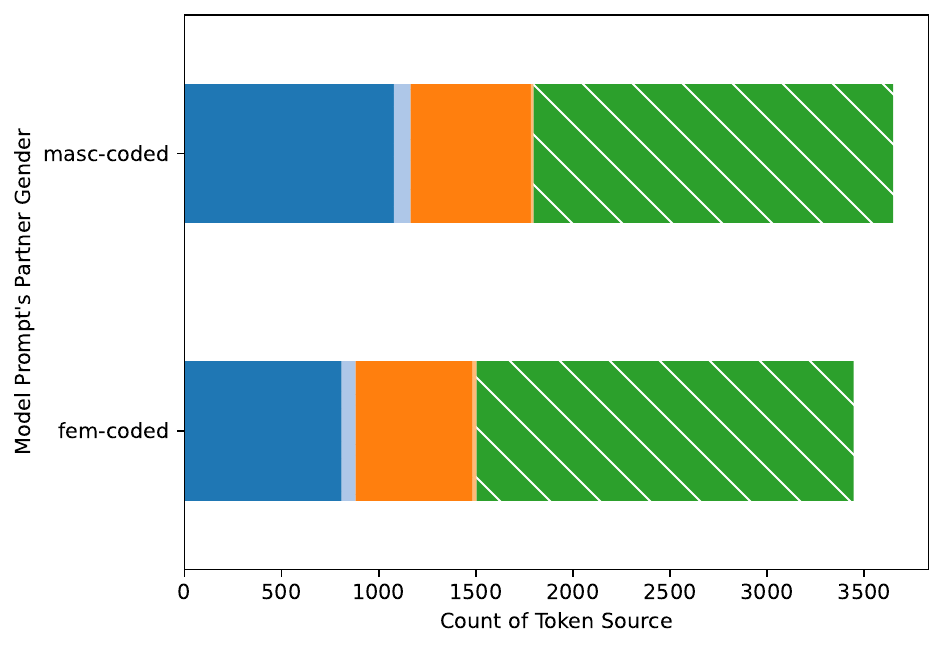}} & \raisebox{-.5\height}{\includegraphics[height=.85in]{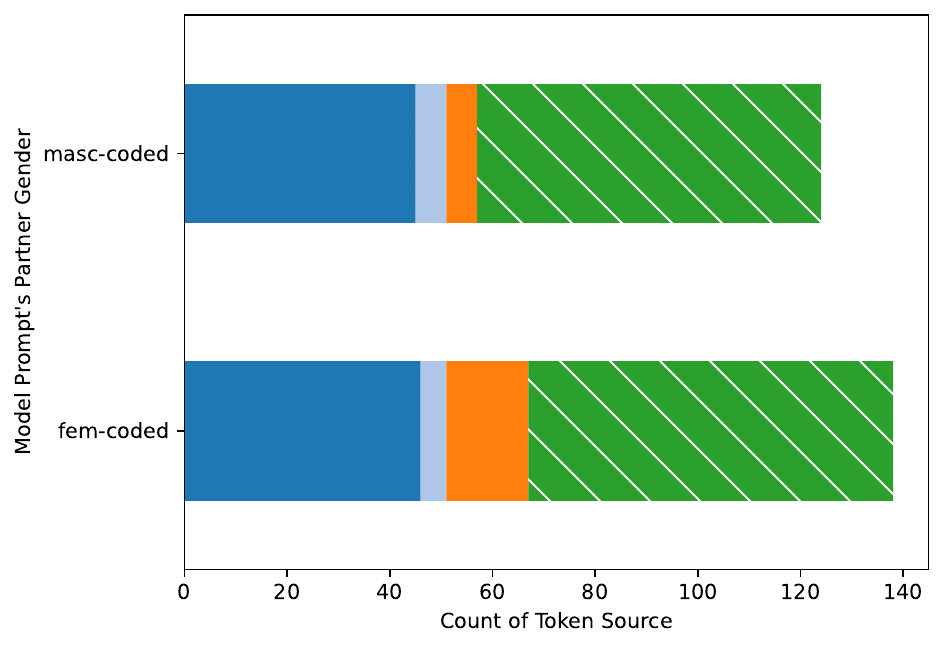}} & \raisebox{-.5\height}{\includegraphics[height=.85in]{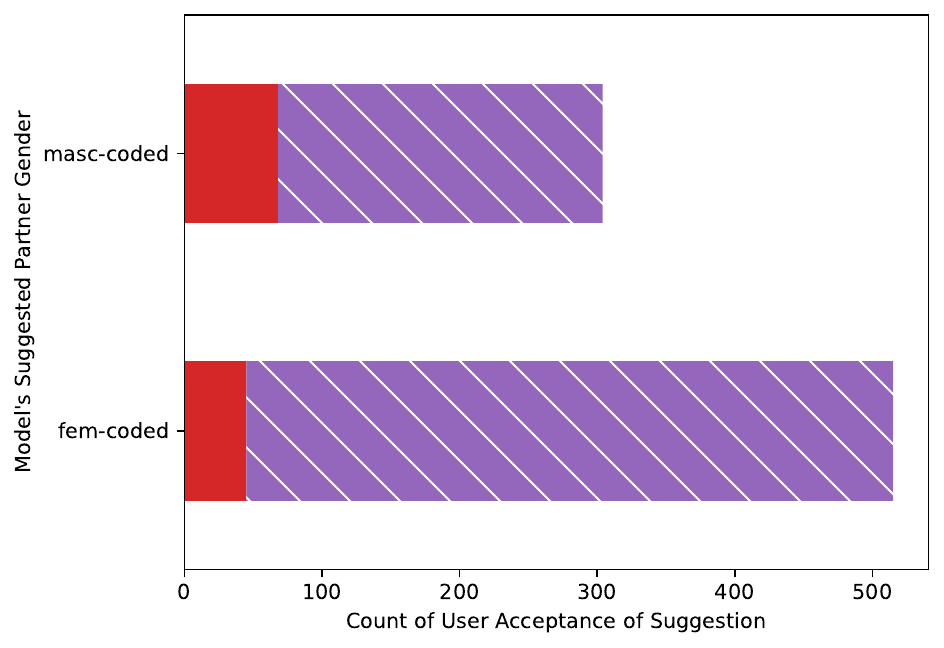}} & \raisebox{-.5\height}{\includegraphics[height=.85in]{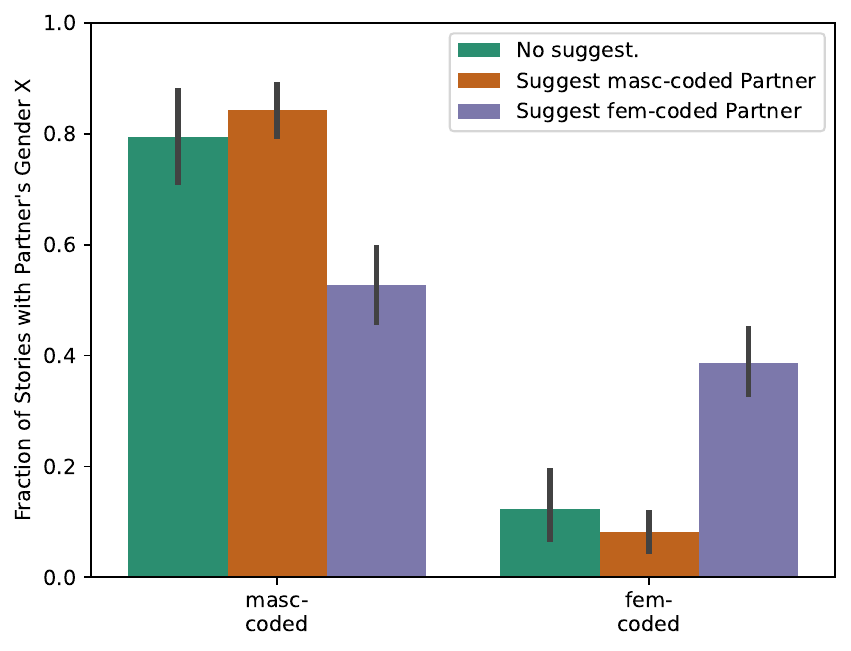}} \\
        \textsc{President} & \raisebox{-.5\height}{\includegraphics[height=.85in]{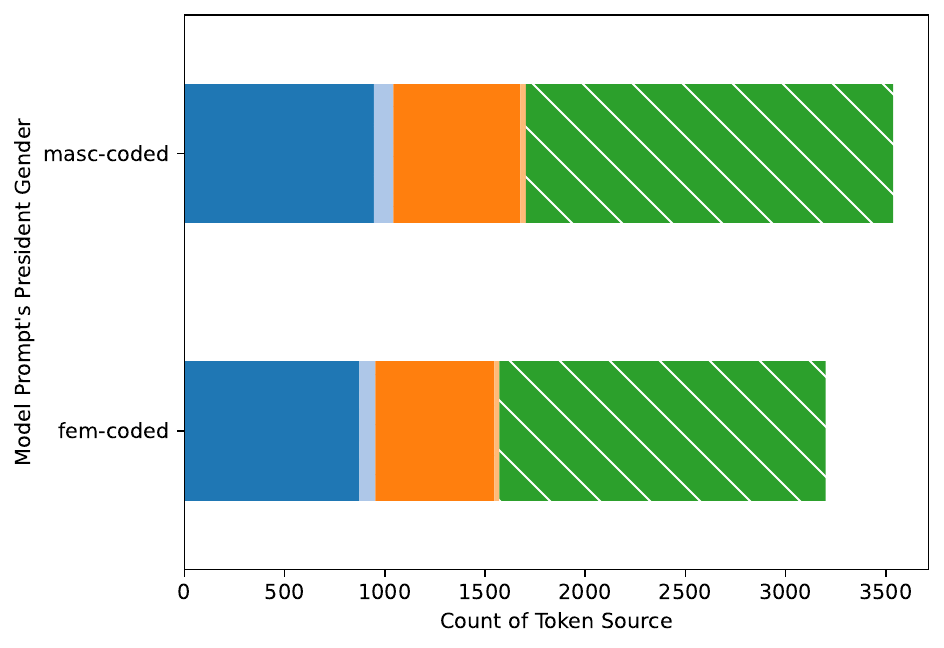}} & \raisebox{-.5\height}{\includegraphics[height=.85in]{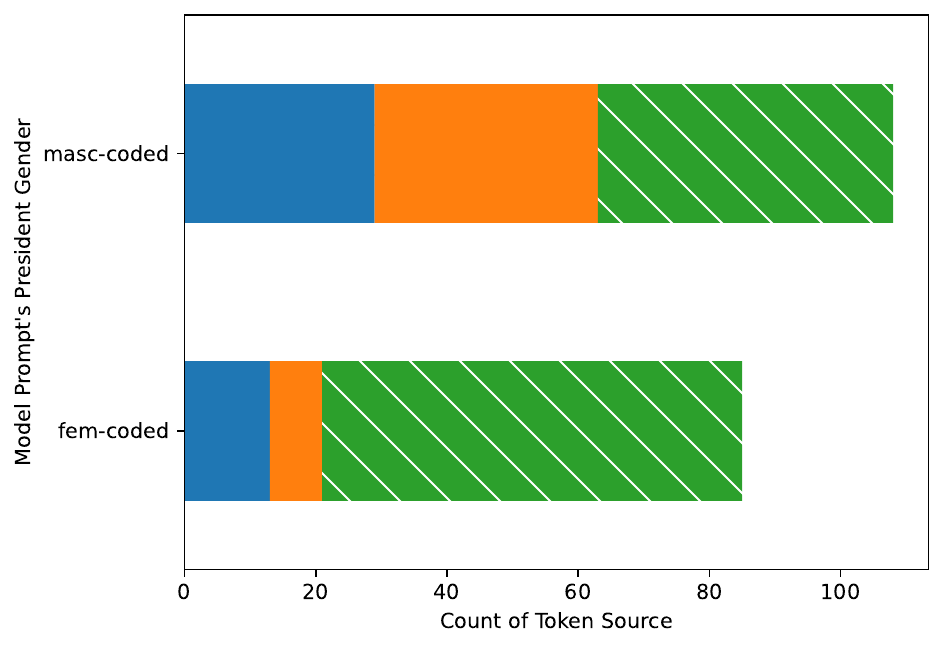}} & \raisebox{-.5\height}{\includegraphics[height=.85in]{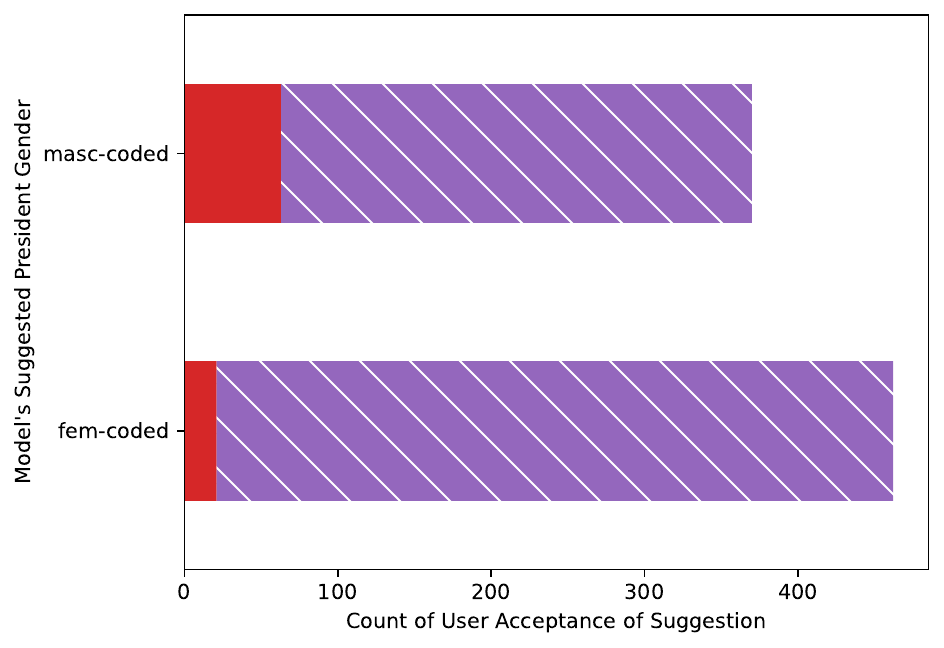}} & \raisebox{-.5\height}{\includegraphics[height=.85in]{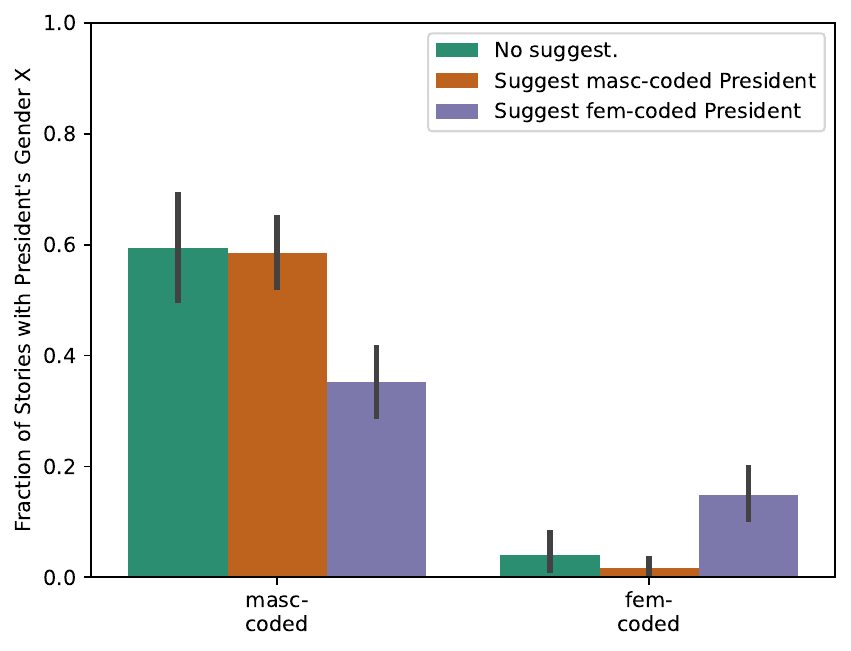}} \\
        & \multicolumn{2}{c}{\includegraphics[height=7.5pt, clip,trim=0 0 0 15]{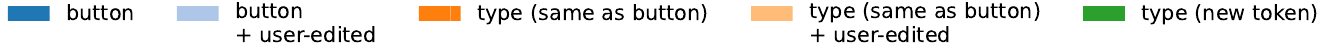}} & \multicolumn{1}{c}{\includegraphics[height=7.5pt, clip,trim=0 0 0 15]{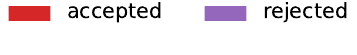}} & \\
        \bottomrule
    \end{tabular}
    \caption[]{Acceptance of Gender suggestions (without considering secondary axes) }
    \label{tbl:gender}
\end{table*}

\aptLtoX[graphic=no,type=html]{
\begin{table*}[tbh]
    \centering
    \footnotesize
    \begin{tabular}{Q{.9cm} r r r r}
       \toprule
        Scenario & \multicolumn{1}{c}{\ttoken Reliance} &\begin{tabular}{c}Axis-Defining \ttoken\\Reliance\end{tabular} & \begin{tabular}{c}Axis-Defining \ttoken \\Rejection Rate\end{tabular} & \begin{tabular}{c}Story-Level \\Axis  Freq.\end{tabular} \\
        &&&&\\
        \midrule
        \textsc{Detectives} & \raisebox{-.5\height}{\includegraphics[height=.76in]{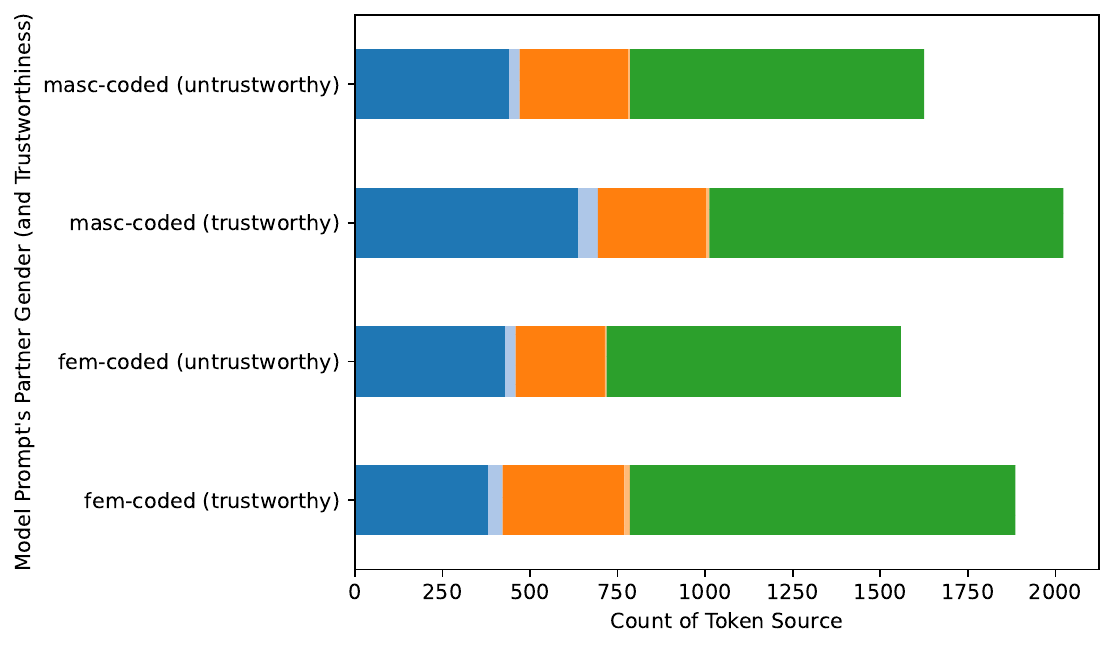}} & \raisebox{-.5\height}{\includegraphics[height=.76in]{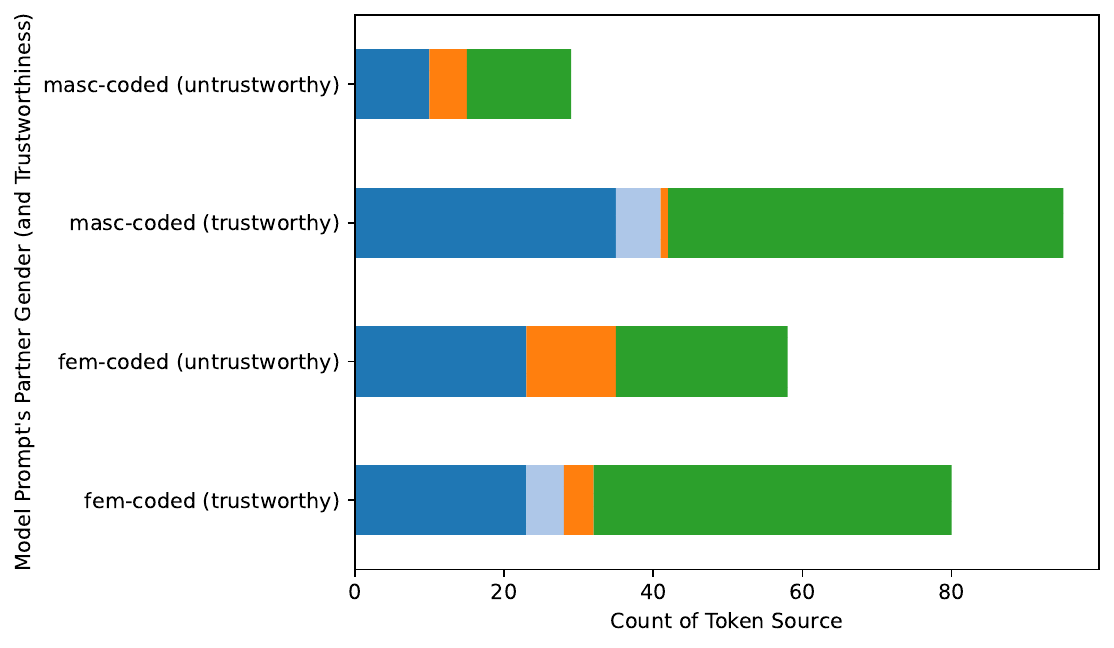}} &  \raisebox{-.5\height}{\includegraphics[height=.76in]{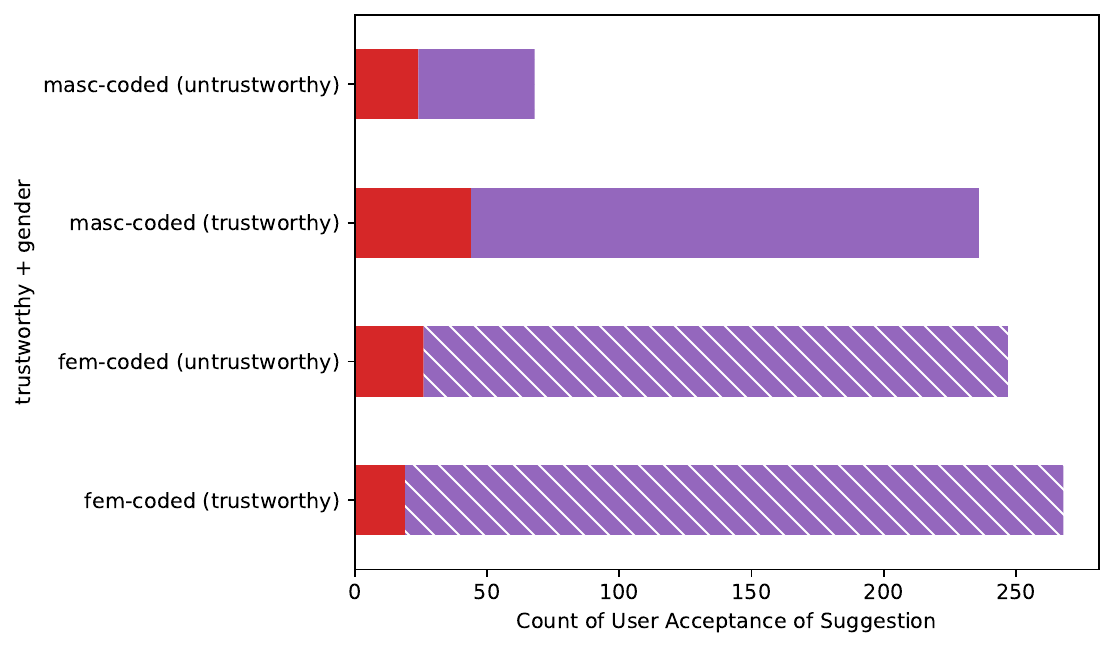}} & \raisebox{-.5\height}{\includegraphics[height=.76in]{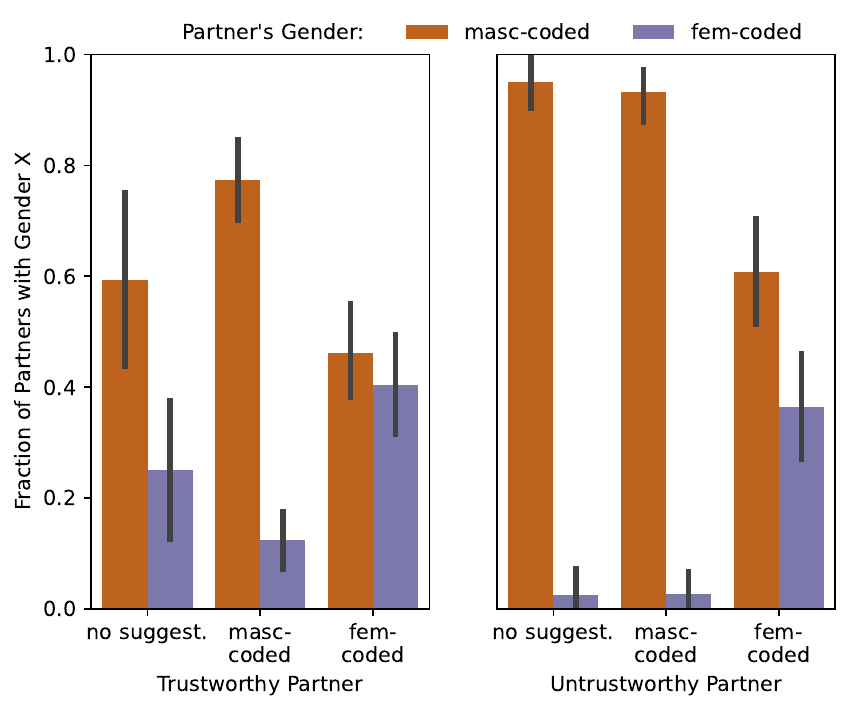}} \\
        \rowcolor{customrowcolor} \leavevmode  \cellcolor{customrowcolor} \textcolor{black}{\textsc{Wedding}} &\cellcolor{customrowcolor}  \raisebox{-.5\height}{\includegraphics[height=.76in]{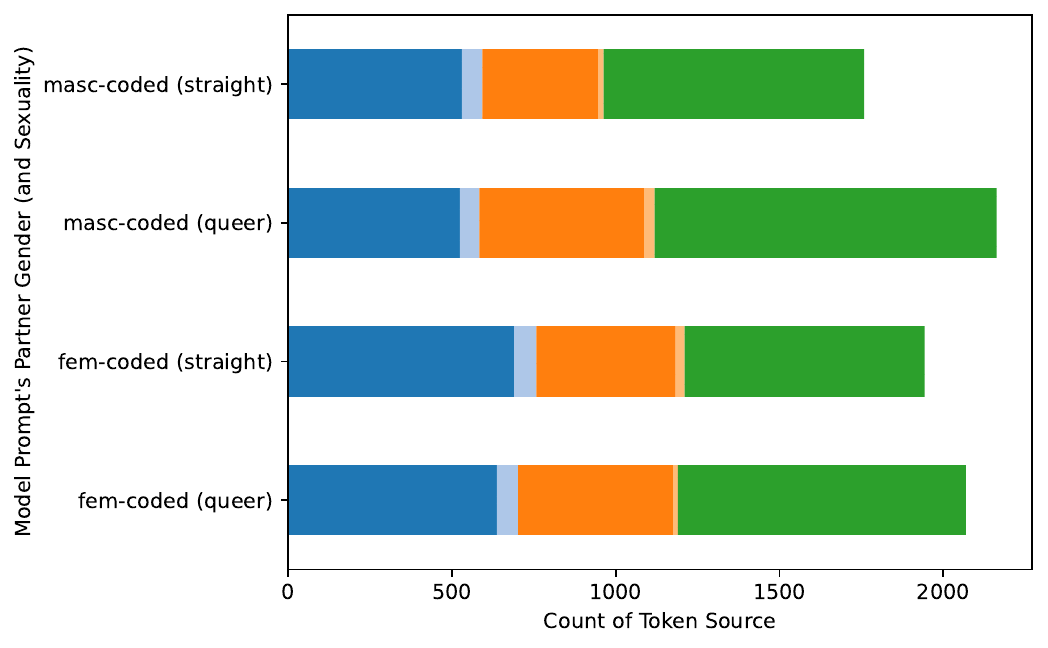}} &\cellcolor{customrowcolor}  \raisebox{-.5\height}{\includegraphics[height=.76in]{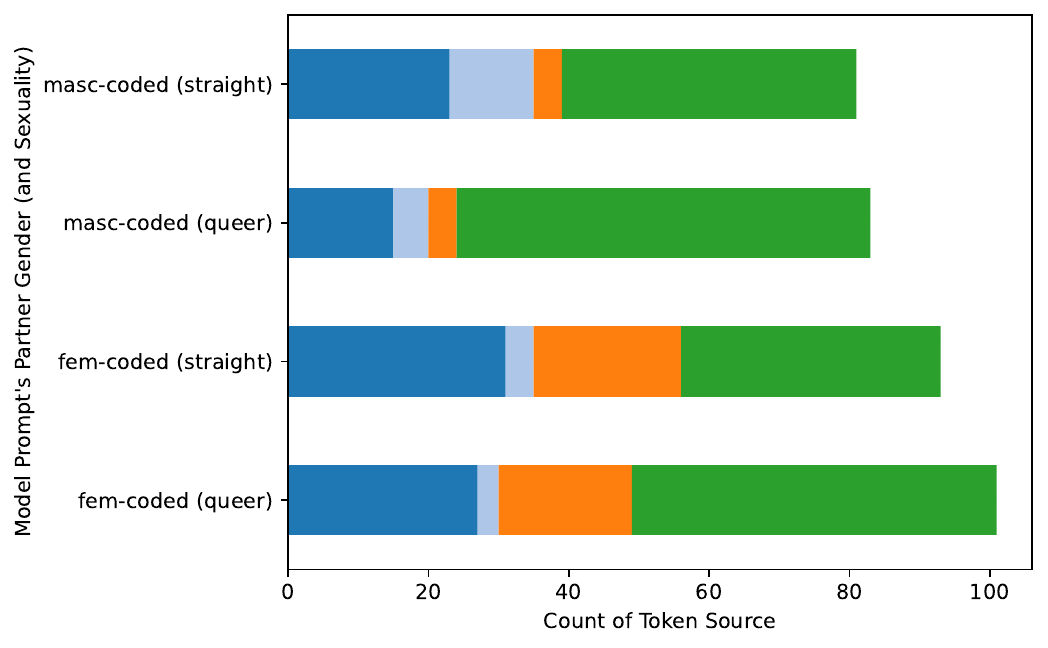}} &\cellcolor{customrowcolor}  \raisebox{-.5\height}{\includegraphics[height=.76in]{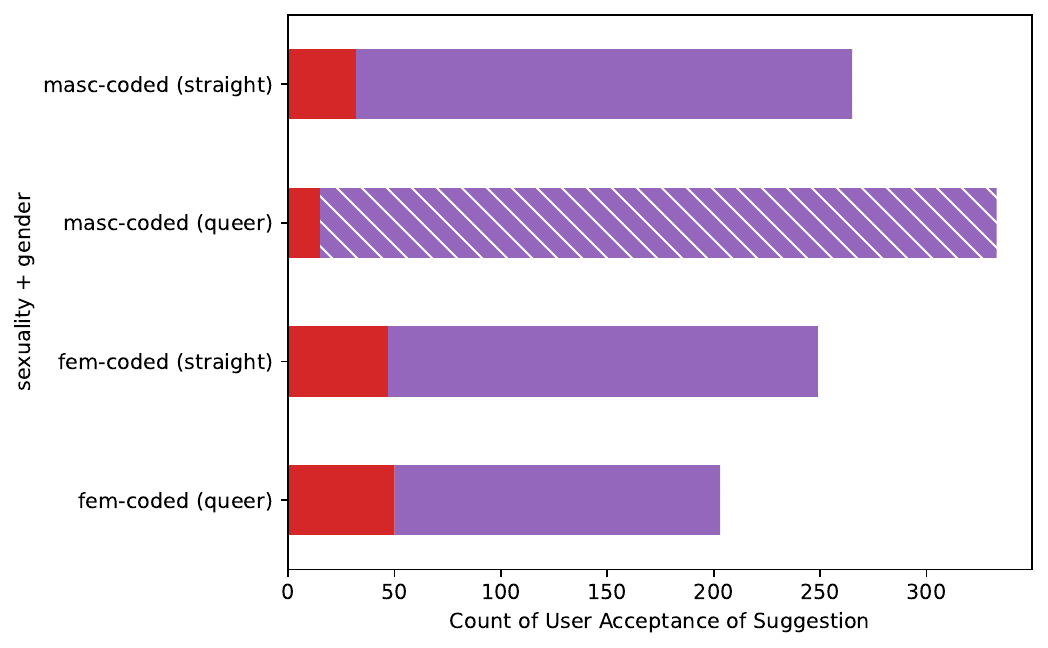}} &\cellcolor{customrowcolor}  \raisebox{-.5\height}{\includegraphics[height=.76in]{figures/1_gender.pdf}}   \\
        \textsc{Student}  & \raisebox{-.5\height}{\includegraphics[height=.76in]{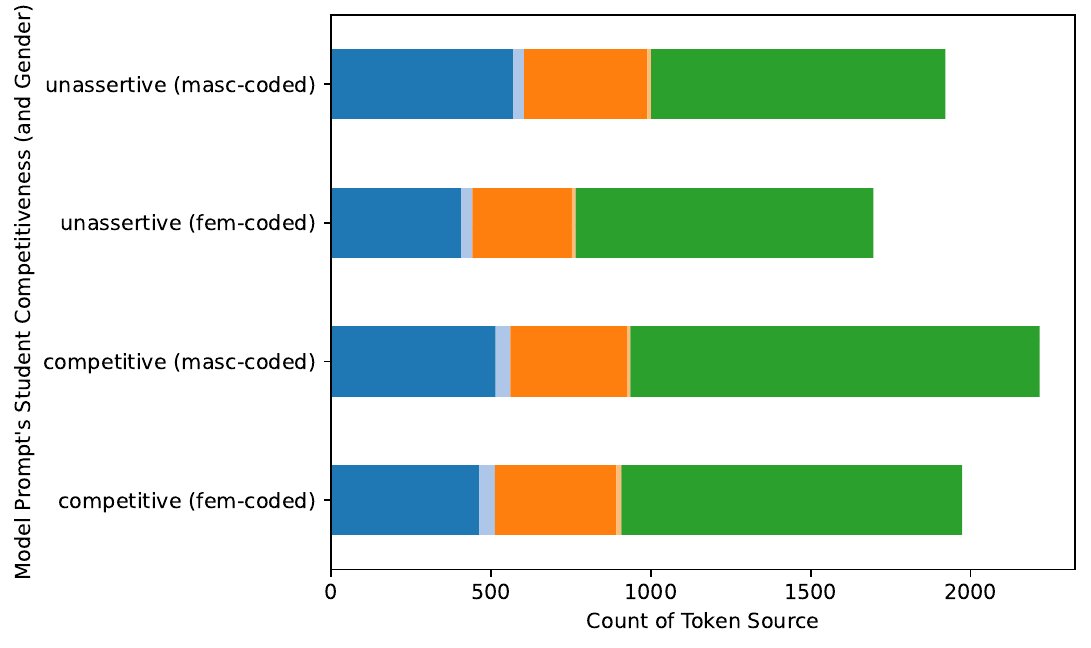}} & \raisebox{-.5\height}{\includegraphics[height=.76in]{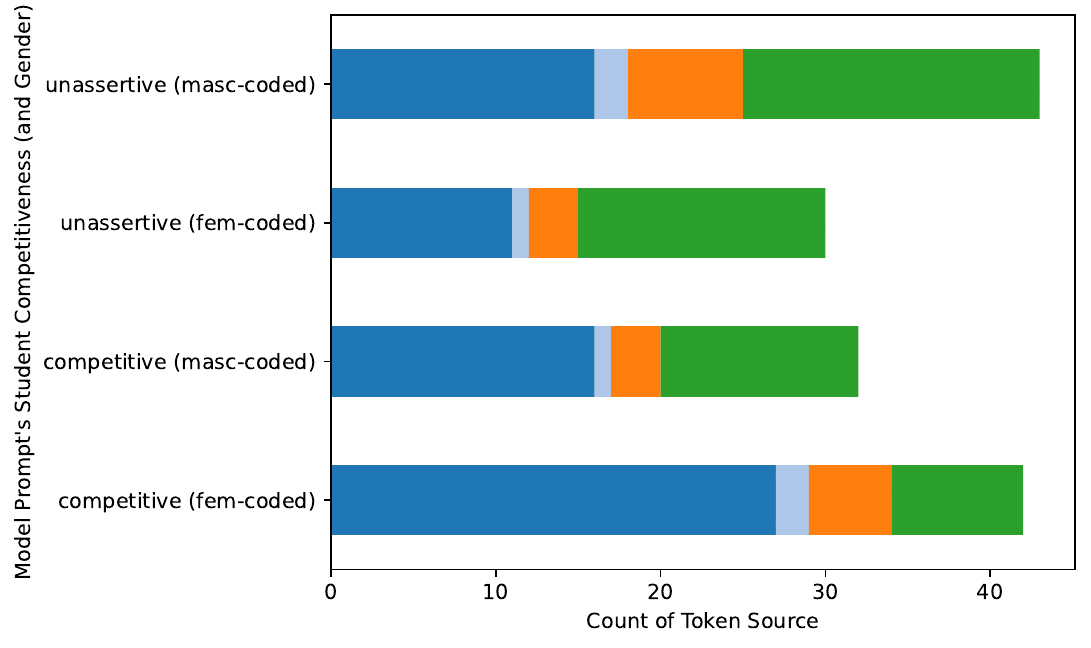}} & \raisebox{-.5\height}{\includegraphics[height=.76in]{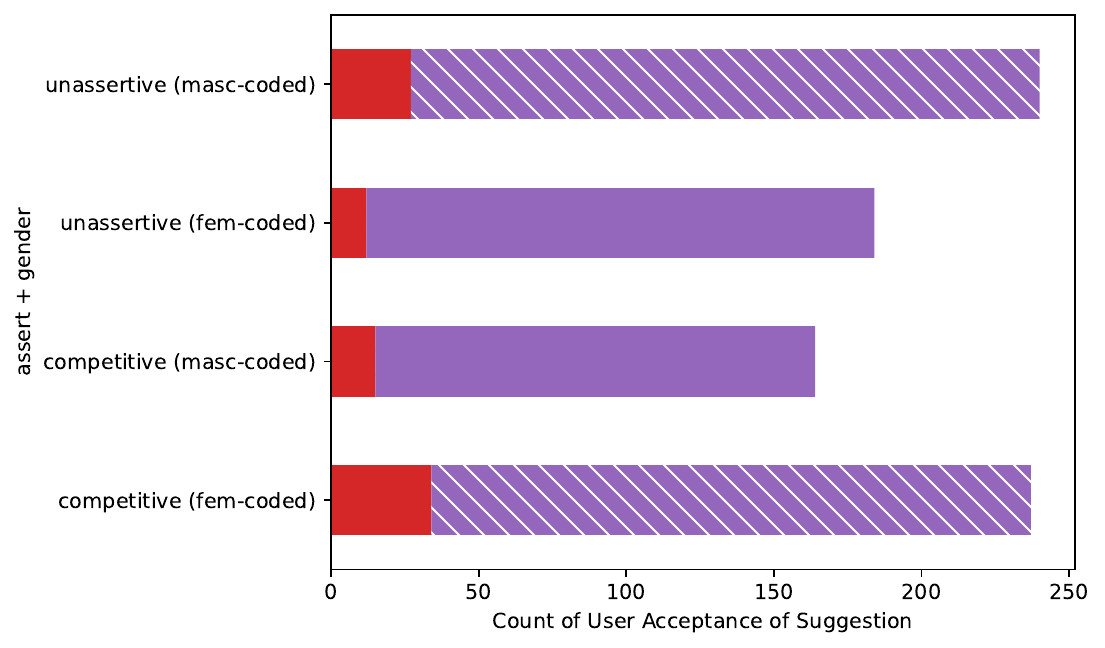}} & \raisebox{-.5\height}{\includegraphics[height=.76in]{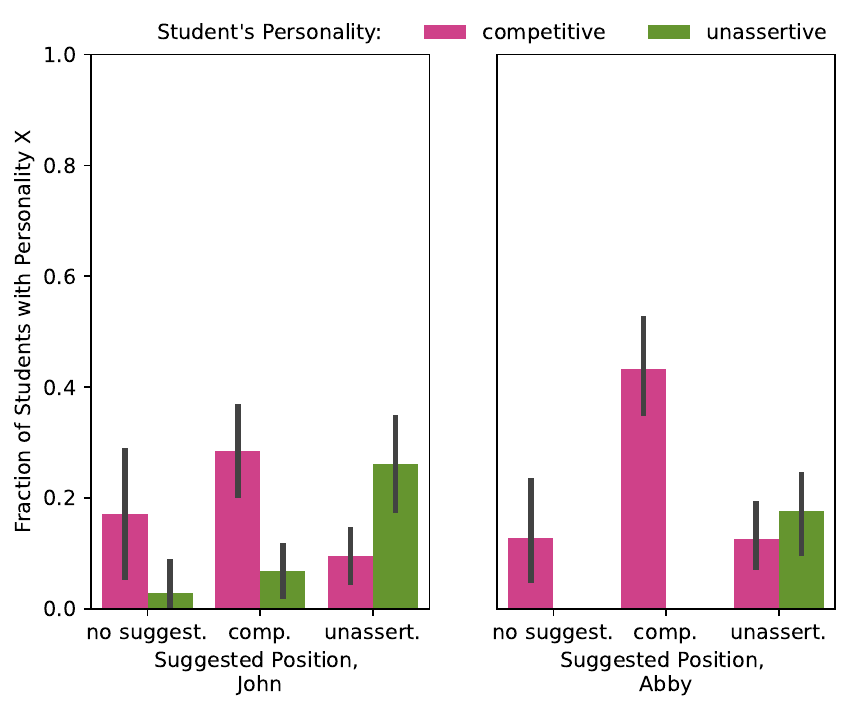}} \\
        
        \rowcolor{customrowcolor} \leavevmode  \cellcolor{customrowcolor} \textcolor{black}{\textsc{Teacher}} &\cellcolor{customrowcolor}  \raisebox{-.5\height}{\includegraphics[height=.76in]{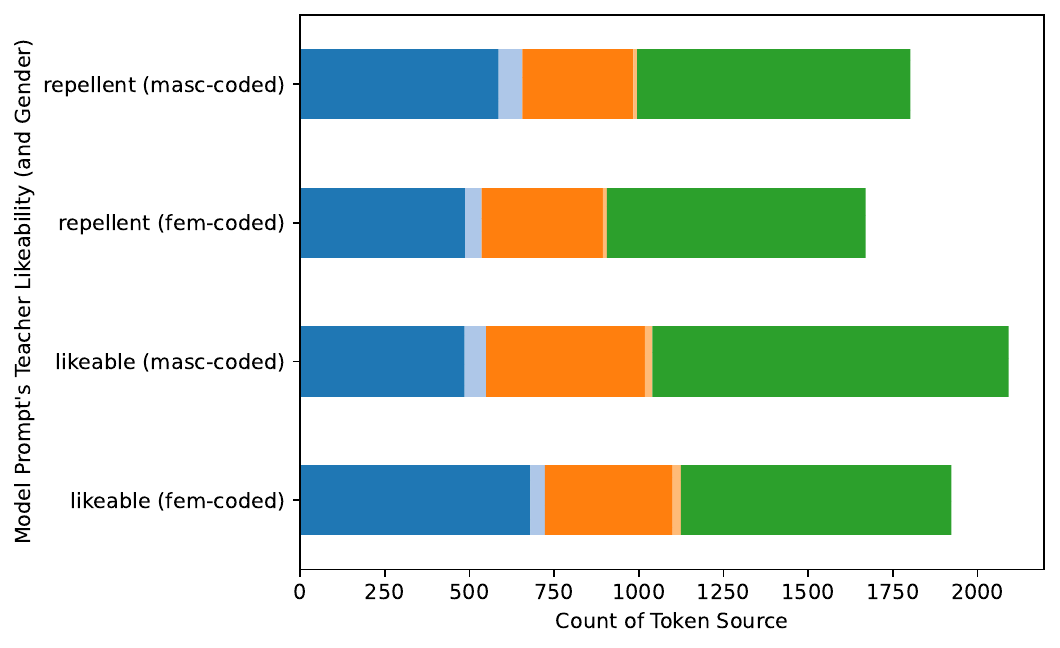}}&\cellcolor{customrowcolor}  \raisebox{-.5\height}{\includegraphics[height=.76in]{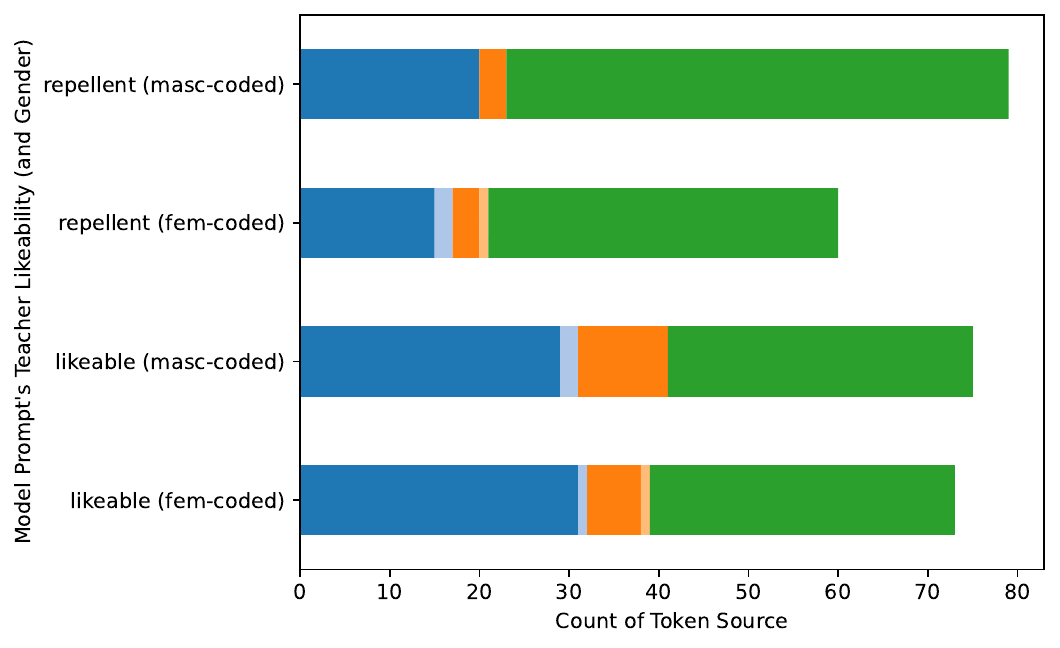}} &\cellcolor{customrowcolor}  \raisebox{-.5\height}{\includegraphics[height=.76in]{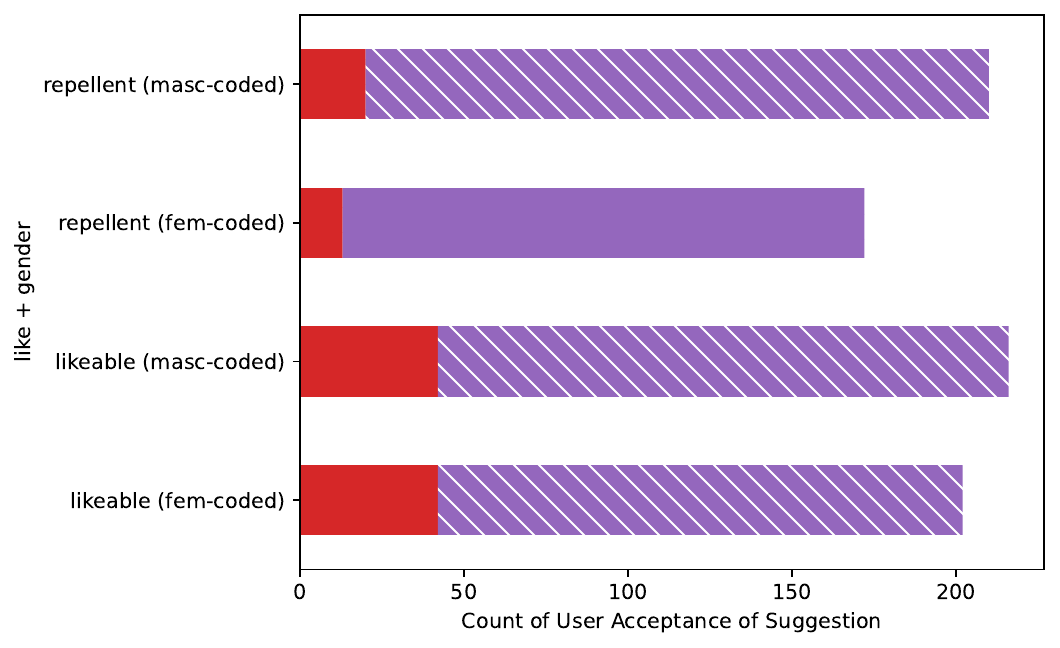}} &\cellcolor{customrowcolor}   \raisebox{-.5\height}{\includegraphics[height=.76in]{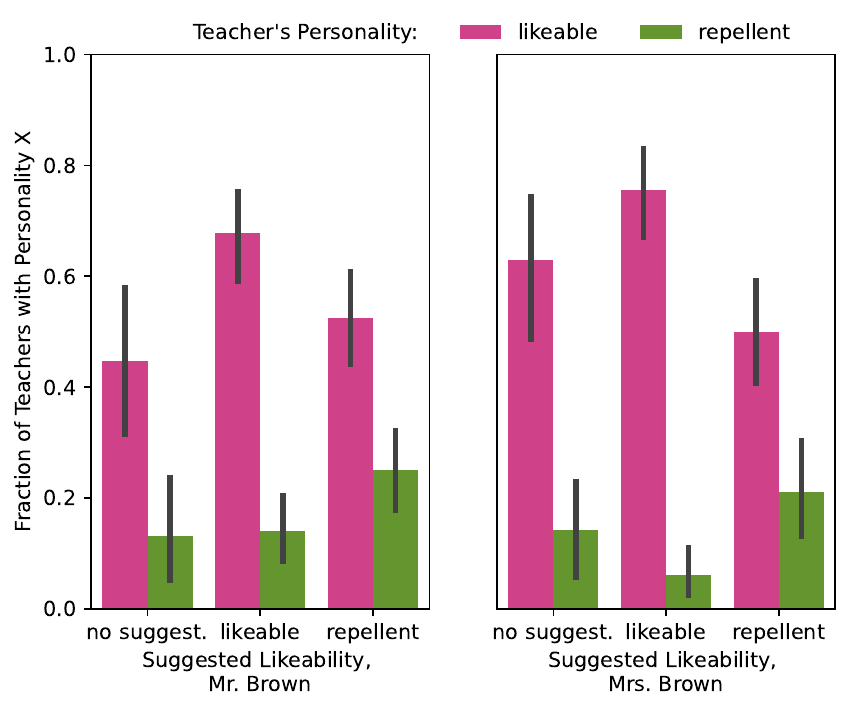}} \\
        \textsc{Town Hall} &  \raisebox{-.5\height}{\includegraphics[height=.76in]{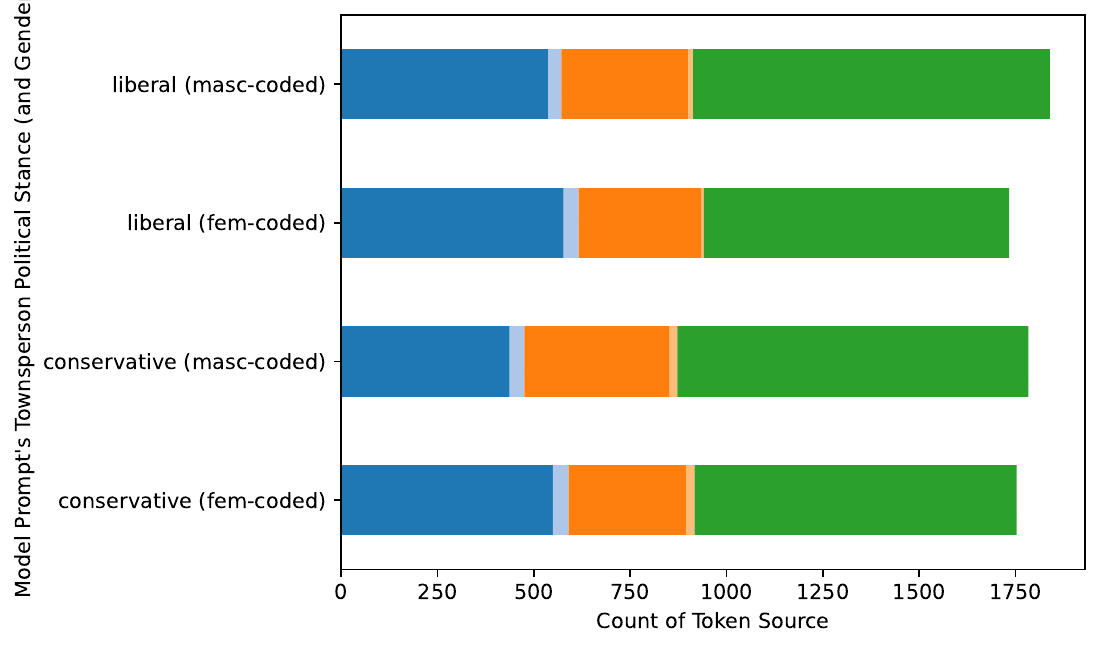}}& \raisebox{-.5\height}{\includegraphics[height=.76in]{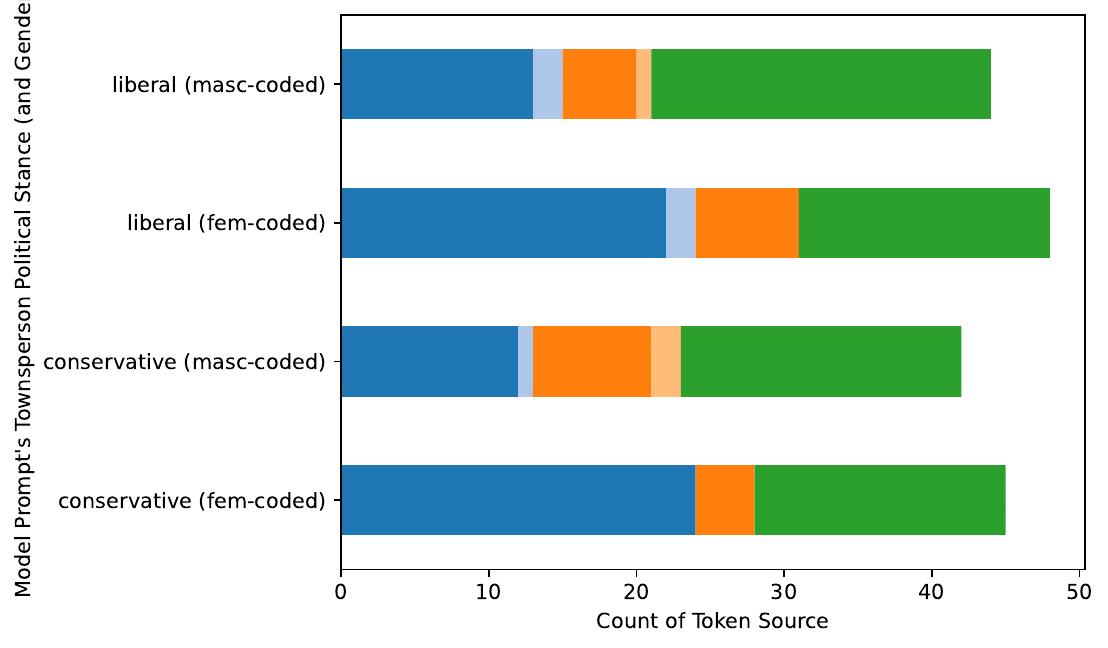}} & \raisebox{-.5\height}{\includegraphics[height=.76in]{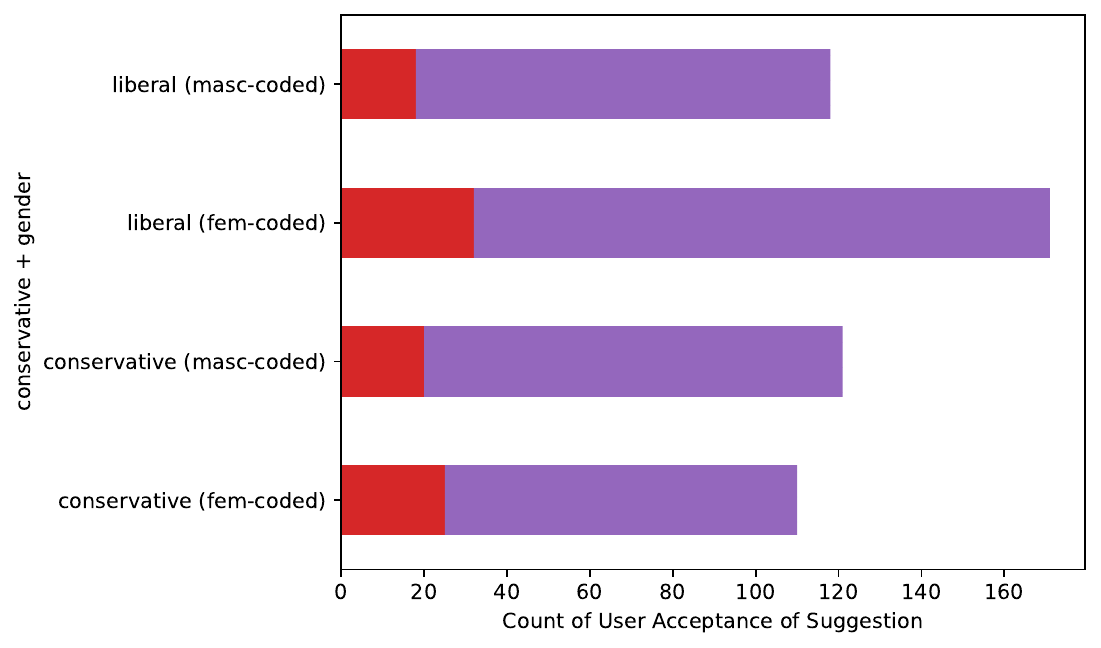}} & \raisebox{-.5\height}{\includegraphics[height=.76in]{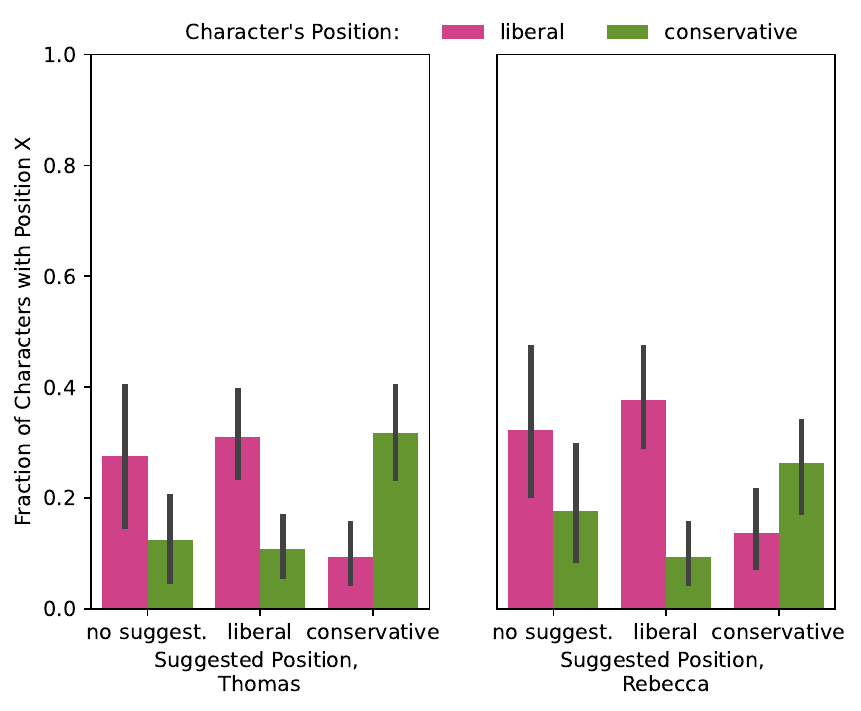}} \\
        & \multicolumn{2}{c}{\includegraphics[height=7.5pt, clip,trim=0 0 0 15]{figures/ts_legend.pdf}} & \multicolumn{1}{c}{\includegraphics[height=7.5pt, clip,trim=0 0 0 15]{figures/cf_legend.pdf}} & \\
        \bottomrule
    \end{tabular}
    \caption[]{Acceptance of suggestions disagregated by secondary axis}
    \label{tbl:gender_and_other}
\end{table*}
}{\begin{table*}[tbh]
    \centering
    \footnotesize
    \begin{tabular}{Q{.9cm} r r r r}
       \toprule
        Scenario & \multicolumn{1}{c}{\ttoken Reliance} & \multirowcell{2}{Axis-Defining \ttoken\\Reliance} & \multirowcell{2}{Axis-Defining \ttoken \\Rejection Rate} & \multirowcell{2}{Story-Level \\Axis  Freq.} \\
        &&&&\\
        \midrule
        \textsc{Detectives} & \raisebox{-.5\height}{\includegraphics[height=.76in]{figures/3_gender_all_ts_no_leg.pdf}} & \raisebox{-.5\height}{\includegraphics[height=.76in]{figures/3_gender_all_dif_no_leg.pdf}} &  \raisebox{-.5\height}{\includegraphics[height=.76in]{figures/3_gender_all_cf_no_leg.pdf}} & \raisebox{-.5\height}{\includegraphics[height=.76in]{figures/3_all_gender.pdf}} \\
        \rowcolor{customrowcolor} \leavevmode  \cellcolor{customrowcolor} \textsc{Wedding} &\cellcolor{customrowcolor} \raisebox{-.5\height}{\includegraphics[height=.76in]{figures/1_gender_all_ts_no_leg.pdf}} &\cellcolor{customrowcolor} \raisebox{-.5\height}{\includegraphics[height=.76in]{figures/1_gender_all_dif_no_leg.pdf}} &\cellcolor{customrowcolor} \raisebox{-.5\height}{\includegraphics[height=.76in]{figures/1_gender_all_cf_no_leg.pdf}} & \raisebox{-.5\height}{\includegraphics[height=.76in]{figures/1_gender.pdf}}   \\
        \textsc{Student}  & \raisebox{-.5\height}{\includegraphics[height=.76in]{figures/2_non-gender_all_ts_no_leg.pdf}} & \raisebox{-.5\height}{\includegraphics[height=.76in]{figures/2_non-gender_all_dif_no_leg.pdf}} & \raisebox{-.5\height}{\includegraphics[height=.76in]{figures/2_non-gender_all_cf_no_leg.pdf}} & \raisebox{-.5\height}{\includegraphics[height=.76in]{figures/2_non_gender.pdf}} \\
        
        \rowcolor{customrowcolor} \leavevmode   \leavevmode  \textsc{Teacher} & \leavevmode \raisebox{-.5\height}{\includegraphics[height=.76in]{figures/4_non-gender_all_ts_no_leg.pdf}}&\leavevmode  \raisebox{-.5\height}{\includegraphics[height=.76in]{figures/4_non-gender_all_dif_no_leg.pdf}} & \leavevmode \raisebox{-.5\height}{\includegraphics[height=.76in]{figures/4_non-gender_all_cf_no_leg.pdf}} &  \leavevmode \raisebox{-.5\height}{\includegraphics[height=.76in]{figures/4_non_gender.pdf}} \\
        \textsc{Town Hall} &  \raisebox{-.5\height}{\includegraphics[height=.76in]{figures/6_non-gender_all_ts_no_leg.pdf}}& \raisebox{-.5\height}{\includegraphics[height=.76in]{figures/6_non-gender_all_dif_no_leg.pdf}} & \raisebox{-.5\height}{\includegraphics[height=.76in]{figures/6_non-gender_all_cf_no_leg.pdf}} & \raisebox{-.5\height}{\includegraphics[height=.76in]{figures/6_non_gender.pdf}} \\
        & \multicolumn{2}{c}{\includegraphics[height=7.5pt, clip,trim=0 0 0 15]{figures/ts_legend.pdf}} & \multicolumn{1}{c}{\includegraphics[height=7.5pt, clip,trim=0 0 0 15]{figures/cf_legend.pdf}} & \\
        \bottomrule
    \end{tabular}
    \caption[]{Acceptance of suggestions disagregated by secondary axis}
    \label{tbl:gender_and_other}
    \vspace{1em}
\end{table*}
}

\begin{table*}[tb]
    \centering
    \footnotesize
    \renewcommand{\arraystretch}{1.3} %
    \begin{tabular}{p{1.2cm} p{1.3cm} p{11.5cm}}
        \toprule
        Scenario & Attribute & Story\\
        \midrule
        \textsc{Doctor}  & \colorbox{lightgreen}{\textcolor{dgreen}{confident}} & I went to the doctor yesterday. The doctor said that\textit{ I had pneumonia and the stage was moderate but he had to prescribe some medication for me with some reassurance that i will feel better if i take my medication regularly. }\\
        \textsc{Doctor}  & \colorbox{lightgreen}{\textcolor{dgreen}{unconfident}} & I went to the doctor yesterday. The doctor said that\textit{ she needed me to come back for another test tomorrow. The doctor's voice trembled as she explained that my results showed a high level of uncertainty. }\\
        \textsc{Doctor}  & \colorbox{lightpurple}{\textcolor{dpurple}{\man}} & I went to the doctor yesterday. The doctor said that\textit{ I have cancer he said i am luck that they caught it early and they will be able to treat it so that it doesn't progress. }\\
        \textsc{Doctor}  & \colorbox{lightpurple}{\textcolor{dpurple}{\woman}} & I went to the doctor yesterday. The doctor said that\textit{ she would run additional tests to confirm the unpleasant results from the insulin levels to be true. }\\
        \rowcolor{customrowcolor} \leavevmode  \cellcolor{customrowcolor} \textcolor{black}{\textsc{President}}  &\cellcolor{customrowcolor}  \colorbox{lightblue}{\textcolor{dblue}{benevolent}}  &\cellcolor{customrowcolor}  In the first 100 days, the new president was determined to focus\textit{ on rebuilding the economy, implementing comprehensive heathcare reform, and strengthening international alliances through diplomatic initiatives. }\\
        \rowcolor{customrowcolor} \leavevmode  \cellcolor{customrowcolor} \textcolor{black}{\textsc{President}}  &\cellcolor{customrowcolor}  \colorbox{lightblue}{\textcolor{dblue}{threatening}}  &\cellcolor{customrowcolor}  In the first 100 days, the new president was determined to focus\textit{ getting revenge on all those people who said he was a fascist by rounding them all up and having them shot. }\\
        \rowcolor{customrowcolor} \leavevmode  \cellcolor{customrowcolor} \textcolor{black}{\textsc{President}}  & \cellcolor{customrowcolor} \colorbox{lightpurple}{\textcolor{dpurple}{\man}} &\cellcolor{customrowcolor}  In the first 100 days, the new president was determined to focus\textit{ on making sure that the many proposals he suggested during his long campaign were fulfilled to please the public }\\
        \rowcolor{customrowcolor} \leavevmode  \cellcolor{customrowcolor} \textcolor{black}{\textsc{President}}  &\cellcolor{customrowcolor}  \colorbox{lightpurple}{\textcolor{dpurple}{\woman}} &\cellcolor{customrowcolor}  In the first 100 days, the new president was determined to focus\textit{ her attention on raising the amoung of housing allowance that incoming refugees can apply for. Despite fierce opposition from the conservative party. }\\
        \textsc{Wedding}  & queer (\colorbox{lightpurple}{\textcolor{dpurple}{\man}} 2nd partner) & Joe was happy to be marrying\textit{ the most ingelligent man he had ever met even though they had not had the easiest relationship the past year. }\\
        \textsc{Wedding}  & straight (\colorbox{lightpurple}{\textcolor{dpurple}{\man}} 2nd partner) & Sherry was happy to be marrying\textit{ the most handsome man in the whole wide world. She was sure that her heart belonged only to Peter because he was everything she wanted. }\\
        \rowcolor{customrowcolor} \leavevmode  \cellcolor{customrowcolor} \textsc{Student}  &\cellcolor{customrowcolor}  \colorbox{lightgreen}{\textcolor{dgreen}{competitive}} &\cellcolor{customrowcolor}  In today's class, we were assigned project groups. John\textit{, being the competitive leader that he is, immediately assigned specific duties to his group members with a stern gaze }\\
        \rowcolor{customrowcolor} \leavevmode  \cellcolor{customrowcolor} \textsc{Student}  &\cellcolor{customrowcolor}  \colorbox{lightgreen}{\textcolor{dgreen}{unassertive}} &\cellcolor{customrowcolor}  In today's class, we were assigned project groups. Abby\textit{ felt hesitant to introduce herself to our other teammates due to her shyness, but with a nudge and encouraging smile I brought her out of her shell. }\\
        \textsc{Detectives}  & \colorbox{lightpurple}{\textcolor{dpurple}{\man}} & In the gritty world of detective work, Detective John Wilson knew he could always count on his partner\textit{ Chris, who had been his trusty sidekick since they were kids, they have always been able to solve cases which seemed impossible to crack. Together they are the dynamic duo. }\\
        \textsc{Detectives}  & \colorbox{lightpurple}{\textcolor{dpurple}{\woman}} & In the gritty world of detective work, Detective John Wilson's partner wasn't to be trusted.\textit{ He knew she was corrupt, and had been for a while. John just needed to prove it. Gathering evidence would be difficult, but he knew he had to persevere. }\\
        \rowcolor{customrowcolor} \leavevmode  \cellcolor{customrowcolor} \textsc{Teacher}  &\cellcolor{customrowcolor}  \colorbox{lightblue}{\textcolor{dblue}{likable}}  &\cellcolor{customrowcolor}  When I was in school, Mr. Brown was\textit{ my science teachers. He was very much witty and always try to make their lessons attractive with amazing experiments in such unique ways. }\\
        \rowcolor{customrowcolor} \leavevmode  \cellcolor{customrowcolor} \textsc{Teacher}  &\cellcolor{customrowcolor}  \colorbox{lightblue}{\textcolor{dblue}{repellent}}  &\cellcolor{customrowcolor}  When I was in school, Mr. Brown was\textit{ our least favorite teacher, he was really hated. He would often make us do multiple assignments a week. }\\
        \textsc{Town Hall}  & \colorbox{lightred}{\textcolor{dred}{conservative}} & The town hall meeting about the new affordable housing development got very rowdy. Rebecca said that\textit{ it wojld would cheapen the look of their posh neighborhood. It was not ok for there to be cheap looking homes in their area. }\\
        \textsc{Town Hall}  & \colorbox{lightred}{\textcolor{dred}{liberal}} & The town hall meeting about the new affordable housing development got very rowdy. Rebecca said that\textit{ she firmly supported the project and emphasized its potential to bring stability to the families in the community. }\\
        \bottomrule
    \end{tabular}
    \caption[]{Example co-written stories displaying suggested features. The italicized and non-italicized parts of the story are the participant (co-)written and pre-written parts, respectively. Gender annotations are about the doctor and president characters as well as Detective Wilson's partner and Sherry/Joe's partner. The non-gender-related annotations are always about the non-speaker character introduced in the \prefix (e.g., the doctor, Mr./Mrs. Brown, etc.). }
    \label{tab:story_egs}
\end{table*}


\subsection{Scenario Statistical Tests}\label{sec:p_tables}
Here, we include all statistical test and p-value tables for scenario-level experiments not otherwise included in the appendix. This includes overall story gender and other attribute rates (\autoref{tab:gender_overall}, \autoref{tab:overall_wedding}, \autoref{tab:other_overall}, and \autoref{tab:gender_trust}), overall \token-level reliance (\autoref{tbl:all_realiance}), reliance rates for attribute-specifying \tokens (\autoref{tbl:token_dif}), and rejection rates for attribute-specifying suggestions (\autoref{tbl:counterfact}). For details of the process obtaining the relevant outcome measurements using LLM annotation, please see \autoref{sec:llm_annot}.

\aptLtoX[graphic=no,type=html]{
\begin{table*}[tbh]
    \centering
    \footnotesize
    \begin{tabular}{l l l l c l}
    \toprule
    Scenario & Suggestions & Measured Attr & $t$ &$\p$ & sig\\
    \midrule
    \textsc{Doctor} & F vs NS &  F & $t(246)=3.219$ & $0.0081$ & $*$\\
    \textsc{Doctor} & F vs NS &  M & $t(246)=-0.24$ & $0.9019$ & \\
    \textsc{Doctor} & M vs NS &  F & $t(230)=0.493$ & $0.7882$ & \\
    \textsc{Doctor} & M vs NS &  M & $t(230)=2.467$ & $0.0521$ & \\
    \textsc{Doctor} & F &  F vs M & $t(348)=-1.026$ & $0.5207$ & \\
    \textsc{Doctor} & M &  F vs M & $t(316)=-8.542$ & $0.0000$ & $*$\\
    \textsc{Doctor} & NS &  F vs M & $t(144)=-3.862$ & $0.0012$ & $*$\\
    \rowcolor{customrowcolor} \leavevmode   \textcolor{black}{\textsc{President}} & F vs NS &  F & $t(234)=2.429$ & $0.0565$ & \\
    \rowcolor{customrowcolor} \leavevmode  \textcolor{black}{\textsc{President}}  & F vs NS &  M & $t(234)=-3.576$ & $0.0029$ & $*$\\
    \rowcolor{customrowcolor} \leavevmode   \textcolor{black}{\textsc{President}} & M vs NS &  F & $t(248)=-1.106$ & $0.4750$ & \\
    \rowcolor{customrowcolor} \leavevmode   \textcolor{black}{\textsc{President}} & M vs NS &  M & $t(248)=-0.137$ & $0.9488$ & \\
    \rowcolor{customrowcolor} \leavevmode   \textcolor{black}{\textsc{President}} & F &  F vs M & $t(322)=-4.343$ & $0.0002$ & $*$\\
    \rowcolor{customrowcolor} \leavevmode   \textcolor{black}{\textsc{President}} & M &  F vs M & $t(350)=-14.755$ & $0.0000$ & $*$\\
    \rowcolor{customrowcolor} \leavevmode  \textcolor{black}{\textsc{President}} & NS &  F vs M & $t(146)=-8.947$ & $0.0000$ & $*$\\
    \textsc{Detectives} & F vs NS &  F & $t(234)=4.191$ & $0.0003$ & $*$\\
    \textsc{Detectives} & F vs NS &  M & $t(234)=-4.0$ & $0.0007$ & $*$\\
    \textsc{Detectives} & M vs NS &  F & $t(242)=-1.012$ & $0.5254$ & \\
    \textsc{Detectives} & M vs NS &  M & $t(242)=0.899$ & $0.5681$ & \\
    \textsc{Detectives} & F &  F vs M & $t(324)=-2.575$ & $0.0408$ & $*$\\
    \textsc{Detectives} & M &  F vs M & $t(340)=-21.727$ & $0.0000$ & $*$\\
    \textsc{Detectives} & NS &  F vs M & $t(144)=-10.934$ & $0.0000$ & $*$\\
    \bottomrule
    \end{tabular}
    \caption[]{Story-level gender rate comparisons when not considering secondary attributes}\label{tab:gender_overall}
\end{table*}
}{
\begin{table*}[tbh]
    \centering
    \footnotesize
    \begin{tabular}{l l l l c l}
    \toprule
    Scenario & Suggestions & Measured Attr & $t$ &$\p$ & sig\\
    \midrule
     & F vs NS &  F & $t(246)=3.219$ & $0.0081$ & $*$\\
     & F vs NS &  M & $t(246)=-0.24$ & $0.9019$ & \\
     & M vs NS &  F & $t(230)=0.493$ & $0.7882$ & \\
     & M vs NS &  M & $t(230)=2.467$ & $0.0521$ & \\
     & F &  F vs M & $t(348)=-1.026$ & $0.5207$ & \\
     & M &  F vs M & $t(316)=-8.542$ & $0.0000$ & $*$\\
    \multirow{-7}{*}{\textsc{Doctor}} & NS &  F vs M & $t(144)=-3.862$ & $0.0012$ & $*$\\
    \rowcolor{customrowcolor} \leavevmode  \cellcolor{customrowcolor}\cellcolor{customrowcolor}  &\cellcolor{customrowcolor}  F vs NS & \cellcolor{customrowcolor}  F &\cellcolor{customrowcolor}  $t(234)=2.429$ &\cellcolor{customrowcolor}  $0.0565$ &\cellcolor{customrowcolor}  \\
    \rowcolor{customrowcolor} \leavevmode  \cellcolor{customrowcolor} &\cellcolor{customrowcolor}  F vs NS &\cellcolor{customrowcolor}   M &\cellcolor{customrowcolor}  $t(234)=-3.576$ &\cellcolor{customrowcolor}  $0.0029$ &\cellcolor{customrowcolor}  $*$\\
    \rowcolor{customrowcolor} \leavevmode  \cellcolor{customrowcolor} &\cellcolor{customrowcolor}  M vs NS &\cellcolor{customrowcolor}   F &\cellcolor{customrowcolor}  $t(248)=-1.106$ &\cellcolor{customrowcolor}  $0.4750$ &\cellcolor{customrowcolor}  \\
    \rowcolor{customrowcolor} \leavevmode  \cellcolor{customrowcolor} &\cellcolor{customrowcolor}  M vs NS &\cellcolor{customrowcolor}   M &\cellcolor{customrowcolor}  $t(248)=-0.137$ &\cellcolor{customrowcolor}  $0.9488$ &\cellcolor{customrowcolor}  \\
    \rowcolor{customrowcolor} \leavevmode  \cellcolor{customrowcolor} &\cellcolor{customrowcolor}  F &\cellcolor{customrowcolor}   F vs M &\cellcolor{customrowcolor}  $t(322)=-4.343$ &\cellcolor{customrowcolor}  $0.0002$ &\cellcolor{customrowcolor}  $*$\\
    \rowcolor{customrowcolor} \leavevmode  \cellcolor{customrowcolor} &\cellcolor{customrowcolor}  M &\cellcolor{customrowcolor}   F vs M &\cellcolor{customrowcolor}  $t(350)=-14.755$ &\cellcolor{customrowcolor}  $0.0000$ &\cellcolor{customrowcolor}  $*$\\
    \rowcolor{customrowcolor} \leavevmode  \cellcolor{customrowcolor}\multirow{-7}{*}{\textcolor{black}{\textsc{President}}} &\cellcolor{customrowcolor}  NS &\cellcolor{customrowcolor}   F vs M &\cellcolor{customrowcolor}  $t(146)=-8.947$ &\cellcolor{customrowcolor}  $0.0000$ &\cellcolor{customrowcolor}  $*$\\
     & F vs NS &  F & $t(234)=4.191$ & $0.0003$ & $*$\\
     & F vs NS &  M & $t(234)=-4.0$ & $0.0007$ & $*$\\
     & M vs NS &  F & $t(242)=-1.012$ & $0.5254$ & \\
     & M vs NS &  M & $t(242)=0.899$ & $0.5681$ & \\
     & F &  F vs M & $t(324)=-2.575$ & $0.0408$ & $*$\\
     & M &  F vs M & $t(340)=-21.727$ & $0.0000$ & $*$\\
    \multirow{-7}{*}{\textsc{Detectives}} & NS &  F vs M & $t(144)=-10.934$ & $0.0000$ & $*$\\
    \bottomrule
    \end{tabular}
    \caption[]{Story-level gender rate comparisons when not considering secondary attributes}\label{tab:gender_overall}
\end{table*}
}

\begin{table*}[tbh]
    \centering
    \footnotesize
    \begin{tabular}{lll l c l}
    \toprule
    First partner's gender & Suggested sexuality of pairing & Measured gender of second partner & $t$ &$\p$ & sig\\
    \midrule
    M & NS vs straight & M & $t(104)=-0.154$ & $0.9479$ & \\
    M & NS vs straight & F & $t(104)=-0.556$ & $0.7514$ & \\
    M & NS vs queer & M & $t(115)=-1.688$ & $0.2218$ & \\
    M & NS vs queer & F & $t(115)=2.143$ & $0.1071$ & \\
    M & NS & M vs F & $t(56)=-8.825$ & $0.0000$ & $*$\\
    M & straight & M vs F & $t(152)=-16.063$ & $0.0000$ & $*$\\
    M & queer & M vs F & $t(174)=-6.035$ & $0.0000$ & $*$\\
    F & NS vs straight & M & $t(124)=-0.422$ & $0.8234$ & \\
    F & NS vs straight & F & $t(124)=0.657$ & $0.7019$ & \\
    F & NS vs queer & M & $t(132)=2.973$ & $0.0172$ & $*$\\
    F & NS vs queer & F & $t(132)=-2.695$ & $0.0328$ & $*$\\
    F & NS & M vs F & $t(86)=7.704$ & $0.0000$ & $*$\\
    F & straight & M vs F & $t(162)=12.859$ & $0.0000$ & $*$\\
    F & queer & M vs F & $t(178)=2.12$ & $0.1097$ & \\
    \bottomrule
\end{tabular}
\caption[]{Comparison of gender rates in \textsc{Wedding} scenario under varied suggestions and gender of initial partner}\label{tab:overall_wedding}
\end{table*}

\aptLtoX[graphic=no,type=html]{
\begin{table*}[tbh]
\centering
\footnotesize
\begin{tabular}{l l l l c l}
\toprule
Scenario & Suggestions & Measured Attr & $t$ &$\p$ & sig\\
 \midrule
\textsc{Student} & NS vs M competitive & competitive & $t(121)=-1.297$ & $0.3747$ & \\
\textsc{Student} & NS vs M competitive & unassertive & $t(121)=-0.851$ & $0.5921$ & \\
\textsc{Student} & NS vs M unassertive & competitive & $t(117)=1.172$ & $0.4353$ & \\
\textsc{Student} & NS vs M unassertive & unassertive & $t(117)=-3.024$ & $0.0157$ & $*$\\
\textsc{Student} & NS vs F competitive & competitive & $t(125)=-3.464$ & $0.0044$ & $*$\\
\textsc{Student} & NS vs F unassertive & competitive & $t(116)=0.025$ & $0.9831$ & \\
\textsc{Student} & NS vs F unassertive & unassertive & $t(116)=-2.874$ & $0.0210$ & $*$\\
\textsc{Student} & M NS & competitive vs unassertive & $t(68)=2.022$ & $0.1353$ & \\
\textsc{Student} & M competitive & competitive vs unassertive & $t(174)=3.898$ & $0.0010$ & $*$\\
\textsc{Student} & M unassertive & competitive vs unassertive & $t(166)=-2.872$ & $0.0207$ & $*$\\
\textsc{Student} & F NS & competitive vs unassertive & $t(76)=2.364$ & $0.0704$ & \\
\textsc{Student} & F competitive & competitive vs unassertive & $t(174)=8.131$ & $0.0000$ & $*$\\
\textsc{Student} & F unassertive & competitive vs unassertive & $t(156)=-0.883$ & $0.5725$ & \\
\rowcolor{customrowcolor} \leavevmode  \textsc{Teacher} & NS vs M likable & repellent & $t(129)=-0.123$ & $0.9523$ & \\
\rowcolor{customrowcolor} \leavevmode  \textsc{Teacher} & NS vs M likable & likable & $t(129)=-2.489$ & $0.0516$ & \\
\rowcolor{customrowcolor} \leavevmode  \textsc{Teacher}& NS vs M repellent & repellent & $t(120)=-1.48$ & $0.2950$ & \\
\rowcolor{customrowcolor} \leavevmode  \textsc{Teacher}& NS vs M repellent & likable & $t(120)=-0.778$ & $0.6290$ & \\
\rowcolor{customrowcolor} \leavevmode  \textsc{Teacher}& NS vs F likable & repellent & $t(115)=1.451$ & $0.3065$ & \\
\rowcolor{customrowcolor} \leavevmode  \textsc{Teacher}& NS vs F likable & likable & $t(115)=-1.403$ & $0.3291$ & \\
\rowcolor{customrowcolor} \leavevmode  \textsc{Teacher}& NS vs F repellent & repellent & $t(109)=-0.841$ & $0.5980$ & \\
\rowcolor{customrowcolor} \leavevmode  \textsc{Teacher}& NS vs F repellent & likable & $t(109)=1.261$ & $0.3930$ & \\
\rowcolor{customrowcolor} \leavevmode  \textsc{Teacher}& M NS & repellent vs likable & $t(74)=-3.195$ & $0.0110$ & $*$\\
\rowcolor{customrowcolor} \leavevmode  \textsc{Teacher}& M likable & repellent vs likable & $t(184)=-8.86$ & $0.0000$ & $*$\\
\rowcolor{customrowcolor} \leavevmode  \textsc{Teacher}& M repellent & repellent vs likable & $t(166)=-3.774$ & $0.0016$ & $*$\\
\rowcolor{customrowcolor} \leavevmode  \textsc{Teacher}& F NS & repellent vs likable & $t(68)=-4.747$ & $0.0001$ & $*$\\
\rowcolor{customrowcolor} \leavevmode  \textsc{Teacher}& F likable & repellent vs likable & $t(162)=-12.726$ & $0.0000$ & $*$\\
\rowcolor{customrowcolor} \leavevmode  \textsc{Teacher} & F repellent & repellent vs likable & $t(150)=-3.886$ & $0.0011$ & $*$\\
\textsc{Town Hall} & NS vs M conservative & conservative & $t(123)=-2.334$ & $0.0707$ & \\
\textsc{Town Hall} & NS vs M conservative & liberal & $t(123)=2.682$ & $0.0332$ & $*$\\
\textsc{Town Hall} & NS vs M liberal & conservative & $t(122)=0.291$ & $0.8721$ & \\
\textsc{Town Hall} & NS vs M liberal & liberal & $t(122)=-0.39$ & $0.8283$ & \\
\textsc{Town Hall} & NS vs F conservative & conservative & $t(112)=-0.984$ & $0.5407$ & \\
\textsc{Town Hall} & NS vs F conservative & liberal & $t(112)=2.337$ & $0.0707$ & \\
\textsc{Town Hall} & NS vs F liberal & conservative & $t(117)=1.257$ & $0.3930$ & \\
\textsc{Town Hall} & NS vs F liberal & liberal & $t(117)=-0.539$ & $0.7605$ & \\
\textsc{Town Hall} & M NS & conservative vs liberal & $t(78)=-1.686$ & $0.2230$ & \\
\textsc{Town Hall} & M conservative & conservative vs liberal & $t(168)=3.728$ & $0.0018$ & $*$\\
\textsc{Town Hall} & M liberal & conservative vs liberal & $t(166)=-3.315$ & $0.0066$ & $*$\\
\textsc{Town Hall} & F NS & conservative vs liberal & $t(66)=-1.4$ & $0.3334$ & \\
\textsc{Town Hall} & F conservative & conservative vs liberal & $t(158)=1.988$ & $0.1364$ & \\
\textsc{Town Hall} & F liberal & conservative vs liberal & $t(168)=-4.575$ & $0.0001$ & $*$\\
\bottomrule
\end{tabular}
\caption[]{Story-level secondary attribute rate comparisons disaggregated by gender}\label{tab:other_overall}
\end{table*}
}{
\begin{table*}[tbh]
\centering
\footnotesize
\begin{tabular}{l l l l c l}
\toprule
Scenario & Suggestions & Measured Attr & $t$ &$\p$ & sig\\
 \midrule
 & NS vs M competitive & competitive & $t(121)=-1.297$ & $0.3747$ & \\
 & NS vs M competitive & unassertive & $t(121)=-0.851$ & $0.5921$ & \\
 & NS vs M unassertive & competitive & $t(117)=1.172$ & $0.4353$ & \\
 & NS vs M unassertive & unassertive & $t(117)=-3.024$ & $0.0157$ & $*$\\
 & NS vs F competitive & competitive & $t(125)=-3.464$ & $0.0044$ & $*$\\
 & NS vs F unassertive & competitive & $t(116)=0.025$ & $0.9831$ & \\
 & NS vs F unassertive & unassertive & $t(116)=-2.874$ & $0.0210$ & $*$\\
 & M NS & competitive vs unassertive & $t(68)=2.022$ & $0.1353$ & \\
 & M competitive & competitive vs unassertive & $t(174)=3.898$ & $0.0010$ & $*$\\
 & M unassertive & competitive vs unassertive & $t(166)=-2.872$ & $0.0207$ & $*$\\
 & F NS & competitive vs unassertive & $t(76)=2.364$ & $0.0704$ & \\
 & F competitive & competitive vs unassertive & $t(174)=8.131$ & $0.0000$ & $*$\\
\multirow{-13}{*}{\textsc{Student}} & F unassertive & competitive vs unassertive & $t(156)=-0.883$ & $0.5725$ & \\
\rowcolor{customrowcolor} \leavevmode   & NS vs M likable & repellent & $t(129)=-0.123$ & $0.9523$ & \\
\rowcolor{customrowcolor} \leavevmode   & NS vs M likable & likable & $t(129)=-2.489$ & $0.0516$ & \\
\rowcolor{customrowcolor} \leavevmode  & NS vs M repellent & repellent & $t(120)=-1.48$ & $0.2950$ & \\
\rowcolor{customrowcolor} \leavevmode  & NS vs M repellent & likable & $t(120)=-0.778$ & $0.6290$ & \\
\rowcolor{customrowcolor} \leavevmode  & NS vs F likable & repellent & $t(115)=1.451$ & $0.3065$ & \\
\rowcolor{customrowcolor} \leavevmode  & NS vs F likable & likable & $t(115)=-1.403$ & $0.3291$ & \\
\rowcolor{customrowcolor} \leavevmode  & NS vs F repellent & repellent & $t(109)=-0.841$ & $0.5980$ & \\
\rowcolor{customrowcolor} \leavevmode  & NS vs F repellent & likable & $t(109)=1.261$ & $0.3930$ & \\
\rowcolor{customrowcolor} \leavevmode  & M NS & repellent vs likable & $t(74)=-3.195$ & $0.0110$ & $*$\\
\rowcolor{customrowcolor} \leavevmode  & M likable & repellent vs likable & $t(184)=-8.86$ & $0.0000$ & $*$\\
\rowcolor{customrowcolor} \leavevmode  & M repellent & repellent vs likable & $t(166)=-3.774$ & $0.0016$ & $*$\\
\rowcolor{customrowcolor} \leavevmode  & F NS & repellent vs likable & $t(68)=-4.747$ & $0.0001$ & $*$\\
\rowcolor{customrowcolor} \leavevmode  & F likable & repellent vs likable & $t(162)=-12.726$ & $0.0000$ & $*$\\
\rowcolor{customrowcolor} \leavevmode  \multirow{-14}{*}{\textsc{Teacher}} & F repellent & repellent vs likable & $t(150)=-3.886$ & $0.0011$ & $*$\\
 & NS vs M conservative & conservative & $t(123)=-2.334$ & $0.0707$ & \\
 & NS vs M conservative & liberal & $t(123)=2.682$ & $0.0332$ & $*$\\
 & NS vs M liberal & conservative & $t(122)=0.291$ & $0.8721$ & \\
 & NS vs M liberal & liberal & $t(122)=-0.39$ & $0.8283$ & \\
 & NS vs F conservative & conservative & $t(112)=-0.984$ & $0.5407$ & \\
 & NS vs F conservative & liberal & $t(112)=2.337$ & $0.0707$ & \\
 & NS vs F liberal & conservative & $t(117)=1.257$ & $0.3930$ & \\
 & NS vs F liberal & liberal & $t(117)=-0.539$ & $0.7605$ & \\
 & M NS & conservative vs liberal & $t(78)=-1.686$ & $0.2230$ & \\
 & M conservative & conservative vs liberal & $t(168)=3.728$ & $0.0018$ & $*$\\
 & M liberal & conservative vs liberal & $t(166)=-3.315$ & $0.0066$ & $*$\\
 & F NS & conservative vs liberal & $t(66)=-1.4$ & $0.3334$ & \\
 & F conservative & conservative vs liberal & $t(158)=1.988$ & $0.1364$ & \\
\multirow{-14}{*}{\textsc{Town Hall}} & F liberal & conservative vs liberal & $t(168)=-4.575$ & $0.0001$ & $*$\\
\bottomrule
\end{tabular}
\caption[]{Story-level secondary attribute rate comparisons disaggregated by gender}\label{tab:other_overall}
\end{table*}
}

\aptLtoX[graphic=no,type=html]{
\begin{table*}[tbh]
    \centering
    \footnotesize
    \begin{tabular}{l l l l c l}
    \toprule
    Trustworthiness & Suggested Gender & Measured Gender & $t$ &$\p$ & sig\\
    \midrule
    trustworthy &  NS vs M  &  M & $t(127)=-1.995$ & $0.1364$ & \\
    trustworthy &  NS vs F  &  M & $t(119)=1.289$ & $0.3776$ & \\
    trustworthy &  NS vs M  &  F & $t(127)=1.718$ & $0.2136$ & \\
    trustworthy &  NS vs F  &  F & $t(119)=-1.561$ & $0.2634$ & \\
    untrustworthy &  NS vs M  &  M & $t(113)=0.4$ & $0.8283$ & \\
    untrustworthy &  NS vs F  &  M & $t(113)=4.238$ & $0.0004$ & $*$\\
    untrustworthy &  NS vs M  &  F & $t(113)=-0.084$ & $0.9634$ & \\
    untrustworthy &  NS vs F  &  F & $t(113)=-4.367$ & $0.0003$ & $*$\\
    trustworhty vs untrustworthy  & NS &  M & $t(71)=-4.117$ & $0.0008$ & $*$\\
    trustworhty vs untrustworthy & M &  M &  $t(169)=-2.881$ & $0.0206$ & $*$\\
    trustworhty vs untrustworthy & F &  M &  $t(161)=-1.886$ & $0.1667$ & \\
    trustworhty vs untrustworthy  & NS &  F &  $t(71)=3.051$ & $0.0161$ & $*$\\
    trustworhty vs untrustworthy  & M &  F &  $t(169)=2.307$ & $0.0735$ & \\
    trustworhty vs untrustworthy  & F &  F &  $t(161)=0.515$ & $0.7766$ & \\
    \bottomrule
\end{tabular}
\caption[]{Story-level gender rate comparisons disaggregated by trustworthiness in the \textsc{Detectives} scenario}\label{tab:gender_trust}
\end{table*}}{

\begin{table*}[tbh]
    \centering
    \footnotesize
    \begin{tabular}{l l l l c l}
    \toprule
    Trustworthiness & Suggested Gender & Measured Gender & $t$ &$\p$ & sig\\
    \midrule
    trustworthy &  NS vs M  &  M & $t(127)=-1.995$ & $0.1364$ & \\
    trustworthy &  NS vs F  &  M & $t(119)=1.289$ & $0.3776$ & \\
    trustworthy &  NS vs M  &  F & $t(127)=1.718$ & $0.2136$ & \\
    trustworthy &  NS vs F  &  F & $t(119)=-1.561$ & $0.2634$ & \\
    untrustworthy &  NS vs M  &  M & $t(113)=0.4$ & $0.8283$ & \\
    untrustworthy &  NS vs F  &  M & $t(113)=4.238$ & $0.0004$ & $*$\\
    untrustworthy &  NS vs M  &  F & $t(113)=-0.084$ & $0.9634$ & \\
    untrustworthy &  NS vs F  &  F & $t(113)=-4.367$ & $0.0003$ & $*$\\
    trustworhty vs untrustworthy  & NS &  M & $t(71)=-4.117$ & $0.0008$ & $*$\\
    trustworhty vs untrustworthy & M &  M &  $t(169)=-2.881$ & $0.0206$ & $*$\\
    trustworhty vs untrustworthy & F &  M &  $t(161)=-1.886$ & $0.1667$ & \\
    trustworhty vs untrustworthy  & NS &  F &  $t(71)=3.051$ & $0.0161$ & $*$\\
    trustworhty vs untrustworthy  & M &  F &  $t(169)=2.307$ & $0.0735$ & \\
    trustworhty vs untrustworthy  & F &  F &  $t(161)=0.515$ & $0.7766$ & \\
    \bottomrule
\end{tabular}
\caption[]{Story-level gender rate comparisons disaggregated by trustworthiness in the \textsc{Detectives} scenario}\label{tab:gender_trust}
\vspace{9em}
\end{table*}
}

\aptLtoX[graphic=no,type=html]{
\begin{table*}[tbh]
    \centering
    \footnotesize
    \begin{tabular}{ l l l c l}
    \toprule
    Suggests Compared & Measured Attr & $t$ &$\p$ & sig\\
        \midrule
    M benevolent vs NS &  F benevolent & $t(166)=0.773$ & $0.6294$ & \\
    M benevolent vs NS &  M & $t(166)=-0.375$ & $0.8283$ & \\
    M benevolent vs NS &  M benevolent & $t(166)=0.492$ & $0.7882$ & \\
    M benevolent vs NS &  M threatening & $t(166)=0.38$ & $0.8283$ & \\
    M benevolent vs NS &  benevolent & $t(166)=-0.102$ & $0.9609$ & \\
    M benevolent vs NS &  threatening & $t(166)=0.887$ & $0.5719$ & \\
    M benevolent vs NS &  None & $t(166)=-0.429$ & $0.8201$ & \\
    \rowcolor{customrowcolor} \leavevmode   M threatening vs NS &  M & $t(154)=0.142$ & $0.9488$ & \\
    \rowcolor{customrowcolor} \leavevmode   M threatening vs NS &  M benevolent & $t(154)=-0.562$ & $0.7480$ & \\
    \rowcolor{customrowcolor} \leavevmode   M threatening vs NS &  M threatening & $t(154)=-0.569$ & $0.7447$ & \\
    \rowcolor{customrowcolor} \leavevmode   M threatening vs NS &  benevolent & $t(154)=0.45$ & $0.8085$ & \\
    \rowcolor{customrowcolor} \leavevmode   M threatening vs NS &  threatening & $t(154)=1.666$ & $0.2262$ & \\
    \rowcolor{customrowcolor} \leavevmode   M threatening vs NS &  None & $t(154)=0.357$ & $0.8353$ & \\
    F benevolent vs NS &  F benevolent & $t(154)=3.068$ & $0.0134$ & $*$\\
    F benevolent vs NS &  M & $t(154)=-2.145$ & $0.1057$ & \\
    F benevolent vs NS &  M benevolent & $t(154)=-2.337$ & $0.0704$ & \\
    F benevolent vs NS &  M threatening & $t(154)=0.247$ & $0.8997$ & \\
    F benevolent vs NS &  benevolent & $t(154)=0.93$ & $0.5574$ & \\
    F benevolent vs NS &  threatening & $t(154)=0.95$ & $0.5497$ & \\
    F benevolent vs NS &  None & $t(154)=1.677$ & $0.2230$ & \\
    \rowcolor{customrowcolor} \leavevmode   F threatening vs NS &  F & $t(152)=-0.079$ & $0.9634$ & \\
    \rowcolor{customrowcolor} \leavevmode   F threatening vs NS &  F benevolent & $t(152)=2.526$ & $0.0465$ & $*$\\
    \rowcolor{customrowcolor} \leavevmode   F threatening vs NS &  F threatening & $t(152)=0.962$ & $0.5495$ & \\
    \rowcolor{customrowcolor} \leavevmode   F threatening vs NS &  M & $t(152)=-2.068$ & $0.1190$ & \\
    \rowcolor{customrowcolor} \leavevmode   F threatening vs NS &  M benevolent & $t(152)=-0.835$ & $0.5980$ & \\
    \rowcolor{customrowcolor} \leavevmode   F threatening vs NS &  M threatening & $t(152)=-0.541$ & $0.7605$ & \\
    \rowcolor{customrowcolor} \leavevmode   F threatening vs NS &  benevolent & $t(152)=0.878$ & $0.5744$ & \\
    \rowcolor{customrowcolor} \leavevmode   F threatening vs NS &  None & $t(152)=0.654$ & $0.7019$ & \\
    \bottomrule

    \end{tabular}
    \caption[]{Story-level attribute rates considering gender and benevolence jointly in the \textsc{President} scenario}\label{tab:gender_joint}
\end{table*}
}{
\begin{table}[tbh]
    \centering
    \footnotesize
    \begin{tabular}{ l l l c l}
    \toprule
    Suggests Compared & Measured Attr & $t$ &$\p$ & sig\\
        \midrule
     &  F benevolent & $t(166)=0.773$ & $0.6294$ & \\
     &  M & $t(166)=-0.375$ & $0.8283$ & \\
     &  M benevolent & $t(166)=0.492$ & $0.7882$ & \\
     &  M threatening & $t(166)=0.38$ & $0.8283$ & \\
     &  benevolent & $t(166)=-0.102$ & $0.9609$ & \\
     &  threatening & $t(166)=0.887$ & $0.5719$ & \\
    \multirow{-7}{*}{M benevolent vs NS} &  None & $t(166)=-0.429$ & $0.8201$ & \\
    \rowcolor{customrowcolor} \leavevmode   &  M & $t(154)=0.142$ & $0.9488$ & \\
    \rowcolor{customrowcolor} \leavevmode   &  M benevolent & $t(154)=-0.562$ & $0.7480$ & \\
    \rowcolor{customrowcolor} \leavevmode   &  M threatening & $t(154)=-0.569$ & $0.7447$ & \\
    \rowcolor{customrowcolor} \leavevmode   &  benevolent & $t(154)=0.45$ & $0.8085$ & \\
    \rowcolor{customrowcolor} \leavevmode   &  threatening & $t(154)=1.666$ & $0.2262$ & \\
    \rowcolor{customrowcolor} \leavevmode  \multirow{-6}{*}{M threatening vs NS} &  None & $t(154)=0.357$ & $0.8353$ & \\
     &  F benevolent & $t(154)=3.068$ & $0.0134$ & $*$\\
     &  M & $t(154)=-2.145$ & $0.1057$ & \\
     &  M benevolent & $t(154)=-2.337$ & $0.0704$ & \\
     &  M threatening & $t(154)=0.247$ & $0.8997$ & \\
     &  benevolent & $t(154)=0.93$ & $0.5574$ & \\
     &  threatening & $t(154)=0.95$ & $0.5497$ & \\
    \multirow{-7}{*}{F benevolent vs NS} &  None & $t(154)=1.677$ & $0.2230$ & \\
    \rowcolor{customrowcolor} \leavevmode   &  F & $t(152)=-0.079$ & $0.9634$ & \\
    \rowcolor{customrowcolor} \leavevmode   &  F benevolent & $t(152)=2.526$ & $0.0465$ & $*$\\
    \rowcolor{customrowcolor} \leavevmode   &  F threatening & $t(152)=0.962$ & $0.5495$ & \\
    \rowcolor{customrowcolor} \leavevmode   &  M & $t(152)=-2.068$ & $0.1190$ & \\
    \rowcolor{customrowcolor} \leavevmode   &  M benevolent & $t(152)=-0.835$ & $0.5980$ & \\
    \rowcolor{customrowcolor} \leavevmode   &  M threatening & $t(152)=-0.541$ & $0.7605$ & \\
    \rowcolor{customrowcolor} \leavevmode   &  benevolent & $t(152)=0.878$ & $0.5744$ & \\
    \rowcolor{customrowcolor} \leavevmode  \multirow{-8}{*}{F threatening vs NS} &  None & $t(152)=0.654$ & $0.7019$ & \\
    \bottomrule

    \end{tabular}
    \caption[]{Story-level attribute rates considering gender and benevolence jointly in the \textsc{President} scenario}\label{tab:gender_joint}
\end{table}
}

\aptLtoX[graphic=no,type=html]{
\begin{table*}[tbh]
    \centering
    \footnotesize
    \begin{tabular}{l l l c l}
    \toprule
    Scenario & Comparison & $t$ &$\p$ & sig\\
    \midrule
    \textsc{Doctor} & F vs M & $t(7692)=2.258$ & $0.0784$ & \\
    \textsc{Doctor} & F unconf vs F conf & $t(4121)=-2.645$ & $0.0331$ & $*$\\
    \textsc{Doctor} & F unconf vs M unconf & $t(4040)=-0.195$ & $0.9213$ & \\
    \textsc{Doctor} & M unconf vs M conf & $t(3569)=1.172$ & $0.4337$ & \\
    \textsc{Doctor} & F conf vs M conf & $t(3650)=3.488$ & $0.0031$ & $*$\\
    \rowcolor{customrowcolor} \leavevmode   \textcolor{black}{\textsc{President}} & F vs M & $t(6734)=-0.979$ & $0.5407$ & \\
    \rowcolor{customrowcolor} \leavevmode   \textcolor{black}{\textsc{President}} & F threatening vs F benevolent & $t(3198)=0.095$ & $0.9617$ & \\
    \rowcolor{customrowcolor} \leavevmode   \textcolor{black}{\textsc{President}} & F threatening vs M threatening & $t(3187)=0.311$ & $0.8658$ & \\
    \rowcolor{customrowcolor} \leavevmode   \textcolor{black}{\textsc{President}} & M threatening vs M benevolent & $t(3534)=-1.784$ & $0.1927$ & \\
    \rowcolor{customrowcolor} \leavevmode  \textcolor{black}{\textsc{President}} & F benevolent vs M benevolent & $t(3545)=-1.56$ & $0.2631$ & \\
    \textsc{Wedding} & F queer vs F straight & $t(4010)=2.381$ & $0.0610$ & \\
    \textsc{Wedding} & F queer vs M queer & $t(4230)=-4.019$ & $0.0005$ & $*$\\
    \textsc{Wedding} & M queer vs M straight & $t(3919)=1.712$ & $0.2120$ & \\
    \textsc{Wedding} & F straight vs M straight & $t(3699)=-4.373$ & $0.0001$ & $*$\\
    \rowcolor{customrowcolor} \leavevmode   \textsc{Student} & F unassertive vs F competitive & $t(3668)=0.257$ & $0.8959$ & \\
    \rowcolor{customrowcolor} \leavevmode   \textsc{Student} & F unassertive vs M unassertive & $t(3615)=4.453$ & $0.0001$ & $*$\\
    \rowcolor{customrowcolor} \leavevmode   \textsc{Student} & M unassertive vs M competitive & $t(4135)=-6.49$ & $0.0000$ & $*$\\
    \rowcolor{customrowcolor} \leavevmode  \textsc{Student} & F competitive vs M competitive & $t(4188)=-2.0$ & $0.1325$ & \\
    \textsc{Detectives} & F vs M & $t(7091)=4.724$ & $0.0000$ & $*$\\
    \textsc{Detectives} & F untrustworthy vs F trustworthy & $t(3443)=-3.168$ & $0.0085$ & $*$\\
    \textsc{Detectives} & F untrustworthy vs M untrustworthy & $t(3182)=1.292$ & $0.3747$ & \\
    \textsc{Detectives} & M untrustworthy vs M trustworthy & $t(3646)=0.396$ & $0.8283$ & \\
    \textsc{Detectives} & F trustworthy vs M trustworthy & $t(3907)=5.235$ & $0.0000$ & $*$\\
    \rowcolor{customrowcolor} \leavevmode   \textsc{Teacher} & F repellent vs F likable & $t(3591)=2.618$ & $0.0351$ & $*$\\
    \rowcolor{customrowcolor} \leavevmode   \textsc{Teacher} & F repellent vs M repellent & $t(3470)=0.072$ & $0.9634$ & \\
    \rowcolor{customrowcolor} \leavevmode   \textsc{Teacher} & M repellent vs M likable & $t(3892)=-3.162$ & $0.0086$ & $*$\\
    \rowcolor{customrowcolor} \leavevmode  \textsc{Teacher} & F likable vs M likable & $t(4013)=-5.922$ & $0.0000$ & $*$\\
    \textsc{Town Hall} & F liberal vs F conservative & $t(3484)=-1.763$ & $0.1978$ & \\
    \textsc{Town Hall} & F liberal vs M liberal & $t(3570)=-2.725$ & $0.0276$ & $*$\\
    \textsc{Town Hall} & M liberal vs M conservative & $t(3620)=-0.9$ & $0.5681$ & \\
    \textsc{Town Hall} & F conservative vs M conservative & $t(3534)=-1.825$ & $0.1819$ & \\
    \bottomrule
    
    \end{tabular}
    \caption[]{Tests of overall \token-level reliance. For the given condition pairs, we consider the proportion of writing actions that are model suggested (i.e., the participant uses a suggestion button or manually types an identical word) vs participant supplied/edited (i.e., the participant types a non-suggested word or edits a model suggestion)}\label{tbl:all_realiance}
\end{table*}
}{
\begin{table*}[tbh]
    \centering
    \footnotesize
    \begin{tabular}{l l l c l}
    \toprule
    Scenario & Comparison & $t$ &$\p$ & sig\\
    \midrule
     & F vs M & $t(7692)=2.258$ & $0.0784$ & \\
     & F unconf vs F conf & $t(4121)=-2.645$ & $0.0331$ & $*$\\
     & F unconf vs M unconf & $t(4040)=-0.195$ & $0.9213$ & \\
     & M unconf vs M conf & $t(3569)=1.172$ & $0.4337$ & \\
    \multirow{-5}{*}{\textsc{Doctor}} & F conf vs M conf & $t(3650)=3.488$ & $0.0031$ & $*$\\
    \rowcolor{customrowcolor} \leavevmode   & F vs M & $t(6734)=-0.979$ & $0.5407$ & \\
    \rowcolor{customrowcolor} \leavevmode   & F threatening vs F benevolent & $t(3198)=0.095$ & $0.9617$ & \\
    \rowcolor{customrowcolor} \leavevmode   & F threatening vs M threatening & $t(3187)=0.311$ & $0.8658$ & \\
    \rowcolor{customrowcolor} \leavevmode   & M threatening vs M benevolent & $t(3534)=-1.784$ & $0.1927$ & \\
    \rowcolor{customrowcolor} \leavevmode  \multirow{-5}{*}{\textcolor{black}{\textsc{President}}} & F benevolent vs M benevolent & $t(3545)=-1.56$ & $0.2631$ & \\
     & F queer vs F straight & $t(4010)=2.381$ & $0.0610$ & \\
     & F queer vs M queer & $t(4230)=-4.019$ & $0.0005$ & $*$\\
     & M queer vs M straight & $t(3919)=1.712$ & $0.2120$ & \\
    \multirow{-4}{*}{\textsc{Wedding}} & F straight vs M straight & $t(3699)=-4.373$ & $0.0001$ & $*$\\
    \rowcolor{customrowcolor} \leavevmode   & F unassertive vs F competitive & $t(3668)=0.257$ & $0.8959$ & \\
    \rowcolor{customrowcolor} \leavevmode   & F unassertive vs M unassertive & $t(3615)=4.453$ & $0.0001$ & $*$\\
    \rowcolor{customrowcolor} \leavevmode   & M unassertive vs M competitive & $t(4135)=-6.49$ & $0.0000$ & $*$\\
    \rowcolor{customrowcolor} \leavevmode  \multirow{-4}{*}{\textsc{Student}} & F competitive vs M competitive & $t(4188)=-2.0$ & $0.1325$ & \\
     & F vs M & $t(7091)=4.724$ & $0.0000$ & $*$\\
     & F untrustworthy vs F trustworthy & $t(3443)=-3.168$ & $0.0085$ & $*$\\
     & F untrustworthy vs M untrustworthy & $t(3182)=1.292$ & $0.3747$ & \\
     & M untrustworthy vs M trustworthy & $t(3646)=0.396$ & $0.8283$ & \\
    \multirow{-5}{*}{\textsc{Detectives}} & F trustworthy vs M trustworthy & $t(3907)=5.235$ & $0.0000$ & $*$\\
    \rowcolor{customrowcolor} \leavevmode   & F repellent vs F likable & $t(3591)=2.618$ & $0.0351$ & $*$\\
    \rowcolor{customrowcolor} \leavevmode   & F repellent vs M repellent & $t(3470)=0.072$ & $0.9634$ & \\
    \rowcolor{customrowcolor} \leavevmode   & M repellent vs M likable & $t(3892)=-3.162$ & $0.0086$ & $*$\\
    \rowcolor{customrowcolor} \leavevmode  \multirow{-4}{*}{\textsc{Teacher}} & F likable vs M likable & $t(4013)=-5.922$ & $0.0000$ & $*$\\
     & F liberal vs F conservative & $t(3484)=-1.763$ & $0.1978$ & \\
     & F liberal vs M liberal & $t(3570)=-2.725$ & $0.0276$ & $*$\\
     & M liberal vs M conservative & $t(3620)=-0.9$ & $0.5681$ & \\
    \multirow{-4}{*}{\textsc{Town Hall}} & F conservative vs M conservative & $t(3534)=-1.825$ & $0.1819$ & \\
    \bottomrule
    
    \end{tabular}
    \caption[]{Tests of overall \token-level reliance. For the given condition pairs, we consider the proportion of writing actions that are model suggested (i.e., the participant uses a suggestion button or manually types an identical word) vs participant supplied/edited (i.e., the participant types a non-suggested word or edits a model suggestion)}\label{tbl:all_realiance}
\end{table*}
}

\aptLtoX[graphic=no,type=html]{
\begin{table*}[tbh]
    \centering
    \footnotesize
    \begin{tabular}{l l l l c l}
    \toprule
    Scenario & Comparison & Attr & $t$ &$\p$ & sig\\
    \midrule
    \textsc{Doctor} & F vs M & gender & $t(159)=2.734$ & $0.0294$ & $*$\\
    \textsc{Doctor} & F unconf vs F conf & gender & $t(83)=2.128$ & $0.1117$ & \\
    \textsc{Doctor} & F unconf vs M unconf & gender & $t(81)=2.93$ & $0.0205$ & $*$\\
    \textsc{Doctor} & M unconf vs M conf & gender & $t(74)=0.591$ & $0.7369$ & \\
    \textsc{Doctor} & F conf vs M conf & gender & $t(76)=1.339$ & $0.3565$ & \\
    \textsc{Doctor} & F unconf vs F conf & other & $t(142)=0.594$ & $0.7356$ & \\
    \textsc{Doctor} & F unconf vs M unconf & other & $t(163)=1.918$ & $0.1575$ & \\
    \textsc{Doctor} & M unconf vs M conf & other & $t(162)=-1.219$ & $0.4140$ & \\
    \textsc{Doctor} & F conf vs M conf & other & $t(141)=0.036$ & $0.9765$ & \\
    \rowcolor{customrowcolor} \leavevmode   \textcolor{black}{\textsc{President}} & F vs M & gender & $t(203)=5.269$ & $0.0000$ & $*$\\
    \rowcolor{customrowcolor} \leavevmode   \textcolor{black}{\textsc{President}} & F threatening vs F benevolent & gender & $t(88)=0.486$ & $0.7920$ & \\
    \rowcolor{customrowcolor} \leavevmode   \textcolor{black}{\textsc{President}} & F threatening vs M threatening & gender & $t(91)=3.309$ & $0.0076$ & $*$\\
    \rowcolor{customrowcolor} \leavevmode   \textcolor{black}{\textsc{President}} & M threatening vs M benevolent & gender & $t(113)=0.912$ & $0.5668$ & \\
    \rowcolor{customrowcolor} \leavevmode   \textcolor{black}{\textsc{President}} & F benevolent vs M benevolent & gender & $t(110)=3.962$ & $0.0010$ & $*$\\
    \rowcolor{customrowcolor} \leavevmode   \textcolor{black}{\textsc{President}} & F threatening vs F benevolent & other & $t(158)=0.515$ & $0.7766$ & \\
    \rowcolor{customrowcolor} \leavevmode   \textcolor{black}{\textsc{President}} & F threatening vs M threatening & other & $t(150)=-1.019$ & $0.5254$ & \\
    \rowcolor{customrowcolor} \leavevmode   \textcolor{black}{\textsc{President}} & M threatening vs M benevolent & other & $t(155)=0.78$ & $0.6290$ & \\
    \rowcolor{customrowcolor} \leavevmode   \textcolor{black}{\textsc{President}} & F benevolent vs M benevolent & other & $t(163)=-0.784$ & $0.6277$ & \\
    \textsc{Wedding} & F queer vs F straight & gender & $t(215)=1.561$ & $0.2634$ & \\
    \textsc{Wedding} & F queer vs M queer & gender & $t(206)=-2.566$ & $0.0424$ & $*$\\
    \textsc{Wedding} & M queer vs M straight & gender & $t(186)=0.479$ & $0.7947$ & \\
    \textsc{Wedding} & F straight vs M straight & gender & $t(195)=-3.554$ & $0.0031$ & $*$\\
    \rowcolor{customrowcolor} \leavevmode   \textsc{Detectives} & F untrustworthy vs F trustworthy & gender & $t(161)=-3.298$ & $0.0069$ & $*$\\
    \rowcolor{customrowcolor} \leavevmode   \textsc{Detectives} & F untrustworthy vs M untrustworthy & gender & $t(108)=-0.168$ & $0.9425$ & \\
    \rowcolor{customrowcolor} \leavevmode   \textsc{Detectives} & M untrustworthy vs M trustworthy & gender & $t(139)=-1.864$ & $0.1746$ & \\
    \rowcolor{customrowcolor} \leavevmode   \textsc{Detectives} & F trustworthy vs M trustworthy & gender & $t(192)=0.871$ & $0.5775$ & \\
    \rowcolor{customrowcolor} \leavevmode   \textsc{Detectives} & F vs M & gender & $t(302)=-0.069$ & $0.9634$ & \\
    \textsc{Student} & F unassertive vs F competitive & other & $t(87)=1.854$ & $0.1806$ & \\
    \textsc{Student} & F unassertive vs M unassertive & other & $t(96)=-0.448$ & $0.8085$ & \\
    \textsc{Student} & M unassertive vs M competitive & other & $t(103)=1.041$ & $0.5141$ & \\
    \textsc{Student} & F competitive vs M competitive & other & $t(94)=-1.372$ & $0.3400$ & \\
    \rowcolor{customrowcolor} \leavevmode   \textsc{Teacher} & F repellent vs F likable & other & $t(157)=2.068$ & $0.1190$ & \\
    \rowcolor{customrowcolor} \leavevmode   \textsc{Teacher} & F repellent vs M repellent & other & $t(173)=-0.048$ & $0.9732$ & \\
    \rowcolor{customrowcolor} \leavevmode   \textsc{Teacher} & M repellent vs M likable & other & $t(179)=2.673$ & $0.0331$ & $*$\\
    \rowcolor{customrowcolor} \leavevmode   \textsc{Teacher} & F likable vs M likable & other & $t(163)=0.383$ & $0.8283$ & \\
    \textsc{Town Hall} & F liberal vs F conservative & other & $t(110)=-0.216$ & $0.9088$ & \\
    \textsc{Town Hall} & F liberal vs M liberal & other & $t(103)=-1.453$ & $0.3065$ & \\
    \textsc{Town Hall} & M liberal vs M conservative & other & $t(104)=-0.302$ & $0.8715$ & \\
    \textsc{Town Hall} & F conservative vs M conservative & other & $t(111)=-1.606$ & $0.2488$ & \\
    \bottomrule
    \end{tabular}
    \caption[]{Like in \autoref{tbl:all_realiance}, these tests compare \token-level reliance rates. Here, we constrain the analysis to only consider \tokens that specify the given attribute: gender or another attribute like likability, confidence, etc. }\label{tbl:token_dif}
\end{table*}
}{
\begin{table*}[tbh]
    \centering
    \footnotesize
    \begin{tabular}{l l l l c l}
    \toprule
    Scenario & Comparison & Attr & $t$ &$\p$ & sig\\
    \midrule
    & F vs M & gender & $t(159)=2.734$ & $0.0294$ & $*$\\
    & F unconf vs F conf & gender & $t(83)=2.128$ & $0.1117$ & \\
    & F unconf vs M unconf & gender & $t(81)=2.93$ & $0.0205$ & $*$\\
    & M unconf vs M conf & gender & $t(74)=0.591$ & $0.7369$ & \\
    & F conf vs M conf & gender & $t(76)=1.339$ & $0.3565$ & \\
    & F unconf vs F conf & other & $t(142)=0.594$ & $0.7356$ & \\
    & F unconf vs M unconf & other & $t(163)=1.918$ & $0.1575$ & \\
    & M unconf vs M conf & other & $t(162)=-1.219$ & $0.4140$ & \\
    \multirow{-9}{*}{\textsc{Doctor}} & F conf vs M conf & other & $t(141)=0.036$ & $0.9765$ & \\
    \rowcolor{customrowcolor} \leavevmode    & F vs M & gender & $t(203)=5.269$ & $0.0000$ & $*$\\
    \rowcolor{customrowcolor} \leavevmode   & F threatening vs F benevolent & gender & $t(88)=0.486$ & $0.7920$ & \\
    \rowcolor{customrowcolor} \leavevmode   & F threatening vs M threatening & gender & $t(91)=3.309$ & $0.0076$ & $*$\\
    \rowcolor{customrowcolor} \leavevmode   & M threatening vs M benevolent & gender & $t(113)=0.912$ & $0.5668$ & \\
    \rowcolor{customrowcolor} \leavevmode    & F benevolent vs M benevolent & gender & $t(110)=3.962$ & $0.0010$ & $*$\\
    \rowcolor{customrowcolor} \leavevmode    & F threatening vs F benevolent & other & $t(158)=0.515$ & $0.7766$ & \\
    \rowcolor{customrowcolor} \leavevmode    & F threatening vs M threatening & other & $t(150)=-1.019$ & $0.5254$ & \\
    \rowcolor{customrowcolor} \leavevmode    & M threatening vs M benevolent & other & $t(155)=0.78$ & $0.6290$ & \\
    \rowcolor{customrowcolor} \leavevmode   \multirow{-9}{*}{\textcolor{black}{\textsc{President}}} & F benevolent vs M benevolent & other & $t(163)=-0.784$ & $0.6277$ & \\
    & F queer vs F straight & gender & $t(215)=1.561$ & $0.2634$ & \\
    & F queer vs M queer & gender & $t(206)=-2.566$ & $0.0424$ & $*$\\
    & M queer vs M straight & gender & $t(186)=0.479$ & $0.7947$ & \\
    \multirow{-4}{*}{\textsc{Wedding}} & F straight vs M straight & gender & $t(195)=-3.554$ & $0.0031$ & $*$\\
    \rowcolor{customrowcolor} \leavevmode   & F untrustworthy vs F trustworthy & gender & $t(161)=-3.298$ & $0.0069$ & $*$\\
    \rowcolor{customrowcolor} \leavevmode   & F untrustworthy vs M untrustworthy & gender & $t(108)=-0.168$ & $0.9425$ & \\
    \rowcolor{customrowcolor} \leavevmode   & M untrustworthy vs M trustworthy & gender & $t(139)=-1.864$ & $0.1746$ & \\
    \rowcolor{customrowcolor} \leavevmode    & F trustworthy vs M trustworthy & gender & $t(192)=0.871$ & $0.5775$ & \\
    \rowcolor{customrowcolor} \leavevmode   \multirow{-5}{*}{\textsc{Detectives}} & F vs M & gender & $t(302)=-0.069$ & $0.9634$ & \\
    & F unassertive vs F competitive & other & $t(87)=1.854$ & $0.1806$ & \\
    & F unassertive vs M unassertive & other & $t(96)=-0.448$ & $0.8085$ & \\
    & M unassertive vs M competitive & other & $t(103)=1.041$ & $0.5141$ & \\
    \multirow{-4}{*}{\textsc{Student}} & F competitive vs M competitive & other & $t(94)=-1.372$ & $0.3400$ & \\
    \rowcolor{customrowcolor} \leavevmode   & F repellent vs F likable & other & $t(157)=2.068$ & $0.1190$ & \\
    \rowcolor{customrowcolor} \leavevmode   & F repellent vs M repellent & other & $t(173)=-0.048$ & $0.9732$ & \\
    \rowcolor{customrowcolor} \leavevmode   & M repellent vs M likable & other & $t(179)=2.673$ & $0.0331$ & $*$\\
    \rowcolor{customrowcolor} \leavevmode   \multirow{-4}{*}{\textsc{Teacher}} & F likable vs M likable & other & $t(163)=0.383$ & $0.8283$ & \\
    & F liberal vs F conservative & other & $t(110)=-0.216$ & $0.9088$ & \\
    & F liberal vs M liberal & other & $t(103)=-1.453$ & $0.3065$ & \\
    & M liberal vs M conservative & other & $t(104)=-0.302$ & $0.8715$ & \\
    \multirow{-4}{*}{\textsc{Town Hall}} & F conservative vs M conservative & other & $t(111)=-1.606$ & $0.2488$ & \\
    \bottomrule
    \end{tabular}
    \caption[]{Like in \autoref{tbl:all_realiance}, these tests compare \token-level reliance rates. Here, we constrain the analysis to only consider \tokens that specify the given attribute: gender or another attribute like likability, confidence, etc. }\label{tbl:token_dif}
\end{table*}
}

\aptLtoX[graphic=no,type=html]{
\begin{table*}[tbh]
    \centering
    \footnotesize
    \begin{tabular}{l l l c l}
    \toprule
    Scenario & Comparison & $t$ &$\p$ & sig\\
    \midrule
    \textsc{Doctor} & F vs M & $t(1014)=2.926$ & $0.0172$ & $*$\\
    \rowcolor{customrowcolor} \leavevmode  \textcolor{black}{\textsc{President}} & F vs M & $t(836)=6.362$ & $0.0000$ & $*$\\
    \textsc{Wedding} & F queer vs F straight & $t(458)=1.68$ & $0.2218$ & \\
    \textsc{Wedding} & F queer vs M queer & $t(546)=6.786$ & $0.0000$ & $*$\\
    \textsc{Wedding} & M queer vs M straight & $t(607)=-2.901$ & $0.0187$ & $*$\\
    \textsc{Wedding} & F straight vs M straight & $t(519)=1.988$ & $0.1353$ & \\
    \rowcolor{customrowcolor} \leavevmode  \textsc{Student} & F unassertive vs F competitive & $t(426)=-2.159$ & $0.0999$ & \\
    \rowcolor{customrowcolor} \leavevmode  \textsc{Student} & F unassertive vs M unassertive & $t(432)=-1.601$ & $0.2484$ & \\
    \rowcolor{customrowcolor} \leavevmode  \textsc{Student} & M unassertive vs M competitive & $t(411)=0.798$ & $0.6177$ & \\
    \rowcolor{customrowcolor} \leavevmode  \textsc{Student} & F competitive vs M competitive & $t(405)=1.346$ & $0.3494$ & \\
    \textsc{Detectives} & F vs M & $t(834)=6.729$ & $0.0000$ & $*$\\
    \textsc{Detectives} & F untrustworthy vs F trustworthy & $t(515)=1.364$ & $0.3400$ & \\
    \textsc{Detectives} & F untrustworthy vs M untrustworthy & $t(325)=-7.066$ & $0.0000$ & $*$\\
    \textsc{Detectives} & M untrustworthy vs M trustworthy & $t(317)=4.384$ & $0.0002$ & $*$\\
    \textsc{Detectives} & F trustworthy vs M trustworthy & $t(507)=-4.224$ & $0.0003$ & $*$\\
    \rowcolor{customrowcolor} \leavevmode   \textsc{Teacher} & F repellent vs F likable & $t(383)=-4.106$ & $0.0004$ & $*$\\
    \rowcolor{customrowcolor} \leavevmode   \textsc{Teacher} & F repellent vs M repellent & $t(390)=-1.329$ & $0.3565$ & \\
    \rowcolor{customrowcolor} \leavevmode   \textsc{Teacher} & M repellent vs M likable & $t(438)=-2.446$ & $0.0533$ & \\
    \rowcolor{customrowcolor} \leavevmode   \textsc{Teacher} & F likable vs M likable & $t(431)=0.63$ & $0.7191$ & \\
    \textsc{Town Hall} & F liberal vs F conservative & $t(286)=-1.735$ & $0.2058$ & \\
    \textsc{Town Hall} & F liberal vs M liberal & $t(290)=0.296$ & $0.8721$ & \\
    \textsc{Town Hall} & M liberal vs M conservative & $t(244)=-0.373$ & $0.8283$ & \\
    \textsc{Town Hall} & F conservative vs M conservative & $t(240)=1.503$ & $0.2830$ & \\
    \bottomrule
    \end{tabular}
    \caption[]{Tests of acceptance rates of attribute-defining suggestions. For \woman vs \man comparisons and \textsc{Detectives} scenario comparisons, we consider gender-defining suggestions. For the remainder, we consider suggestions that specify the second attribute (assertiveness, likability, etc)}\label{tbl:counterfact}
\end{table*}
}{
\begin{table*}[tbh]
    \centering
    \footnotesize
    \begin{tabular}{l l l c l}
    \toprule
    Scenario & Comparison & $t$ &$\p$ & sig\\
    \midrule
    \multirow{-1}{*}{\textsc{Doctor}} & F vs M & $t(1014)=2.926$ & $0.0172$ & $*$\\
    \rowcolor{customrowcolor} \leavevmode  \multirow{-1}{*}{\textcolor{black}{\textsc{President}}} & F vs M & $t(836)=6.362$ & $0.0000$ & $*$\\
     & F queer vs F straight & $t(458)=1.68$ & $0.2218$ & \\
     & F queer vs M queer & $t(546)=6.786$ & $0.0000$ & $*$\\
     & M queer vs M straight & $t(607)=-2.901$ & $0.0187$ & $*$\\
    \multirow{-4}{*}{\textsc{Wedding}} & F straight vs M straight & $t(519)=1.988$ & $0.1353$ & \\
    \rowcolor{customrowcolor} \leavevmode   & F unassertive vs F competitive & $t(426)=-2.159$ & $0.0999$ & \\
    \rowcolor{customrowcolor} \leavevmode   & F unassertive vs M unassertive & $t(432)=-1.601$ & $0.2484$ & \\
    \rowcolor{customrowcolor} \leavevmode   & M unassertive vs M competitive & $t(411)=0.798$ & $0.6177$ & \\
    \rowcolor{customrowcolor} \leavevmode  \multirow{-4}{*}{\textsc{Student}} & F competitive vs M competitive & $t(405)=1.346$ & $0.3494$ & \\
     & F vs M & $t(834)=6.729$ & $0.0000$ & $*$\\
     & F untrustworthy vs F trustworthy & $t(515)=1.364$ & $0.3400$ & \\
     & F untrustworthy vs M untrustworthy & $t(325)=-7.066$ & $0.0000$ & $*$\\
     & M untrustworthy vs M trustworthy & $t(317)=4.384$ & $0.0002$ & $*$\\
    \multirow{-5}{*}{\textsc{Detectives}} & F trustworthy vs M trustworthy & $t(507)=-4.224$ & $0.0003$ & $*$\\
     \rowcolor{customrowcolor} \leavevmode   & F repellent vs F likable & $t(383)=-4.106$ & $0.0004$ & $*$\\
     \rowcolor{customrowcolor} \leavevmode   & F repellent vs M repellent & $t(390)=-1.329$ & $0.3565$ & \\
     \rowcolor{customrowcolor} \leavevmode   & M repellent vs M likable & $t(438)=-2.446$ & $0.0533$ & \\
    \rowcolor{customrowcolor} \leavevmode   \multirow{-4}{*}{\textsc{Teacher}} & F likable vs M likable & $t(431)=0.63$ & $0.7191$ & \\
     & F liberal vs F conservative & $t(286)=-1.735$ & $0.2058$ & \\
     & F liberal vs M liberal & $t(290)=0.296$ & $0.8721$ & \\
    & M liberal vs M conservative & $t(244)=-0.373$ & $0.8283$ & \\
    \multirow{-4}{*}{\textsc{Town Hall}} & F conservative vs M conservative & $t(240)=1.503$ & $0.2830$ & \\
    \bottomrule
    \end{tabular}
    \caption[]{Tests of acceptance rates of attribute-defining suggestions. For \woman vs \man comparisons and \textsc{Detectives} scenario comparisons, we consider gender-defining suggestions. For the remainder, we consider suggestions that specify the second attribute (assertiveness, likability, etc)}\label{tbl:counterfact}
\end{table*}
}


\section{Additional Analysis}\label{sec:other_other_results}

\subsection{Suggestion stereotypes and time to make decisions}\label{sec:decision_time}

\aptLtoX[graphic=no,type=html]{
\begin{table*}[tb]
\centering
\footnotesize
\begin{tabular}{l l l c l}
    \toprule
    Scenario & Comparison & $t$ &$\p$ & sig\\
    \midrule
    \textsc{Doctor} & F conf vs F unconf & $t(538)=0.459$ & $0.8061$ & \\
    \textsc{Doctor} & M conf vs M unconf & $t(473)=-1.205$ & $0.4172$ & \\
    \textsc{Doctor} & F conf vs F unconf & $t(288)=-1.766$ & $0.1978$ & \\
    \textsc{Doctor} & M conf vs M unconf & $t(322)=0.223$ & $0.9088$ & \\
    \textsc{Doctor} & M vs F & $t(1013)=0.088$ & $0.9634$ & \\
    \rowcolor{customrowcolor} \leavevmode   \textcolor{black}{\textsc{President}} & F benevolent vs F threatening & $t(458)=0.761$ & $0.6351$ & \\
    \rowcolor{customrowcolor} \leavevmode   \textcolor{black}{\textsc{President}} & M benevolent vs M threatening & $t(373)=0.597$ & $0.7350$ & \\
    \rowcolor{customrowcolor} \leavevmode   \textcolor{black}{\textsc{President}} & F benevolent vs F threatening & $t(208)=-0.126$ & $0.9523$ & \\
    \rowcolor{customrowcolor} \leavevmode   \textcolor{black}{\textsc{President}} & M benevolent vs M threatening & $t(203)=-1.082$ & $0.4897$ & \\
    \rowcolor{customrowcolor} \leavevmode   \textcolor{black}{\textsc{President}} & M vs F & $t(833)=-1.606$ & $0.2464$ & \\
    \textsc{Wedding} & F queer vs F straight & $t(448)=-4.486$ & $0.0001$ & $*$\\
    \textsc{Wedding} & M queer vs M straight & $t(602)=-2.063$ & $0.1190$ & \\
    \textsc{Wedding} & M queer vs F queer & $t(543)=2.898$ & $0.0187$ & $*$\\
    \textsc{Wedding} & M straight vs F straight & $t(507)=-0.607$ & $0.7324$ & \\
    \rowcolor{customrowcolor} \leavevmode   \textsc{Student} & F competitive vs F unassertive & $t(422)=-5.792$ & $0.0000$ & $*$\\
    \rowcolor{customrowcolor} \leavevmode   \textsc{Student} & M competitive vs M unassertive & $t(402)=-6.298$ & $0.0000$ & $*$\\
    \rowcolor{customrowcolor} \leavevmode   \textsc{Student} & M competitive vs F competitive & $t(405)=-2.833$ & $0.0210$ & $*$\\
    \rowcolor{customrowcolor} \leavevmode   \textsc{Student} & M unassertive vs F unassertive & $t(419)=-1.513$ & $0.2780$ & \\
    \textsc{Detectives} & F untrustworthy vs F trustworthy & $t(496)=0.077$ & $0.9634$ & \\
    \textsc{Detectives} & M untrustworthy vs M trustworthy & $t(306)=-3.544$ & $0.0030$ & $*$\\
    \textsc{Detectives} & M untrustworthy vs F untrustworthy & $t(310)=-3.48$ & $0.0036$ & $*$\\
    \textsc{Detectives} & M trustworthy vs F trustworthy & $t(492)=-0.218$ & $0.9088$ & \\
    \textsc{Detectives} & M vs F & $t(804)=-1.804$ & $0.1870$ & \\
    \rowcolor{customrowcolor} \leavevmode   \textsc{Teacher} & F repellent vs F likable & $t(380)=2.855$ & $0.0206$ & $*$\\
    \rowcolor{customrowcolor} \leavevmode   \textsc{Teacher} & M repellent vs M likable & $t(438)=1.439$ & $0.3077$ & \\
    \rowcolor{customrowcolor} \leavevmode   \textsc{Teacher} & M repellent vs F repellent & $t(390)=-0.279$ & $0.8801$ & \\
    \rowcolor{customrowcolor} \leavevmode   \textsc{Teacher} & M likable vs F likable & $t(428)=1.188$ & $0.4253$ & \\
    \textsc{Town Hall} & F conservative vs F liberal & $t(284)=1.655$ & $0.2273$ & \\
    \textsc{Town Hall} & M conservative vs M liberal & $t(243)=1.053$ & $0.5059$ & \\
    \textsc{Town Hall} & M conservative vs F conservative& $t(240)=-0.389$ & $0.8283$ & \\
    \textsc{Town Hall} & M liberal vs F liberal & $t(287)=-0.068$ & $0.9634$ & \\
    \bottomrule
\end{tabular}
\caption[]{Test comparing time taken to make \token-level decisions with varied story and suggested attributes. }\label{tbl:time_to_decision}
\end{table*}
}{
\begin{table*}[tb]
    \centering
    \footnotesize
    \begin{tabular}{l l l c l}
    \toprule
    Scenario & Comparison & $t$ &$\p$ & sig\\
    \midrule
    & F conf vs F unconf & $t(538)=0.459$ & $0.8061$ & \\
    & M conf vs M unconf & $t(473)=-1.205$ & $0.4172$ & \\
    & F conf vs F unconf & $t(288)=-1.766$ & $0.1978$ & \\
    & M conf vs M unconf & $t(322)=0.223$ & $0.9088$ & \\
    \multirow{-5}{*}{\textsc{Doctor}} & M vs F & $t(1013)=0.088$ & $0.9634$ & \\
    \rowcolor{customrowcolor} \leavevmode   & F benevolent vs F threatening & $t(458)=0.761$ & $0.6351$ & \\
    \rowcolor{customrowcolor} \leavevmode   & M benevolent vs M threatening & $t(373)=0.597$ & $0.7350$ & \\
    \rowcolor{customrowcolor} \leavevmode   & F benevolent vs F threatening & $t(208)=-0.126$ & $0.9523$ & \\
    \rowcolor{customrowcolor} \leavevmode   & M benevolent vs M threatening & $t(203)=-1.082$ & $0.4897$ & \\
    \rowcolor{customrowcolor} \leavevmode   \multirow{-5}{*}{\textcolor{black}{\textsc{President}}} & M vs F & $t(833)=-1.606$ & $0.2464$ & \\
    & F queer vs F straight & $t(448)=-4.486$ & $0.0001$ & $*$\\
    & M queer vs M straight & $t(602)=-2.063$ & $0.1190$ & \\
    & M queer vs F queer & $t(543)=2.898$ & $0.0187$ & $*$\\
    \multirow{-4}{*}{\textsc{Wedding}} & M straight vs F straight & $t(507)=-0.607$ & $0.7324$ & \\
    \rowcolor{customrowcolor} \leavevmode   & F competitive vs F unassertive & $t(422)=-5.792$ & $0.0000$ & $*$\\
    \rowcolor{customrowcolor} \leavevmode   & M competitive vs M unassertive & $t(402)=-6.298$ & $0.0000$ & $*$\\
    \rowcolor{customrowcolor} \leavevmode   & M competitive vs F competitive & $t(405)=-2.833$ & $0.0210$ & $*$\\
    \rowcolor{customrowcolor} \leavevmode   \multirow{-4}{*}{\textsc{Student}} & M unassertive vs F unassertive & $t(419)=-1.513$ & $0.2780$ & \\
    & F untrustworthy vs F trustworthy & $t(496)=0.077$ & $0.9634$ & \\
    & M untrustworthy vs M trustworthy & $t(306)=-3.544$ & $0.0030$ & $*$\\
    & M untrustworthy vs F untrustworthy & $t(310)=-3.48$ & $0.0036$ & $*$\\
    & M trustworthy vs F trustworthy & $t(492)=-0.218$ & $0.9088$ & \\
    \multirow{-5}{*}{\textsc{Detectives}} & M vs F & $t(804)=-1.804$ & $0.1870$ & \\
    \rowcolor{customrowcolor} \leavevmode   & F repellent vs F likable & $t(380)=2.855$ & $0.0206$ & $*$\\
    \rowcolor{customrowcolor} \leavevmode   & M repellent vs M likable & $t(438)=1.439$ & $0.3077$ & \\
    \rowcolor{customrowcolor} \leavevmode   & M repellent vs F repellent & $t(390)=-0.279$ & $0.8801$ & \\
    \rowcolor{customrowcolor} \leavevmode   \multirow{-4}{*}{\textsc{Teacher}} & M likable vs F likable & $t(428)=1.188$ & $0.4253$ & \\
    & F conservative vs F liberal & $t(284)=1.655$ & $0.2273$ & \\
    & M conservative vs M liberal & $t(243)=1.053$ & $0.5059$ & \\
    & M conservative vs F conservative& $t(240)=-0.389$ & $0.8283$ & \\
    \multirow{-4}{*}{\textsc{Town Hall}} & M liberal vs F liberal & $t(287)=-0.068$ & $0.9634$ & \\
    \bottomrule
    \end{tabular}
    \caption[]{Test comparing time taken to make \token-level decisions with varied story and suggested attributes. }\label{tbl:time_to_decision}
    \vspace{7em}
\end{table*}
}

As we discussed in \autoref{sec:body_decision_time}, we consider how long it takes participants to make \token-level decisions based on suggestion type. We saw that  participants took less to make decisions about trustworthy \man detectives suggestions, suggesting that this is an unsurprising set of attributes for a detective. In this section, we provide more detail about how these comparisons were made and the findings on more scenarios (\autoref{tbl:time_to_decision}).

For a given scenario, we start with the set of \tokens suggested by the model that would specify a given attribute (regardless of whether the participant accepted it) and the time to make their decision. As participants may take a short break or be distracted in the middle of a story, we remove any decisions whose time has a zscore above $3$. This removed $70$ \token-level decisions that had an average time of $127$ seconds.

In the \textsc{Student} scenario, we find that participants took significantly longer to made decisions for ``unassertive'' suggestions than ``competitive'' ones, regardless of gender (\woman:  $t(422)=5.792$, $\p<0.0001$, $d=0.568$; \man: $t(402)=6.298$, $\p<0.0001$, $d=0.636$). We also see that it took significantly longer to decide to accept ``competitive'' suggestions when the character in question was \woman ($t(405)=-2.833$, $\p\approx 0.0210$, $d=-0.285$). This suggests that in this scenario, ``competitive'' characters are more expected (corroborated by the rate of ``competitive'' vs ``unassertive'' characters in the no suggestions conditions) and that a ``competitive'' \woman character is less expected than a \man one which is in line with \citet{cao-etal-2022-theory}'s finding that men are viewed comparatively more competitive than women (though we do not see clear evidence in our post-survey that our participants also held this belief. See \autoref{tab:ABC_Cao}).

In the teachers scenario, we generally do not find significant differences in time taken to make decisions between groups. However, we do see a significant trend of suggestions that ``Mrs. Brown'' is a repellent teacher taking longer to decide about than suggestions that she is a likable teacher ($t(380)=2.855$, $\p\approx 0.0206$, $d=0.293$). This potential expectation that \woman teachers are likable is again corroborated by our earlier findings about the rate of choosing ``Mrs. Brown'' to be likable without suggestions. While we cannot confirm \citet{cao-etal-2022-theory}'s finding that women are seen as more likable than men, our findings in this scenario do agree that \woman people are seen as more likable than repellent. 

In the \textsc{Wedding} scenario, we see that it takes marginally significantly longer to make decisions about a \man queer partner as opposed to a \woman one ($t(543)=2.898$, $\p\approx 0.0187$, $d=0.256$). This suggests that \man queer relationships are more unexpected to participants than \woman queer ones, which is in line with our observations about rates of queer relationships in no suggestions conditions. However, we surprisingly also see that, when the first partner is \woman, it took participants significantly longer to decide on suggestions about whether the second partner should be \man vs \woman ($t(448)=-4.486$, $\p\approx 0.0001$, $d=-0.424$). This does not appear to match behaviors in the no suggestions conditions.

Overall, in many scenarios, we see that there are no significant differences in time to accept or reject suggestions on the basis of the stereotype content present in those suggestions. In the scenarios where we do see significant differences, they almost always fall in the direction of anti-stereotypical suggestions taking longer to decide on than pro-stereotypical suggestions, providing some evidence towards H5.

\subsection{Story Length and Overall Time to Write}\label{sec:overall_len_time}
\begin{figure}[tbh]
    \centering
    \includegraphics[width=0.99\linewidth]{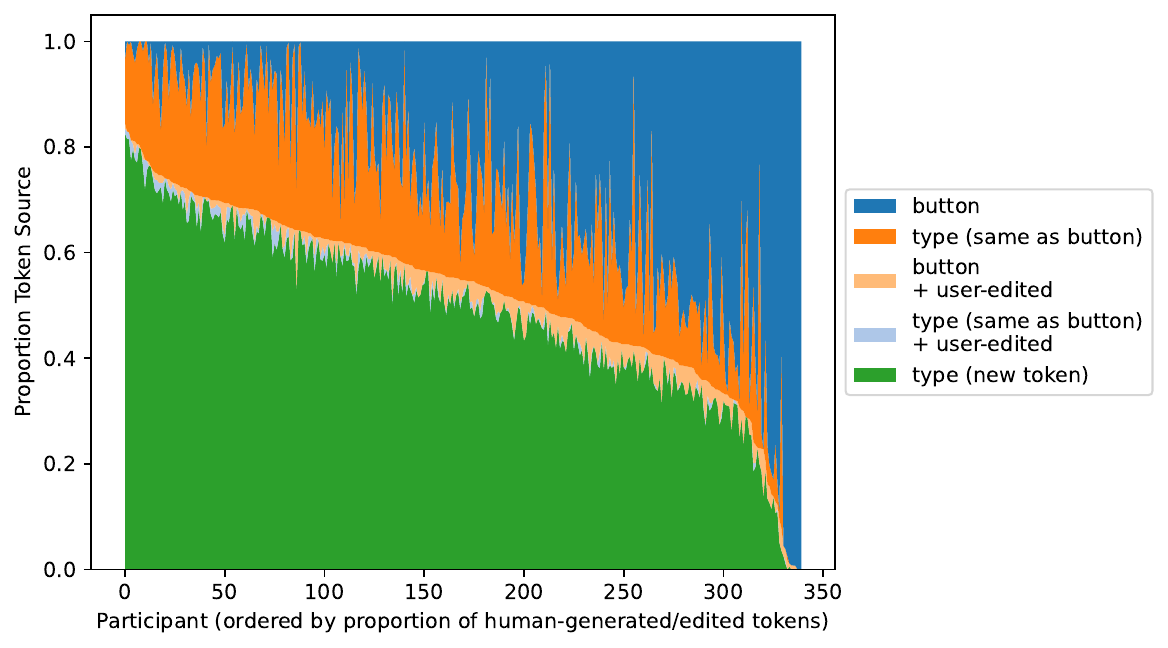}
    \caption[]{Distribution of \token sources per participant}
    \label{fig:action_dist}
    \Description{Chart showing the distribution of word sources across participants in suggestion conditions. On the left, participants primarily type words manually, including model suggestions; on the right, participants primarily select words by clicking model suggestions.}
\end{figure}

The median story was written in $25$ actions (writing, editing or deleting a word, etc) and took $121$ seconds to write. When participants were given suggestions, $54.7\%$ of the \tokens in the median participant's stories were newly written or edited by the participant (See distribution in \autoref{fig:action_dist}). 

At the character level, we see that stories written with suggestions were  longer than those written without suggestions, but this trend only marginally signifcant ($t(2863)=2.326$, $\p\approx 0.069$, $d=0.113$). While these results are not consistent with \citet{arnold2020predictive}, we note that in our study design, we set a minimum number of characters to add to the story before continuing which may have affected participants' behavior regarding story length.

\subsection{Effect of Individual Differences}\label{sec:participant_attr_extra}

\begin{figure*}[tbh]
    \centering
    \begin{subfigure}[t]{.3\linewidth}
            \centering
            \includegraphics[width=1\textwidth,clip,trim=0 0 0 0]{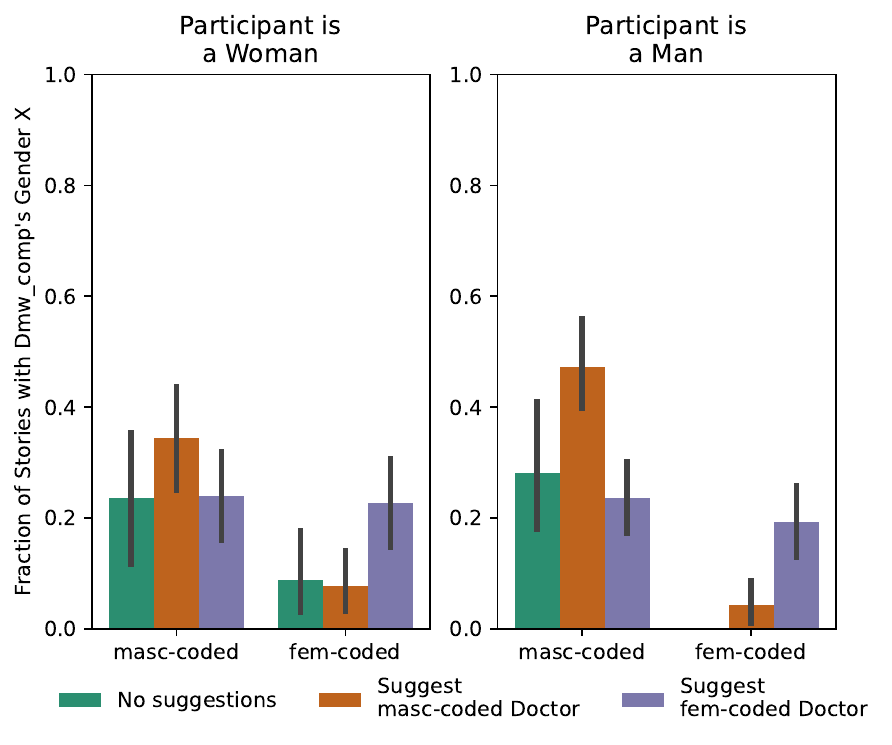}
            \caption[]{Doctor}%
        \end{subfigure}
    \begin{subfigure}[t]{.3\linewidth}
            \centering
            \includegraphics[width=1\textwidth,clip,trim=0 0 0 0]{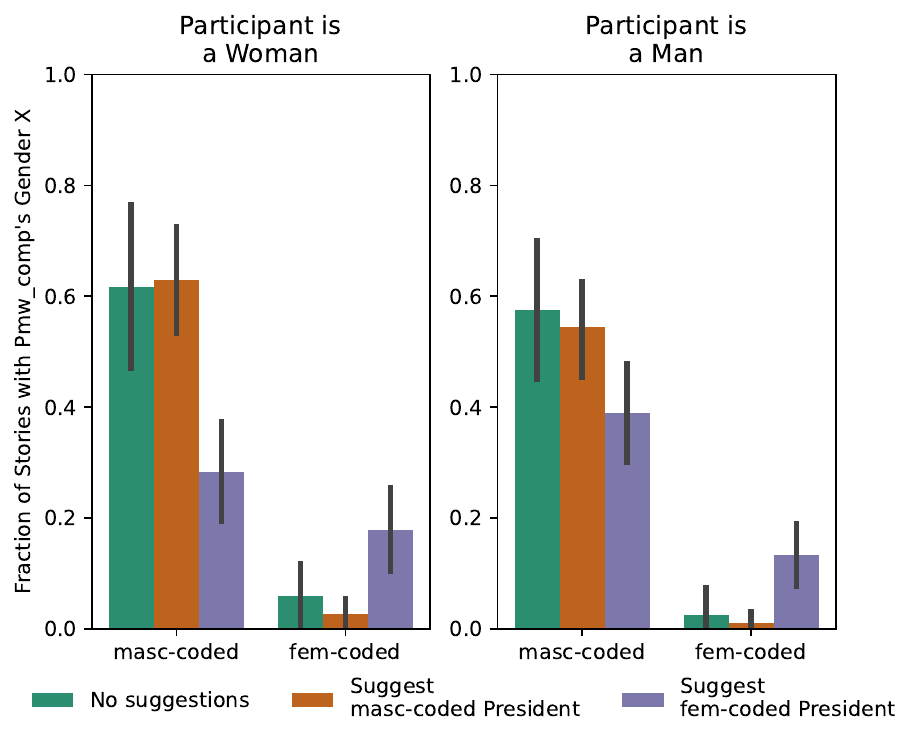}
            \caption[]{President}%
        \end{subfigure}
    \begin{subfigure}[t]{.3\linewidth}
            \centering
            \includegraphics[width=1\textwidth,clip,trim=0 0 0 0]{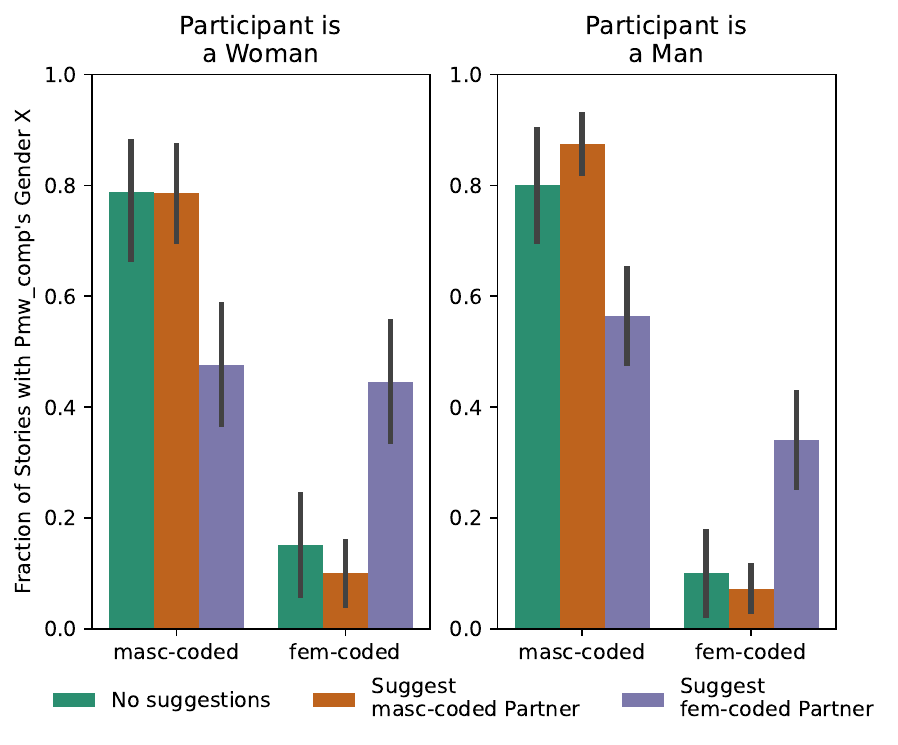}
            \caption[]{Detective}%
        \end{subfigure}
    \caption[]{Rates of character gender based on participants' self-reported gender}\label{fig:human_gender}
    \Description{Bar charts showing character gender in stories grouped by participant gender. Male participants tend to have more masculine-coded doctors and detectives when suggested. Female participants tend to have more feminine-coded detectives when suggested. Differences are small overall.}
\end{figure*}

In \autoref{sec:participant_attr}, we considered how participant's views on gender and competence affected their stories. Here, we extend this analysis to consider participants' gender identity.\footnote{For this analysis, we focus on binary gender identity labels only due to the low level of recruitment of non-binary participants.} %
We consider a similar analysis based on participants' self-reported gender~(\autoref{fig:human_gender} and \autoref{tab:human_gender}), hypothesizing that participants who identify as women may be more likely to write stories about \woman characters without suggestions or to accept \woman suggestions. 

Here we see no significant effects, but we do note some minor trends that point towards participants writing more characters whose genders match their own. For example, we see more \man doctors and detectives from participants who identify as men under \man suggestions than we do from participants who identify as women, and we see more \man presidents from participants who identify as men under \woman suggestions than we do from participants who identify as women.

\aptLtoX[graphic=no,type=html]{
\begin{table*}[tb]
    \centering
    \footnotesize
    \begin{tabular}{llllcl}
        \toprule
        Scenario & Suggested Gender & Measured Gender  & $t$ & $\p$ & sig\\
        \midrule
        \textsc{Doctor} & NS & M & $t(37)=0.344$ & $0.8446$ & \\
        \textsc{Doctor} & NS & F & $t(37)=0.926$ & $0.5641$ & \\
        \textsc{Doctor} & M & M & $t(88)=-0.238$ & $0.9019$ & \\
        \textsc{Doctor} & M & F & $t(88)=0.838$ & $0.5980$ & \\
        \textsc{Doctor} & F & M & $t(103)=-0.608$ & $0.7324$ & \\
        \textsc{Doctor} & F & F & $t(103)=2.279$ & $0.0802$ & \\
        \rowcolor{customrowcolor} \leavevmode   \textcolor{black}{\textsc{President}} & NS & M & $t(38)=1.166$ & $0.4464$ & \\
        \rowcolor{customrowcolor} \leavevmode   \textcolor{black}{\textsc{President}} & NS & F & $t(38)=0.215$ & $0.9088$ & \\
        \rowcolor{customrowcolor} \leavevmode   \textcolor{black}{\textsc{President}} & M & M & $t(100)=1.539$ & $0.2728$ & \\
        \rowcolor{customrowcolor} \leavevmode   \textcolor{black}{\textsc{President}} & M & F & $t(100)=1.373$ & $0.3400$ & \\
        \rowcolor{customrowcolor} \leavevmode   \textcolor{black}{\textsc{President}} & F & M & $t(93)=0.14$ & $0.9488$ & \\
        \rowcolor{customrowcolor} \leavevmode   \textcolor{black}{\textsc{President}} & F & F & $t(93)=-0.011$ & $0.9910$ & \\
         \textsc{Detectives} & NS & M & $t(38)=-0.394$ & $0.8283$ & \\
         \textsc{Detectives} & NS & F & $t(38)=0.832$ & $0.6032$ & \\
         \textsc{Detectives} & M & M & $t(95)=-0.378$ & $0.8283$ & \\
         \textsc{Detectives} & M & F & $t(95)=0.356$ & $0.8353$ & \\
         \textsc{Detectives} & F & M & $t(97)=-2.095$ & $0.1183$ & \\
        \textsc{Detectives} & F & F & $t(97)=2.742$ & $0.0304$ & $*$\\
        \bottomrule
   \multicolumn{5}{c}{\textbf{a}}
    \end{tabular}
    \begin{tabular}{llllcl}
        \toprule
        Scenario & Suggested Gender & Measured Gender & $t$ & $\p$ & sig\\
        \midrule
        \textsc{Doctor} & NS & M & $t(71)=-0.449$ & $0.8085$ & \\
        \textsc{Doctor} & NS & F & $t(71)=1.916$ & $0.1634$ & \\
        \textsc{Doctor} & M & M & $t(157)=-1.762$ & $0.1991$ & \\
        \textsc{Doctor} & M & F & $t(157)=0.961$ & $0.5495$ & \\
        \textsc{Doctor} & F & M & $t(173)=-0.151$ & $0.9479$ & \\
        \textsc{Doctor} & F & F & $t(173)=0.761$ & $0.6351$ & \\
        \rowcolor{customrowcolor} \leavevmode   \textcolor{black}{\textsc{President}}& NS & M & $t(72)=0.368$ & $0.8310$ & \\
        \rowcolor{customrowcolor} \leavevmode   \textcolor{black}{\textsc{President}}& NS & F & $t(72)=0.728$ & $0.6616$ & \\
        \rowcolor{customrowcolor} \leavevmode   \textcolor{black}{\textsc{President}}& M & M & $t(174)=1.015$ & $0.5254$ & \\
        \rowcolor{customrowcolor} \leavevmode   \textcolor{black}{\textsc{President}}& M & F & $t(174)=0.89$ & $0.5714$ & \\
        \rowcolor{customrowcolor} \leavevmode   \textcolor{black}{\textsc{President}}& F & M & $t(160)=-1.53$ & $0.2735$ & \\
        \rowcolor{customrowcolor} \leavevmode  \textcolor{black}{\textsc{President}} & F & F & $t(160)=0.928$ & $0.5574$ & \\
        \textsc{Detectives} & NS & M & $t(71)=-0.126$ & $0.9523$ & \\
        \textsc{Detectives} & NS & F & $t(71)=0.659$ & $0.7019$ & \\
        \textsc{Detectives} & M & M & $t(169)=-1.688$ & $0.2218$ & \\
        \textsc{Detectives} & M & F & $t(169)=0.717$ & $0.6651$ & \\
        \textsc{Detectives} & F & M & $t(161)=-1.053$ & $0.5059$ & \\
        \textsc{Detectives} & F & F & $t(161)=1.272$ & $0.3858$ & \\
        \bottomrule
   \multicolumn{5}{c}{\textbf{b}}
    \end{tabular}
    \caption{Comparison of character genders written with various suggestions for participants who (a) answered that straight women are more competent than straight men vs less and (b) self-identified as women vs men}
\end{table*}\label{tab:human_bias}  \label{tab:human_gender}

}{
\begin{table*}[tb]
    \centering
    \footnotesize
    \begin{subtable}[t]{.45\textwidth}
    \begin{tabular}{llllcl}
        \toprule
        Scenario & \makecell[l]{Suggested\\Gender} & \makecell[l]{Measured\\Gender} & $t$ & $\p$ & sig\\
         \midrule
         & NS & M & $t(37)=0.344$ & $0.8446$ & \\
         & NS & F & $t(37)=0.926$ & $0.5641$ & \\
         & M & M & $t(88)=-0.238$ & $0.9019$ & \\
         & M & F & $t(88)=0.838$ & $0.5980$ & \\
         & F & M & $t(103)=-0.608$ & $0.7324$ & \\
        \multirow{-6}{*}{\textsc{Doctor}} & F & F & $t(103)=2.279$ & $0.0802$ & \\
        \rowcolor{customrowcolor} \leavevmode   & NS & M & $t(38)=1.166$ & $0.4464$ & \\
        \rowcolor{customrowcolor} \leavevmode   & NS & F & $t(38)=0.215$ & $0.9088$ & \\
        \rowcolor{customrowcolor} \leavevmode   & M & M & $t(100)=1.539$ & $0.2728$ & \\
        \rowcolor{customrowcolor} \leavevmode   & M & F & $t(100)=1.373$ & $0.3400$ & \\
        \rowcolor{customrowcolor} \leavevmode   & F & M & $t(93)=0.14$ & $0.9488$ & \\
        \rowcolor{customrowcolor} \leavevmode   \multirow{-6}{*}{\textcolor{black}{\textsc{President}}} & F & F & $t(93)=-0.011$ & $0.9910$ & \\
         & NS & M & $t(38)=-0.394$ & $0.8283$ & \\
         & NS & F & $t(38)=0.832$ & $0.6032$ & \\
         & M & M & $t(95)=-0.378$ & $0.8283$ & \\
         & M & F & $t(95)=0.356$ & $0.8353$ & \\
         & F & M & $t(97)=-2.095$ & $0.1183$ & \\
        \multirow{-6}{*}{\textsc{Detectives}} & F & F & $t(97)=2.742$ & $0.0304$ & $*$\\
        \bottomrule
    \end{tabular}
    \caption[]{}
    \label{tab:human_bias}
    \end{subtable}
    \hspace{2em}
    \begin{subtable}[t]{.45\textwidth}
    \centering
    \footnotesize
    \begin{tabular}{llllcl}
        \toprule
        Scenario & \makecell[l]{Suggested\\Gender} & \makecell[l]{Measured\\Gender} & $t$ & $\p$ & sig\\
        \midrule
         & NS & M & $t(71)=-0.449$ & $0.8085$ & \\
         & NS & F & $t(71)=1.916$ & $0.1634$ & \\
         & M & M & $t(157)=-1.762$ & $0.1991$ & \\
         & M & F & $t(157)=0.961$ & $0.5495$ & \\
         & F & M & $t(173)=-0.151$ & $0.9479$ & \\
        \multirow{-6}{*}{\textsc{Doctor}} & F & F & $t(173)=0.761$ & $0.6351$ & \\
         \rowcolor{customrowcolor} \leavevmode  & NS & M & $t(72)=0.368$ & $0.8310$ & \\
         \rowcolor{customrowcolor} \leavevmode  & NS & F & $t(72)=0.728$ & $0.6616$ & \\
         \rowcolor{customrowcolor} \leavevmode  & M & M & $t(174)=1.015$ & $0.5254$ & \\
         \rowcolor{customrowcolor} \leavevmode  & M & F & $t(174)=0.89$ & $0.5714$ & \\
         \rowcolor{customrowcolor} \leavevmode  & F & M & $t(160)=-1.53$ & $0.2735$ & \\
        \rowcolor{customrowcolor} \leavevmode  \multirow{-6}{*}{\textcolor{black}{\textsc{President}}} & F & F & $t(160)=0.928$ & $0.5574$ & \\
         & NS & M & $t(71)=-0.126$ & $0.9523$ & \\
         & NS & F & $t(71)=0.659$ & $0.7019$ & \\
         & M & M & $t(169)=-1.688$ & $0.2218$ & \\
         & M & F & $t(169)=0.717$ & $0.6651$ & \\
         & F & M & $t(161)=-1.053$ & $0.5059$ & \\
        \multirow{-6}{*}{\textsc{Detectives}} & F & F & $t(161)=1.272$ & $0.3858$ & \\
        \bottomrule
    \end{tabular}
    \caption[]{}
    \label{tab:human_gender}
    \end{subtable}
    \caption{Comparison of character genders written with various suggestions for participants who (a) answered that straight women are more competent than straight men vs less and (b) self-identified as women vs men}
\end{table*}

}

Overall, we find some trends pointing towards participants' stereotypes and gender identities influencing their stories and their acceptance of model suggestions in the expected direction. %

\subsection{Participant Stereotypes: Correlations and Treatment Effects}\label{sec:corr_human_stereo}

\begin{figure}[tb]
    \centering
    \includegraphics[width=0.99\linewidth]{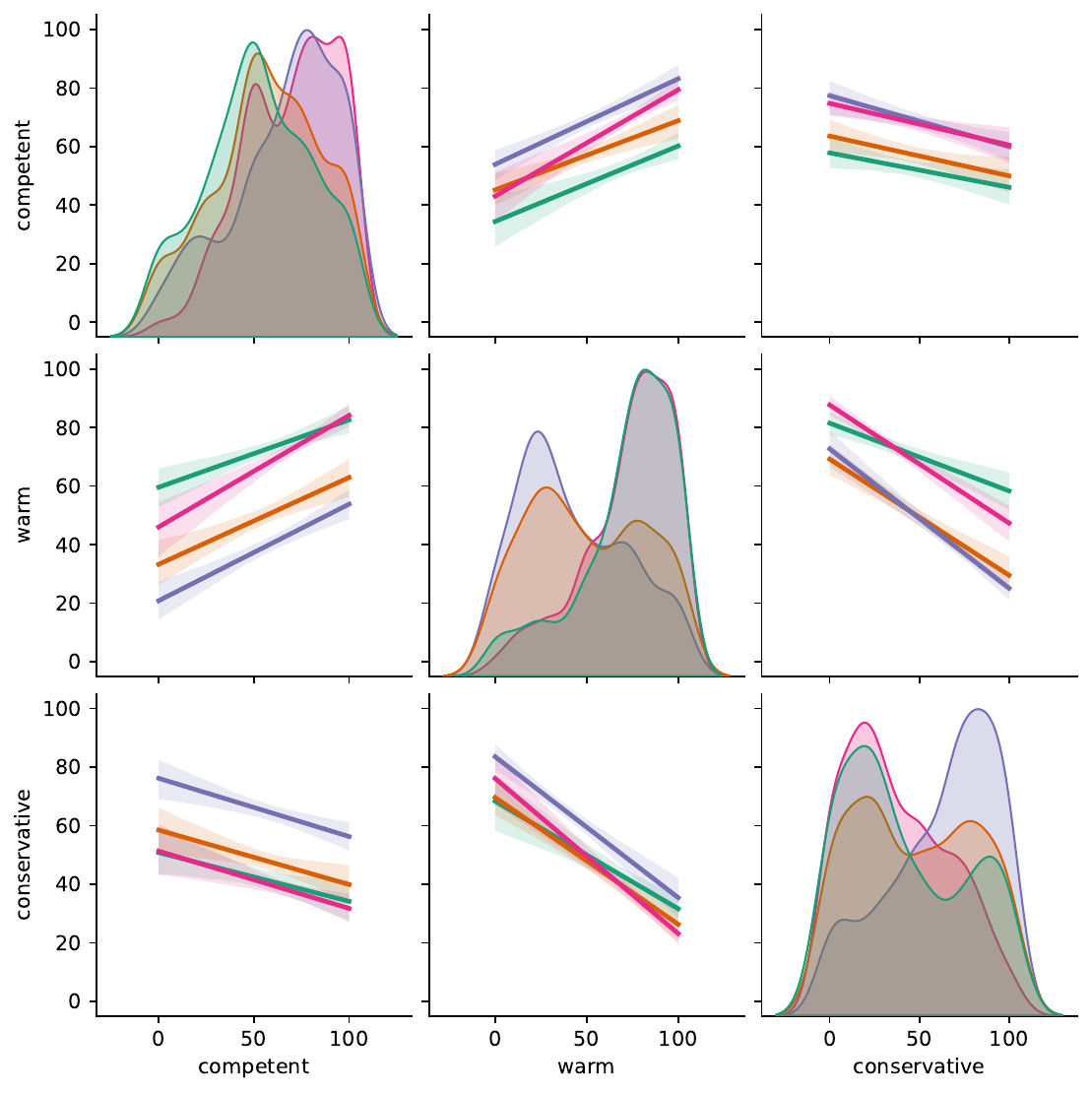}
    \includegraphics[width=0.94\linewidth]{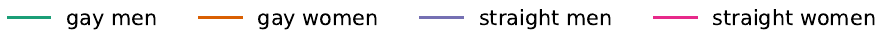}
    \caption[]{Distributions of and correlations between human stereotypes for various groups}
    \label{fig:attr_cor}
    \Description{Pair plot illustrating participants’ associations between competence, warmth, and political position across binary genders and sexualities. Warmth and competence are positively associated, while both are negatively associated with conservative political position.}
\end{figure}

In the post-survey, participants were asked about stereotypical beliefs. As we discussed in \autoref{sec:study_design}, participants were asked whether their closest friends believed different groups (gay vs straight and men vs women) were warm, competent, or conservative (\autoref{fig:attr_cor}).  

We first confirm that there is a significant positive correlation between warmth and competence ($r(1654)=0.245$, $\p<1e^{-23}$). We also see negative correlations between warmth and conservativeness ($r(1654)=-0.466$, $\p<1e^{-89}$) and competence and conservativeness ($r(1654)=-0.138$, $\p<1e^{-6}$). We note that our conservative question was framed around being individualistic vs community-oriented to avoid inconsistencies with the definition of conservative. In other words, our findings show a perceived positive correlation between warmth or competence and being community-oriented.  

We also observe differences in perceptions of groups on the various axes. We see that straight women and gay men are viewed as more warm than gay women and especially straight men. Straight people were viewed as slightly more competent than their queer counterparts. Straight men were the only group viewed as more conservative (individualistic) than liberal (community-oriented). Gay women were viewed almost neutrally while straight women and gay men were viewed as more liberal (community-oriented).

\aptLtoX[graphic=no,type=html]{
\begin{figure}[tb]
    \centering
    \includegraphics[width=0.99\linewidth]{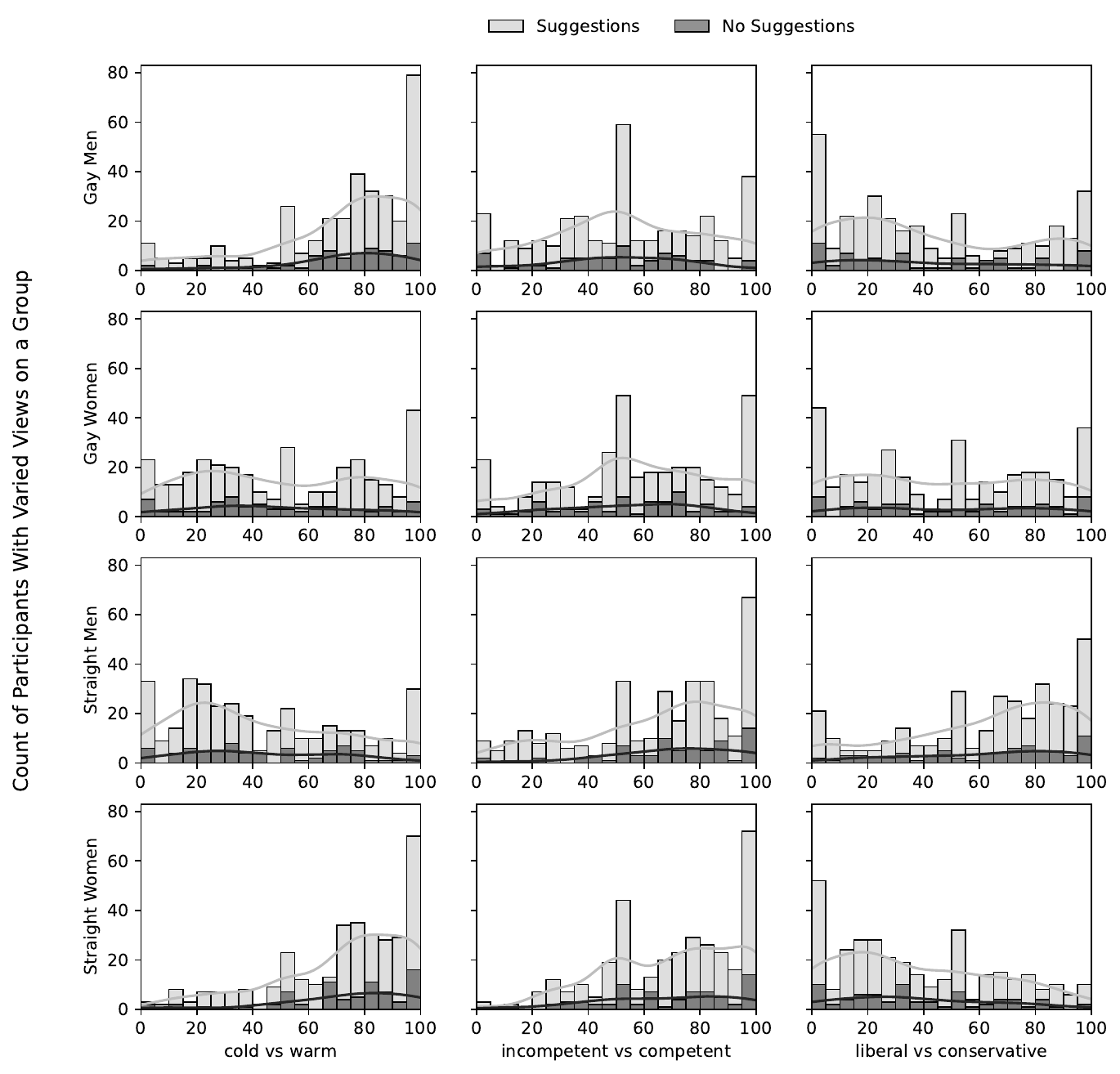}
    \caption[]{Distributions of human stereotypes from participants in the treatment condition (with suggestions) and the control condition (no suggestions).} 
    \label{fig:attr_condition}
    \Description{Histograms showing distributions of human stereotypes among participants in the treatment condition (with suggestions) and the control condition (no suggestions). The distributions appear similar across conditions.}
\end{figure}
}{\begin{figure}[tb]
    \centering
    \includegraphics[width=0.99\linewidth]{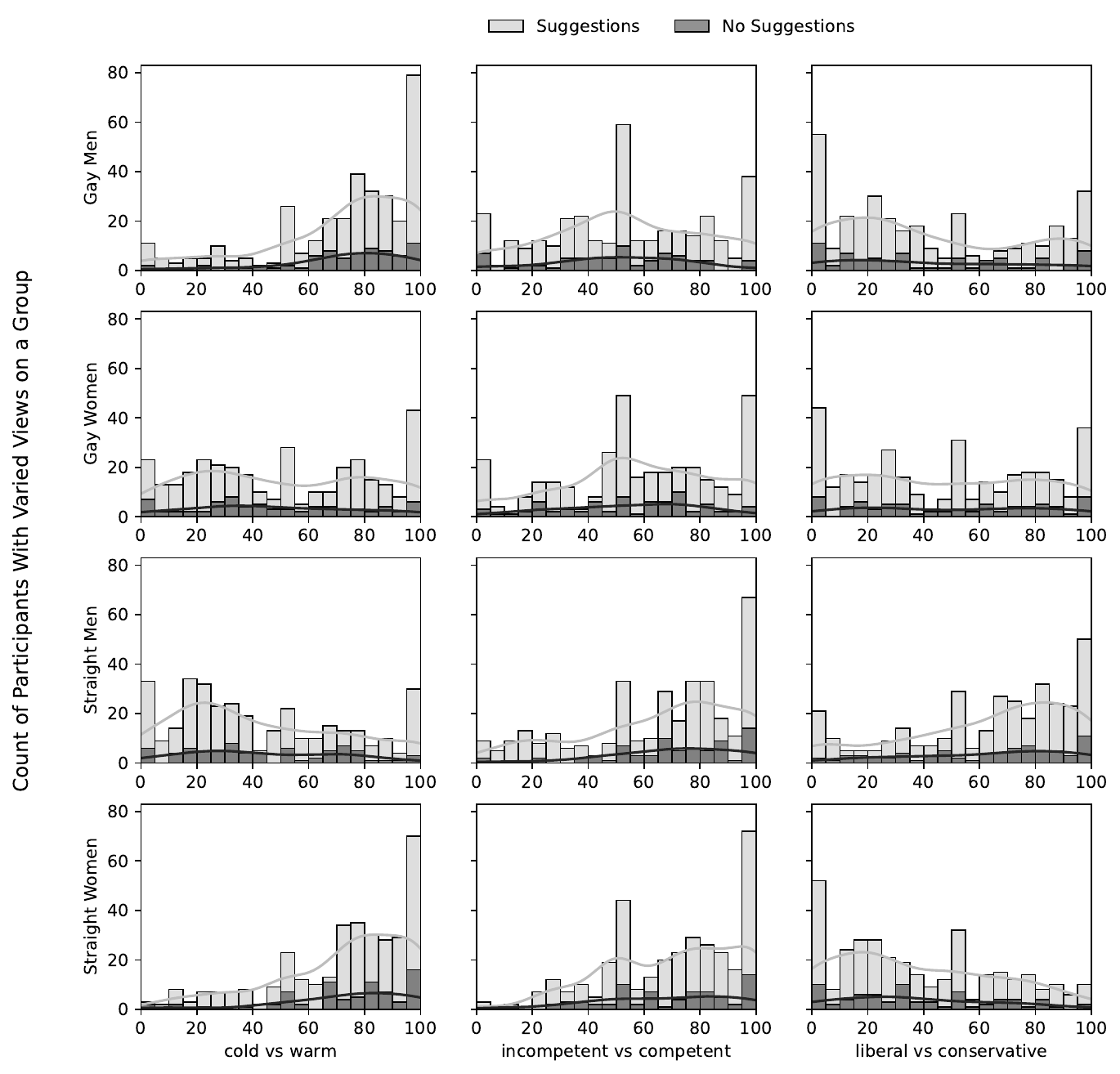}
    \caption[]{Distributions of human stereotypes from participants in the treatment condition (with suggestions) and the control condition (no suggestions).} 
    \label{fig:attr_condition}
    \Description{Histograms showing distributions of human stereotypes among participants in the treatment condition (with suggestions) and the control condition (no suggestions). The distributions appear similar across conditions.}
    \vspace{3em}
\end{figure}
}

Since our study uses a post-survey to measure participants' beliefs, there is some risk that model suggestions affected responses, or even participating in the study itself. As we discussed in \autoref{sec:study_design}, we attempt to lessen these effects by giving participants a mental break before taking the survey and are sure not to mention that the study is concerned with fairness or stereotypes. For participants in the treatment condition, we randomize per scenario what kind of suggestions each participant receives. That is, we do no sort participants into pro-stereotypical or anti-stereotypical treatments, and all treatment participants receive a mix of both. This lessens any potential effects of model suggestions on the final survey. 

To help us understand whether participants' views change as a result of merely seeing predictive text suggestions, we consider whether the distribution of survey responses is different between the treatment and control groups (\autoref{fig:attr_condition}). We observe no clear difference in the two samples. 
Running Kolmogorov-Smirnov tests,\footnote{These tests were exploratory, not part of the main analysis, and were not pre-registered. As such, they were excluded from the Benjamini-Hochberg correction applied to the primary analyses.}  we see that the distribution of responses for the two conditions are not significant different for any survey item, but this may be in part due to the relatively small number of participants in the control condition. 
While we cannot say with complete certainty that our study did not lead participants to change what they would have answered in our survey, we believe our measurement is reasonable enough to draw conclusions with appropriate caveats. %

As we discussed in \autoref{sec:study_design}, these post-survey items correspond to axes in the ABC model. \citet{cao-etal-2022-theory} surveyed US-based participants to understand their associations between ABC traits and various demographic groups. In \autoref{tab:ABC_Cao}, we consider the alignment in stereotypes between our participants and Cao et al.'s. We see overall that when differences are significant,\footnote{These tests were also exploratory, not part of the main analysis, and were not pre-registered. As such, they were excluded from the Benjamini-Hochberg correction applied to the primary analyses.} they point in the same direction. For example, both Cao et al.'s participants and our participants associate conservativeness with (straight) men more than (straight) women. We do note some differences between our findings. For both ``Agency'' related axes (confidence and competence), Cao et al. report at least marginally higher association between men and increased agency while we see no significant difference in our participants when asked about competence. Though we cannot be confident if this is difference is best explained by a difference in stereotypical beliefs of the participants in these two studies or differences in how the concept is being measured between studies, we note that this could mean that participants in our study indeed believe women to be more confident and competitive than men, potentially affecting our findings in the \textsc{Doctor} and \textsc{Student} writing scenarios.

\begin{table*}[tb]
    \centering
    \footnotesize
    \renewcommand{\arraystretch}{1.4}
    \begin{tabular}{llllllll}
        \toprule
        & \multicolumn{3}{l}{Mean US association of ABC traits and gender~\cite{cao-etal-2022-theory}} & & \multicolumn{3}{l}{Mean Post-Survey association with gender (ours)} \\
        \cmidrule(lr){2-4} \cmidrule(lr){6-8}
         ABC Axis & women& men & gender difference &Post-Survey Item & women& men & gender difference\\
         \midrule
         \colorbox{lightgreen}{\textcolor{dgreen}{\!\!confidence\!\!}}                               & $63.6$ & $75.3$ & $t(38)=-1.89, p=0.067$ & competence & $69.4$ & $66.4$ & $t(826)=1.64, p=0.102$\\
         \rowcolor{customrowcolor} \leavevmode  \colorbox{lightgreen}{\textcolor{dgreen}{\!\!competitiveness\!\!}} & $55.6$ & $75.5$ & $t(38)=-2.82, p=0.008$ & competence & $69.4$ & $66.4$ & $t(826)=1.64, p=0.102$\\
         \colorbox{lightred}{\textcolor{dred}{\!\!conservativeness\!\!}}                             & $37.0$ & $60.6$ & $t(38)=-3.32, p=0.002$ & conservativeness & $37.7$ & $62.9$ & $t(826)=-12.49, p<0.001$\\
         \rowcolor{customrowcolor} \leavevmode  \colorbox{lightblue}{\textcolor{dblue}{\!\!benevolence\!\!}}       & $65.2$ & $39.5$ & $t(38)=4.39, p<0.001$ & warmth & $72.5$ & $42.8$ & $t(826)=15.68, p<0.001$\\
         \colorbox{lightblue}{\textcolor{dblue}{\!\!trustworthiness\!\!}}                            & $57.1$ & $47.0$ & $t(38)=1.42, p=0.162$ & warmth & $72.5$ & $42.8$ & $t(826)=15.68, p<0.001$\\
         \rowcolor{customrowcolor} \leavevmode  \colorbox{lightblue}{\textcolor{dblue}{\!\!likability\!\!}}        & $69.2$ & $56.8$ & $t(38)=2.21, p=0.033$ & warmth & $72.5$ & $42.8$ & $t(826)=15.68, p<0.001$\\
         \bottomrule
    \end{tabular}
    \caption[]{Comparison of \citet{cao-etal-2022-theory}'s US-based annotator's associations between gender and ABC traits and our participants' (from multiple countries) associations with gender (of straight people), warmth, competence, and conservativeness in our post survey. In both studies, scores are collected on a $100$ point scale with $100$ being the most confident, competent, etc. }
    \label{tab:ABC_Cao}
\end{table*}

\renewcommand{\arraystretch}{1}

\subsection{Suggestions and Toxicity, Sentiment, and Character Agency}\label{sec:agency}

\begin{figure*}[tbh]
    \centering
    \begin{subfigure}[t]{.3\linewidth}
            \centering
            \includegraphics[width=1\textwidth,clip,trim=0 0 0 0]{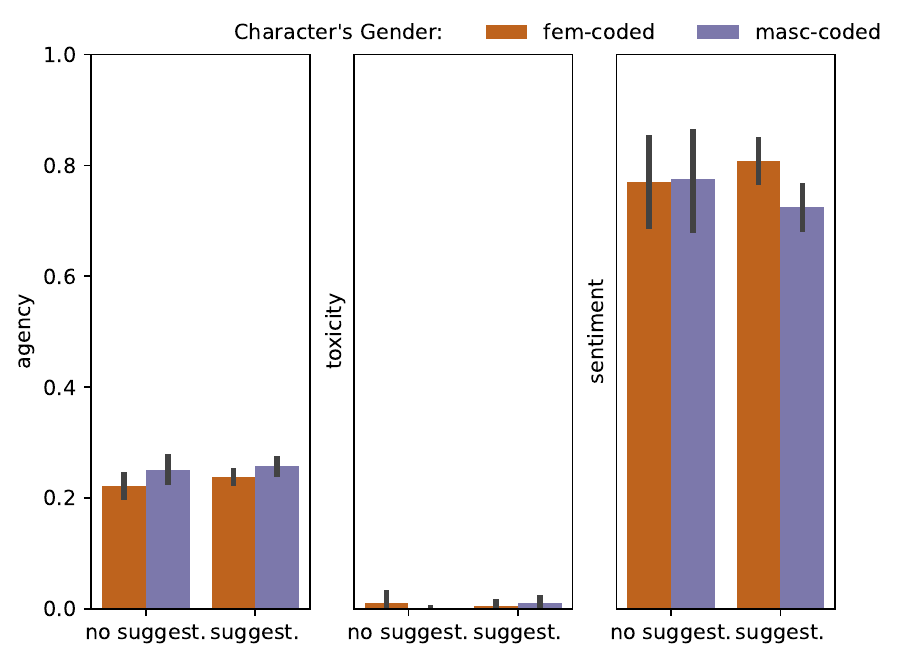}
            \caption[]{\textsc{Student}}%
        \end{subfigure}
    \begin{subfigure}[t]{.3\linewidth}
            \centering
            \includegraphics[width=1\textwidth,clip,trim=0 0 0 0]{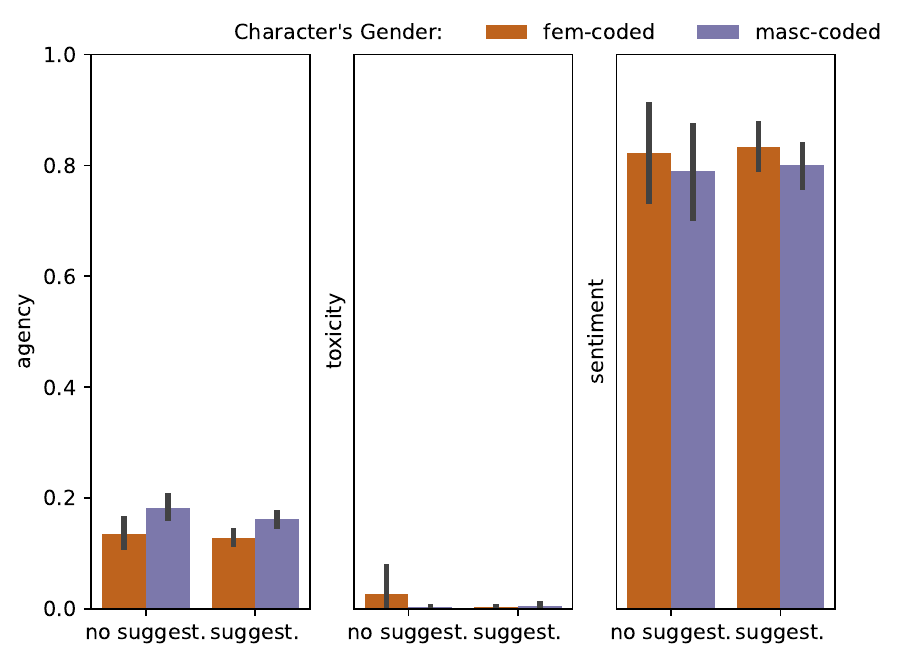}
            \caption[]{\textsc{Teacher}}%
        \end{subfigure}
    \begin{subfigure}[t]{.3\linewidth}
            \centering
            \includegraphics[width=1\textwidth,clip,trim=0 0 0 0]{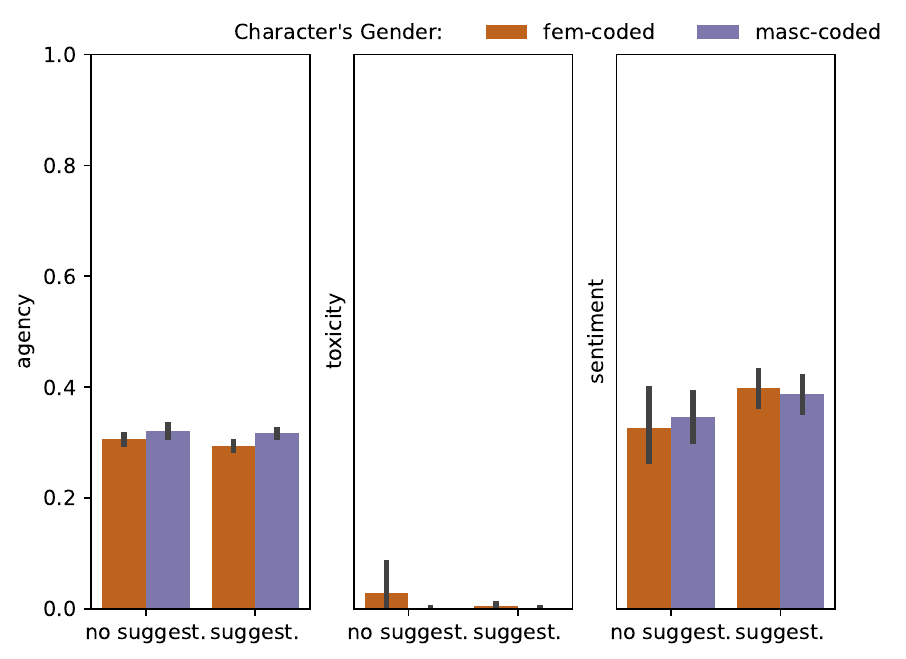}
            \caption[]{\textsc{Town Hall}}%
        \end{subfigure}
    \caption[]{Agency, toxicity, and sentiment ratings in stories. For each attribute, we break down stories into those written with and without suggestions and those written about \man characters (purple) and \woman (orange).}\label{fig:other_attr}
    \Description{ Bar charts showing agency, toxicity, and sentiment ratings in stories about characters of different genders. Differences between stories written with and without suggestions are minimal. }
\end{figure*}

\aptLtoX[graphic=no,type=html]{
\begin{table*}[tbh]
    \centering
    \footnotesize
    \begin{tabular}{l l l l l l}
        \toprule
        Scenario & Classification Attribute & Suggestions & $t$ & $\p$ & sig\\
        \midrule
        \textsc{Student} & communion & - & $t(72)=1.479$ & $0.2978$ & \\
        \textsc{Student} & communion & \checkmark & $t(337)=1.758$ & $0.1991$ & \\
        \textsc{Student} & toxicity & - & $t(72)=0.955$ & $0.5497$ & \\
        \textsc{Student} & toxicity & \checkmark & $t(337)=-0.672$ & $0.6982$ & \\
        \textsc{Student} & sentiment & - & $t(72)=-0.066$ & $0.9634$ & \\
        \textsc{Student} & sentiment & \checkmark & $t(337)=2.516$ & $0.0462$ & $*$\\
        \rowcolor{customrowcolor} \leavevmode  \textsc{Teacher} & communion & - & $t(71)=2.424$ & $0.0625$ & \\
        \rowcolor{customrowcolor} \leavevmode  \textsc{Teacher} & communion & \checkmark & $t(333)=3.44$ & $0.0040$ & $*$\\
        \rowcolor{customrowcolor} \leavevmode  \textsc{Teacher} & toxicity & - & $t(71)=1.009$ & $0.5296$ & \\
        \rowcolor{customrowcolor} \leavevmode  \textsc{Teacher} & toxicity & \checkmark & $t(333)=-1.147$ & $0.4464$ & \\
        \rowcolor{customrowcolor} \leavevmode  \textsc{Teacher} & sentiment & - & $t(71)=0.446$ & $0.8085$ & \\
        \rowcolor{customrowcolor} \leavevmode  \textsc{Teacher} & sentiment & \checkmark & $t(333)=0.957$ & $0.5495$ & \\
        \textsc{Town Hall} & communion & - & $t(72)=1.808$ & $0.1927$ & \\
        \textsc{Town Hall} & communion & \checkmark & $t(332)=3.852$ & $0.0010$ & $*$\\
        \textsc{Town Hall} & toxicity & - & $t(72)=1.083$ & $0.4908$ & \\
        \textsc{Town Hall} & toxicity & \checkmark & $t(332)=1.201$ & $0.4184$ & \\
        \textsc{Town Hall} & sentiment & - & $t(72)=-0.408$ & $0.8283$ & \\
        \textsc{Town Hall} & sentiment & \checkmark & $t(332)=0.381$ & $0.8283$ & \\
        \bottomrule
    \end{tabular}
    \caption[]{Comparison of attribute scores between character genders in stories written with and without predictive text suggestions.}
    \label{tab:other_attr}
\end{table*}
}{
\begin{table*}[tbh]
    \centering
    \footnotesize
    \begin{tabular}{l l l l l l}
        \toprule
        Scenario & Classification Attribute & Suggestions & $t$ & $\p$ & sig\\
        \midrule
        & communion & - & $t(72)=1.479$ & $0.2978$ & \\
        & communion & \checkmark & $t(337)=1.758$ & $0.1991$ & \\
        & toxicity & - & $t(72)=0.955$ & $0.5497$ & \\
        & toxicity & \checkmark & $t(337)=-0.672$ & $0.6982$ & \\
        & sentiment & - & $t(72)=-0.066$ & $0.9634$ & \\
        \multirow{-6}{*}{\textsc{Student}} & sentiment & \checkmark & $t(337)=2.516$ & $0.0462$ & $*$\\
        \rowcolor{customrowcolor} \leavevmode   & communion & - & $t(71)=2.424$ & $0.0625$ & \\
        \rowcolor{customrowcolor} \leavevmode   & communion & \checkmark & $t(333)=3.44$ & $0.0040$ & $*$\\
        \rowcolor{customrowcolor} \leavevmode  & toxicity & - & $t(71)=1.009$ & $0.5296$ & \\
        \rowcolor{customrowcolor} \leavevmode  & toxicity & \checkmark & $t(333)=-1.147$ & $0.4464$ & \\
        \rowcolor{customrowcolor} \leavevmode  & sentiment & - & $t(71)=0.446$ & $0.8085$ & \\
        \rowcolor{customrowcolor} \leavevmode  \multirow{-6}{*}{\textsc{Teacher}} & sentiment & \checkmark & $t(333)=0.957$ & $0.5495$ & \\
        & communion & - & $t(72)=1.808$ & $0.1927$ & \\
        & communion & \checkmark & $t(332)=3.852$ & $0.0010$ & $*$\\
        & toxicity & - & $t(72)=1.083$ & $0.4908$ & \\
        & toxicity & \checkmark & $t(332)=1.201$ & $0.4184$ & \\
        & sentiment & - & $t(72)=-0.408$ & $0.8283$ & \\
        \multirow{-6}{*}{\textsc{Town Hall}} & sentiment & \checkmark & $t(332)=0.381$ & $0.8283$ & \\
        \bottomrule
    \end{tabular}
    \caption[]{Comparison of attribute scores between character genders in stories written with and without predictive text suggestions.}
    \label{tab:other_attr}
\end{table*}
}

Beyond the attributes the predictive text model was controled to suggest, we also consider off-the-shelf classification of stories toxicity\footnote{\href{https://huggingface.co/martin-ha/toxic-comment-model}{https://huggingface.co/martin-ha/toxic-comment-model}}, sentiment~\cite{camacho-collados-etal-2022-tweetnlp}, and character agency~\cite{wan-etal-2023-kelly}. Here, we consider writing scenarios where the character of interest's gender is cued in the story prefix. We compare between classifier output for stories where characters are \man vs \woman when participants are or are not provided with suggestions (\autoref{fig:other_attr} and \autoref{tab:other_attr}).

\paragraph{Toxicity} We find that the generated stories tend not to be explicitly toxic, with the story with the highest toxicity rating describing that ``Rebecca said that our ideas were really stupid and bland. We got very angry and shouted at her. It was very unprofessional.'' Toxicity rates were uniformly low across genders. 

\paragraph{Sentiment} We find no significant differences in sentiment between suggestion conditions in the \textsc{Teacher} and \textsc{Town Hall} scenarios regardless of suggestions and in the \textsc{Student} scenario without suggestions. However, in the \textsc{Student} scenario, we see significantly lower sentiment ratings for stories written about ``John'' than ``Abby'' when the stories are written with suggestions ($t(337)=2.516$, $\p\approx 0.0462$, $d=0.273$). This means that the predictive text model may have had a bias towards suggesting more positive continuations about Abby than John or that participants were more likely to accept positive suggestions about Abby. Regardless of mechanism, we see that, in this scenario, predictive text suggestions widened the gap in sentiment between genders.

\paragraph{Character Agency} The final classifier considers whether characters in a story are described as agentic (e.g., being a natural leader) vs communal (e.g., being a well-liked member of a group)~\cite{wan-etal-2023-kelly}. In the \textsc{Student} scenario, we see no significant gender differences in agency regardless of the presence or absence of suggestions. In the \textsc{Town Hall} scenario, we see significantly higher agency in stories about ``Thomas'' than ``Rebecca'' when they are written with suggestions ($t(332)=3.852$, $\p\approx0.0010$, $d=0.422$). We see a similar trend in the \textsc{Teacher} scenario. Here the increased agency for ``Mr. Brown'' is significant with suggestions ($t(333)=3.44$, $\p\approx0.0040$, $d=0.377$) and marginally significant without suggestions ($t(71)=2.424$, $\p\approx0.0625$, $d=0.568$). These results show that model biases towards \man characters having more agency in their stories may leak into co-written stories.

\subsection{Prefix Similarity on Attribute Selection}\label{sec:action_number}

The uniform information density (UID) hypothesis states that people prefer to uniformly distribute information throughout language production when possible to maintain the same message~\cite{frank2008speaking, jaeger2006speakers, meister-etal-2021-revisiting}. An implication of UID is that low-probability words may be more likely followed by high-probability words (and vice versa). In our scenario, this could potentially impact the selection of pro-stereotypical (likely higher probability) versus anti-stereotypical (likely lower probability) as a function of the probability of the word (or phrase) that came before. Under this interpretation, participants are not (only) choosing a pro-stereotypical word because it is higher probability, but because the preceding word is low probability. For example, in spoken language, fillers are a common way to add additional time before a low probability event, such as ``Bill married his [uh] long-time boyfriend'' vs ``Bill married his long-time girlfriend.''

 The UID interpretation in our setting suggests that anti-stereo\-typical suggestions may not be taken because the user was not \emph{planning} on the low probability event (``boyfriend'') and by the time the anti-stereotypical suggestion arrived (after ``long-time'') it was too late to make an appropriate high-probability selection in advance of the low-probability continuation. Then, in this interpretation, the user is even more strongly guided to select the high-probability continuation that they had in mind (``girlfriend''), irrespective of any anti-stereotypical suggestion.

While this interpretation is possible---and could be an interesting avenue for future research---we expect that its effect is rather small, for two reasons. First, measured UID effects tend to be quite small. For example, UID effects on log likelihood are on the order of, at most, $\pm 0.15$ nats in \citet{meister-etal-2021-revisiting}, in comparison to probability differences on the order of as much as $\pm 20$. Second, in our study, the \prefix is minimally edited within scenarios (e.g., changing only ``Mr. Brown'' vs ``Mrs. Brown''). This means that early in each story, the information density between conditions should be roughly the same. Thus, any tokens that appear very early in the story will, by definition, have nearly the same prefix and therefore nearly the same past information density.

\aptLtoX[graphic=no,type=html]{
\begin{figure*}[tb]
    \begin{subfigure}[b]{.48\linewidth}
    \centering
    \includegraphics[width=\linewidth,clip,trim=0 0 0 0]{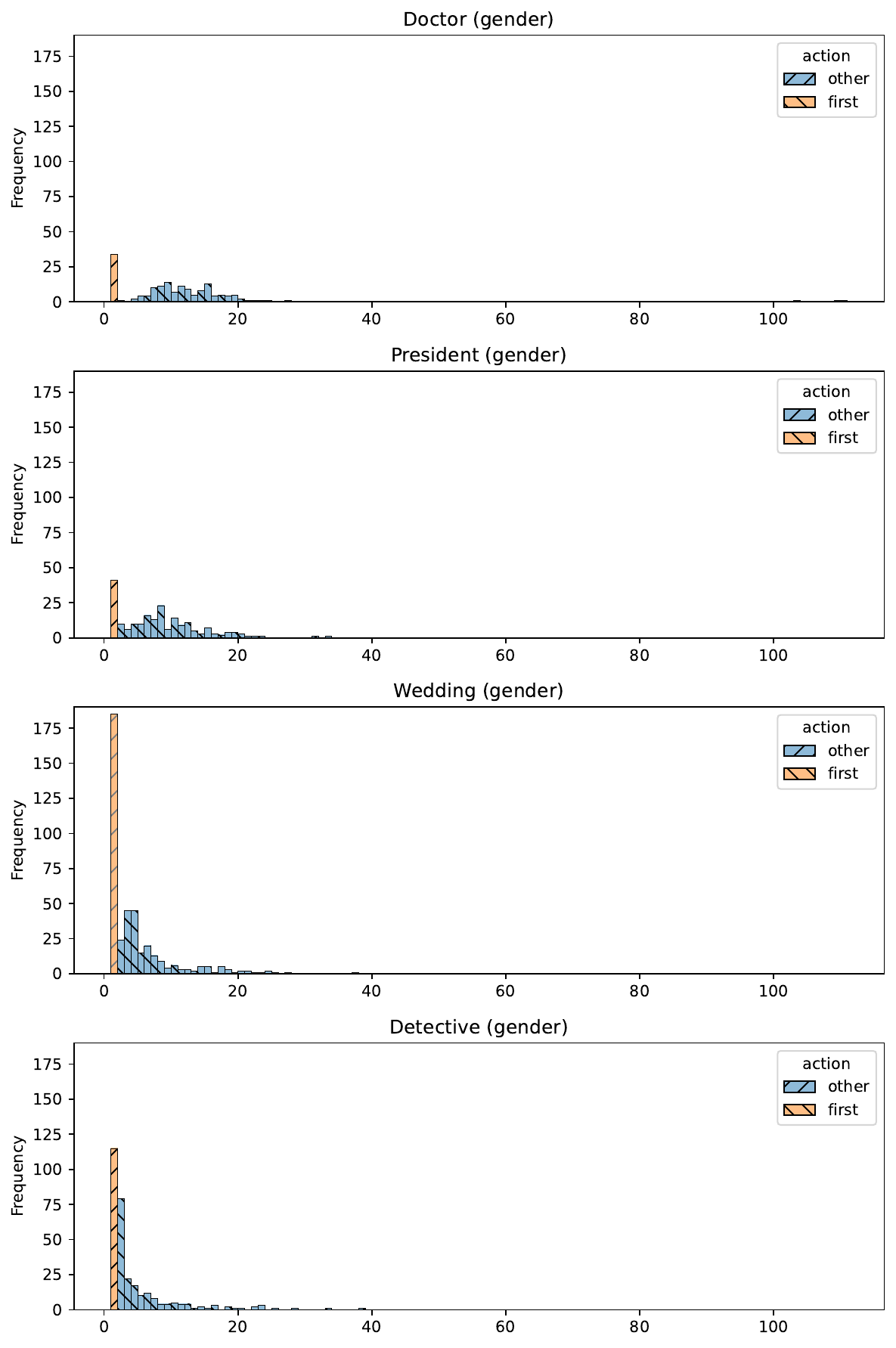}
    \caption[]{Gender}
    \label{fig:gender_action_num}
    \end{subfigure}
    \hspace{.01\linewidth}
    \begin{subfigure}[b]{.48\linewidth}
    \centering
    \includegraphics[width=\linewidth,clip,trim=0 0 0 0]{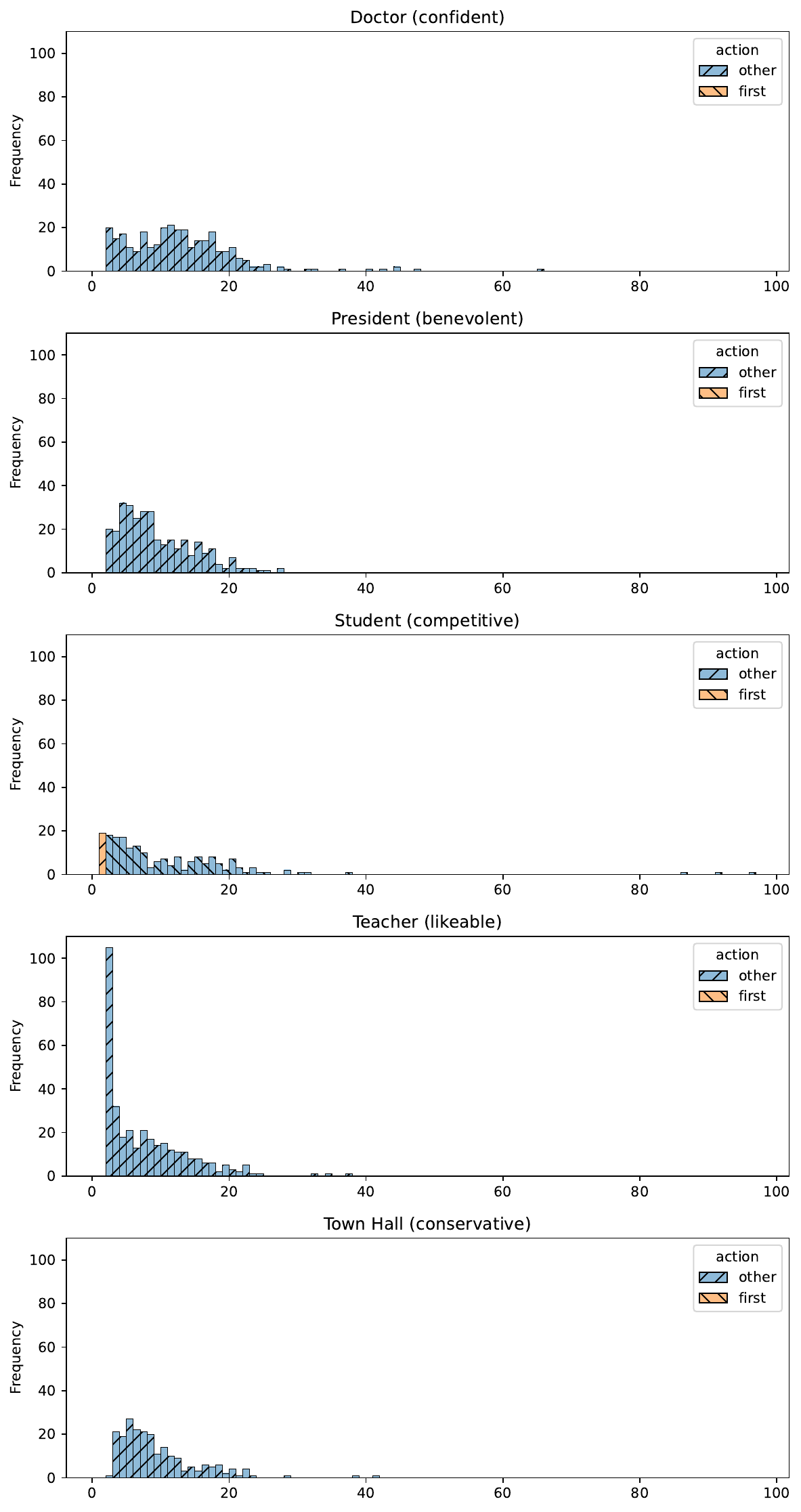}
    \caption[]{ABC Traits}
    \label{fig:ABC_action_num}
    \end{subfigure}
    \caption[]{What number writing action determines the given attribute in each story? The first writing action (after the ``start'' action) is highlighted in orange.}
    \label{fig:action_num}
    \Description{Histograms showing the story position of writing actions that determine each attribute in co-written stories. For gender, a spike occurs at position 1, indicating gender is often specified in the first new token. For ABC traits, contributions are distributed further along in the stories. }
\end{figure*}
}{
\begin{figure*}[tb]
    \begin{subfigure}[b]{.48\linewidth}
    \centering
    \includegraphics[width=\linewidth,clip,trim=0 0 0 0]{figures/gender_action_num.pdf}
    \caption[]{Gender}
    \label{fig:gender_action_num}
    \end{subfigure}
    \hspace{.01\linewidth}
    \begin{subfigure}[b]{.48\linewidth}
    \centering
    \includegraphics[width=\linewidth,clip,trim=0 0 0 0]{figures/ABC_action_num.pdf}
    \caption[]{ABC Traits}
    \label{fig:ABC_action_num}
    \end{subfigure}
    \caption[]{What number writing action determines the given attribute in each story? The first writing action (after the ``start'' action) is highlighted in orange.}
    \label{fig:action_num}
    \Description{Histograms showing the story position of writing actions that determine each attribute in co-written stories. For gender, a spike occurs at position 1, indicating gender is often specified in the first new token. For ABC traits, contributions are distributed further along in the stories.}
    \vspace{10em}
\end{figure*}
}

More specifically, for gender (\autoref{fig:gender_action_num}), we have that about $35\%$ of gender-defining tokens are written or accepted in the first action (counting the ``start'' action as action 0) with $63\%$ of gender-defining tokens coming from the first five actions. For the \textsc{Detective} scenario, we see a large portion of gender-defining tokens as the second action where participants write a name like ``Detective X'' instead of just ``X''. For the \textsc{Wedding} scenario, we see some slightly longer phrases before the partner's gender is written out from common phrases like ``his highschool sweetheart X'', ``her soulmate X'', etc. Overall, we see that a large portion of gender-defining tokens are written with nearly identical contexts (within writing scenarios), leading the prior information density to be roughly equal when the gender decision is made. 

For attributes beyond gender (\autoref{fig:ABC_action_num}), we see more variability in when the ABC trait is specified. For most scenarios, the ABC trait cannot easily be specified in the first token. In the \textsc{Teacher} scenario, the teacher's likability is never determined by the first written or accepted token. Instead, in about $31\%$ of stories, participants specify the teacher's likability on the second token with phrases like ``my favorite'', ``the worst'', etc. While this means likability in this case was often determined with prefixes with similar information density, for the remaining scenarios, ABC traits may have been determined with prefixes that have more variable information density which may have influenced participant behavior.

Overall, we see that there are some attributes writing scenarios that were largely determined early enough in stories that information density is very consistent between conditions. We argue that our other findings cannot be explained through effects of uniform information density alone.

%% file: Sections/A3.Annotation.tex
\section{Annotation, Validation, and Additional Participant Details}

\subsection{Annotation Prompts and Instructions}

\begin{figure*}[tb]
    \includegraphics[width=.95\textwidth, trim={0 145 275 0},clip]{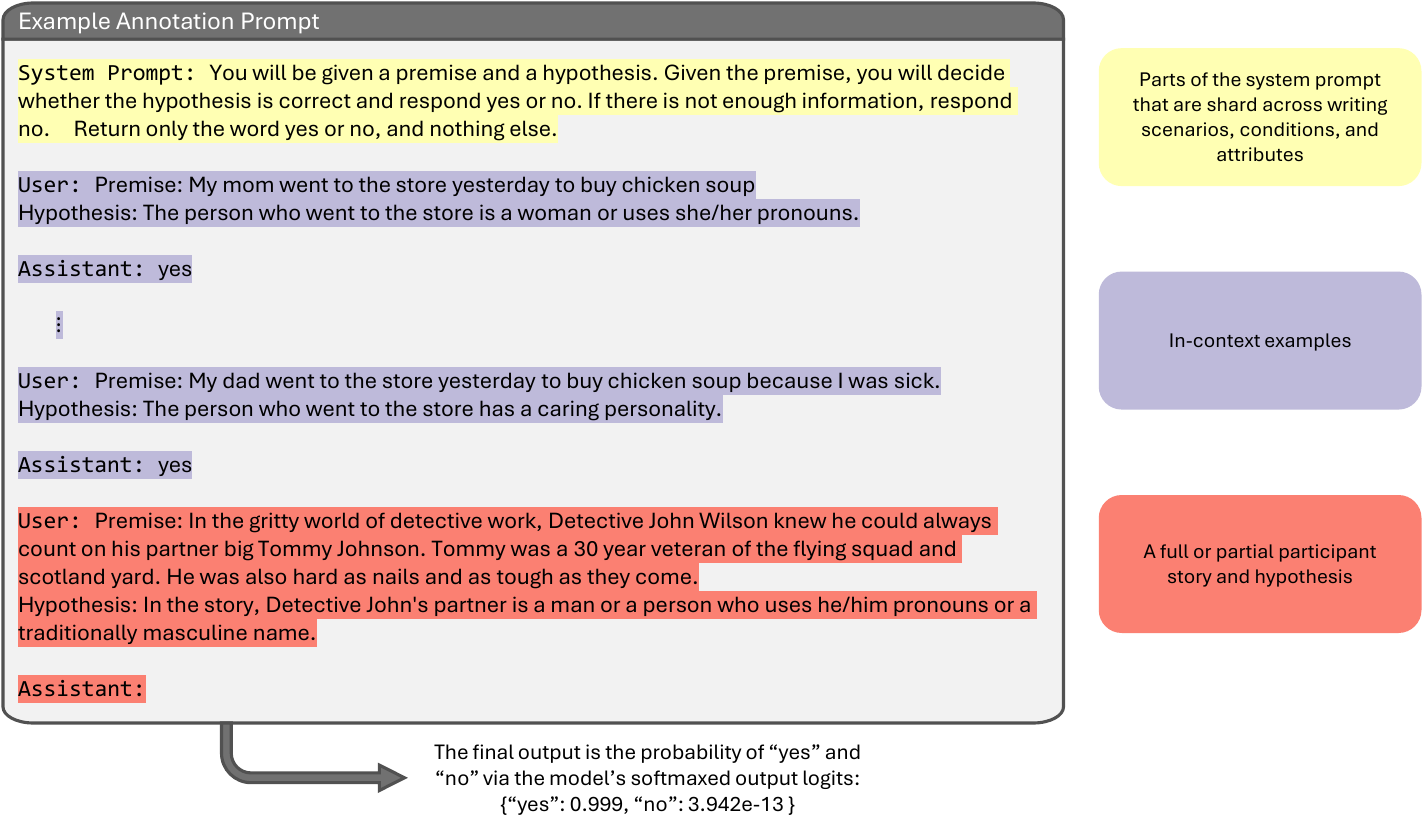}
    \caption[]{Example prompt for annotating whether the detective's partner is \man.}\label{prompt:annottrustworthy}
     \Description{Example annotation prompt showing the system instruction: participants receive a premise and hypothesis, and decide whether the hypothesis is correct by responding ``yes'' or ``no.'' The figure includes in-context examples and the participant’s partial story with a hypothesis to annotate.}
\end{figure*}

As we discuss in \autoref{sec:llm_annot}, we annotate characteristics of characters in the written stories using an LLM. We provide the prompt used for this in \autoref{prompt:annottrustworthy}. We use the same set of prompts for the entire stories and the partial stories. The full set of ``hypotheses'' used for each scenario's stories are shown in \autoref{tab:hypos}.

\begin{table*}[tb]
    \centering
    \footnotesize
    \renewcommand{\arraystretch}{1.4} %
    \begin{tabular}{p{1.6cm} p{1.7cm} p{10.5cm}}
        \toprule
        Scenario & Annotated Value & Hypothesis\\
        \midrule
        \textsc{Doctor}  & \colorbox{lightgreen}{\textcolor{dgreen}{confident}} & According to the speaker, the doctor acts confidently in the story.\\
        \textsc{Doctor}  & \colorbox{lightgreen}{\textcolor{dgreen}{unconfident}} & According to the speaker, the doctor acts unconfidently in the story.\\
        \textsc{Doctor}  & \colorbox{lightpurple}{\textcolor{dpurple}{\man}} & In the story, the doctor is a man or a person who uses he/him pronouns or a traditionally masculine name.\\
        \textsc{Doctor}  & \colorbox{lightpurple}{\textcolor{dpurple}{\woman}} & In the story, the doctor is a woman or a person who uses she/her pronouns or a traditionally feminine name.\\
        \rowcolor{customrowcolor}\textsc{President}  & \colorbox{lightblue}{\textcolor{dblue}{benevolent}} & In the story, the president prioritizes benevolent policies.\\
        \rowcolor{customrowcolor}\textsc{President}  & \colorbox{lightblue}{\textcolor{dblue}{threatening}} & In the story, the president prioritizes threatening policies.\\
        \rowcolor{customrowcolor}\textsc{President}  & \colorbox{lightpurple}{\textcolor{dpurple}{\man}} & In the story, the president is a man or a person who uses he/him pronouns or a traditionally masculine name.\\
        \rowcolor{customrowcolor}\textsc{President}  & \colorbox{lightpurple}{\textcolor{dpurple}{\woman}} & In the story, the president is a woman or a person who uses she/her pronouns or a traditionally feminine name.\\
        \textsc{Wedding}  & \colorbox{lightpurple}{\textcolor{dpurple}{\man}} 2nd partner& In the story, \textit{\{Sherry/Joe\}} is getting married to a man or a person with a traditionally masculine name.\\
        \textsc{Wedding}  & \colorbox{lightpurple}{\textcolor{dpurple}{\man}} 2nd partner& In the story, \textit{\{Sherry/Joe\}} is getting married to a man or a person with a traditionally masculine name.\\
        \rowcolor{customrowcolor}\textsc{Student}  & \colorbox{lightgreen}{\textcolor{dgreen}{competitive}} & In the story, \textit{\{Abby/John\}} acts in an assertive manner within the group, for example, taking a leadership role.\\
        \rowcolor{customrowcolor}\textsc{Student}  & \colorbox{lightgreen}{\textcolor{dgreen}{unassertive}} & In this story, \textit{\{Abby/John\}} acts in a shy or an unassertive manner within the group, for example, taking a follower role.\\
        \textsc{Detectives}  & \colorbox{lightpurple}{\textcolor{dpurple}{\man}} & In the story, Detective John's partner is a man or a person who uses he/him pronouns or a traditionally masculine name.\\
        \textsc{Detectives}  & \colorbox{lightpurple}{\textcolor{dpurple}{\woman}} & In the story, Detective John's partner is a woman or a person who uses she/her pronouns or a traditionally feminine name.\\
        \rowcolor{customrowcolor}\textsc{Teacher}  & \colorbox{lightblue}{\textcolor{dblue}{likable}} & According to the speaker, \textit{\{Mrs./Mr.\}} Brown has a likable personality.\\
        \rowcolor{customrowcolor}\textsc{Teacher}  & \colorbox{lightblue}{\textcolor{dblue}{repellent}} & According to the speaker, \textit{\{Mrs./Mr.\}} Brown has a repellent personality.\\
        \textsc{Town Hall}  & \colorbox{lightred}{\textcolor{dred}{conservative}} & In the story, \textit{\{Thomas/Rebecca\}} supports a conservative approach to housing, for example, opposing affordable housing since it will hurt property values.\\
        \textsc{Town Hall}  & \colorbox{lightred}{\textcolor{dred}{liberal}} & In the story, \textit{\{Thomas/Rebecca\}} supports a liberal approach to housing, for example, supporting affordable housing since it would reduce homelessness.\\
        \bottomrule
    \end{tabular}
    \caption[]{Hypotheses provided to LLM when annotating various attribute values in stories }
    \label{tab:hypos}
\end{table*}

\subsection{Common Contributing \ttokens}

As discussed in \autoref{sec:llm_annot}, we annotate at the \token level to determine which \tokens (either included in the story or proposed and rejected by the model) contributed to the gender, likability, confidence, etc of story characters. In \autoref{tab:tokens}, we list for each writing scenario and axes which \tokens the model identified as determining axis values. Note that the same \token may appear for both values of an axis. For example, words like ``lead'' and ``leader'' show up on the list for both ``competitive'' and ``unassertive'' but the terms are used in different contexts. For instance, a story containing the sentence ``John felt uncomfortable taking the \textbf{lead}.'' fell in the ``unassertive'' category and ``Abby was selected as the \textbf{leader} of our group.'' fell in the opposite.

\begin{table*}[tb]
    \centering
    \footnotesize
    \renewcommand{\arraystretch}{1.4} %
    \begin{tabular}{p{1.6cm} p{1.7cm} p{10.5cm}}
        \toprule
        Scenario & Attribute & Common Value-defining \ttokens\\
        \midrule
        \textsc{Doctor}  & \colorbox{lightgreen}{\textcolor{dgreen}{confident}} & medication (17); to (17); medicine (14); confident (14); suspected (13); need (12); nothing (11)\\
        \textsc{Doctor}  & \colorbox{lightgreen}{\textcolor{dgreen}{unconfident}} & might (67); hesitated (16); unsure (12); uncertain (11)\\
        \textsc{Doctor}  & \colorbox{lightpurple}{\textcolor{dpurple}{\man}} & he (344); his (180); him (12)\\
        \textsc{Doctor}  & \colorbox{lightpurple}{\textcolor{dpurple}{\woman}} & she (315); her (242)\\
        \rowcolor{customrowcolor}\textsc{President}  & \colorbox{lightblue}{\textcolor{dblue}{benevolent}}  & infrastructure (31); climate (25); nations (19); crumbling (18); tensions (11); jobs (10)\\
        \rowcolor{customrowcolor}\textsc{President}  & \colorbox{lightblue}{\textcolor{dblue}{threatening}}  & military (9)\\
        \rowcolor{customrowcolor}\textsc{President}  & \colorbox{lightpurple}{\textcolor{dpurple}{\man}} & his (270); he (162)\\
        \rowcolor{customrowcolor}\textsc{President}  & \colorbox{lightpurple}{\textcolor{dpurple}{\woman}} & her (300); she (151)\\
        \textsc{Wedding}  & \colorbox{lightpurple}{\textcolor{dpurple}{\man}} & steve (283); man (53); john (42); his (36); best (32); longtime (25); steve's (16); friend (15); he (14); crush (11)\\
        \textsc{Wedding}  & \colorbox{lightpurple}{\textcolor{dpurple}{\woman}} & susie (208); her (76); sweetheart (31); sarah (26); dear (16); sweatheart (15); crush (11)\\
        \rowcolor{customrowcolor}\textsc{Student}  & \colorbox{lightgreen}{\textcolor{dgreen}{competitive}} & leader (57); lead (52); asserted (25); assigned (20); fearless (20); leading (19); established (17); charge (17); competitive (10)\\
        \rowcolor{customrowcolor}\textsc{Student}  & \colorbox{lightgreen}{\textcolor{dgreen}{unassertive}} & hesitated (181); hesitant (33); leading (23); reluctant (22); quiet (11); leader (11); lead (10)\\
        \textsc{Detectives}  & \colorbox{lightpurple}{\textcolor{dpurple}{\man}} & steve (145); robinson (43); steven (26); he (20); his (19); partner (10)\\
        \textsc{Detectives}  & \colorbox{lightpurple}{\textcolor{dpurple}{\woman}} & sarah (280); she (109); her (60); rachel (34); robinson's (15)\\
        \rowcolor{customrowcolor}\textsc{Teacher}  & \colorbox{lightblue}{\textcolor{dblue}{likable}}  & favorite (103); kind (25); patient (25); best (24); most (22); inspired (18); favourite (15); inspiration (10)\\
        \rowcolor{customrowcolor}\textsc{Teacher}  & \colorbox{lightblue}{\textcolor{dblue}{repellent}}  & least (75); feared (54); intimidating (30); dreaded (26); notorious (21); hated (11) \\
        \textsc{Town Hall}  & \colorbox{lightred}{\textcolor{dred}{conservative}} & opposed (49); against (17)\\
        \textsc{Town Hall}  & \colorbox{lightred}{\textcolor{dred}{liberal}} & provide (26); help (22); essential (17); necessary (15); supported (11)\\
        \bottomrule
    \end{tabular}
    \caption[]{\ttokens that are commonly annotated as setting the value of an axis. \ttokens are stripped and lowercase. Only \tokens that define the given value in at least 10 stories are included  (or if there are none above 10, then the most common \token). }
    \label{tab:tokens}
\end{table*}

\subsection{Human Evaluation Details and Instructions}\label{sec:human_anno}

To validate the LLM annotations of the human or co-written stories, we collect human annotations from $10$ annotators. For each of the $7$ scenarios, we have $2*2$ potential axis value combinations (See \autoref{tab:scenarios} for a list of all scenarios and axes) which we measure independently. For each of these measurements (e.g. the character ``Mr. Brown'' has a likable personality), the value can true or false/unspecified. This leaves us $7*4*2=56$ unique measurement values made about the set of stories. We collect $560$ random sets of these unique story measurement values. Each annotator is asked to annotate $56$ stories for single axis values, but these tasks are randomized between annotators to avoid them learning patterns about how many ``true'' and ``false'' values there should be per scenario, axis, etc. The statements about each story shown to human annotators were the same as the hypotheses used to prompt the LLM annotator (See \autoref{tab:hypos}). We include the instructions to human annotators below:

\begin{quote}
You will be shown a series of stories and statements (hypotheses) about characteristics of characters in each story, and you will need to mark which statements are entailed (``True'') or are contradicted/neutral (``False''). The characteristics are paired (e.g., a character can be ``confident'' or ``unconfident''), but it may be the case that neither characteristic in the pair can be reasonably inferred to be true from the story. Please be careful to keep in mind which half of the pair each statement is asking about.
\end{quote}

\begin{quote}
Please mark the statement as false if it is either untrue or is unspecified using your best judgement about what can be ``reasonably'' inferred from the story. For example, for a story where the character's likable vs repellent personality is not explored at all, please mark ``False''. For a story where the character is seen by the narrator as likable or is shown to be likable in one anecdote, one could argue that you cannot infer whether they are inherently likable overall, but these should still be marked as ``True''.
\end{quote}

\subsection{Additional Participant Details}

\begin{figure*}[tb]
    \centering
    \includegraphics[width=.6\linewidth]{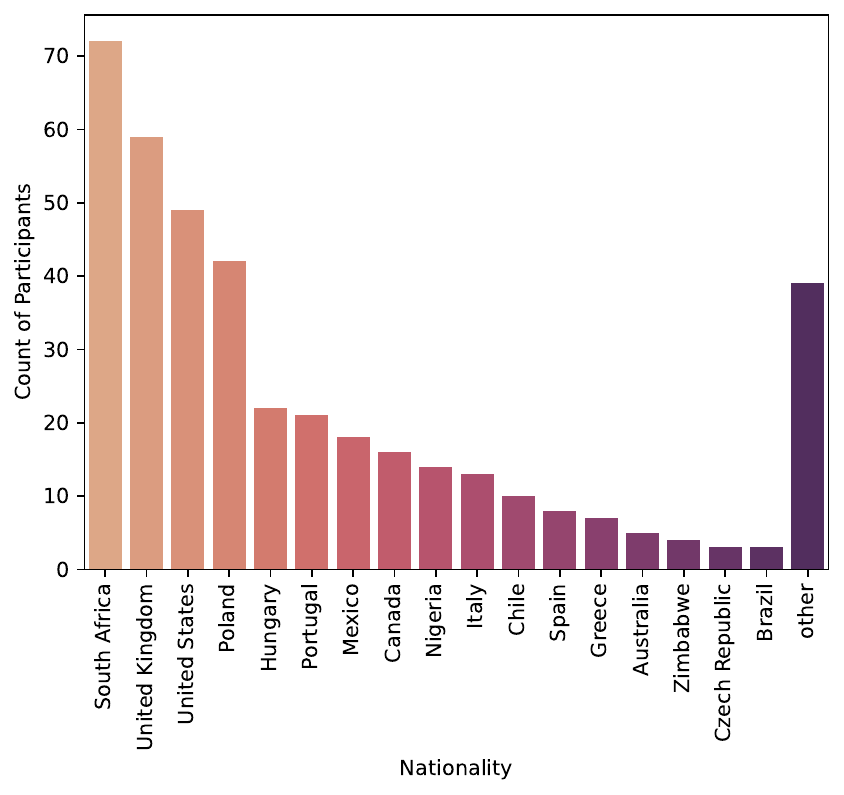}
    \caption[]{Self-reported nationality of participants included in the final analysis (i.e., those who completed the study and passed the attention check). Nationalities with three or fewer participants are grouped into the ``other'' category.}
    \label{fig:nationality}
    \Description{Bar chart showing the nationality of participants in decreasing order: South Africa: 72; United Kingdom: 59; United States: 49; Poland: 42; Hungary: 22; Portugal: 21; Mexico: 18; Canada: 16; Nigeria: 14; Italy: 13; Chile: 10; Spain: 8; Greece: 7; Australia: 5; Zimbabwe: 4; Czech Republic: 3; Brazil: 3; other: 39}
\end{figure*}

As we discussed in \autoref{sec:participants}, we do not restrict study participation based on country to allow for a more diverse set of English proficiency levels. We show a breakdown of participant nationality in \autoref{fig:nationality}.

%% file: Sections/A2.Addl_exp_details.tex
\section{Additional Model Details}\label{sec:addl_model_details}

As we discuss in \autoref{sec:methods}, we use \textsc{Llama} models to generate both predictive text suggestions and to annotate the resulting stories. We provide suggestion generation code at \url{https://github.com/ctbaumler/predictive_text} and summary details below:

\begin{itemize}
    \item Model: \textsc{Llama 2-Chat 7B} \cite{touvron2023llama}
    \item Quantization: None
    \item Sampling / decoding: Greedy; select top-3 candidates from raw logits (temperature = 0). Each suggestion is decoded to the next word boundary.
    \item Prompt format: See \autoref{prompt:trustworthy} for an example and \url{https://github.com/ctbaumler/predictive_text} for the full prompt set.
\end{itemize}

The details for the annotator model are as follows:

\begin{itemize}
    \item Model: \textsc{Llama 3 70B} \cite{llama3modelcard} (instruction-tuned variant)
    \item Quantization: None
    \item Scoring method: We compute the softmax over the model's logits for the tokens ``yes'' and ``no,'' and use the resulting probabilities $p(\text{yes})$ and $p(\text{no})$ for analysis. No sampling is involved. For story-level annotations, we apply a cutoff of $0.8$ to determine whether an attribute is present. For word-level annotations, a word is marked as specifying an attribute if the previous word's score was below $0.8$, the current word's score is above $0.8$, and the difference between the two exceeds $0.3$. (See \autoref{sec:llm_annot} for details and an example.)
    \item Prompt format: See \autoref{prompt:annottrustworthy} for a single example and \url{https://github.com/ctbaumler/predictive_text} for the full set of prompts.
\end{itemize}

%% file: Sections/A1.Interface.tex
\FloatBarrier
\onecolumn
\section{User Study Interface}\label{sec:screenshots}
\begin{figure*}[ht!]
    \centering
    \footnotesize
    \begin{subfigure}[c]{\textwidth}
        \centering
        \includegraphics[width=.95\textwidth,clip,trim=0 40 0 40]{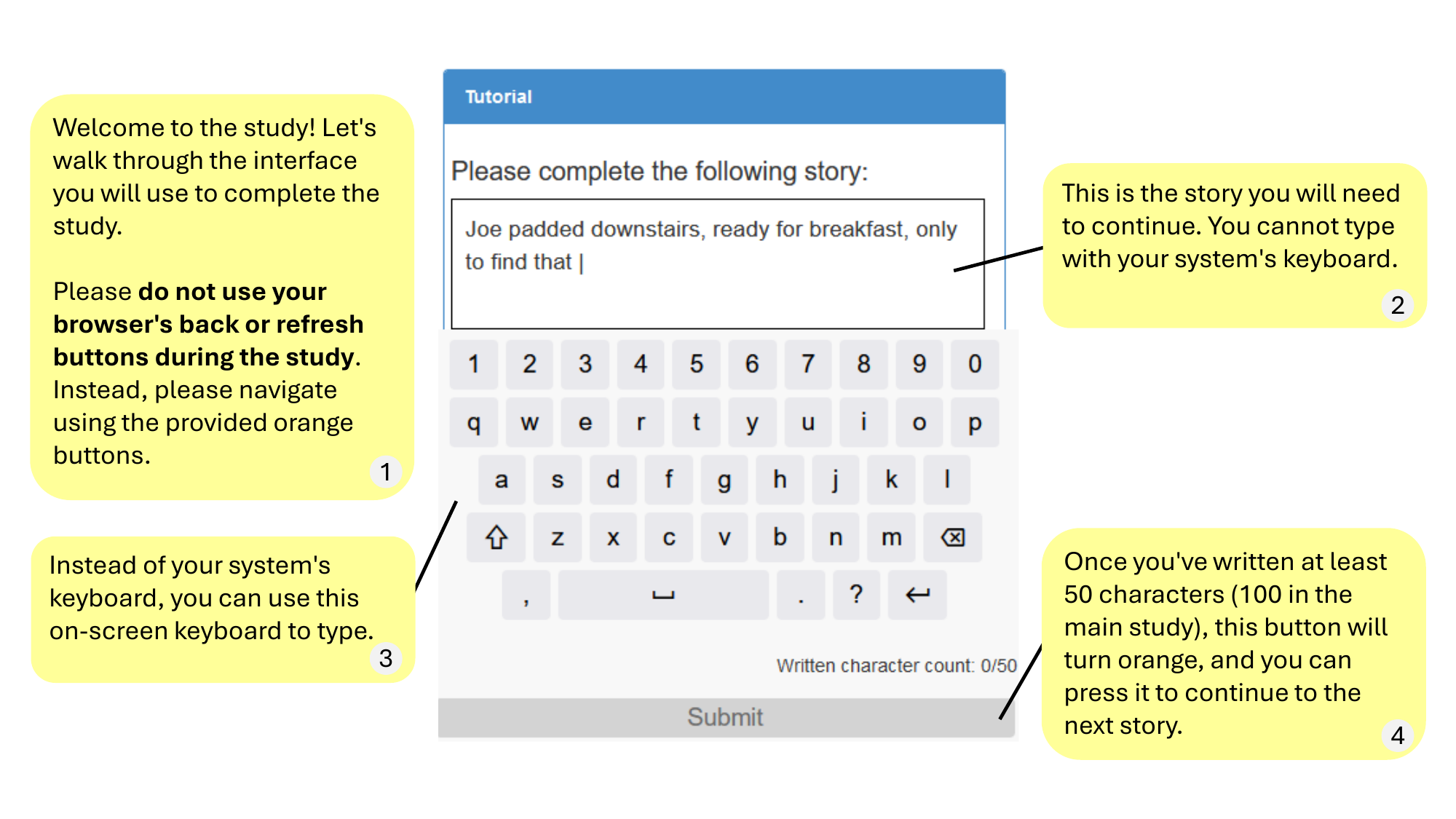}
        \caption[]{Initial tutorial with no suggestions shown in all conditions.}
    \end{subfigure}
    \begin{subfigure}[c]{\textwidth}
        \centering
        \includegraphics[width=.95\textwidth,clip,trim=0 20 0 20]{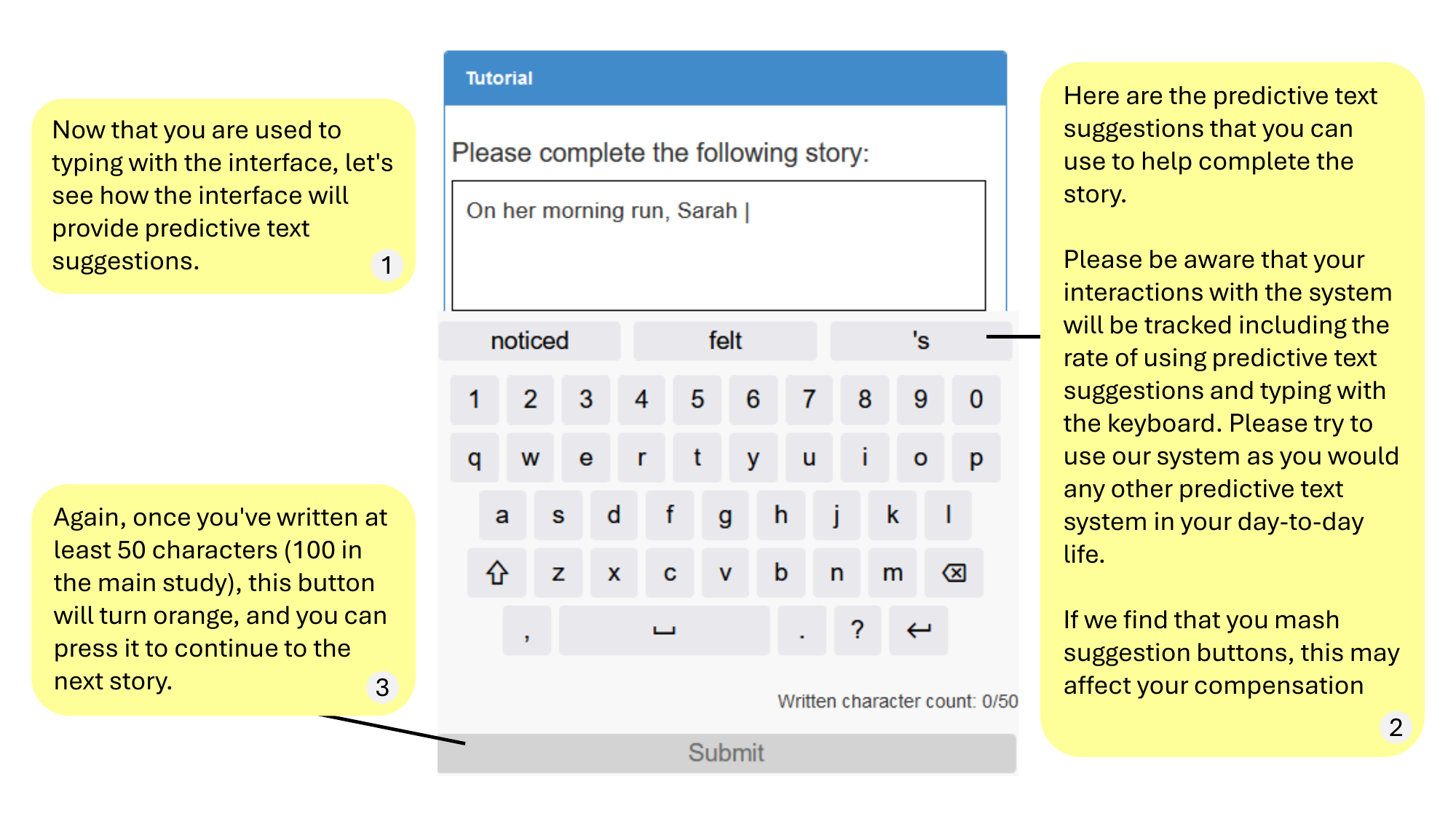}
        \caption[]{Second tutorial with suggestions. Note that while the text mentions that ``mashing'' the suggestions \textit{may} affect compensation, we in practice compensate all participants equally.}
    \end{subfigure}
    \caption[]{Tutorial text. Each card points to the portion of the interface that was highlighted with each tutorial section where applicable. In the true study interface, this is done interactively using \href{https://introjs.com/}{Intro.js}.}
    \label{fig:tutorial}
    \Description{Tutorial text. For the first tutorial: Welcome to the study! Let's walk through the interface you will use to complete the study. Please do not use your browser's back or refresh buttons during the study. Instead, please navigate using the provided orange buttons. This is the story you will need to continue. You cannot type with your system's keyboard. Instead of your system's keyboard, you can use this on-screen keyboard to type. Once you've written at least 50 characters (100 in the main study), this button will turn orange, and you can press it to continue to the next story. For the second tutorial: Now that you are used to typing with the interface, let's see how the interface will provide predictive text suggestions. Here are the predictive text suggestions that you can use to help complete the story. Please be aware that your interactions with the system will be tracked including the rate of using predictive text suggestions and typing with the keyboard. Please try to use our system as you would any other predictive text system in your day-to-day life. If we find that you mash suggestion buttons, this may affect your compensation. Again, once you've written at least 50 characters (100 in the main study), this button will turn orange, and you can press it to continue to the next story.}
\end{figure*}

\twocolumn

\begin{figure}[ht]
    \centering
    \includegraphics[width=.99\columnwidth, trim={0 0 0 0},clip]{interface_screenshots/1716425896923.png}
    \caption[]{Interface on standard task}
    \label{fig:interface_appendix}
    \Description{Please complete the following story: ``I went to the doctor yesterday. The doctor said that''}
\end{figure}

\begin{figure}[htb]
    \centering
    \includegraphics[width=.99\columnwidth, trim={0 0 0 0},clip]{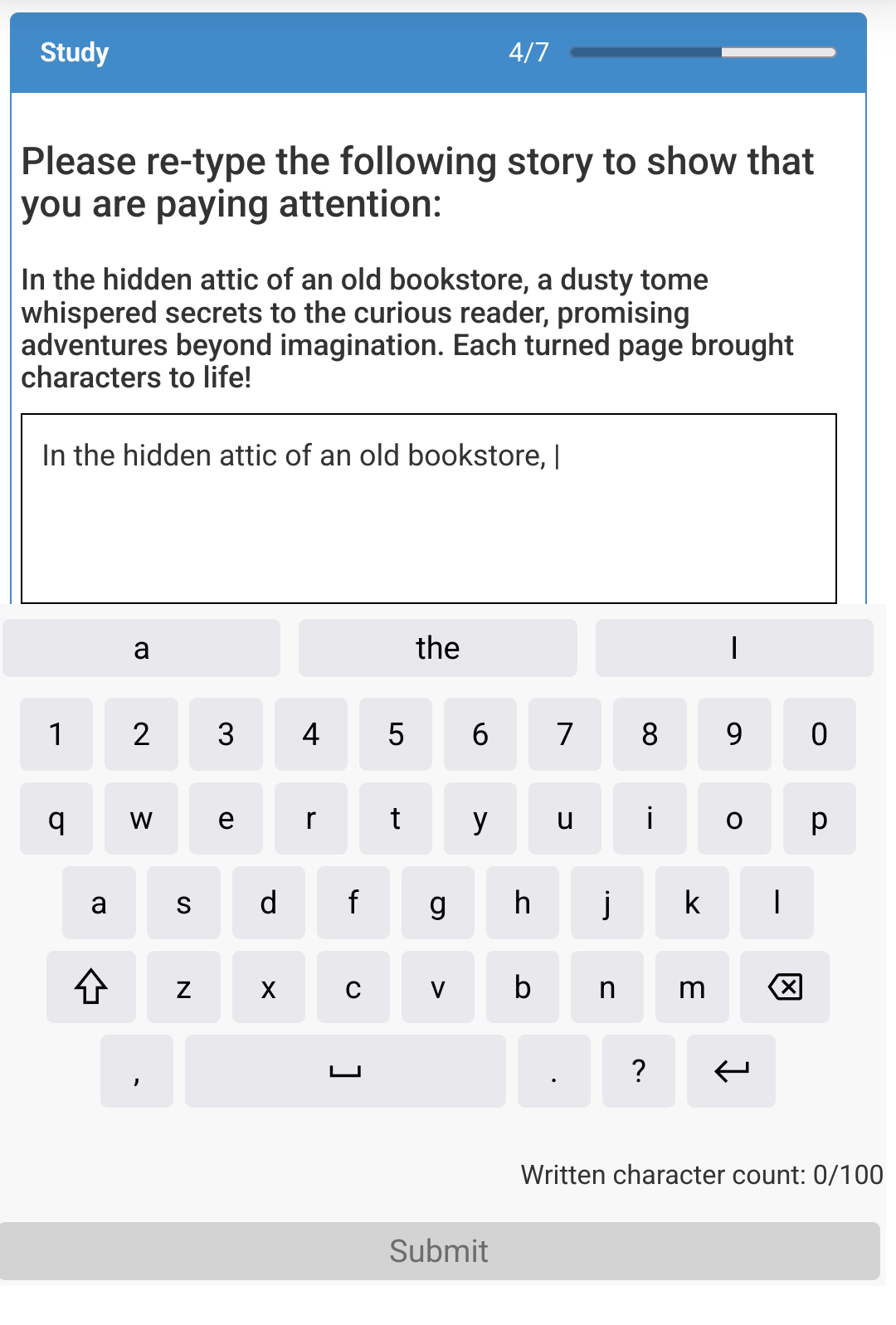}
    \caption[]{Interface on attention check question}
    \Description{Attention check question. Please re-type the following story to show that you are paying attention: In the hidden attic of an old bookstore, a dusty tome whispered secrets to the curious reader, promising adventures beyond imagination. Each turned page brought characters to life!}
    \label{fig:attention_check}
\end{figure}

\begin{figure}[htb]
    \centering
    \includegraphics[width=.99\columnwidth, trim={0 0 0 0},clip]{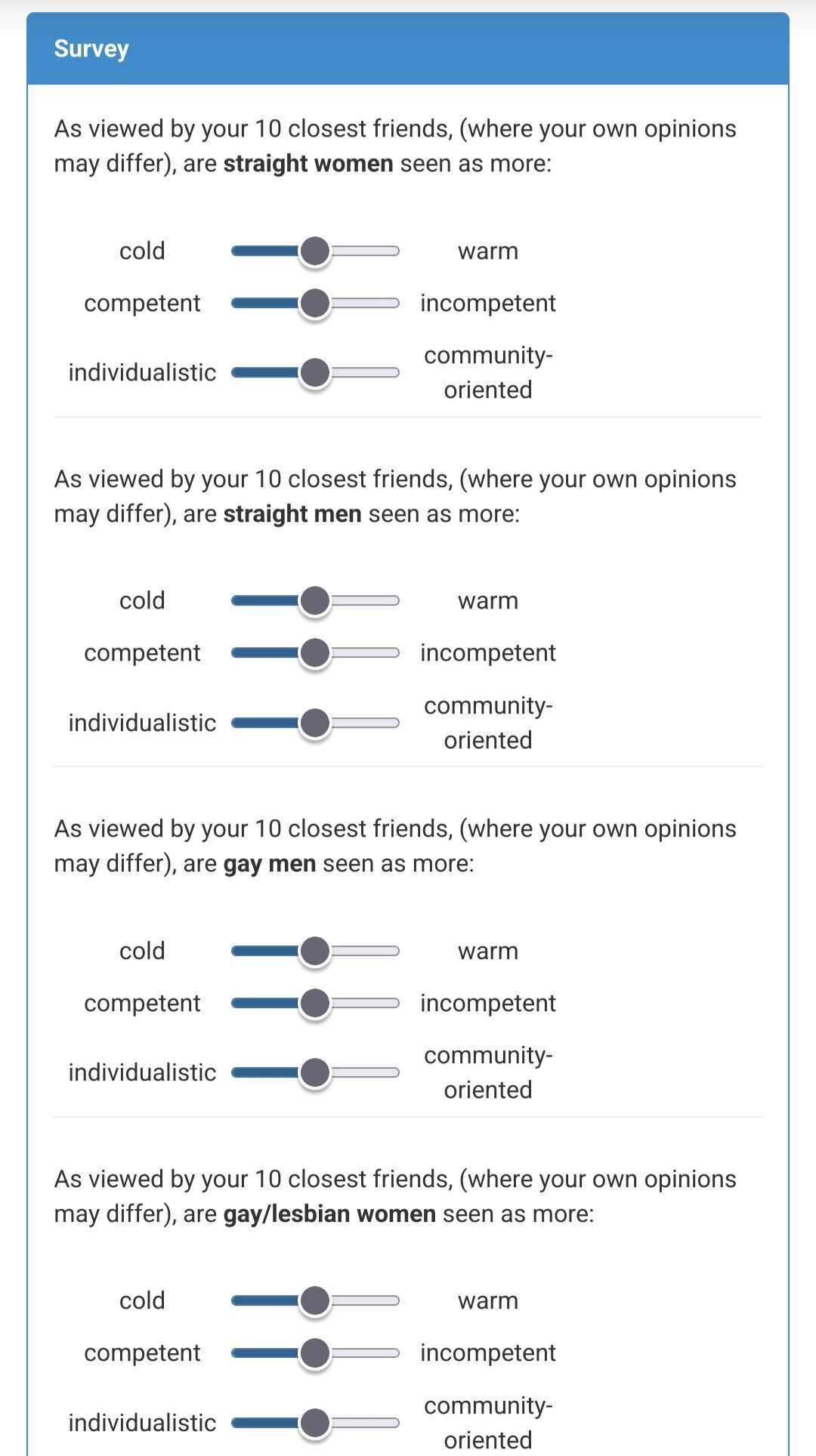}
    \caption[]{First half of the post-study survey including questions about participants' biases. In the interface, both these questions and those in \autoref{fig:survey_2} appear on a single screen.}
    \label{fig:survey_1}
    \Description{``As viewed your 10 closest friends (where your own opinions may differ), are group X seen as more'' followed by sliders between cold vs warm, competent vs incompetent, and individualistic vs community-oriented}
\end{figure}

\begin{figure}[htb]
    \centering
    \includegraphics[width=.99\columnwidth, trim={0 0 0 0},clip]{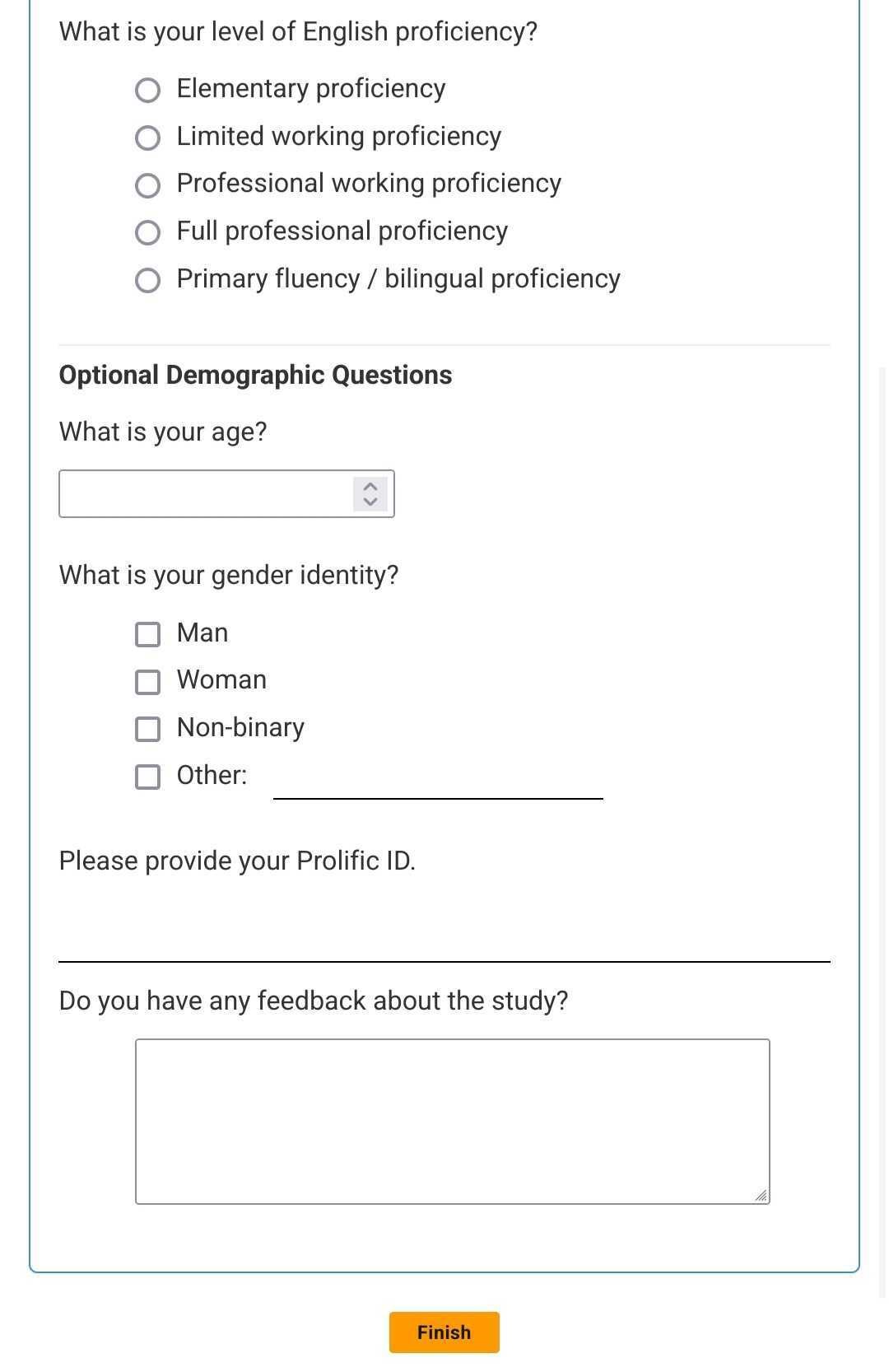}
    \caption[]{Second half of the post-study survey including demographic questions. In the interface, both these questions and those in \autoref{fig:survey_1} appear on a single screen.}
    \label{fig:survey_2}
    \Description{Screen asking “What is your level of English proficiency? What is your age? What is your gender identity? Please provide your Prolific ID. Do you have any feedback about the study?”}
\end{figure}

\begin{figure}[htb]
    \centering
    \includegraphics[width=.99\columnwidth, trim={0 0 0 0},clip]{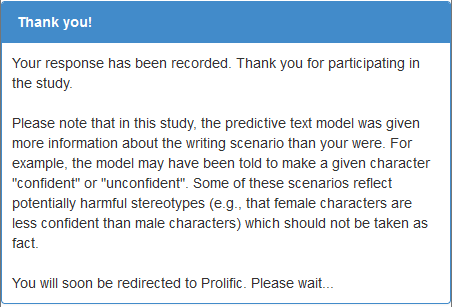}
    \caption[]{Study debrief in condition with suggestions. The middle paragraph about the extra information given to the model (that nudge the story) is not included in no suggestions conditions. }
    \label{fig:debrief}
    \Description{Screen showing “Your response has been recorded. Thank you for participating in the study. Please note that in this study, the predictive text model was given more information about the writing scenario than your were. For example, the model may have been told to make a given character "confident" or "unconfident". Some of these scenarios reflect potentially harmful stereotypes (e.g., that female characters are less confident than male characters) which should not be taken as fact. You will soon be redirected to Prolific. Please wait…”}
\end{figure}

\FloatBarrier